\documentclass[12pt,authoryear,review]{elsarticle}

\usepackage[bookmarks=true, colorlinks=true]{hyperref}
\usepackage{mathrsfs}
\usepackage{url}

\usepackage{makecell}
\usepackage{array}
\usepackage{arydshln}
\usepackage{caption}
\usepackage{amssymb,amsmath}
\usepackage{subfigure}
\usepackage{algorithm}
\usepackage{algorithmicx}
\usepackage{setspace}
\usepackage{algpseudocode}
\usepackage{multirow}
\usepackage{booktabs}
\usepackage{mathrsfs}
\usepackage{lineno}
\usepackage{multirow}\usepackage{makecell}
\usepackage{tikz}
\usepackage[export]{adjustbox}
\usepackage{array}
\usepackage{float}
\usepackage{placeins}
\usepackage{graphicx}
\usepackage{tabularx}

\journal{ISPRS Journal of Photogrammetry and Remote Sensing}

\begin{document}

\begin{frontmatter}

\title{RoofSeg: An edge-aware transformer-based network for end-to-end roof plane segmentation}
\author[1]{Siyuan You } 
\author[1]{Guozheng Xu } 
\author[1]{Pengwei Zhou } 
\author[3]{Qiwen Jin } 
\author[1,2]{Jian Yao}
\author[1,2]{Li Li\corref{CorAuthor}}

\cortext[CorAuthor]{Corresponding author.}
\ead{li.li@whu.edu.cn}	
\address[1]{School of Remote Sensing and Information Engineering, Wuhan University, Wuhan
	430079, China}
\address[2]{Wuhan Unversity Shenzhen Research Institute, Shenzhen 518057, China.}
\address[3]{School of Information Engineering, Wuhan University of Technology, Wuhan 430070, China}

\begin{abstract}
Roof plane segmentation is a crucial technique for reconstructing 3D building models (LoD~2–-3) from airborne LiDAR point clouds. Although traditional approaches can be effective for regular planar scenes, their handcrafted features are mainly computed from local neighborhoods, making them sensitive to noise, occlusions and point density variations. Learning-based approaches alleviate these limitations by modeling local geometry and global contextual information to learn more robust point-wise representations. However, current deep learning-based approaches still have three unsolved problems: (1) most are not fully end-to-end, leading to error accumulation across multiple processing stages; (2) low feature discriminability near edge regions results in inaccurate edge segmentation; and (3) in training step, geometric constraints (e.g. point-to-plane consistency) are not sufficiently considered, which may reduce geometric fidelity of plane segments. To address these issues, we propose RoofSeg, an edge-aware transformer-based network that segment roof plane instances in a truly end-to-end manner. RoofSeg employs a transformer encoder-decoder architecture with learnable plane queries, enhanced by a carefully designed Edge-Aware Mask Module (EAMM) for accurate edge segmentation. Additionally, to eliminate misclassified points and improve geometric fidelity, an adaptive weighting mask loss and a plane geometric loss are proposed to further supervise network training. Extensive experimental results show RoofSeg achieves accurate edges, minimal misclassifications, and high geometric fidelity, outperforming existing segmentation approaches. The source code is available at \url{https://github.com/Li-Li-Whu/RoofSeg}.
\end{abstract}

\begin{keyword}
	Roof plane segmentation \sep 3D building reconstruction \sep transformer  \sep aiborne LiDAR  \sep instance segmentation
\end{keyword}

\end{frontmatter}
%

\section{Introduction}
\label{sec:introduction}

Airborne light detection and ranging (LiDAR) can rapidly acquire large-scale three-dimensional (3D) point clouds with high geometric precision, and has been widely used in 3D building reconstruction~\citep{li2016manhattan, nan2017polyfit, zhang2018large, huang2022city3d, liu2024point2building, sun2024semantic} at levels of detail (LoD) 2 and 3. Since the roof structure of an urban building is generally composed of geometrically varied planar patches, roof plane segmentation is a key procedure for 3D building reconstruction. Accurate roof plane segmentation can significantly improve the final reconstruction quality. However, despite many studies~\citep{khaloo2017robust, wang2019improved, liu2023roof, li2024boundary}, plane instance segmentation from roof point clouds remains challenging for existing 3D reconstruction systems~\citep{yu2022finding}, especially under complex roof topology, uneven point density, noise, occlusions, and ambiguous edge regions.

Recently, the end-to-end building reconstruction method Point2Building~\citep{liu2024point2building} has achieved remarkable progress through a vertex-prediction followed by face-inference paradigm. However, it may be less effective on sparse point clouds or highly complex roofs, where vertex prediction can be incomplete and geometric recovery may be limited. Robust geometric priors, such as accurate roof plane segments, can help guide the inference process, further motivating effective roof plane segmentation methods.

Traditional roof plane segmentation approaches are mainly based on point feature grouping mechanisms. Guided by manually designed features, such as normal vectors or geometric distances, they use region growing~\citep{khaloo2017robust, wu2019accurate}, RANdom SAmple Consensus (RANSAC)~\citep{xu2015investigation, canaz2020improved}, or clustering~\citep{miao2023single, chen2024point} to group coplanar points with similar geometric properties. However, these manually designed features may lack robustness and discriminability in complex roof scenes.
To alleviate these limitations, several deep learning-based approaches~\citep{zhang2022improved, li2024boundary, zeng2025sppsformer} have been proposed to learn more representative point features from roof point clouds. Nevertheless, existing learning-based methods still face three main issues. First, many of them are not fully end-to-end, since they still rely on geometric voting, clustering~\citep{zhang2022improved, li2024boundary}, or post-processing such as edge refinement~\citep{zeng2025sppsformer} to generate final roof segments, which may introduce accumulated errors, computational overhead, and hyperparameter sensitivity. Second, they still struggle to learn instance-discriminative features near roof plane edges, leading to inaccurate edge segmentation. Third, explicit planar geometric constraints, such as point-to-plane consistency, and effective outlier suppression are often insufficiently incorporated into network training, which may reduce geometric fidelity and cause misclassified points across adjacent planes.

To address these issues, we develop an end-to-end edge-aware transformer-based network, named RoofSeg, for segmenting roof planes from airborne LiDAR point clouds. Inspired by transformer-based instance segmentation~\citep{cheng2021per, schult2023mask3d, lai2023mask, roh2024edge}, RoofSeg uses learnable plane queries in an encoder-decoder architecture to directly detect potential planes and predict their instance masks. By avoiding time-consuming and parameter-sensitive post-processing, the proposed end-to-end pipeline improves efficiency, robustness, and segmentation accuracy.
Furthermore, we design an Edge-Aware Mask Module (EAMM) to improve edge segmentation. Since learned edge features in high-dimensional embedding space can be ambiguous, EAMM explicitly incorporates point-to-plane distance as a geometric cue to guide edge feature learning in Euclidean space, thereby producing more accurate plane masks near edge regions. In addition, we introduce a novel loss function with an adaptive-weighting strategy to suppress misclassified points and a plane geometric loss to enforce coplanarity in predicted plane masks. Extensive experiments show that RoofSeg achieves state-of-the-art performance across multiple benchmarks, producing roof plane segments with accurate edges, minimal outliers, and high geometric fidelity. Its generalization capability is further verified on two generic plane segmentation benchmarks.The main contributions of this work are summarized as below:

\begin{itemize}
\item We introduce the first fully end-to-end transformer-based framework, RoofSeg, for roof plane segmentation from airborne LiDAR point clouds. It fundamentally overcomes the limitations of efficiency, robustness, and accuracy inherent in multi-stage pipelines that rely on time-costing and parameter-sensitive post-processing techniques.
\item To address the challenge of inaccurate edge segmentation, we design an Edge-Aware Mask Module (EAMM). Its core innovation lies in explicitly incorporating the key geometric cue, i.e., point-to-plane distance, into the feature learning process. This effectively guides the network to learn more discriminative features in Euclidean space for edge points, leading to refined plane masks with accurate edges.
\item To supervise network training, we propose a novel loss function with two key components: (1) an adaptive-weighting strategy for effective outlier suppression, and (2) a plane geometric loss to constrain coplanarity of the predicted planar points. This dual design ensures the final predictions achieve both minimal outliers and high geometric fidelity.
\item Extensive evaluations on multiple benchmarks (including three roof-specific and two generic plane benchmarks) show that RoofSeg establishes the new state-of-the-art results, evidencing its superior performance and strong generalization capability.
\end{itemize}
%

\section{Related work}
\label{sec:related work}
								
In general, the existing approaches developed for roof plane segmentation can be classified as two categories, namely handcrafted features-based and deep-learning-based approaches. Moreover, we discuss some end-to-end transformers for instance segmentation, which has significantly inspired our work.

\subsection{Handcrafted feature-based approaches}

We classify the current handcrafted features-based approaches as four categories: region growing-based, model-fitting-based, clustering-based and optimization-based approaches. 

Region growing-based approaches~\citep{vo2015octree, khaloo2017robust, wu2019accurate, poux2022automatic, wang2023indoor, chen2025facet} gradually grow the planar patches based on some initial seeds, according to certain geometric growing criteria.~\cite{vo2015octree} proposed to split initial points into octree-based voxels and extract features of each voxel based on point normal analysis, then they performed voxel-based growing to achieve point cloud instance segmentation. To achieve robust normal estimation,~\cite{khaloo2017robust} applied statistical outlier analysis to detect and remove outliers from the neighborhood of each point. Furthermore,~\cite{wang2023indoor} proposed a robust estimator-based Principal Component Analysis (PCA) for normal estimation to improve efficiency while ensuring accuracy.~\cite{chen2025facet} first segmented points into hypervoxel with varied scales, and then performed average distance analysis to split the non-planar voxels. After that, they applied an improved region growing algorithm to generate the fine plane segments. Nevertheless, these approaches tend to fail in the edge regions, because of their sensitivity to the parameter setting in the growing criteria.

Model-fitting-based approaches can be classified into two categories: Hough Transform (HT)-based~\citep{overby2004automatic, liu2005building, tarsha2007hough} and RANSAC-based approaches~\citep{schnabel2007efficient, xu2015investigation, li2017improved, canaz2020improved, xu2023optimal, cao2025enhanced}. HT-based approaches attempt to estimate the best plane models that maximally matches input points in the parameter space, then they allocate points to their nearest plane for generating plane segments. However, according to~\cite{tarsha2007hough}, these approaches are less efficient than RANSAC-based approaches and also yield inferior segmentation accuracy.
RANSAC-based approaches randomly sample points for plane fitting, and then determine the best fitted plane with the most inliers.~\cite{xu2015investigation} added a weighted loss function, based on two geometric constraints, into RANSAC for higher-quality plane fitting.~\cite{canaz2020improved} improved RANSAC by using an outlier-detecting algorithm to remove unreliable sampling points.~\cite{cao2025enhanced} introduced a point filtering method into the RANSAC for eliminating the noises, and then applied Kd-Tree algorithm to remove the non-planar points for segmentation refinement.
Nevertheless, these approaches are prone to generate spurious planar patches when handling roofs with complex structures, as the algorithms are prone to converge to the local optima.

Clustering-based approaches~\citep{he2013line, czerniawski20186d, wang2019improved, cao2022fec, miao2023single, chen2024point} perform roof point cloud instance segmentation by using some clustering techniques. For example,~\cite{he2013line} first extracted line segments from LiDAR point clouds, and then performed line-based clustering in the spectral space to produce the optimal fitted plane segments.~\cite{czerniawski20186d} applied 6D Density Based Spatial Clustering of Applications with Noise (DBSCAN) to group points in both coordinate and normal vector spaces. Besides,~\cite{cao2022fec} proposed a Fast Euclidean Clustering (FEC) method to reduce the computational complexity of conventional Euclidean Clustering (EC).~\cite{chen2024point} first extracted collision points in boundaries. Then, they used adaptive EC method to generate initial clusters and reinserted the collision points to achieve complete instance segmentation of roof point clouds. However, these approaches heavily  rely on the fine-tuned parameters, such as clustering radius, and are limited by the low quality of handcrafted features.

Optimization-based approaches~\citep{yan2014global, gu2017hierarchical, li2020roof, wang2021roof, yu2022finding, weixiao2023pssnet, liu2023roof, xu2025coarse} attempt to minimize certain carefully designed energy function for refining segmentation results produced by the above mentioned approaches.~\cite{yan2014global} constructed a global multi-label energy function to measure the quality of plane segments, and optimized the energy function with graph cuts technique~\citep{boykov2006graph}. In addition,~\cite{yu2022finding} proposed to optimize a multi-objective energy function in an efficient mechanism with five geometric operators. However, the global optimization-based refinement strategies are relatively time-consuming and resource-intensive. To alleviate that, ~\cite{li2020roof}, ~\cite{liu2023roof} and ~\cite{xu2025coarse} shifted focus from global to local domains, specifically targeting at edge regions. They gradually refined edges until obtaining the optimal edge point assignment with minimal energy cost. Yet, they still require setting complex convergence criteria to terminate the optimization of energy functions. And, the improper thresholds can lead to premature termination (sub-optimal segmentation results) or excessive iterations (increased runtime without meaningful gain).

\subsection{Deep learning-based approaches}

With the rapid advancement of deep learning, several approaches have been developed for primitive instance segmentation~\citep{jiang2020pointgroup, huang2021primitivenet, yan2021hpnet, du2022push, li2023surface, wang2023probabilistic, fang2025bgpseg} and roof plane segmentation~\citep{zhao2022extracting, zhang2022improved, li2024boundary, zeng2025sppsformer} from 3D point clouds. Generally, these methods first learn point-wise feature representations with deep networks and then apply feature grouping or clustering techniques to generate final segments. PrimitiveNet~\citep{huang2021primitivenet} jointly learned explicit and implicit point-wise features for distinguishing different primitives, while HPNet~\citep{yan2021hpnet} extracted hybrid representations integrating semantic information, geometric consistency and local smoothness, followed by mean-shift clustering for segmentation. For roof plane segmentation,~\cite{zhang2022improved} proposed a joint-learning network to learn semantic and instance features, where coplanar points are encouraged to be close in the embedding space for subsequent clustering.
However, these methods still have limited ability to exploit edge cues. To improve boundary quality,~\cite{li2024boundary} designed a three-branch boundary-refinement network, which first identifies edge points, groups non-edge points in Euclidean and embedding spaces, and then assigns edge points to nearby planes. More recently,~\cite{zeng2025sppsformer} proposed a superpoint-based transformer for roof plane segmentation, where high-quality superpoints and manually designed features are used to generate initial segments, followed by plane completion and edge refinement. Nevertheless, these learning-based roof segmentation methods still rely on clustering or post-processing steps, and thus cannot achieve truly end-to-end segmentation. In addition, learning instance-discriminative features for edge points remains challenging.

Transformer, first proposed by~\cite{vaswani2017attention} for Natural Language Processing, has achieved remarkable success in vision tasks due to its attention mechanism. Recently, end-to-end transformer-based methods~\citep{cheng2021per, cheng2022masked, schult2023mask3d, sun2023superpoint, lai2023mask, roh2024edge, xiao2025sapformer} have been introduced for 2D and 3D segmentation. Mask2Former~\citep{cheng2022masked} uses query-pixel cross-attention to support multiple image segmentation tasks, while Mask3D~\citep{schult2023mask3d} extends this idea to 3D instance segmentation by interacting transformer queries with multi-resolution point features. SPFormer~\citep{sun2023superpoint} further introduces superpoints to aggregate point features for query-based instance prediction.
Despite their effectiveness, existing transformer-based segmentation methods still face difficulties in accurately segmenting edge regions. Edge-aware designs have been explored by~\cite{roh2024edge}, but the edge guidance remains relatively implicit. More recently,~\cite{xiao2025sapformer} demonstrated the importance of shape-aware features for point cloud segmentation. Inspired by these studies, we develop an end-to-end edge-aware transformer-based network for roof plane segmentation in this paper.

\section{Our approach}
\label{sec:our approach}
\begin{figure*}[h]
	\centering
	\includegraphics[width= 1.00\linewidth]{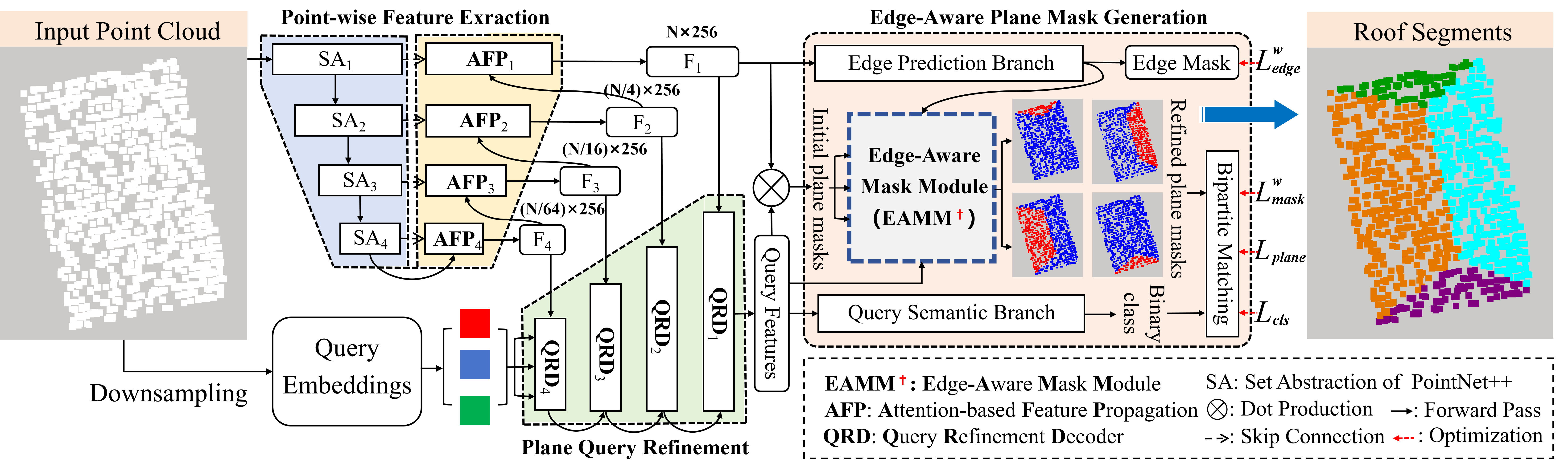}
	\caption{The overview of the end-to-end edge-aware transformer-based network (RoofSeg). AFP is a variant of the Feature Propagation (FP) layer from PointNet++, which incorporates attention-based operations, while Set Abstraction (SA) follows the original PointNet++ design. QRD is a plane query decoder constructed with standard transformer blocks. EAMM is the core module designed for edge-aware plane mask generation, guided by key geometric cues, i.e., point-to-plane distance. Notably, the outputs of each QRD are used for plane mask prediction. For readability, these processes are omitted in this figure.}
	\label{Fig:framework}
\end{figure*}

The architecture of RoofSeg is illustrated in Fig.~\ref{Fig:framework}. The overall objective of RoofSeg is to directly predict individual roof plane instances with accurate edges and high geometric fidelity. To this end, the PointNet++\citep{qi2017pointnet++}-based backbone is applied to first extract multi-scale point features for roof plane discrimination (Section~\ref{sec:point feature edxdtraction}). To generate initial plane masks, the Query Refinement Decoders (QRDs) are applied to progressively integrate multi-scale contextual information into the query embeddings (generated from $K$ sammpled query points) to obtain refined query features $\mathbf{Q}_{ref}$. After that, $\mathbf{Q}_{ref}$ are used to interact with full-resolution point features $\mathbf{F}_{1}$ to derive the corresponding planar instance mask for each query (Section~\ref{sec:plane query refinement}). To further improve edge accuracy, based on the potential edge points identified by the edge prediction branch, the Edge-Aware Mask Module (EAMM) is applied to refine each plane mask by incorporating edge-related geometric priors to enhance edge discriminability (Section~\ref{sec:edge aware mask generation}). Finally, the query semantic branch predicts the binary class (positive/negative) of each refined mask, and bipartite matching is used during training to associate predictions with ground-truth plane instances and suppress duplicate predictions (Section~\ref{sec:loss functions}). During inference, all positive refined masks are merged to obtain the final roof plane segmentation results.


%

\subsection{Point-wise feature extraction}
\label{sec:point feature edxdtraction}
PointNet++~\citep{qi2017pointnet++} is a prevalent feature extraction backbone for processing 3D point clouds. Its hierarchical architecture, composed of Set Abstraction (SA) and Feature Propagation (FP) modules, can effectively extract point features that integrate both local and global information. Based on that, we introduce attention mechanism into the FP modules to learn more robust features through point-wise self-attention operations.

\textbf{Point set abstraction.} The input point clouds can be denoted as $\mathbf{P}=\{\mathbf{p}_{i}\}_{i=1}^{N}$, where $N$ denotes the count of input points. We take the spatial coordinates of $\mathbf{P}$ as its initial features $\mathbf{F}_{init} \in \mathbb{R}^{N \times 3}$. Then, four standard SA modules are employed to progressively perform point downsampling and feature aggregation, with a consistent sampling scale of 1/4 at each level. In this way, the global information are gradually encoded into the features of downsampled points.

\textbf{Attention-based feature propagation.} To recover the point features across different scales, we employ four Attention-based Feature Propagation (AFP) layers for feature decoding. The functional role of AFP is to produce high-quality, discriminative point-level features that are aware of both local geometry and broader contextual relationships. Specifically, in each layer,
as shown in Fig.~\ref{Fig:submodule} (left), the low resolution features and the skip connection features (from the associated SA layer) are fed into the standard FP layer for generating upsampled features. Then, an attention-based module is applied to learn point-wise correlations, thus facilitating more robust feature propagation. Through four AFP layers, we derive multi-scale point features with 256 dimensions, denoted as $\mathbf{F}_{multi} = \{\mathbf{F}_{l} \in \mathbb{R}^{(N/4^{l-1}) \times 256}\}_{l=1}^{L}$, where $L=4$ refers to the layer numbers and $\mathbf{F}_{1}$ is the 
 full resolution point features with the smallest scale.
 \subsection{Plane query generation and refinement}
 \label{sec:plane query refinement}
Inspired by query-based transformers for instance segmentation, we attempt to utilize a fixed number of queries to gain the information of potential planes included in the roof point clouds. In practice, to avoid insufficient plane predictions, the query number is ensured to be larger than the actual plane number through a statistical analysis of the entire dataset.

 \textbf{Query embedding generation.} Through Farthest Point Sampling (FPS) method, we primarily sampled $K$ query points from the initial point clouds. This sampling method guarantees globally maximized sampling coverage. After that, we encode the spatial coordinates of query points with the Fourier encoding~\citep{misra2021end} to generate initial query embeddings $\mathbf{Q}_{emb} \in \mathbb{R}^{K \times 256}$. 
 Notably, compared to traditional positional encoding designed for ordered data (e.g., images or text), Fourier encoding demonstrates superior performance in processing unordered and unstructured point clouds~\citep{schult2023mask3d}. 
 
\begin{figure*}[h]
 	\centering
 	\includegraphics[width= 1.00\linewidth]{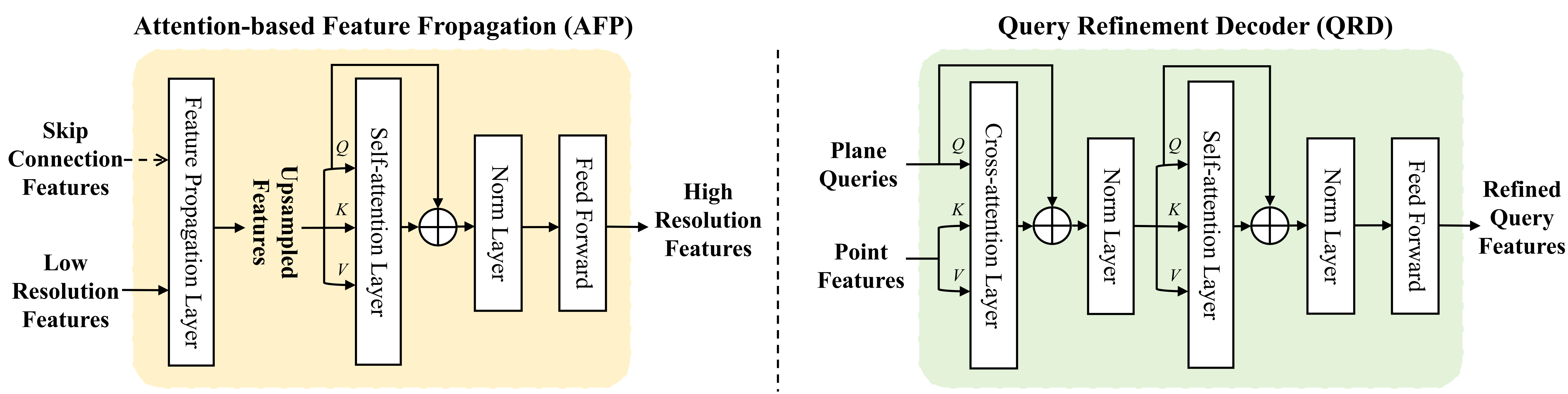}
 	\caption{The illustration of the Attention-based Feature Propagation (AFP) and Query Refinement Decoder (QRD). $Q$, $K$ and $V$ denotes the standard query, key and value embeddings produced for the subsequent attention operation. Notably, the outputs of AFP are point-wise features (with varied resolution across different layers), forming a multi-scale representation for the entire point clouds, while the outputs of QRD serve as a set of representations for specific plane instances.}
 	\label{Fig:submodule}
 \end{figure*}

\textbf{Hierarchical query refinement.}
To enable plane-level information learning, we apply four Query Refinement Decoders (QRDs) to gradually integrate multi-scale features from $\mathbf{F}_{multi}$ into the query embeddings $\mathbf{Q}_{emb}$. As a query-based decoding module, QRD transforms dense point-wise features into a sparse set of instance-aware query representations, thereby bridging per-point feature extraction and per-plane mask prediction. Specifically, $\mathbf{Q}_{emb}$ is used as the initial input of the first decoder and is progressively refined through four decoder layers. For the $(l-1)$-th decoder, its output query embeddings are denoted as $\mathbf{Q}_l$, which are then fed into the $l$-th decoder.Each decoder, as shown in Fig.~\ref{Fig:submodule} (right), follows a standard transformer block, including two attention layers and a feed-forward network. The first attention layer performs cross-attention between plane queries $\mathbf{Q}_{l} \in \mathbb{R}^{K \times 256}$ and point features $\mathbf{F}_{l} \in \mathbb{R}^{(N/4^{l-1}) \times 256}$, which can be described as:
\begin{equation}
	\mathbf{Q}_{cross\_att} = softmax(\frac{QK^{T}}{\sqrt{D}})V,
\end{equation}
where $Q$ is linearly projected from $\mathbf{Q}_{l}$, $K$ and $V$ are linearly projected from $\mathbf{F}_{l}$, and $D=256$ is the feature dimension. The obtained $\mathbf{Q}_{cross\_att}$ is concatenated with $\mathbf{Q}_{l}$ and then fed into the self-attention layer to enhance inter-query communication and suppress redundant predictions of the same plane instance. Finally, residual connection, layer normalization, and a feed-forward network are applied to output the refined query features. After four-stage refinement, the final query features $\mathbf{Q}_{ref} \in \mathbb{R}^{K \times 256}$ are used for subsequent plane mask prediction.
 
 \subsection{Edge-aware mask generation}
 \label{sec:edge aware mask generation}
 Based on the refined query features $\mathbf{Q}_{ref}$ and full resolution point features $\mathbf{F}_{1}$, we generate the initial plane mask for each plane query, and predict the edge mask simultaneously. Then, through an edge-aware mask module, we achieve further plane mask refinement with critical edge geometric cues.
 
 \textbf{Initial plane masks.} We computes the dot products between $\mathbf{Q}_{ref} \in \mathbb{R}^{K \times 256}$ and $\mathbf{F}_{1} \in \mathbb{R}^{N \times 256}$ to generate an affinity matrix $\mathbf{A} \in \mathbb{R}^{K \times N}$, where each element $a_{ij}$ represents the affinity between the $i$-th plane query and $j$-th point. This process can be formulated as bellow:
  \begin{equation}
 	\mathbf{A} = \{a_{ij} = (\mathbf{Q}_{ref} \otimes \mathbf{F}_{1}^{T})_{ij}\}.
 \end{equation}
 Through a sigmoid activation function, $\mathbf{A}$ is then mapped to a normalized value space, where $a_{ij} \in [0, 1]$. After that, we binarize the matrix using a threshold of 0.5, thereby deriving binary plane masks $\mathcal{M} \in \mathbb{R}^{K \times N}$, where a value of 1 indicates points inside the plane, and 0 denotes those outside. This procedure can be expressed as:
 \begin{equation}
 	\mathcal{M} = \{\mathbf{M}_{i} = \{[sigmoid(\mathbf{A}_{ij}) \geq 0.5]\}_{j=1}^{N}\}_{i=1}^{K},
 \end{equation}
 where $[\cdot]$ refers to an indicator for returning 1 or 0, according to whether the condition is satisfied or not. $\mathbf{M}_{i} \in \mathbb{R}^{1 \times N}$ refers to the $i$-th predicted plane mask belong to $\mathcal{M}$.

 \textbf{Edge masks.} To detect the edge points included in point clouds, we input $\mathbf{F}_{1}$ into the edge prediction branch, which is essentially a Multi-Layer Perceptron (MLP) with output dimension set as 1, for predicting the edge probability heatmap $\mathbf{H}_{edge} \in \mathbb{R}^{1 \times N}$. Then, similar to the plane mask generation, we apply normalization and binarization to $\mathbf{H}_{edge}$ for producing the binary edge mask $\mathbf{M}_{edge} \in \mathbb{R}^{1 \times N}$.
  
  \begin{figure*}[h]
  	\centering
  	\includegraphics[width= 1.00\linewidth]{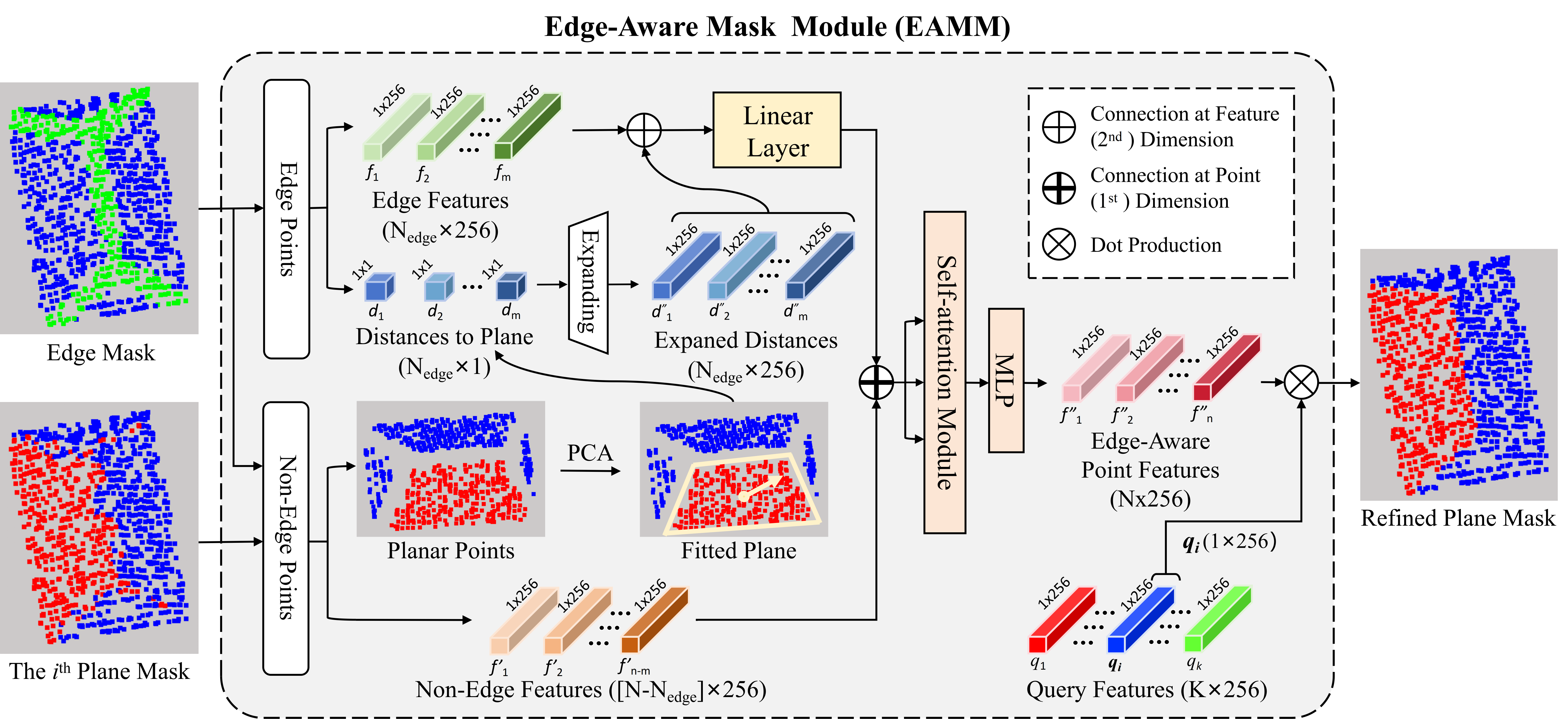}
  	\caption{The illustration of the Edge-Aware Mask Module (EAMM). The green dots in the edge mask represent the predicted edge points in point clouds. The red and blue dots denote the points inside and outside the predicted plane. For simplicity, this figure only illustrates the refinement of one plane mask as an example.}
  	\label{Fig:edge_aware}
  \end{figure*}
  
 \textbf{Edge-aware plane masks.}
As shown in Fig.~\ref{Fig:edge_aware}, the initial plane masks may be inaccurate in edge regions due to the limited discriminability of learned edge features. To address this issue, we introduce point-to-plane distance as a geometric cue and design an Edge-Aware Mask Module (EAMM) for mask refinement. Specifically, we first extract $N_{edge}$ edge points $\mathbf{P}_{edge} = \{e_i\}_{i=1}^{N_{edge}}$ and their features $\mathbf{F}_{edge} \in \mathbb{R}^{N_{edge} \times 256}$ from the edge mask $\mathbf{M}_{edge}$ and point-wise features $\mathbf{F}_{1}$. The planar interior points, excluding edge points, are then used to estimate the plane through PCA. Based on the fitted plane, the tangent distance $d_i$ from each edge point $e_i$ to the plane is computed and expanded into a 256-dimensional vector, forming $\mathbf{F}_{dist} \in \mathbb{R}^{N_{edge} \times 256}$.
Next, $\mathbf{F}_{edge}$ and $\mathbf{F}_{dist}$ are concatenated and passed through a fully connected layer to obtain fused edge features $\mathbf{F}_{edge\_fused} \in \mathbb{R}^{N_{edge} \times 256}$. These features are merged with non-edge features $\mathbf{F}_{non\_edge} \in \mathbb{R}^{(N-N_{edge}) \times 256}$ and processed by a self-attention module to generate edge-aware point features $\mathbf{F}_{edge\_aware} \in \mathbb{R}^{N \times 256}$. Finally, $\mathbf{F}_{edge\_aware}$ and the corresponding query feature $q_i$ are used to produce the refined plane mask.  Notably, edge information is used as auxiliary guidance rather than a rigid decision criterion. By combining learned edge features with point-to-plane geometric cues, EAMM improves edge discriminability and enables more robust plane mask refinement.

 \textbf{Plane mask merging.} Since the number of predicted plane masks $\mathcal{M} \in \mathbb{R}^{K \times N}$ may exceed the number of ground-truth planes, we first select positive masks according to their confidence scores $\mathbf{S}_{cls}=\{s_i\in[0,1]\}_{i=1}^{K}$ predicted by the query classification branch. Masks with $s_i \geq 0.5$ are retained and merged into a multi-label plane mask $\mathbf{M}_{merged} \in \mathbb{R}^{1 \times N}$, where each point label corresponds to a predicted plane instance rather than a binary foreground/background value. Following~\citep{schult2023mask3d}, each point $p_j \in \mathbf{P}$ is assigned to the most suitable positive plane by maximizing the product of the mask confidence score and the point-to-plane affinity:
 \begin{equation}
 	\mathbf{M}_{merged} = \{l_{p_{j}} = \mathop{\arg\max}\limits_{1 \leq i \leq N_{pos}} (s_i \cdot a_{ij})\}_{j=1}^{N},
 \end{equation}
 where $N_{pos}$ is the number of positive masks, $s_i$ is the confidence score of the $i$-th positive mask, and $a_{ij} \in \mathbf{A}$ denotes the affinity between $p_j$ and this plane. The merging operation is only used during inference for visualization and final segmentation output. During training, each predicted mask $\mathbf{M}_{i}$ is supervised independently by the plane mask loss without mask merging.

 \subsection{Loss functions}
 \label{sec:loss functions}
 RoofSeg is trained with a unified loss function $\mathcal{L}$, which is a combination of four parts: adaptive weighting plane mask loss $\mathcal{L}_{mask}^{w}$, plane geometric loss $\mathcal{L}_{plane}$, semantic class loss $\mathcal{L}_{cls}$, edge mask loss $\mathcal{L}_{edge}^{w}$ and outlier mask loss $\mathcal{L}_{outlier}^{w}$ (optional). We formulate it as follows: 
 \begin{equation}
 	\mathcal{L} = \mathcal{L}_{mask}^{w} + \mathcal{L}_{plane} + \mathcal{L}_{cls} + \mathcal{L}_{edge}^{w} + \mathcal{L}_{outlier}^{w}.
 \end{equation}
 
\begin{figure*}[h]
 	\centering
 	\includegraphics[width= 1.00\linewidth]{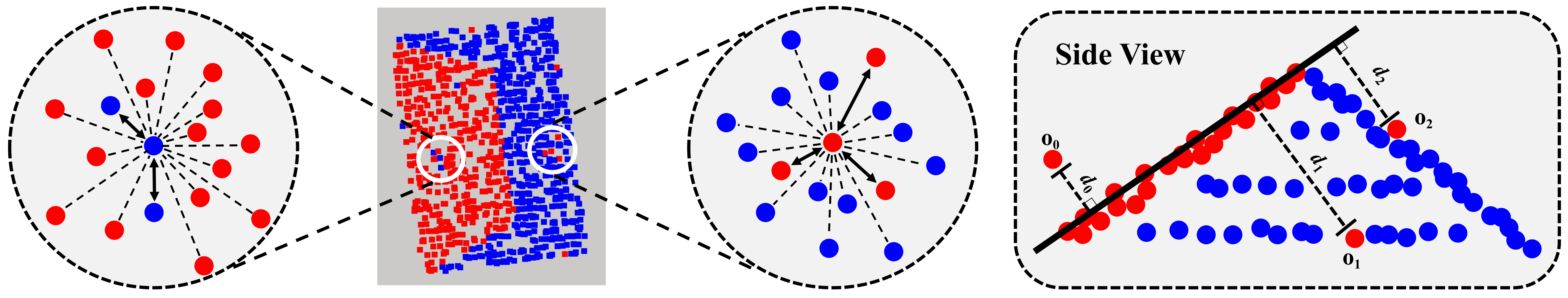}
 	\caption{The illustration of outliers. The red and blue dots represent the predicted in-plane points (with label 1) and out-of-plane points (with label 0). In highlighted circles, the dashed lines refer to the mismatching with inconsistent labels, while the double-headed solid lines refer to the correct matching with consistent labels. The dots located at circle centers represent the detected outliers. In side view, the long solid line denotes the fitted plane. The $o_{0}$, $o_{1}$ and $o_{2}$ represent three outliers existed in the predicted in-plane points.}
 	\label{Fig:mask_outliers}
 \end{figure*}

 \textbf{Plane mask loss.} The classic instance mask loss $\mathcal{L}_{mask}$, proposed by~\citep{schult2023mask3d}, consists of binary cross-entropy loss $\mathcal{L}_{BCE}$ and Dice loss $\mathcal{L}_{Dice}$:
 \begin{equation}
 	\mathcal{L}_{mask} = \sum_{\mathbf{M}_{i} \in \mathcal{M}} \mathcal{L}_{BCE} (\mathbf{M}_{i}) + \mathcal{L}_{Dice} (\mathbf{M}_{i}).
 \end{equation}
 However, this formulation treats all points equally and is therefore less effective in suppressing outliers within predicted masks, as shown in Fig.~\ref{Fig:mask_outliers}. To address this issue, we introduce an adaptive weighting mechanism into $\mathcal{L}_{mask}$, so that potential outliers receive stronger constraints during training and the network is encouraged to predict plane segments with fewer misclassified points. Specifically, we perform neighborhood analysis based on the predicted plane mask. For each point $p_j \in \mathbf{P}$, we calculate $N_{dif}^{j}$, which denotes the number of neighbors with labels different from that of $p_j$. If $N_{dif}^{j}$ exceeds half of the total number of neighbors $N_{nbr}$, $p_j$ is identified as a potential outlier. Through this process, $N_{out}$ outliers are detected and denoted as $\mathbf{O}=\{o_i\}_{i=1}^{N_{out}}$. The weight of each outlier is computed by combining a global factor, i.e., the ratio of inlier count $N-N_{out}$ to outlier count $N_{out}$, and a local factor, i.e., the ratio of $N_{dif}^{j}$ to $N_{nbr}$. The weights of all other points are set to one. In this way, mispredicted outliers generate larger gradients during backpropagation, enabling the network to better correct these cases. The adaptive weighting plane mask loss $\mathcal{L}_{mask}^{w}$ is formulated as:
 \begin{equation}
 	\mathcal{L}_{mask}^{w} = \sum_{\mathbf{M}_{i} \in \mathcal{M}} \mathcal{L}_{BCE}^{w} (\mathbf{M}_{i}) + \mathcal{L}_{Dice}^{w} (\mathbf{M}_{i}), 
 \end{equation}
 \begin{equation}
 	\mathcal{L}_{BCE}^{w} (\mathbf{M}_{i}) = -\frac{1}{N} \sum_{j=1}^{N} w_{ij} \left[l_{gt_{ij}} \cdot \log(a_{ij}) + (1 - l_{gt_{ij}}) \cdot \log(1 - a_{ij}) \right], 
 \end{equation}
 \begin{equation}
 	\mathcal{L}_{Dice}^{w} (\mathbf{M}_{i}) = 1 - \frac{2\sum_{j=1}^{N} w_{ij} (a_{ij} \cdot l_{gt_{ij}}) + \epsilon}{\sum_{j=1}^{N} w_{ij} a_{ij} + \sum_{j=1}^{N} w_{ij} l_{gt_{ij}} + \epsilon},
 \end{equation}
 \begin{equation}
 	w_j = 
 	\begin{cases} 
 		\dfrac{N - N_{out}}{N_{out}} \cdot \dfrac{N_{dif}^{j}}{N_{nbr}}, & \text{if } p_j \in \mathbf{O}, \\
 		1, & \text{otherwise},
 	\end{cases}
 \end{equation}
 where $w_j$ denotes the weight of the $j$-th point, $a_{ij}$ is the affinity of the $j$-th point in the $i$-th predicted plane mask $\mathbf{M}_{i}$, and $l_{gt_{ij}}$ is the label of the $j$-th point in the corresponding ground-truth mask. $\epsilon = 1$ is used as a smoothing term to avoid division by zero. Different from post-processing smoothing methods such as Conditional Random Field (CRF)~\citep{tchapmi2017segcloud}, which may produce jagged edges due to hand-crafted features and the difficulty of balancing smoothness and data terms, our adaptive weighting loss acts as an additional training constraint. Combined with EAMM, it helps preserve accurate edge regions while further reducing residual outliers near edges or within planes.

 \begin{algorithm}[!htb]
 	\setstretch{1.0}
 	\caption{Outlier detection \& adaptive weighting method.}
 	\label{Alg:adaptive weighting}
 	\renewcommand{\algorithmicrequire}{\textbf{Inputs:}}
 	\renewcommand{\algorithmicensure}{\textbf{Outputs:}}
 	\begin{algorithmic}[1]
 		\Require Point clouds $\mathbf{P}$; Plane mask $\mathbf{M}_{i} =\{m_{j}\}_{j=1}^{N}$; Number of neighbors $N_{nbr}$. \vspace{2.5 pt}
 		\Ensure Detected outliers $\mathbf{O}$; Weights of all points $\mathbf{W}=\{w_{1}, \cdots, w_{N}\}$. \vspace{2.5 pt}    
 		\\ \textbf{Initialization}: An empty set $\mathbf{O}$; A weight array $\mathbf{W} = \{w_1, \cdots, w_{N}\}$ with all zeros.
 		\\ \textbf{for} $j=0$ to $N$ \textbf{do}
 		\\  \qquad set $l_{j} = m_{j}$; calculate $\mathcal{N} (p_j) = \text{KNN}(p_j,\mathbf{P},N_{nbr})$; set $N_{dif}^{j} = 0$
 		\\  \qquad \textbf{for} $p_i$ in $\mathcal{N} (p_j)$ \textbf{do}
 		\\  \qquad \qquad set $l_{i} = m_{i}$  
 		\\  \qquad \qquad \textbf{if} $l_{i} == l_{j}$ \textbf{then} continue
 		\\  \qquad \qquad \textbf{else} set $N_{dif}^{j} = N_{dif}^{j}+ 1$
 		\\  \qquad \qquad \textbf{end if}
 		\\  \qquad \textbf{end for}
 		\\  \qquad \textbf{if} $N_{dif}^{j} > N_{nbr} / 2$ \textbf{then}
 		\\  \qquad \qquad insert $p_{j}$ to $\mathbf{O}$; set $w_{j} = N_{dif}^{j} / N_{nbr}$ 
 		\\  \qquad \textbf{end if}
 		\\  \textbf{end for}
 		\\  calculate $N_{out}$ = $|\mathbf{O}|$ 
 		\\  \textbf{for} $j=1$ to $N$ do
 		\\  \qquad \textbf{if} $w_{j}==0$ \textbf{then} set $w_{j}=1$
 		\\  \qquad \textbf{else} set $w_{j} = (N / N_{out} - 1) \cdot w_{j}$ 
 		\\  \qquad \textbf{end if}
 		\\  \textbf{end for}
 		\\  \textbf{return} $\mathbf{O}$, $\mathbf{W}$.
 		
 	\end{algorithmic}
 \end{algorithm}  
 
 \textbf{Plane geometric loss.} 
 In the current studies for instance segmentation, the geometric characteristics of instance are not sufficiently considered to constrain the network training, which may lead to the predicted instance segments with low geometric fidelity. To overcome this disadvantage, we naturally consider to add the plane geometric constrants, namely point-to-plane distance, into the network traning to ensure the high geometric fidelity of the plane predictions.
 For each predicted plane mask $\mathbf{M}_{i} =\{m_j\}_{j=1}^{N}$, we extract in-plane points, denoted as $\mathcal{P}=\{p_j \in \mathbf{P} \mid m_j = 1\}$, and apply PCA to fit the planar parametric equation based on in-plane points excluding outliers. Then, we compute the plane geometric loss, which can be formulated as:  
 \begin{equation}
 	\mathcal{L}_{plane} = \sum_{\mathbf{M}_{i} \in \mathcal{M}} \frac{1}{|\mathcal{P}|} \sum_{p_j \in \mathcal{P}} Dist(p_j, \Pi_{i}),
 \end{equation}
 where $Dist(\cdot)$ refers to point-to-plane distance, $\Pi_{i}$ denotes the $i$-th fitted plane based on $\mathbf{M}_{i}$. It should be noted that this loss formulation can also enhance the suppression of outliers. This is because, compared to in-plane points, the outliers generally exhibit larger distances to the fitted plane, as illustrated in Fig.~\ref{Fig:mask_outliers} (right).
 
 \textbf{Bipartite matching.} To eliminate the duplicated predictions, before loss calculation, we allocate the most suitable plane mask prediction to each ground truth mask through bipartite matching~\citep{schult2023mask3d}. Firstly, we match $K$ predicted plane masks with $N_{gt}$ ground truth masks, thus deriving $K \times N_{gt}$ matched mask pairs. Then, for each mask pair, we compute its matching cost $\mathcal{C}_{match}$ as:  
 \begin{equation}
 	\mathcal{C}_{match} = \mathcal{L}_{mask}^{w}(\mathbf{M}_{i}, \mathbf{\hat{M}}_{i}) + \mathcal{L}_{plane}(\mathbf{M}_{i}) - \mathbf{S}_{cls}^{\mathbf{M}_{i}},
 \end{equation}
 where $\mathbf{M}_{i}$ and $\mathbf{\hat{M}}_{i}$ refer to the predicted mask of the $i$-th plane and its matched ground truth mask. $\mathbf{S}_{cls}^{\mathbf{M}_{i}}$ denotes the confidence score of $\mathbf{M}_{i}$. After that, Hungarian algorithm~\citep{date2016gpu} is used to search for the best $N_{gt}$ mask pairs with the minimal total matching cost.
   
 \textbf{Binary class loss.} We aim to ensure that the predicted plane masks within the optimal matching pairs are identified as positive masks, while the other predicted plane masks are identified as negative masks. To achieve that, we use binary cross-entropy loss to construct binary class loss $\mathcal{L}_{cls}$, which can be described as:
  \begin{equation}
 	\mathcal{L}_{cls} = - \frac{1}{K} \sum_{i=1}^{K} \left[l_{cls_{i}} \cdot \log(s_i) + (1 - l_{cls_{i}}) \cdot \log(1 - s_i) \right],
 \end{equation}
 where $s_i$ and $l_{cls_{i}}$ refers to the confidence score and label of $i$-th predicted plane mask. Here, $l_{cls_i} = 1$ when the $i$-th predicted mask is matched with ground truth, and 0 otherwise.
 
 \textbf{Edge mask loss.} Similar to plane mask loss, we use $\mathcal{L}_{BCE}^{w}$ and $\mathcal{L}_{	Dice}^{w}$ to formulate the edge mask loss $\mathcal{L}_{edge}^{w}$, which can be defined as:
  \begin{equation}
 	\mathcal{L}_{edge}^{w} = \mathcal{L}_{BCE}^{w}(\mathbf{M}_{edge},\mathbf{\hat{M}}_{edge}) + \mathcal{L}_{Dice}^{w}(\mathbf{M}_{edge}, \mathbf{\hat{M}}_{edge}),
 \end{equation}
 where $\mathbf{M}_{edge}$ and $\mathbf{\hat{M}}_{edge}$ refer to the predicted and ground truth edge mask. To generate $\mathbf{\hat{M}}_{edge}$, we perform neighborhood analysis for each point. If there exist neighbors with different labels, the point is identified as an edge point assigned with label 1; otherwise, it is considered a non-edge point assigned with label 0.
 
 \textbf{Outlier mask loss.} Considering that real-world point cloud data may contain noise points far from planar surfaces, namely outliers, these points should be robustly identified rather than being assigned to any plane. To address this, we introduce an optional outlier detection branch in parallel with the edge prediction branch. Similar to $\mathcal{L}_{edge}^{w}$, the outlier mask loss $\mathcal{L}_{outlier}^{w}$ is calculated as:
  \begin{equation}
 	\mathcal{L}_{outlier}^{w} = \mathcal{L}_{BCE}^{w}(\mathbf{M}_{outlier},\mathbf{\hat{M}}_{outlier}) + \mathcal{L}_{Dice}^{w}(\mathbf{M}_{outlier}, \mathbf{\hat{M}}_{outlier}),
 \end{equation}
where $\mathbf{M}_{outlier}$ and $\mathbf{\hat{M}}_{outlier}$ denote the predicted and ground truth outlier masks. To generate $\mathbf{\hat{M}}_{outlier}$, we randomly inject Gaussian noise into a subset of points (see Section 4.5 for details), and assign the perturbed points that are far from any plane instance with a label of 1 (outliers), while all other points with slight perturbations are assigned a label of 0 (non-outliers).

\section{Experiments} 
\subsection{Datasets preparation} 
To comprehensively evaluate the performance and generalization ability of RoofSeg, we conducted experiments on five benchmarks, including three roof-specific benchmarks and two generic plane segmentation benchmarks. The three roof benchmarks are RoofNTNU~\citep{zhang2022improved}, Roofpc3D~\citep{li2024boundary}, and Building3D~\citep{wang2023building3d, li2024boundary}, which cover small-scale real-world roof point clouds, large-scale synthetic roof models, and complex real-world urban building roofs, respectively. These datasets are used to evaluate the performance of RoofSeg on roof plane segmentation under different data sources, roof types, and scene complexities.
In addition, we used ABC~\citep{koch2019abc} and KSR42~\citep{yu2022finding} to further evaluate the generalization ability of RoofSeg on generic plane segmentation tasks. ABC provides CAD models with accurate planar face annotations, while KSR42 contains challenging real-world models from different scenarios. For KSR42, since point-level plane annotations are not originally provided, we manually annotated representative models for evaluation. The implementation details are give in Appendix~\ref{appendix:implementation_details},  while detailed dataset descriptions and train/test split configurations are provided in Appendix~\ref{appendix:dataset_configuration}.

\subsection{Evaluation metrics} 
For comprehensive evaluation of roof plane segmentation results at both point and instance levels, we utilize four metrics: mean coverage (mCov), mean weighted coverage (mWCov), mean precision (mPrec), and mean recall (mRec). The complete definitions of these metrics are provided in ~\cite{li2024boundary}. Larger values of all four metrics indicate better segmentation performance. To further evaluate the over- and under-segmentation of planes, we introduced two metrics, $O_{seg}$ and $U_{seg}$, which are defined as follows:
\begin{equation}
	O_{seg} = N_{os} / N_{gt}, 
\end{equation}
\begin{equation}
	U_{seg} = N_{us} / N_{pr},
\end{equation}
where $N_{gt}$ and $ N_{pr}$ denotes the plane numbers of ground truth and prediction. $N_{os}$ denotes the number of ground-truth planes that are covered by multiple predicted planes (over-segmentation), while $N_{us}$ refers to the number of predicted planes that cover multiple ground-truth planes (under-segmentation). Smaller values of these two metrics indicate less over- and under-segmentation, respectively. Moreover, to evaluate the edge quality of segmentation, we design an edge accuracy metric $E_{acc}$, which is formulated as:
\begin{equation}
	E_{acc} = \frac{
		\sum_{(A,B) \in \mathcal{E}} \left| Correct(E_A) + Correct(E_B) \right|
	}{
		\sum_{(A,B) \in \mathcal{E}} |E_{AB}|
	}
\end{equation}
where $\mathcal{E}$ denotes the set of adjacent plane patch pairs, $(A,B)$ is one such pair, $|E_{AB}|$ represents the total number of edge points between plane $A$ and $B$, $Correct(E_A)$ indicates the number of edge points belonging to plane A that are correctly predicted, while $Correct(E_B)$ has a similar meaning. Larger values of $E_{acc}$ indicate better edge quality.

\subsection{Comparative experiments}
\textbf{Roof benchmark results.} To demonstrate the superiority of RoofSeg, we conducted comparative experiments on three roof benchmarks with three traditional approaches: Region Growing~\citep{vo2015octree}, RANSAC~\citep{xu2015investigation}, and GoCoPP~\citep{yu2022finding}, as well as four deep learning-based approaches: ParseNet~\citep{sharma2020parsenet}, HPNet~\citep{yan2021hpnet}, Mask3D~\citep{schult2023mask3d}, and DeepRoofPlane~\citep{li2024boundary}. ParseNet, HPNet and Mask3D are developed for primitive or instance segmentation, while DeepRoofPlane is specifically designed for roof plane segmentation. The traditional approaches were implemented using public implementations, and their detailed parameter settings are provided in the footnote.\footnote{\parbox[t]{\linewidth}{\raggedright
		\textbf{Region Growing and RANSAC:} implemented based on CGAL, available at \url{https://github.com/CGAL/cgal/}. \\
		\textbf{Region Growing:} epsilon=0.05, norm\_threshold=0.90, min\_region\_size=50. \\
		\textbf{RANSAC:} probability=0.99, epsilon=0.05, cluster\_epsilon=0.10, min\_points=50. \\
		\textbf{GoCoPP:} implemented based on the authors' public implementation. epsilon=0.8\%, sigma=50, nn=15, normal\_deviation=0.90, s=1.0, c=1.0, norm=hybrid, max\_steps=10, constraint=True.}}
The source codes of the four deep learning-based approaches can be obtained from the links provided in the corresponding papers. The quantitative assessment of all comparative approaches and RoofSeg on three benchmarks are presented in Table~\ref{Tab:comparative_appraoches} and Table~\ref{Tab:comparative_appraoches_add_metrics}.

\begin{table} [!htb] \renewcommand{\tabcolsep}{1.95 pt}
	\scriptsize
	\renewcommand{\arraystretch}{1.25}
	\newcommand{\tabincell}[2]{\begin{tabular}{@{}#1@{}}#2\end{tabular}}
	\begin{center}
		\caption{Quantitative assessment (mCov, mWCov, mPrec and mRec) of all comparative approaches and RoofSeg on three roof benchmarks. The values of all metrics are presented as percentages.}
		\label{Tab:comparative_appraoches}
		\begin{tabular}{c|cccc|cccc|cccc} 
			\hline
			\multirow{2}{*}{Different approaches} & \multicolumn{4}{c}{RoofNTNU benchmark}  &  \multicolumn{4}{c}{Roofpc3D benchmark} & \multicolumn{4}{c}{Building3D benchmark}\\
			\cline{2-5} \cline{6-9} \cline{10-13} 
			 & mCov       & mWCov  &   mPrec   &    mRec      
			& mCov       & mWCov  &   mPrec   &    mRec   
			& mCov       & mWCov  &   mPrec   &    mRec    \\ 
			\hline
			Region growing  & 85.97  & 86.49 & 90.52 & 87.37 & 86.32 & 87.11 & 91.32 & 88.33 & 84.41 & 85.03 & 88.49 & 83.65 \\ 
			
			RANSAC  & 86.15  & 87.39 & 91.09 & 86.53 & 87.41 & 88.29 & 92.57 & 89.69 & 85.42 & 85.61 & 88.07 & 86.32  \\ 
			
			GoCoPP  & 90.37  & 91.36 & 95.79 & 93.94 & 91.45 & 93.43 & 96.08 & 92.92 & 87.77 & 89.39 & 93.72 & 91.84  \\ 
			
			ParseNet  & 87.15  & 88.37 & 90.35 & 87.66 & 88.43 & 90.72 & 92.15 & 89.41 & 86.47 & 87.65 & 90.17 & 88.41  \\ 
			
			HPNet  & 87.29  & 88.95 & 92.19 & 90.01 & 89.40 & 90.11 & 93.65 & 91.13 & 86.81 & 87.75 & 90.23 & 88.62  \\ 
			
			Mask3D  & 88.03  & 89.51 & 93.00 & 90.18 & 90.47 & 91.47 & 93.25 & 89.83 & 87.17 & 88.62 & 91.92 & 88.43  \\ 
			
			DeepRoofPlane  & 91.13  & 93.74 & 97.69 & 94.38 & 92.34 & 94.75 & 97.27 & 94.01 & 88.34 & 91.49 & 96.60 & 93.14  \\ 
			
			RoofSeg (ours)  & \textbf{95.89}  & \textbf{96.82} & \textbf{99.60} & \textbf{98.18} & \textbf{96.01} & \textbf{97.56} & \textbf{99.98} & \textbf{98.54} & \textbf{93.74} & \textbf{96.37} & \textbf{98.79} & \textbf{96.58}     \\ 
			\hline
		\end{tabular}
	\end{center}
\end{table}

\begin{table} [!htb] \renewcommand{\tabcolsep}{6.5 pt}
	\scriptsize
	\renewcommand{\arraystretch}{1.25}
	\newcommand{\tabincell}[2]{\begin{tabular}{@{}#1@{}}#2\end{tabular}}
	\begin{center}
		\caption{Quantitative assessment ($O_{seg}$, $U_{seg}$ and $E_{acc}$) of all comparative approaches and RoofSeg on three roof benchmarks. The values of all metrics are presented as percentages.}
		\label{Tab:comparative_appraoches_add_metrics}
		\begin{tabular}{c|ccc|ccc|ccc} 
			\hline
			\multirow{2}{*}{Different approaches} & \multicolumn{3}{c}{RoofNTNU benchmark}  &  \multicolumn{3}{c}{Roofpc3D benchmark} & \multicolumn{3}{c}{Building3D benchmark}\\
			\cline{2-4} \cline{5-7} \cline{8-10} 
			& $O_{seg}\downarrow$  & $U_{seg}\downarrow$   &   $E_{acc}\uparrow$      
			& $O_{seg}\downarrow$ & $U_{seg}\downarrow$   &   $E_{acc}\uparrow$  
			& $O_{seg}\downarrow$  & $U_{seg}\downarrow$   &   $E_{acc}\uparrow$      \\ 
			\hline
			Region growing  & 7.16  & 6.55 & 84.86 & 5.47 & 7.15 & 85.58 & 9.83 & 10.10 & 82.84  \\ 
			
			RANSAC  & 5.51  & 6.34 & 85.57 & 7.75 & 6.19 & 86.07 & 8.76 & 9.88 & 83.35  \\ 
			
			GoCoPP  & 4.36  & 3.47 & 89.60 & 3.19 & 4.88 & 90.59 & 5.40 & 6.75 & 86.63   \\ 
			
			ParseNet  & 5.64  & 5.21 & 86.97 & 4.28 & 7.02 & 87.28 & 6.19 & 7.40 & 84.40  \\ 
			
			HPNet  & 4.97  & 6.88 & 86.51 & 3.59 & 6.28 & 88.03 & 4.43 & 6.92 & 85.92  \\ 
			
			Mask3D  & 4.03  & 5.96 & 87.15 & 4.36 & 5.90 & 89.47 & 6.25 & 8.73 & 86.55   \\ 
			
			DeepRoofPlane  & 3.83  & 4.79 & 90.65 & 5.77 & 6.26 & 91.75 & 5.79 & 7.51  & 87.73    \\ 
			
			RoofSeg (ours)  & \textbf{1.21}  & \textbf{2.39} & \textbf{94.90} & \textbf{2.45} & \textbf{2.69} & \textbf{95.36} & \textbf{3.57} & \textbf{4.38} & \textbf{92.53}  \\ 
			\hline
		\end{tabular}
	\end{center}
\end{table}

From the results in Table~\ref{Tab:comparative_appraoches}, it can be found that our network (RoofSeg) achieves the highest scores across all four evaluation metrics. Compared to DeepRoofPlane~\citep{li2024boundary}, which is the most competitive among all the comparative approaches, RoofSeg leads by 3 to 4 percentage points in Cov, WCov, and mRec, and also shows an improvement of approximately 1.5 percentage points in mPrec. In comparison with the other five approaches, RoofSeg outperforms them significantly in all metrics on three benchmarks. High scores in Cov and WCov indicate that RoofSeg achieves high accuracy in point-level segmentation, while the high scores in mPrec and mRec demonstrate the superior performance of RoofSeg in plane instance-level segmentation. As reported in Table~\ref{Tab:comparative_appraoches_add_metrics}, RoofSeg achieves the lowest scores on both the $O_{seg}$ and $U_{seg}$ metrics, demonstrating its effectiveness in mitigating over- and under-segmentation issues. Moreover, its strong performance on the $E_{acc}$ metric further validates the accuracy in edge region segmentation.
\begin{figure}[h!]
	\scriptsize
	\centering
	\begin{tabular}{ccccc} 	
		\multicolumn{1}{c}{\includegraphics[width=0.17\linewidth]{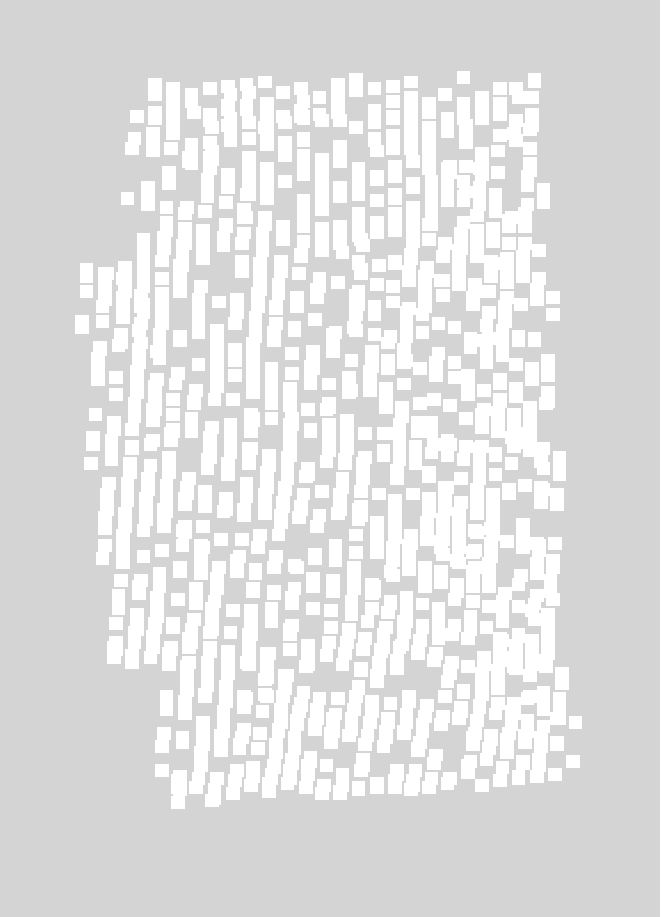}} &
		\multicolumn{1}{c}{\includegraphics[width=0.17\linewidth]{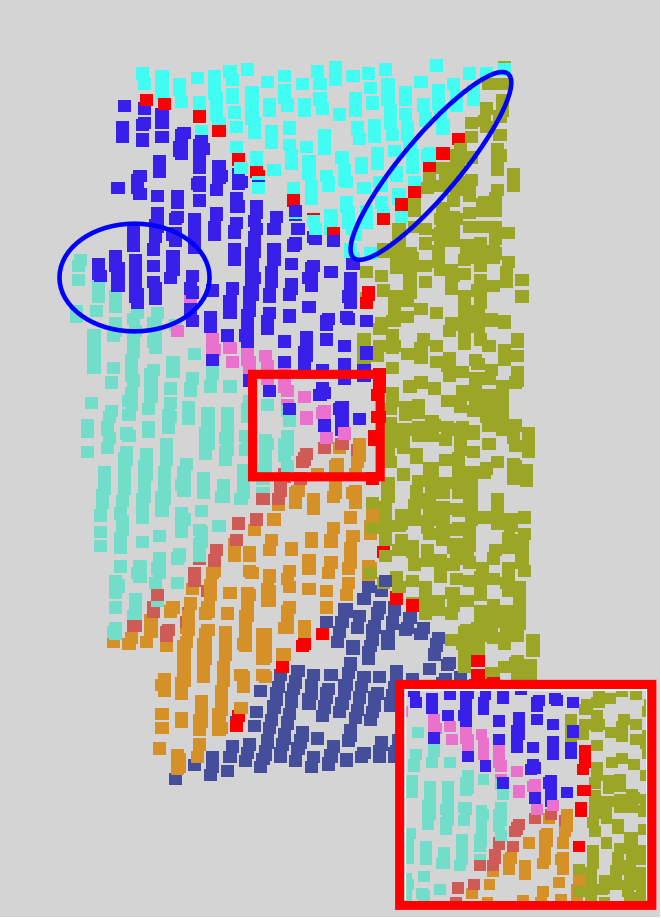}} &
		\multicolumn{1}{c}{\includegraphics[width=0.17\linewidth]{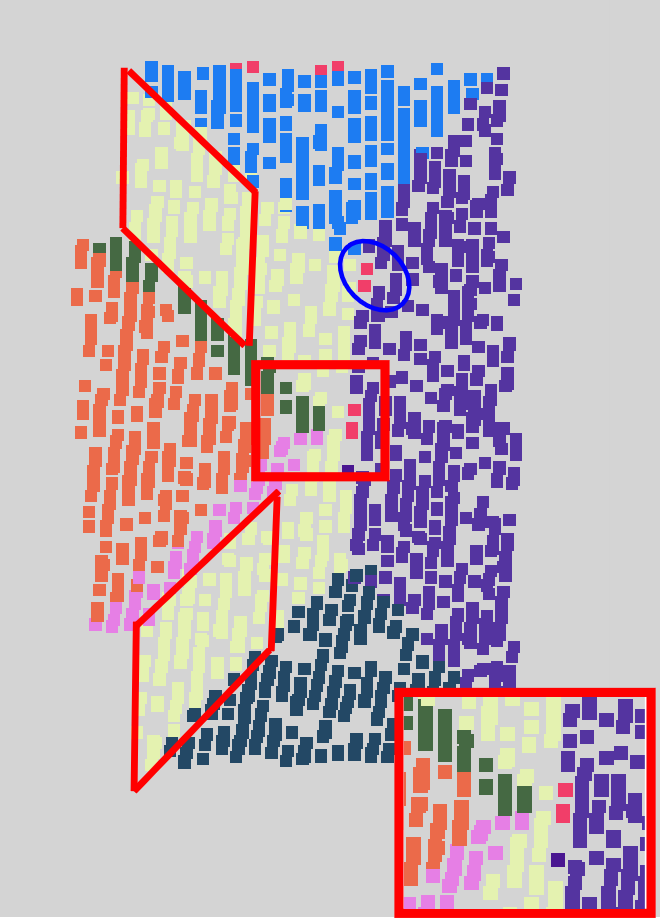}} &
		\multicolumn{1}{c}{\includegraphics[width=0.17\linewidth]{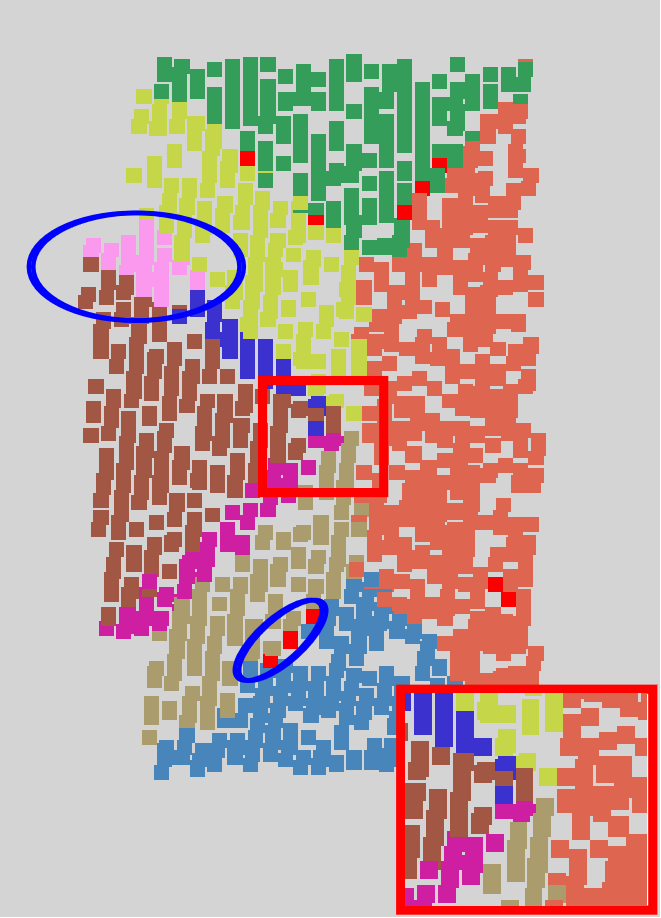}} &
		\multicolumn{1}{c}{\includegraphics[width=0.17\linewidth]{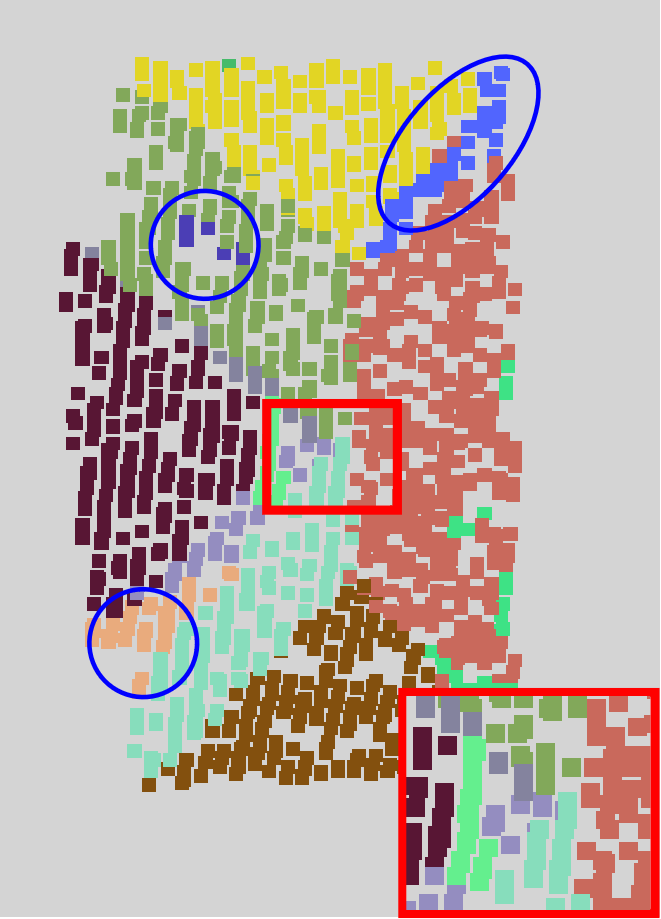}}
		\\
		($\text{A}_{1}$) Input &
		($\text{B}_{1}$) Region growing &
		($\text{C}_{1}$) RANSAC &
		($\text{D}_{1}$) GoCoPP &
		($\text{E}_{1}$) ParseNet
		\\
		
		\multicolumn{1}{c}{\includegraphics[width=0.17\linewidth]{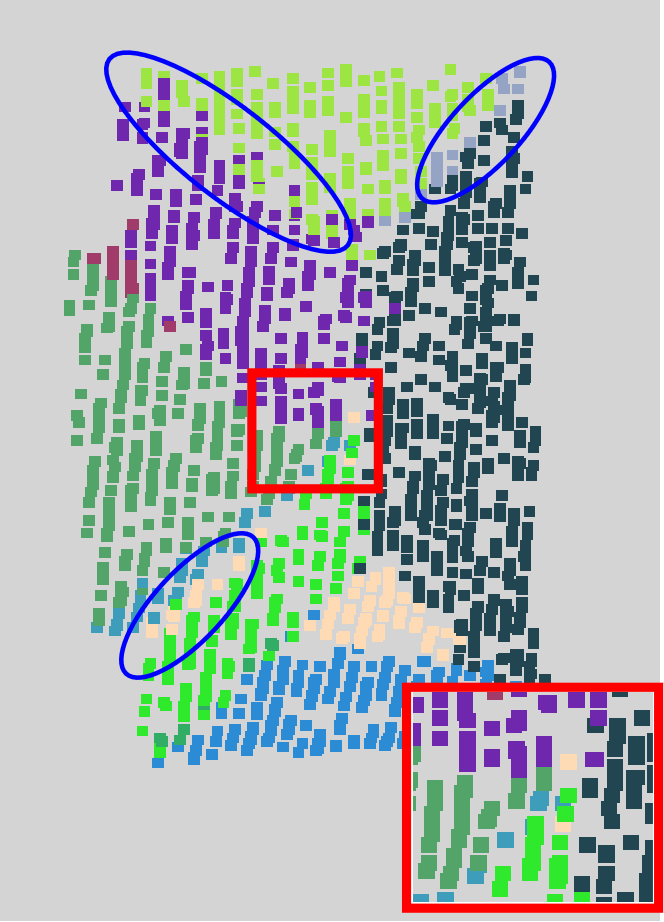}} &
		\multicolumn{1}{c}{\includegraphics[width=0.17\linewidth]{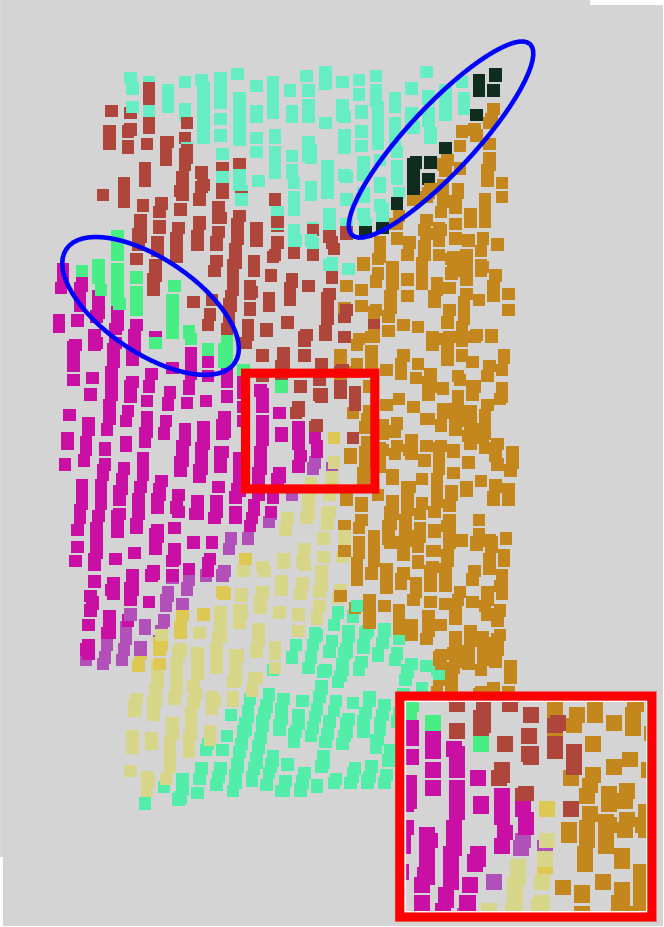}} &
		\multicolumn{1}{c}{\includegraphics[width=0.17\linewidth]{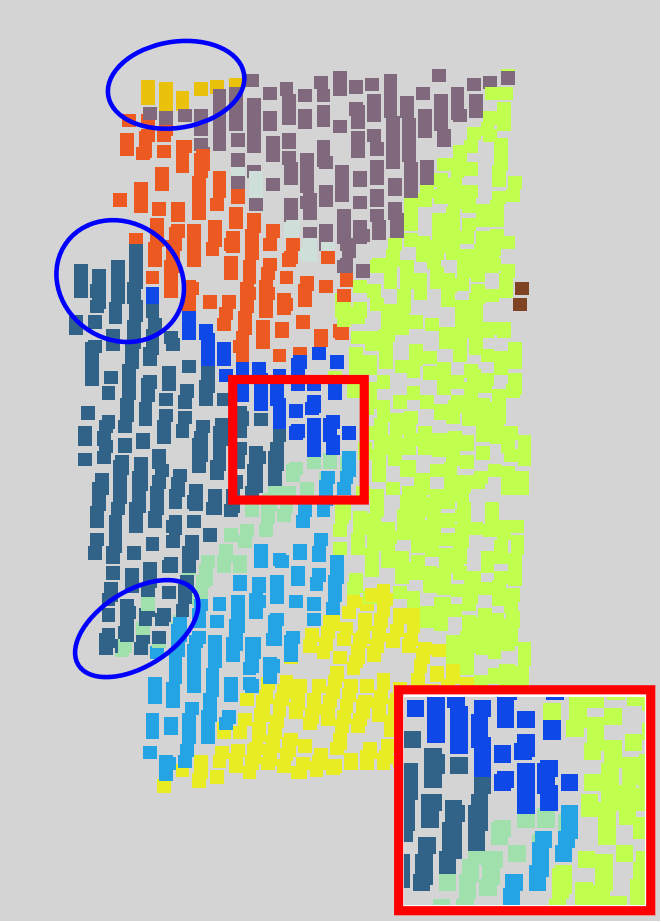}} &
		\multicolumn{1}{c}{\includegraphics[width=0.17\linewidth]{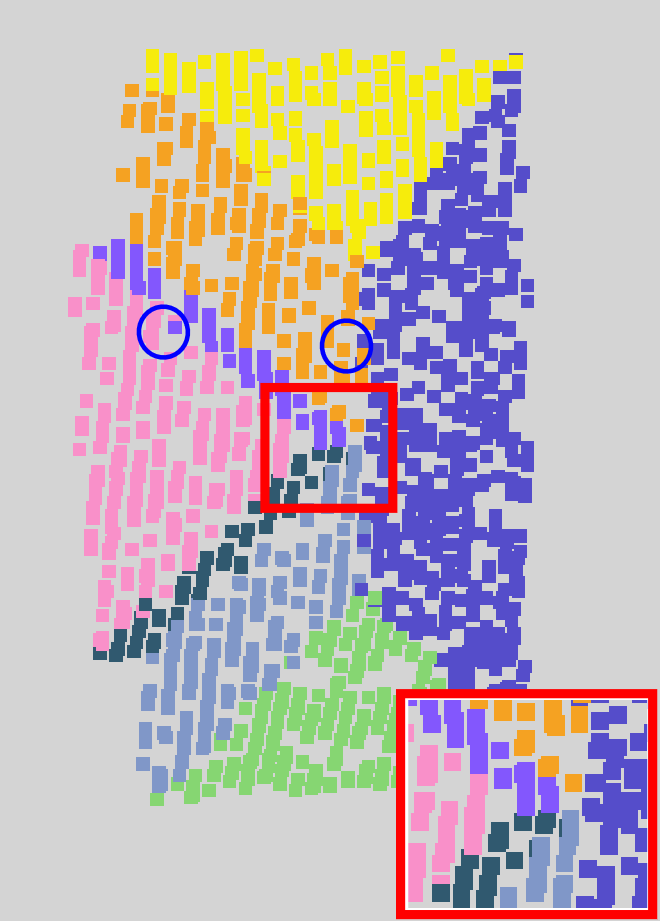}} &
		\multicolumn{1}{c}{\includegraphics[width=0.17\linewidth]{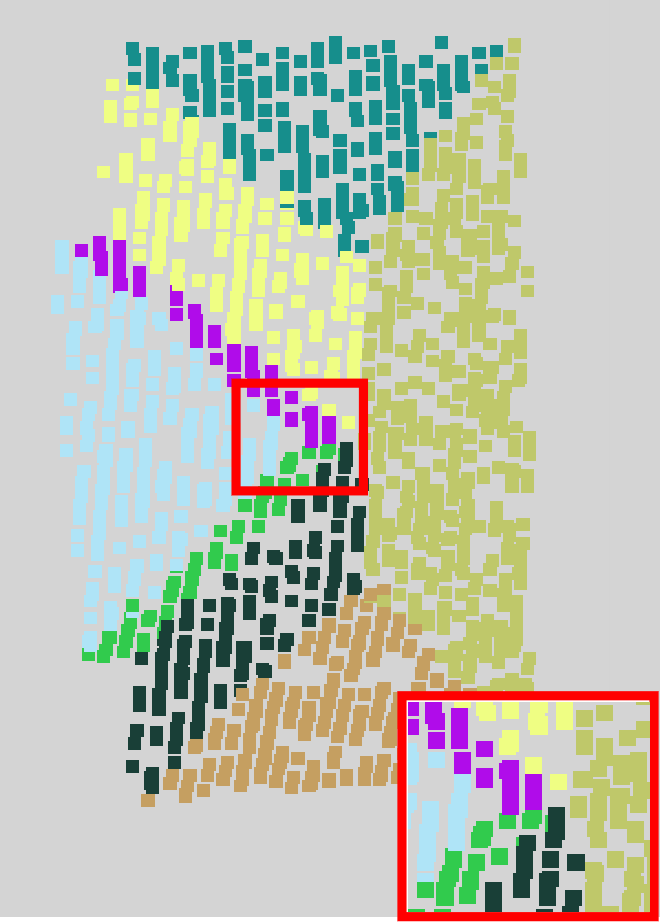}}
		\\
		($\text{F}_{1}$) HPNet &
		($\text{G}_{1}$) Mask3D &
		($\text{H}_{1}$) DeepRoofPlane &
		($\text{I}_{1}$) RoofSeg (ours) &
		($\text{J}_{1}$) GT
		\\
		
		& & & & \\
		\hdashline 
		& & & &\\
		
		\multicolumn{1}{c}{\includegraphics[width=0.17\linewidth]{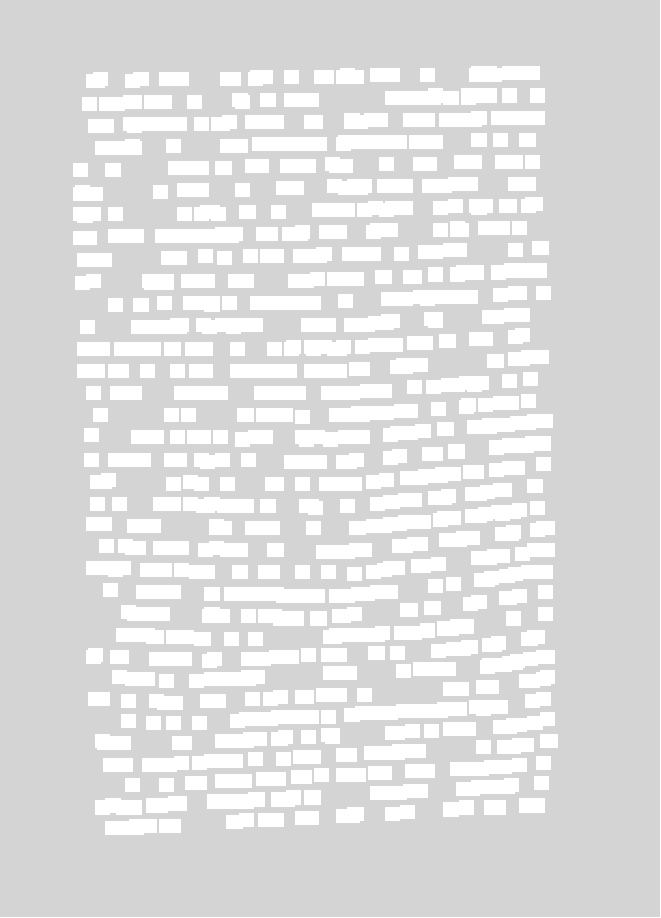}} &
		\multicolumn{1}{c}{\includegraphics[width=0.17\linewidth]{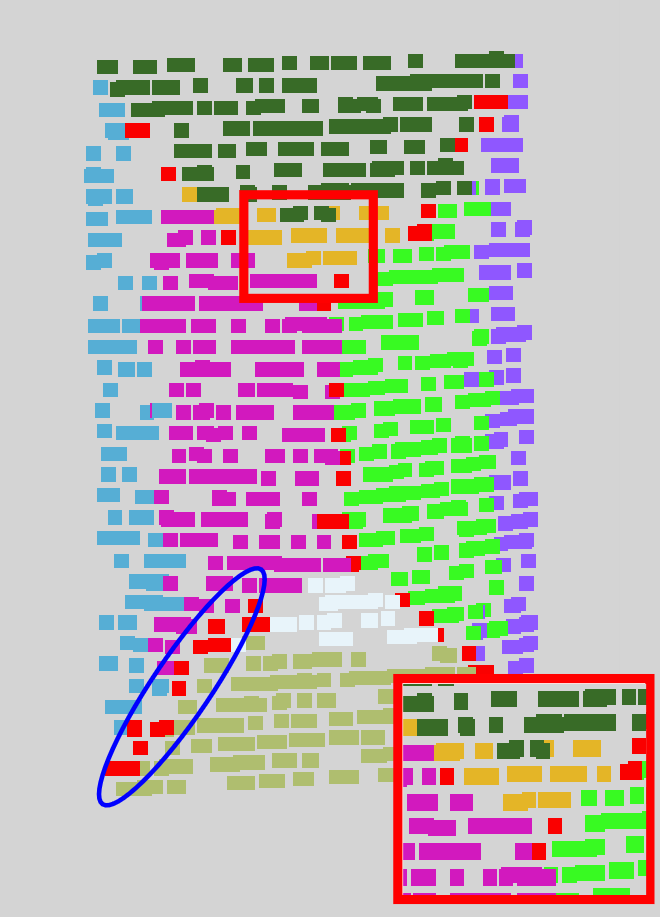}} &
		\multicolumn{1}{c}{\includegraphics[width=0.17\linewidth]{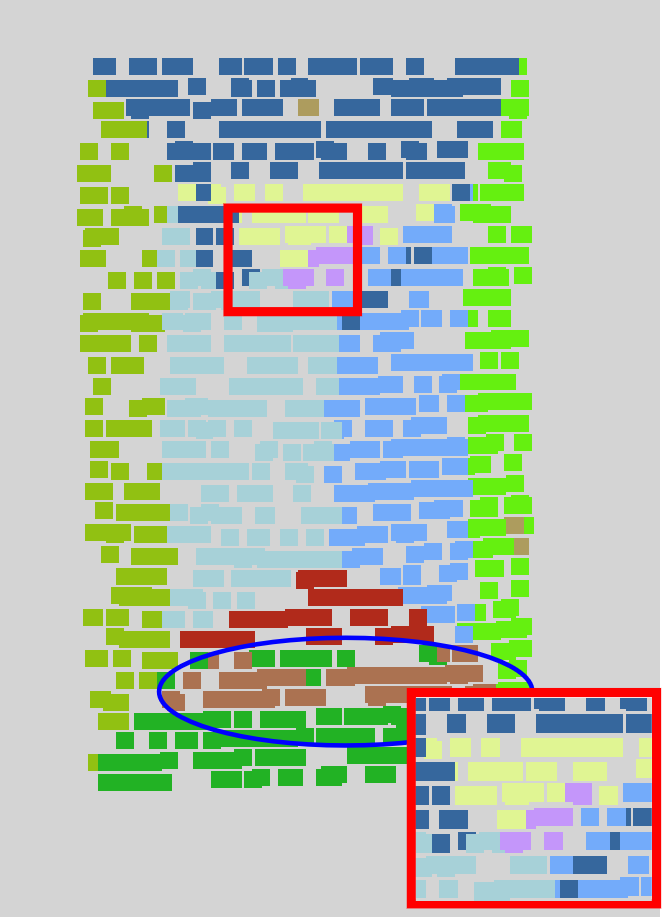}} &
		\multicolumn{1}{c}{\includegraphics[width=0.17\linewidth]{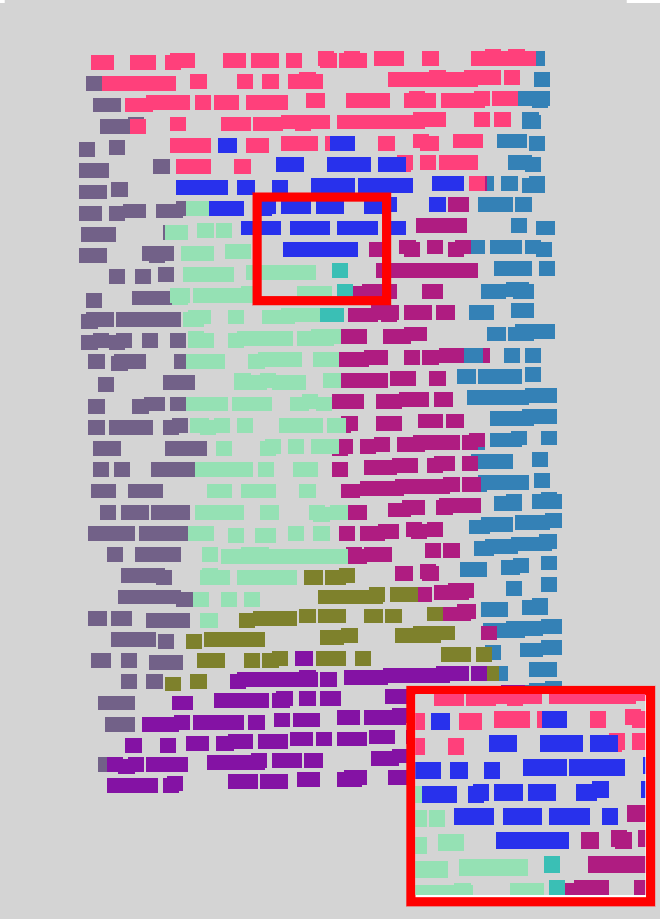}} &
		\multicolumn{1}{c}{\includegraphics[width=0.17\linewidth]{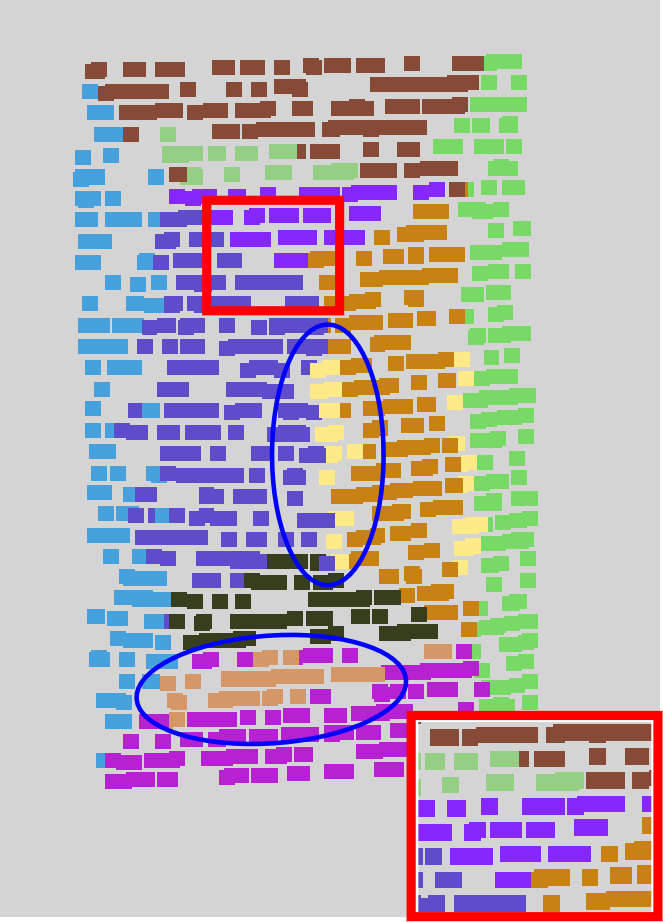}}
		\\
		($\text{A}_{2}$) Input &
		($\text{B}_{2}$) Region growing &
		($\text{C}_{2}$) RANSAC &
		($\text{D}_{2}$) GoCoPP &
		($\text{E}_{2}$) ParseNet
		\\
		
		\multicolumn{1}{c}{\includegraphics[width=0.17\linewidth]{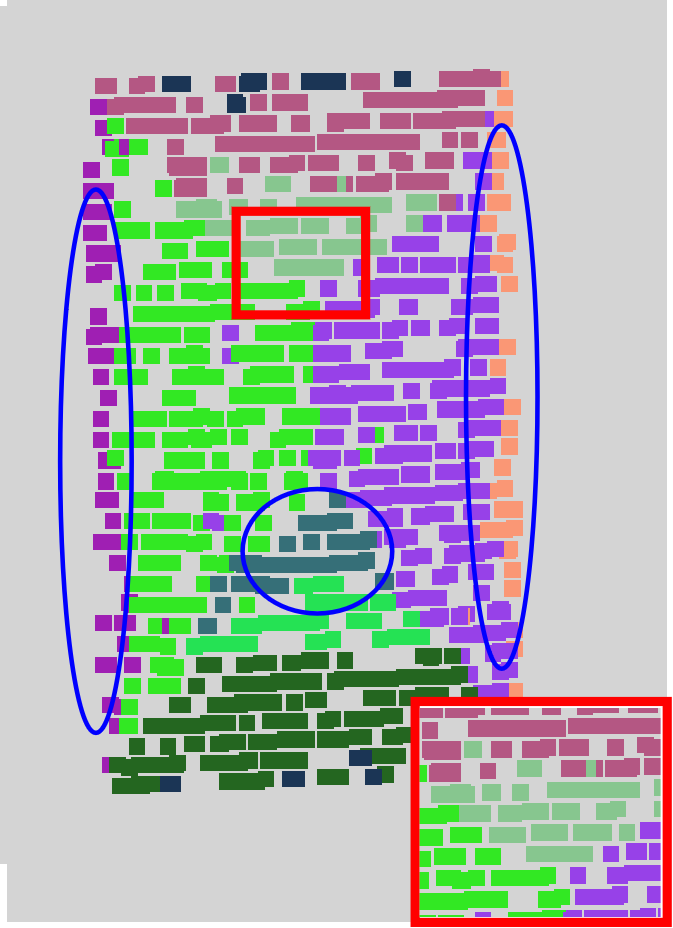}} &
		\multicolumn{1}{c}{\includegraphics[width=0.17\linewidth]{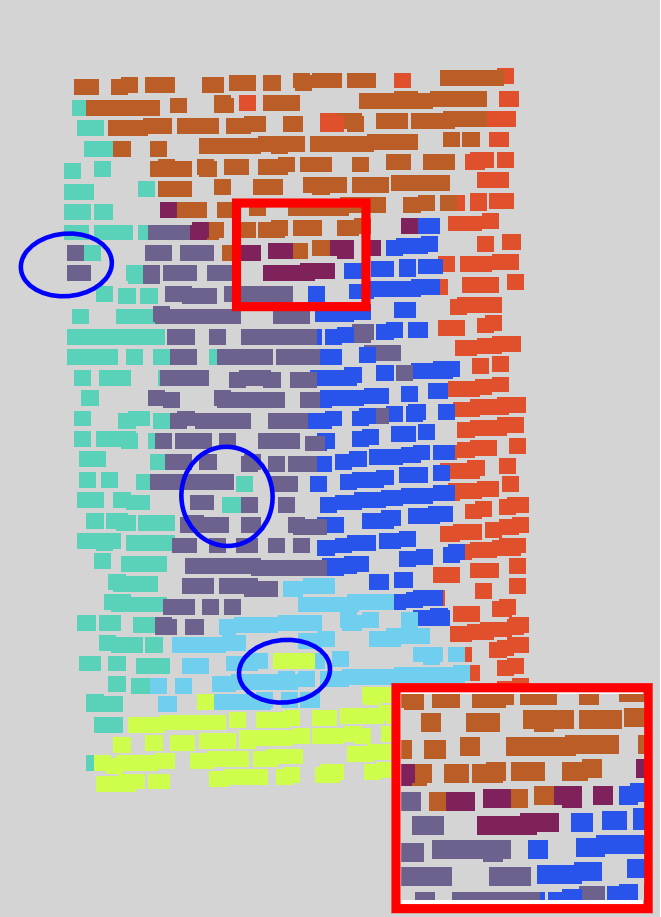}} &
		\multicolumn{1}{c}{\includegraphics[width=0.17\linewidth]{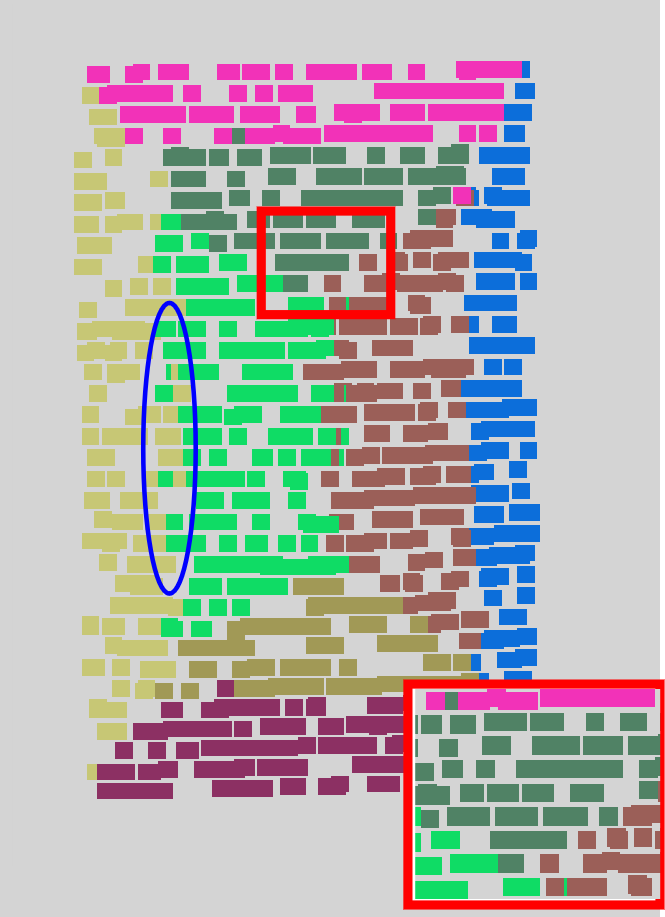}} &
		\multicolumn{1}{c}{\includegraphics[width=0.17\linewidth]{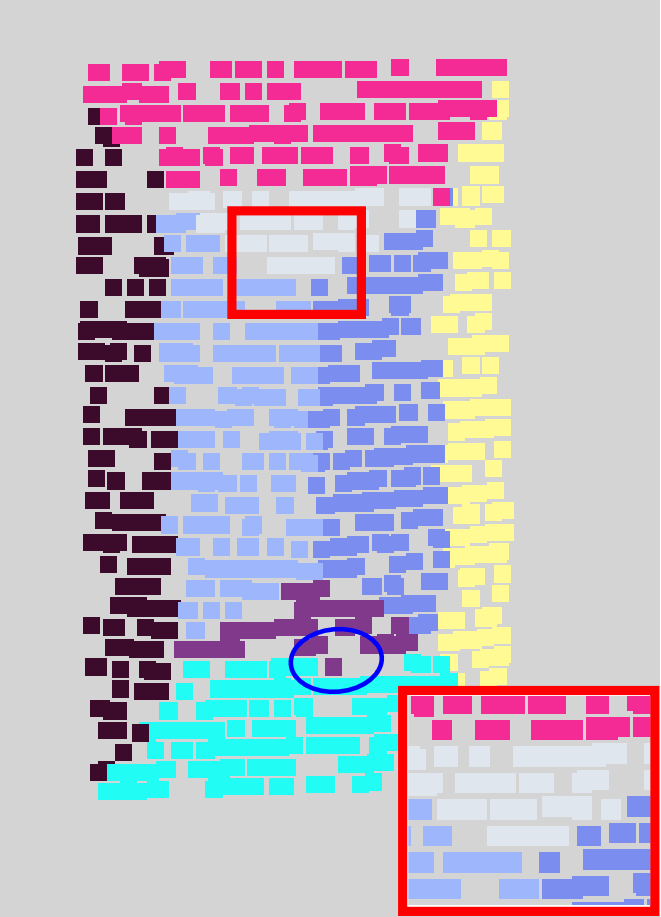}} &
		\multicolumn{1}{c}{\includegraphics[width=0.17\linewidth]{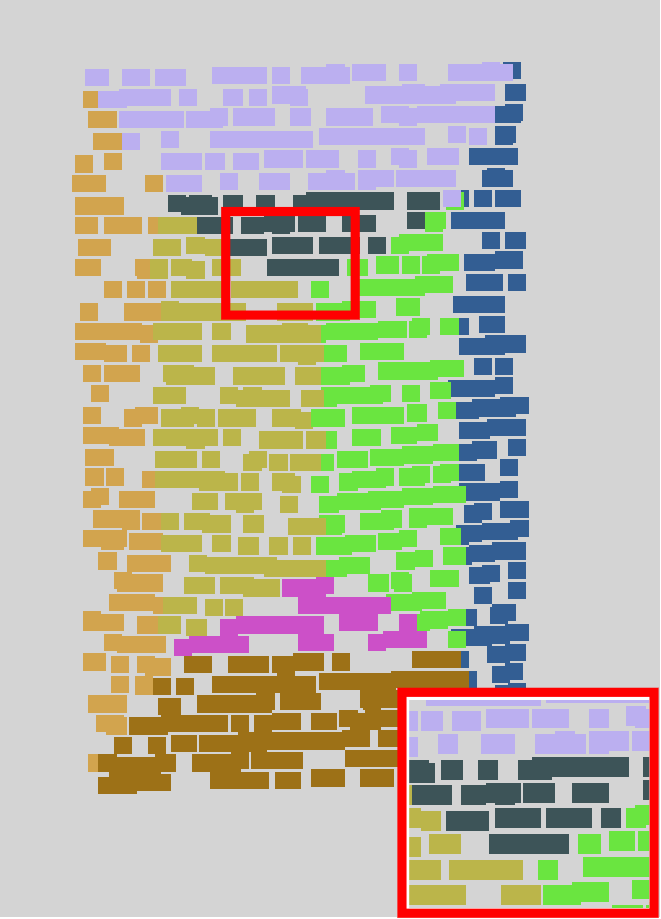}}
		\\
		($\text{F}_{2}$) HPNet &
		($\text{G}_{2}$) Mask3D &
		($\text{H}_{2}$) DeepRoofPlane &
		($\text{I}_{2}$) RoofSeg (ours) &
		($\text{J}_{2}$) GT
	\end{tabular} 
	\caption{Visual results of roof plane segmentation for all comparative approaches and RoofSeg on the Roofpc3D benchmark. The blue ellipses point out some incorrectly segmented regions. The red wireframes point out some regions with under-segmentation.}
	\vspace{0.0em}
	\label{Fig:comparative_Roofpc3D}
\end{figure}

\begin{figure}[h!]
	\scriptsize
	\centering
	\begin{tabular}{ccccc} 	
		
		\multicolumn{1}{c}{\includegraphics[width=0.17\linewidth]{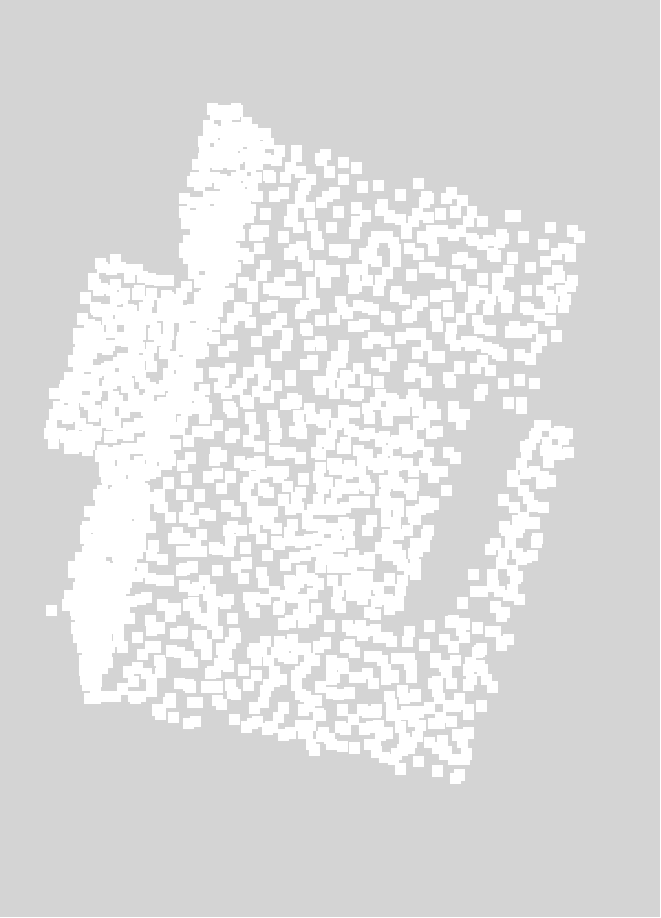}} &
		\multicolumn{1}{c}{\includegraphics[width=0.17\linewidth]{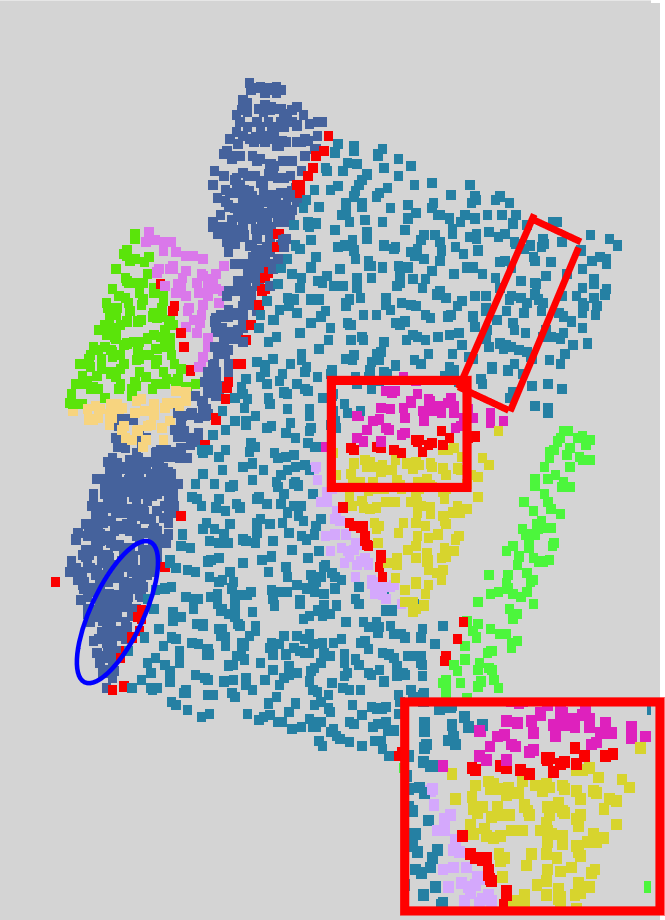}} &
		\multicolumn{1}{c}{\includegraphics[width=0.17\linewidth]{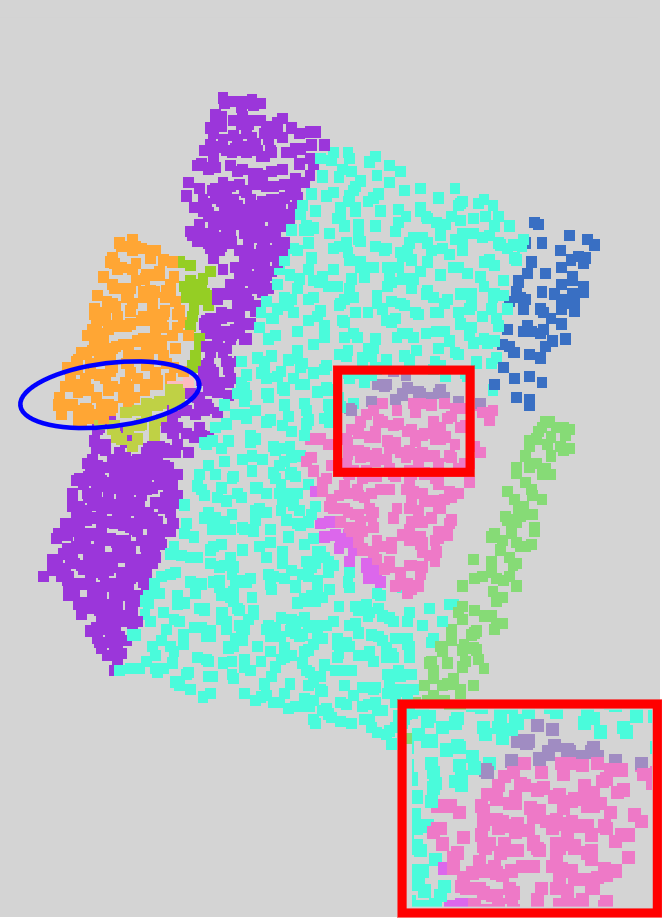}} &
		\multicolumn{1}{c}{\includegraphics[width=0.17\linewidth]{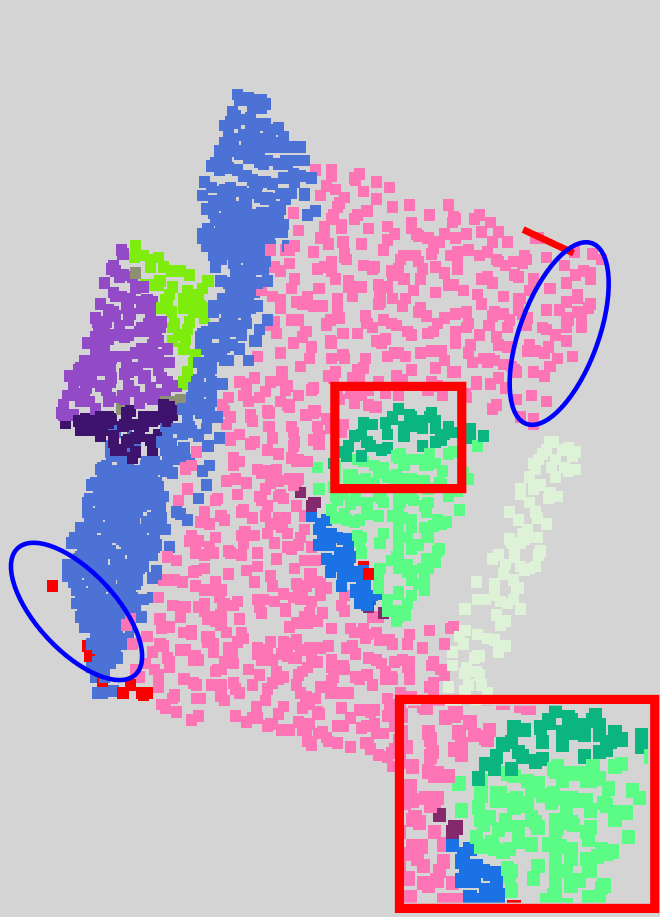}} &
		\multicolumn{1}{c}{\includegraphics[width=0.17\linewidth]{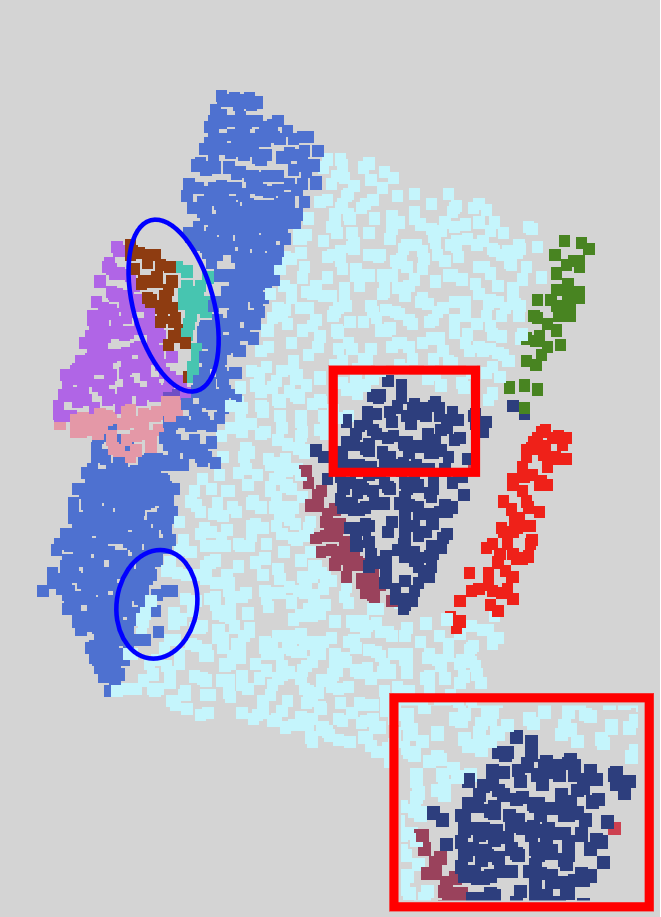}}
		\\
		($\text{A}_{1}$) Input & ($\text{B}_{1}$) Region growing & ($\text{C}_{1}$) RANSAC & ($\text{D}_{1}$) GoCoPP & ($\text{E}_{1}$) ParseNet
		\\
		
		\multicolumn{1}{c}{\includegraphics[width=0.17\linewidth]{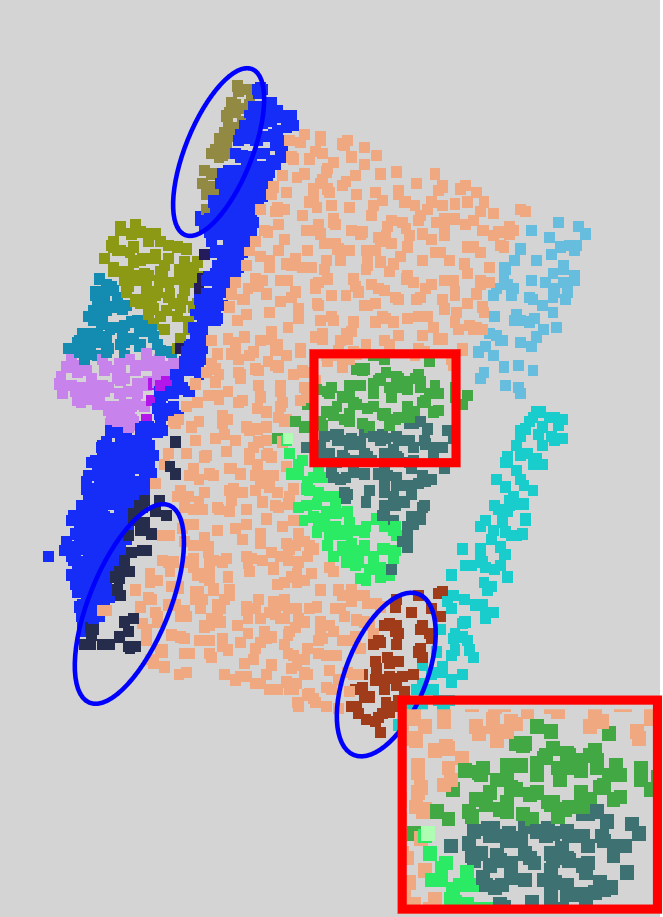}} &
		\multicolumn{1}{c}{\includegraphics[width=0.17\linewidth]{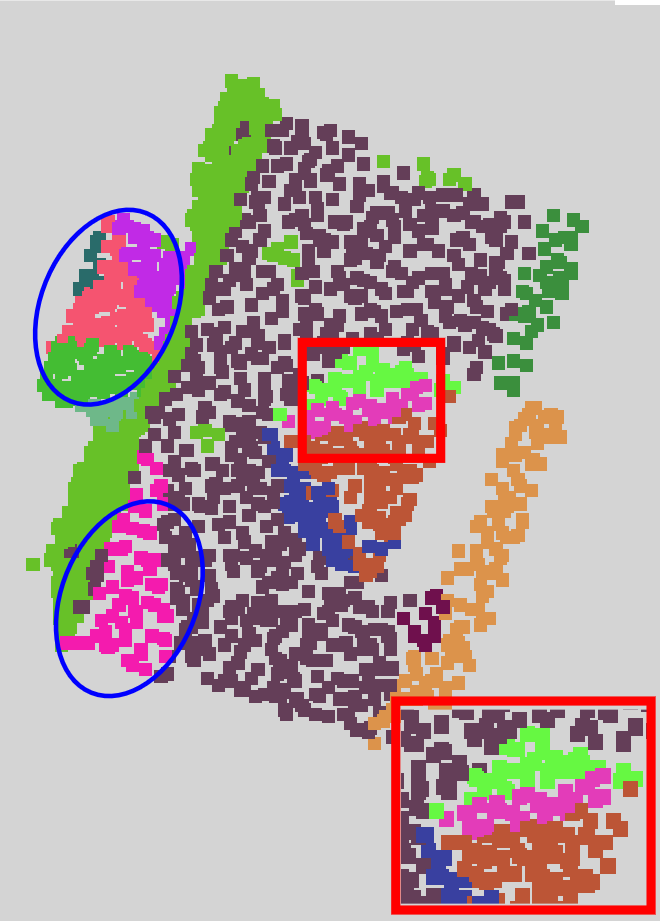}} &
		\multicolumn{1}{c}{\includegraphics[width=0.17\linewidth]{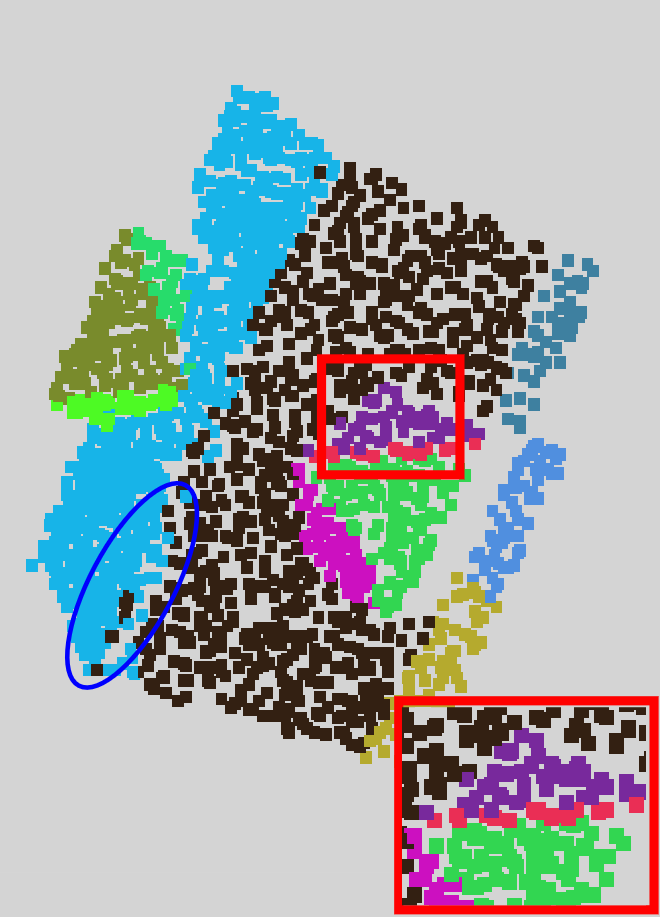}} &
		\multicolumn{1}{c}{\includegraphics[width=0.17\linewidth]{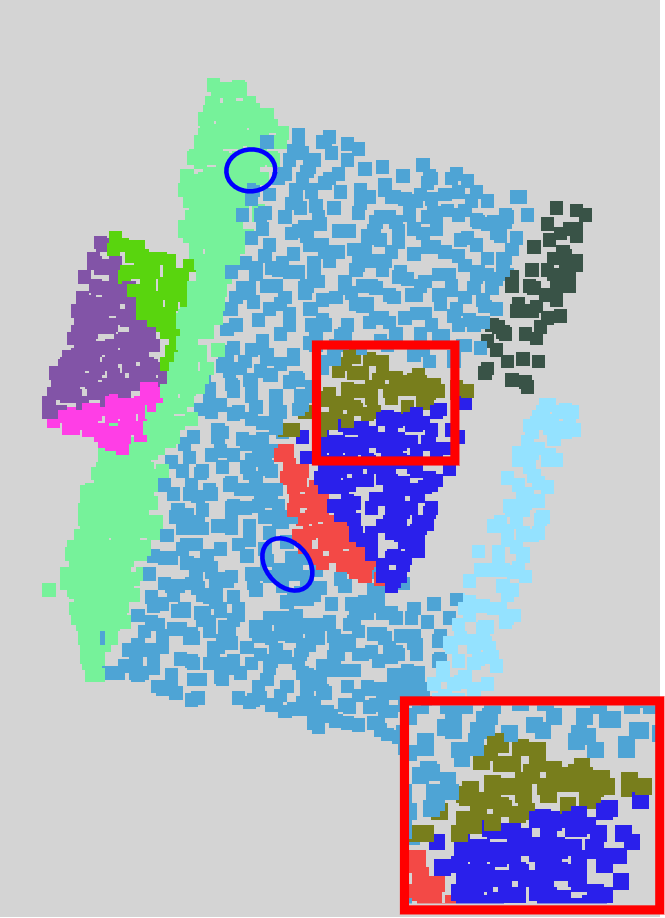}} &
		\multicolumn{1}{c}{\includegraphics[width=0.17\linewidth]{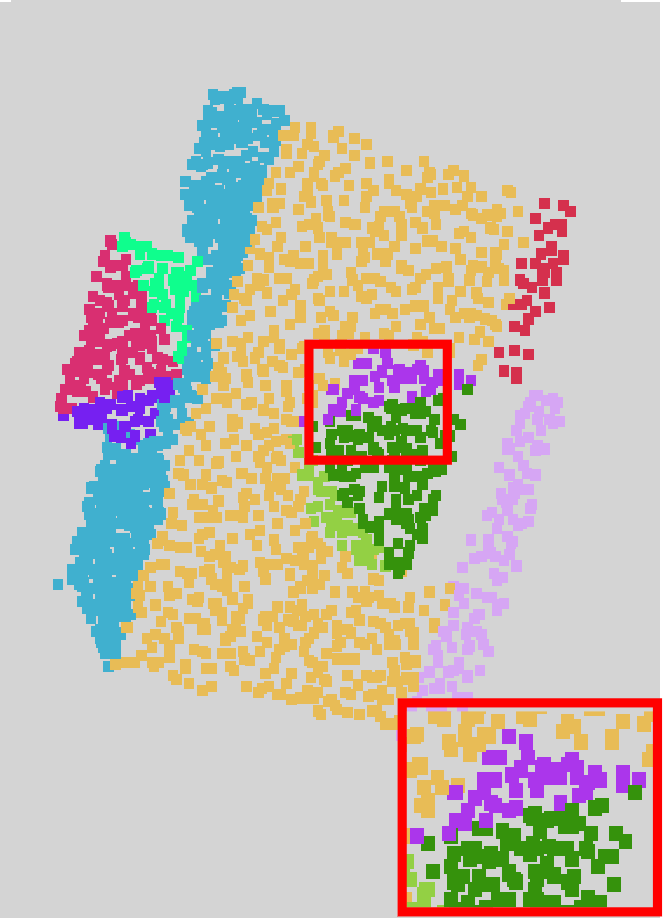}}
		\\
		($\text{F}_{1}$) HPNet & ($\text{G}_{1}$) Mask3D & ($\text{H}_{1}$) DeepRoofPlane & ($\text{I}_{1}$) RoofSeg (ours) & ($\text{J}_{1}$) GT
		\\
		
		& & & &\\
		\hdashline 
		& & & &\\
		
		\multicolumn{1}{c}{\includegraphics[width=0.17\linewidth]{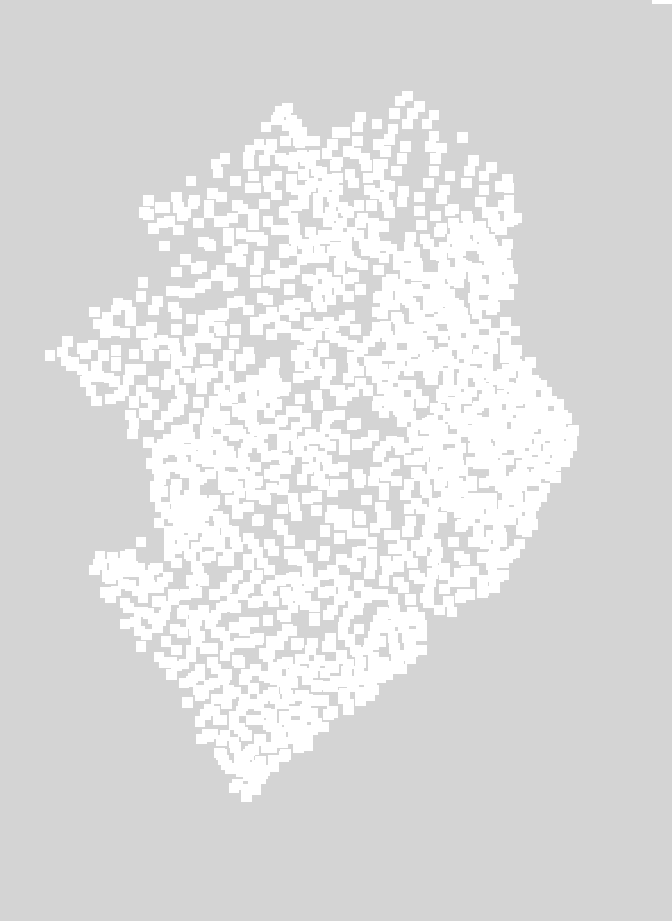}} &
		\multicolumn{1}{c}{\includegraphics[width=0.17\linewidth]{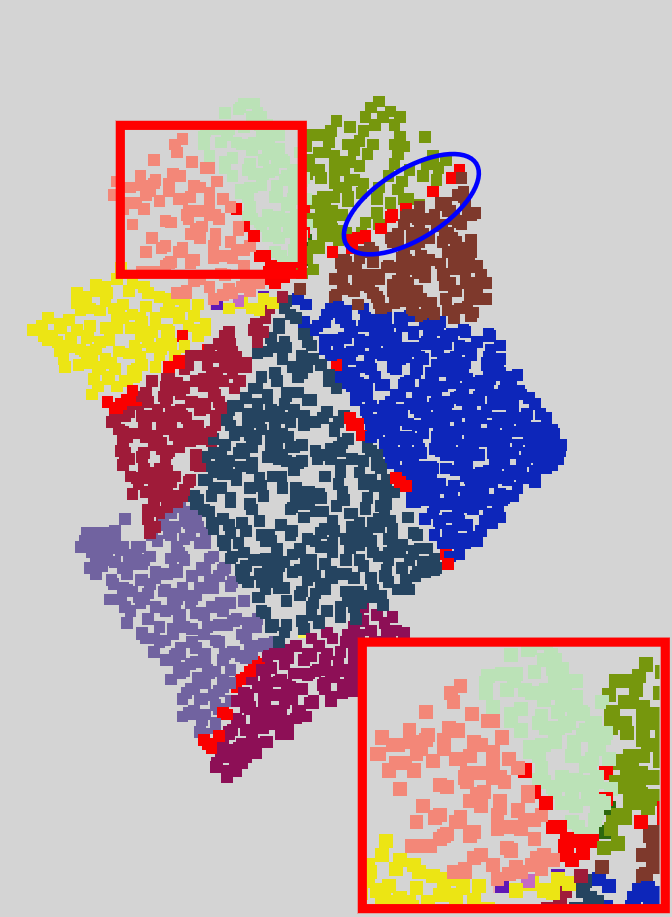}} &
		\multicolumn{1}{c}{\includegraphics[width=0.17\linewidth]{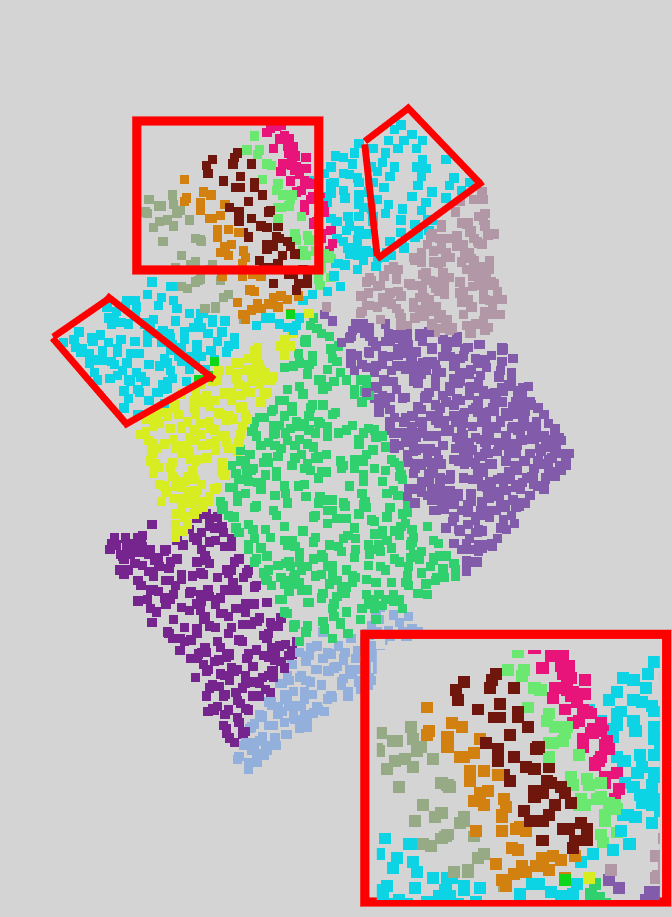}} &
		\multicolumn{1}{c}{\includegraphics[width=0.17\linewidth]{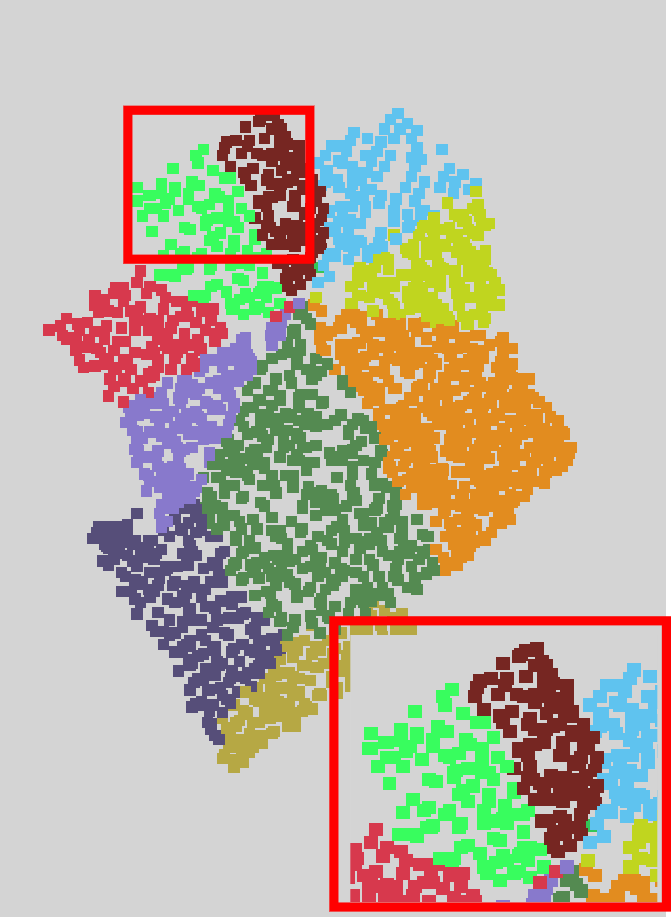}} &
		\multicolumn{1}{c}{\includegraphics[width=0.17\linewidth]{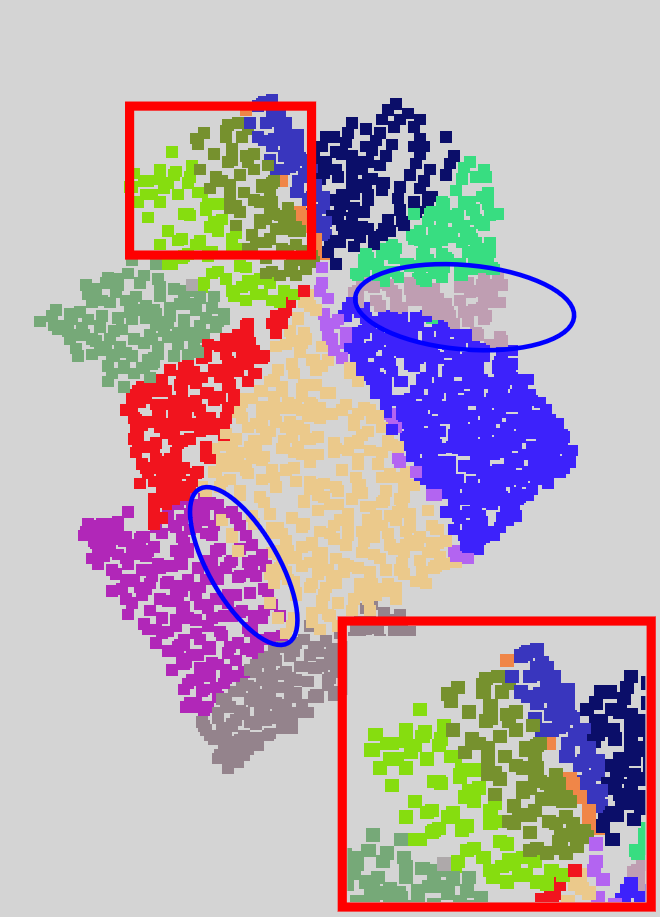}}
		\\
		($\text{A}_{2}$) Input & ($\text{B}_{2}$) Region growing & ($\text{C}_{2}$) RANSAC & ($\text{D}_{2}$) GoCoPP & ($\text{E}_{2}$) ParseNet
		\\
		
		\multicolumn{1}{c}{\includegraphics[width=0.17\linewidth]{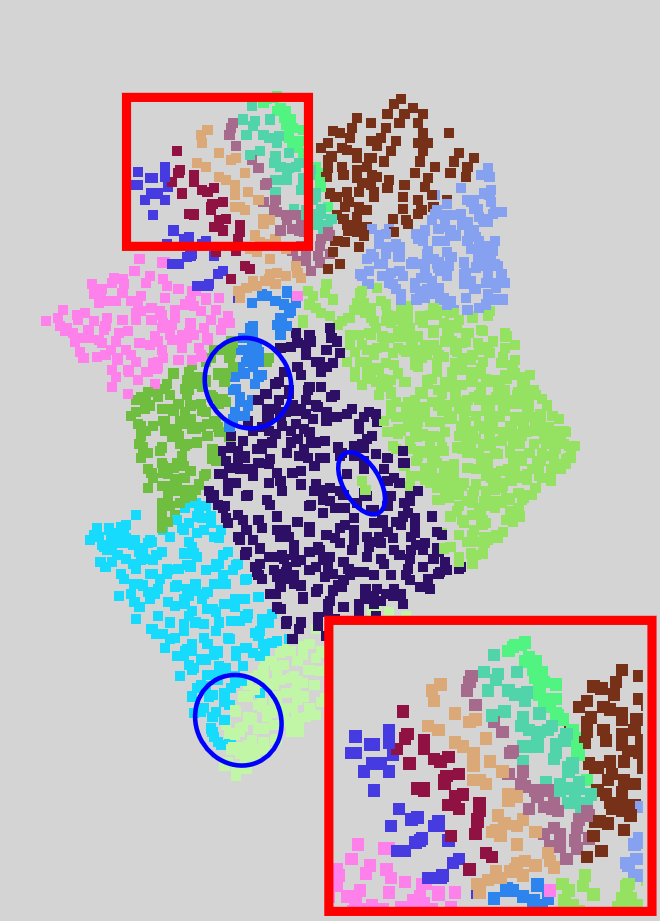}} &
		\multicolumn{1}{c}{\includegraphics[width=0.17\linewidth]{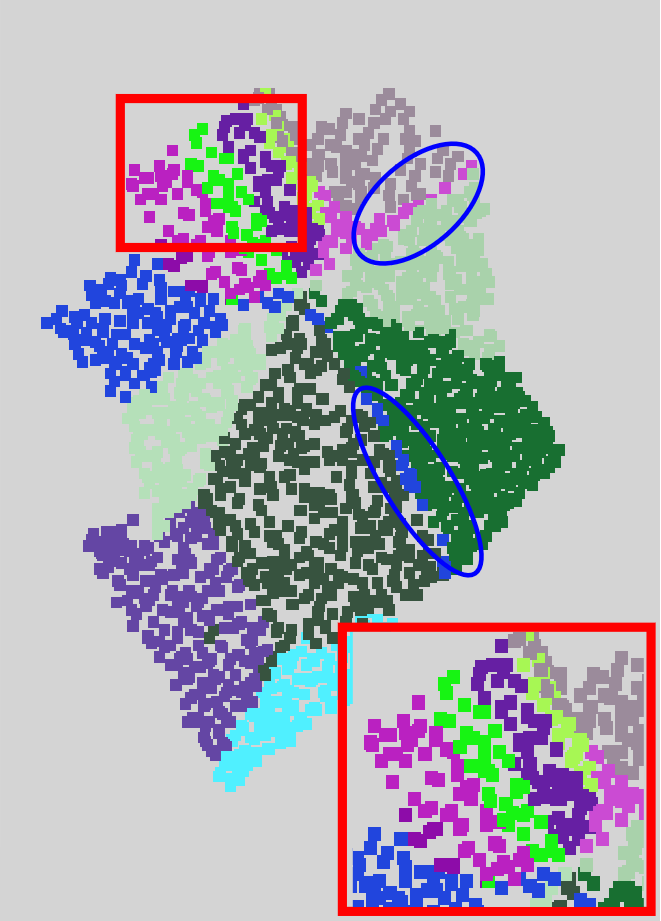}} &
		\multicolumn{1}{c}{\includegraphics[width=0.17\linewidth]{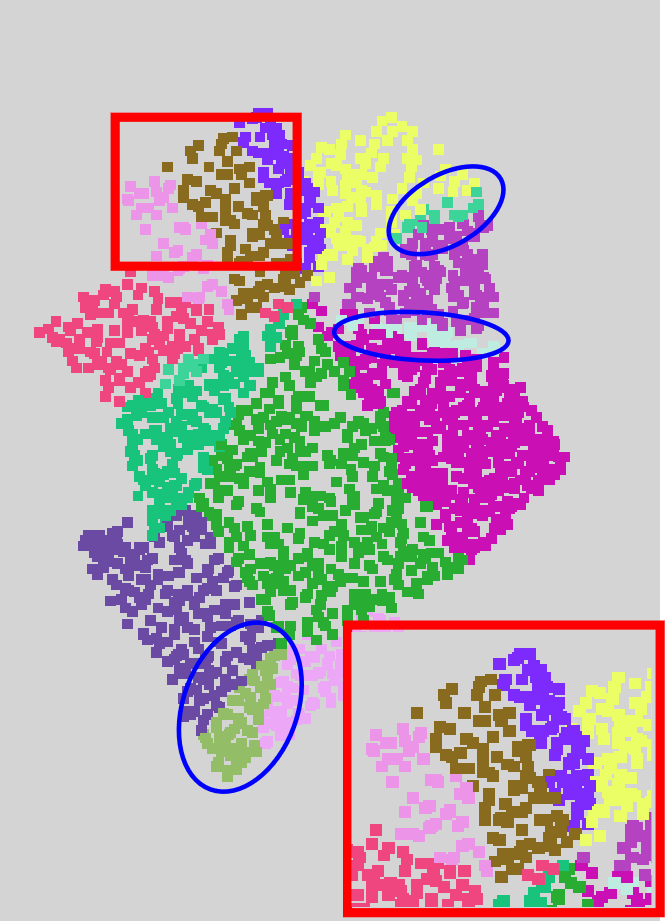}} &
		\multicolumn{1}{c}{\includegraphics[width=0.17\linewidth]{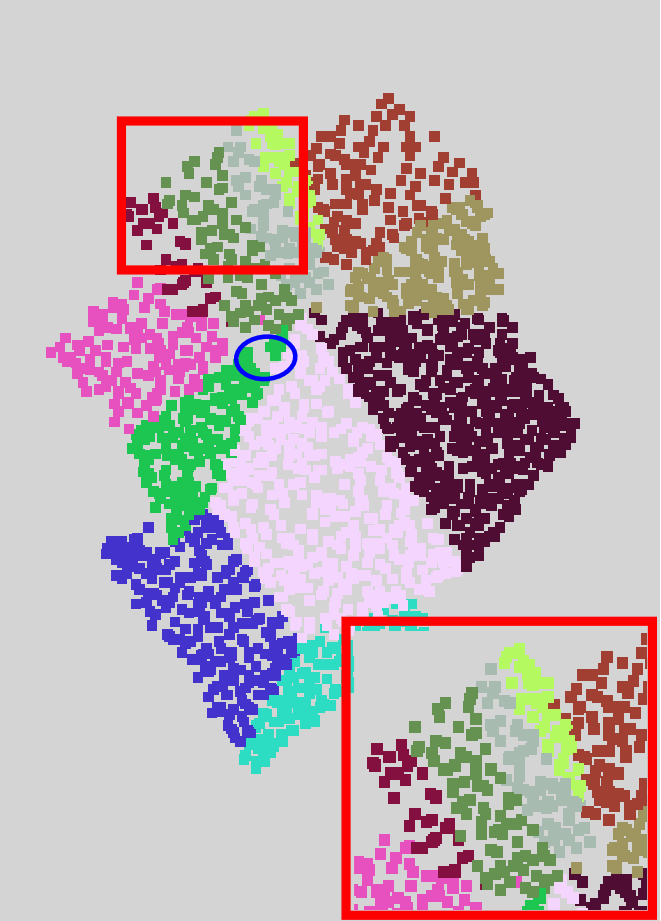}} &
		\multicolumn{1}{c}{\includegraphics[width=0.17\linewidth]{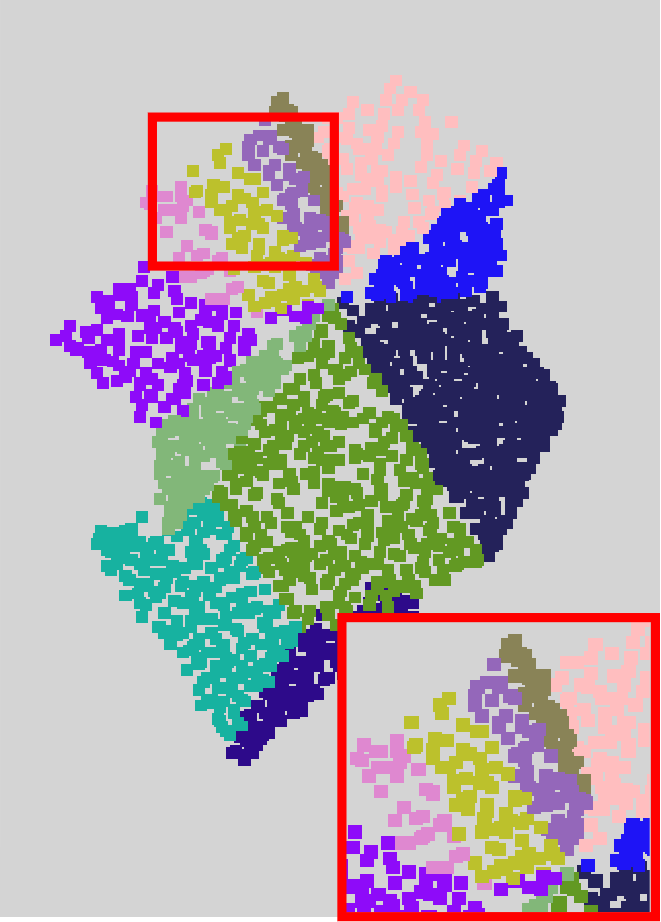}}
		\\
		($\text{F}_{2}$) HPNet & ($\text{G}_{2}$) Mask3D & ($\text{H}_{2}$) DeepRoofPlane & ($\text{I}_{2}$) RoofSeg (ours) & ($\text{J}_{2}$) GT
		
	\end{tabular} 
	\caption{Visual results of roof plane segmentation for all comparative approaches and RoofSeg on the RoofNTNU benchmark. The blue ellipses point out some incorrectly segmented regions. The red wireframes point out some regions with under-segmentation.}
	\vspace{0.0em}
	\label{Fig:comparative_RoofNTNU}
\end{figure}

\begin{figure}[h!]
	\scriptsize
	\centering
	\begin{tabular}{ccccc} 	
		
		\multicolumn{1}{c}{\includegraphics[width=0.17\linewidth]{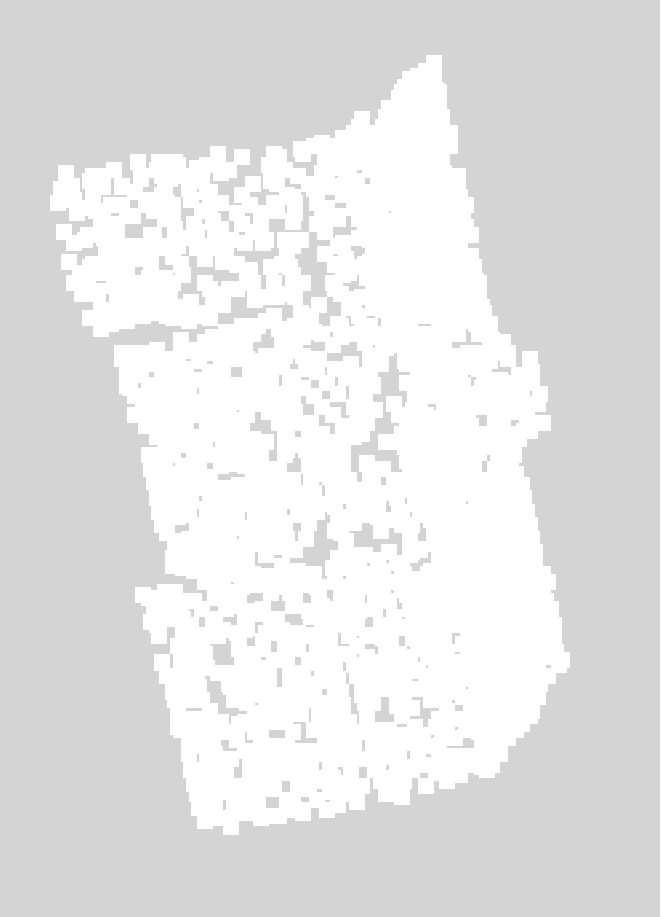}} &
		\multicolumn{1}{c}{\includegraphics[width=0.17\linewidth]{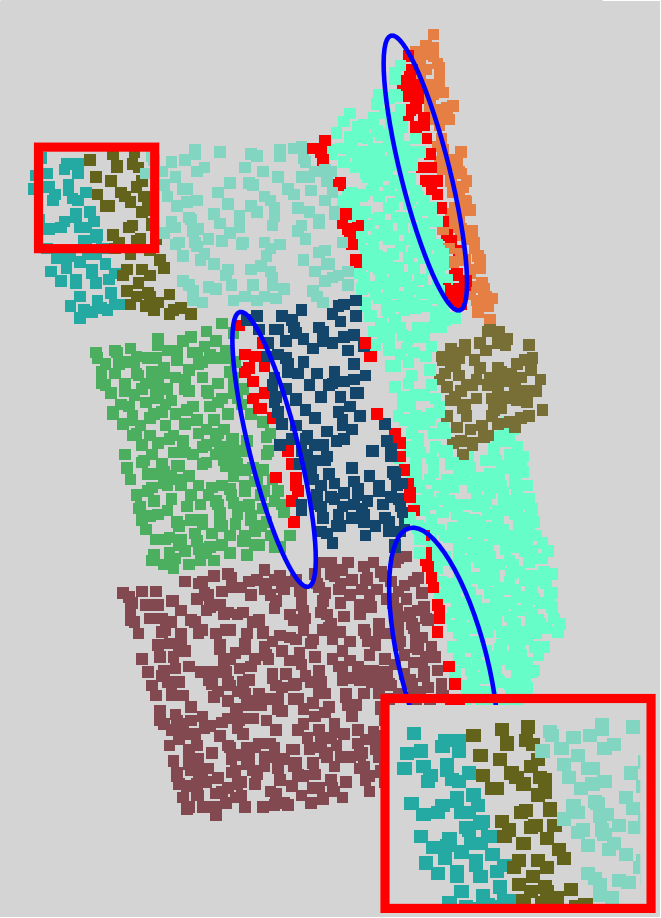}} &
		\multicolumn{1}{c}{\includegraphics[width=0.17\linewidth]{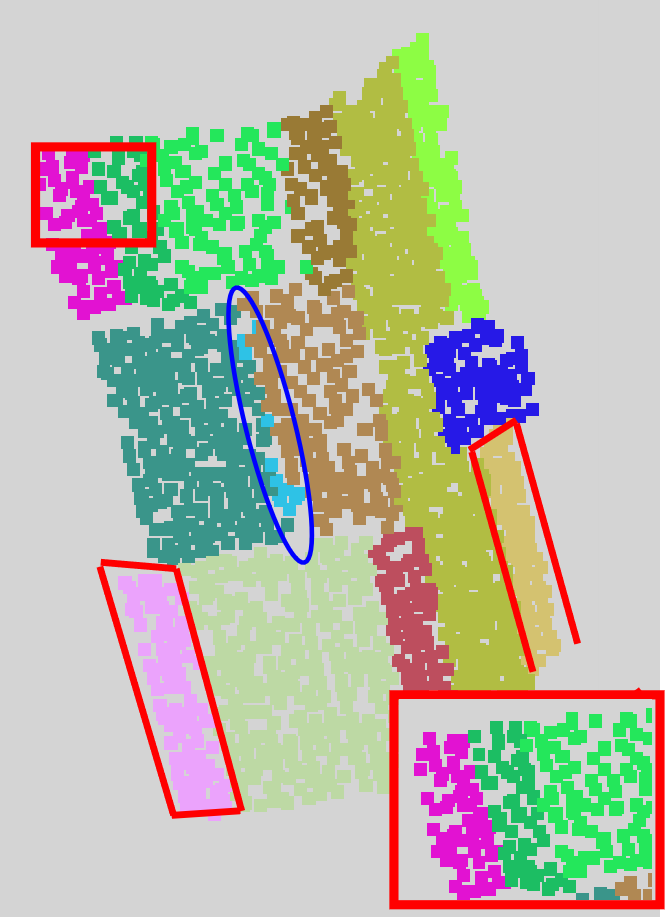}} &
		\multicolumn{1}{c}{\includegraphics[width=0.17\linewidth]{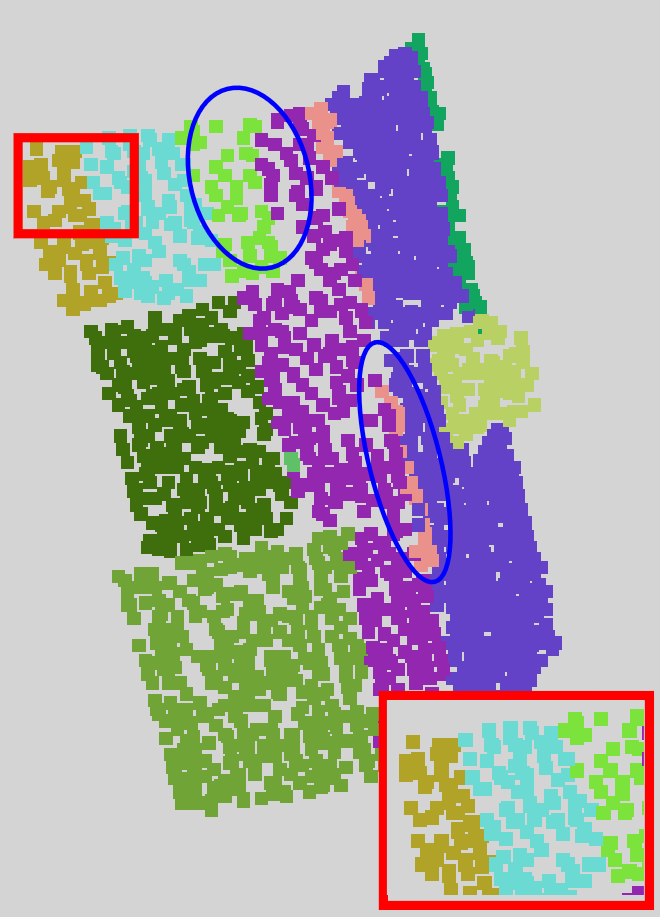}} &
		\multicolumn{1}{c}{\includegraphics[width=0.17\linewidth]{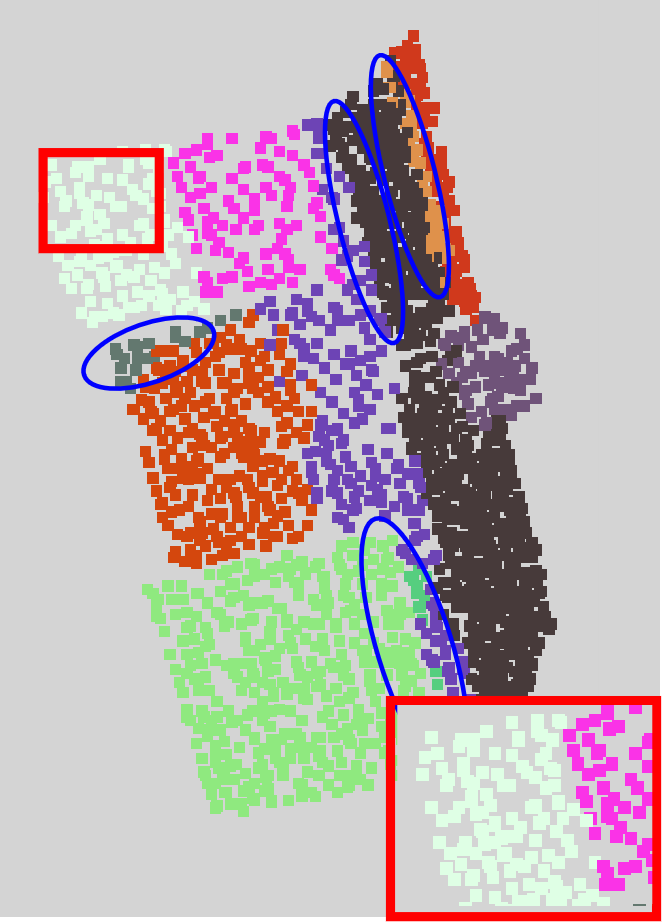}}
		\\
		($\text{A}_{1}$) Input & ($\text{B}_{1}$) Region growing & ($\text{C}_{1}$) RANSAC & ($\text{D}_{1}$) GoCoPP & ($\text{E}_{1}$) ParseNet
		\\
		
		\multicolumn{1}{c}{\includegraphics[width=0.17\linewidth]{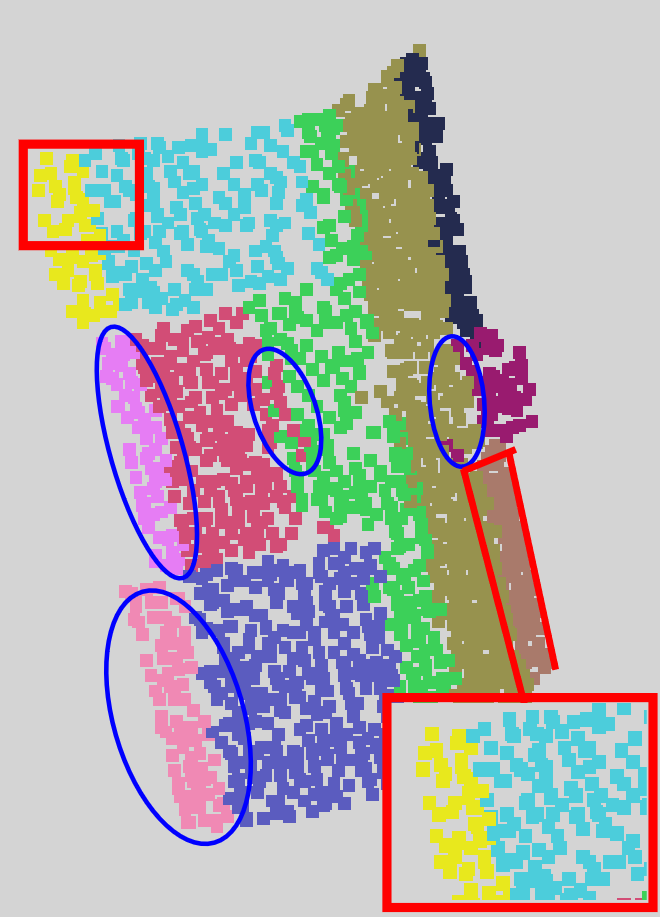}} &
		\multicolumn{1}{c}{\includegraphics[width=0.17\linewidth]{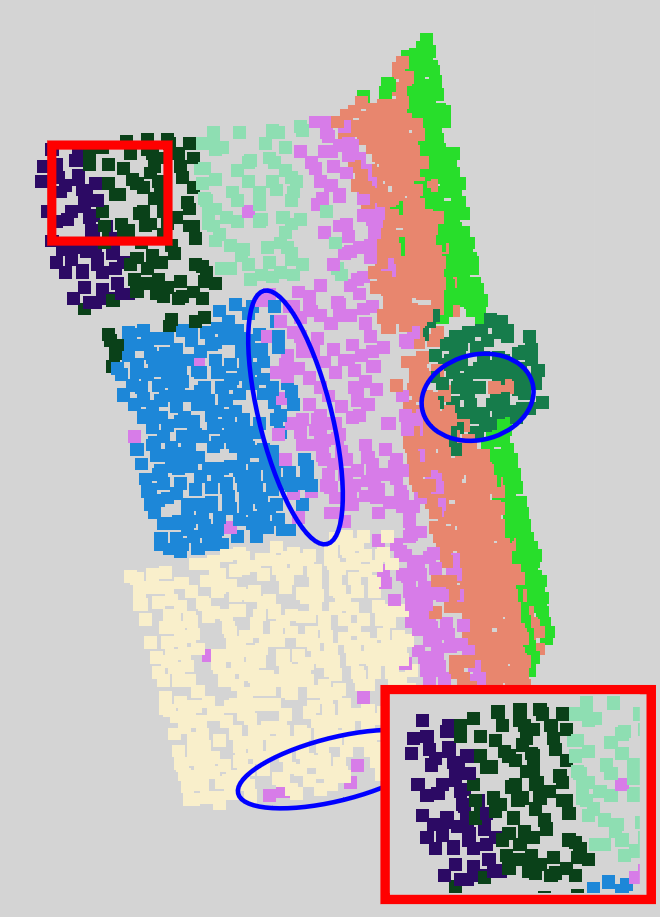}} &
		\multicolumn{1}{c}{\includegraphics[width=0.17\linewidth]{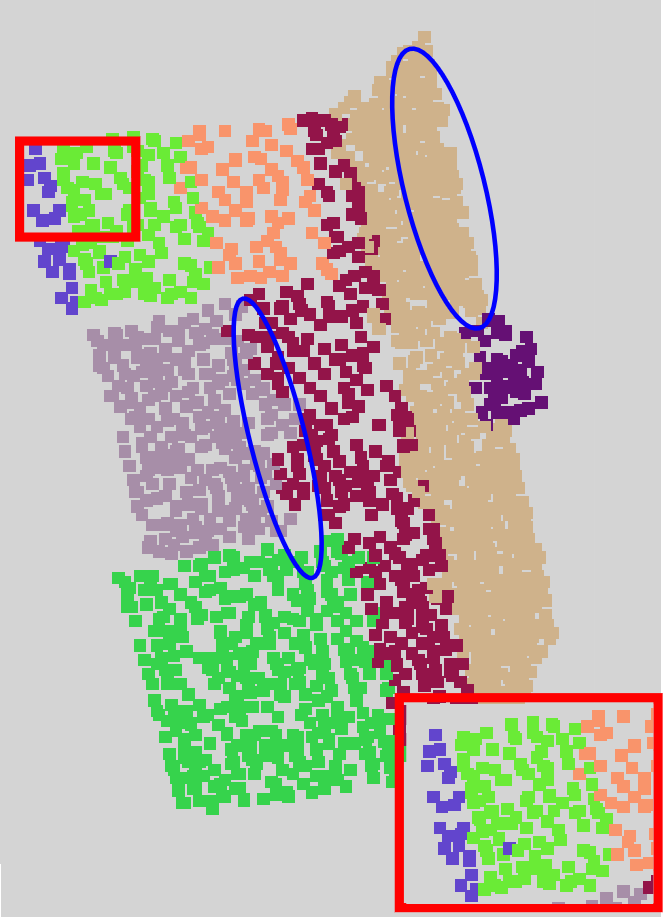}} &
		\multicolumn{1}{c}{\includegraphics[width=0.17\linewidth]{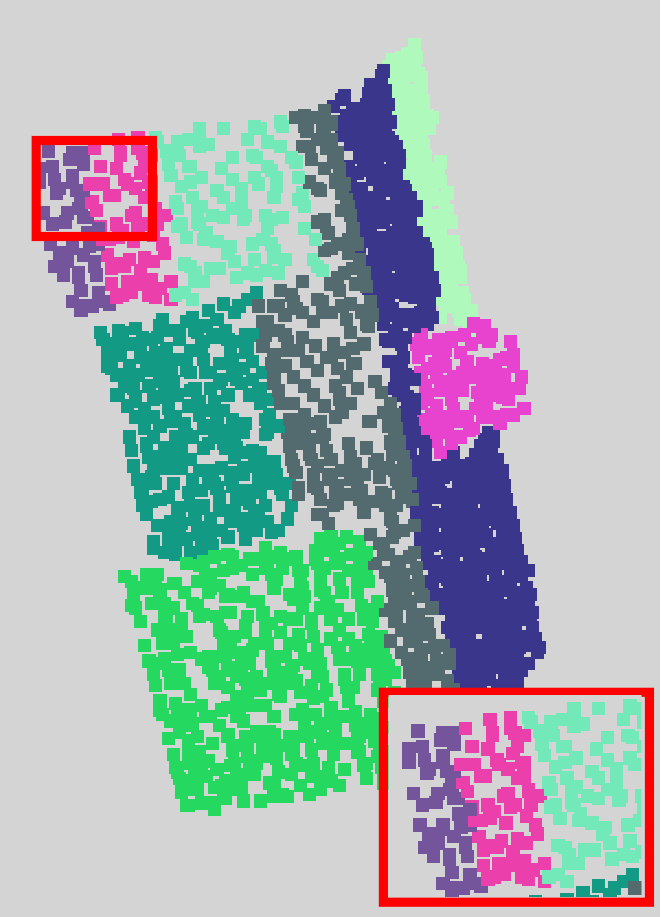}} &
		\multicolumn{1}{c}{\includegraphics[width=0.17\linewidth]{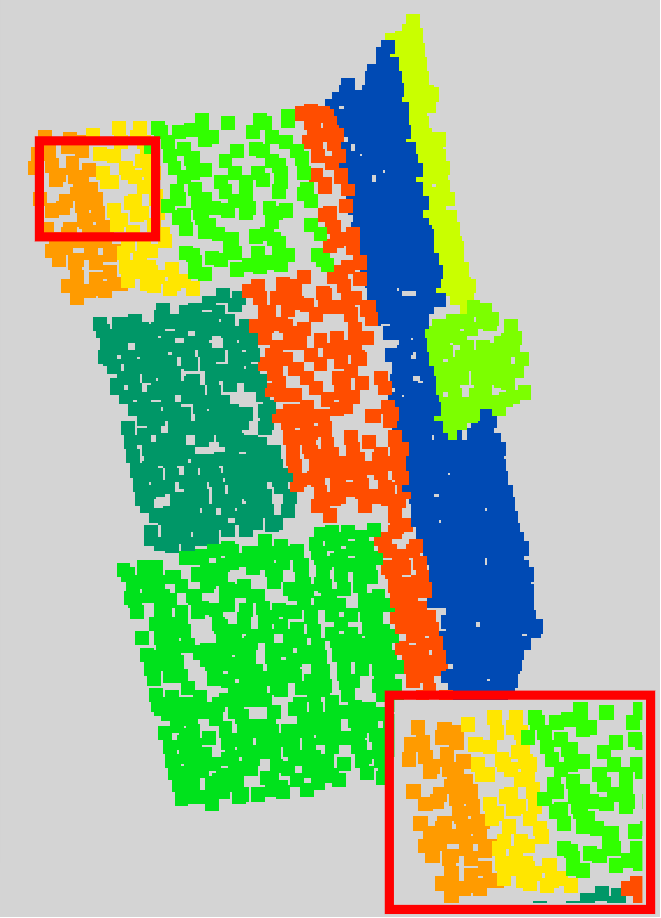}}
		\\
		($\text{F}_{1}$) HPNet & ($\text{G}_{1}$) Mask3D & ($\text{H}_{1}$) DeepRoofPlane & ($\text{I}_{1}$) RoofSeg (ours) & ($\text{J}_{1}$) GT
		\\
		
		& & & &\\
		\hdashline 
		& & & &\\
		
		\multicolumn{1}{c}{\includegraphics[width=0.17\linewidth]{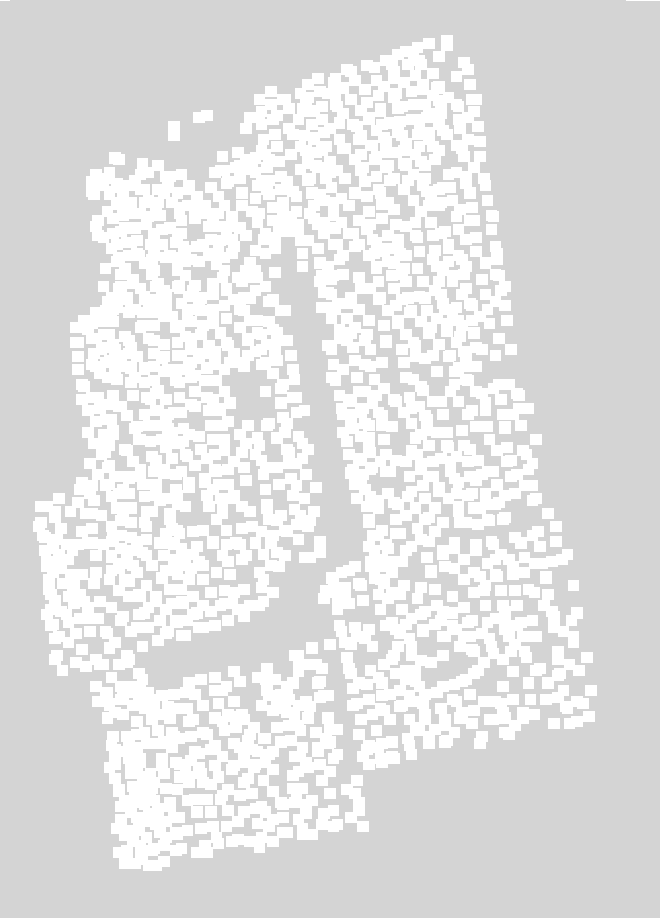}} &
		\multicolumn{1}{c}{\includegraphics[width=0.17\linewidth]{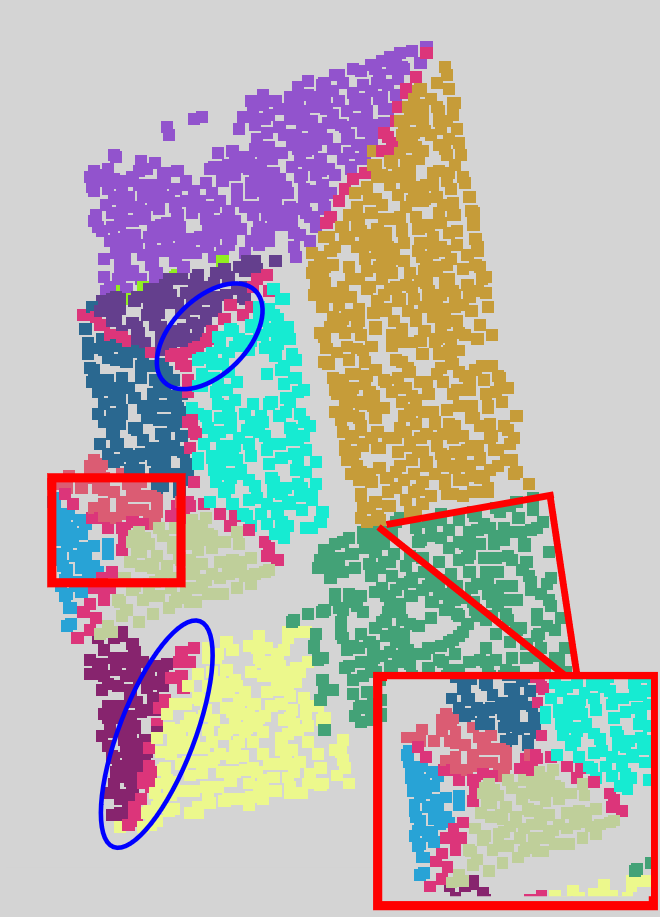}} &
		\multicolumn{1}{c}{\includegraphics[width=0.17\linewidth]{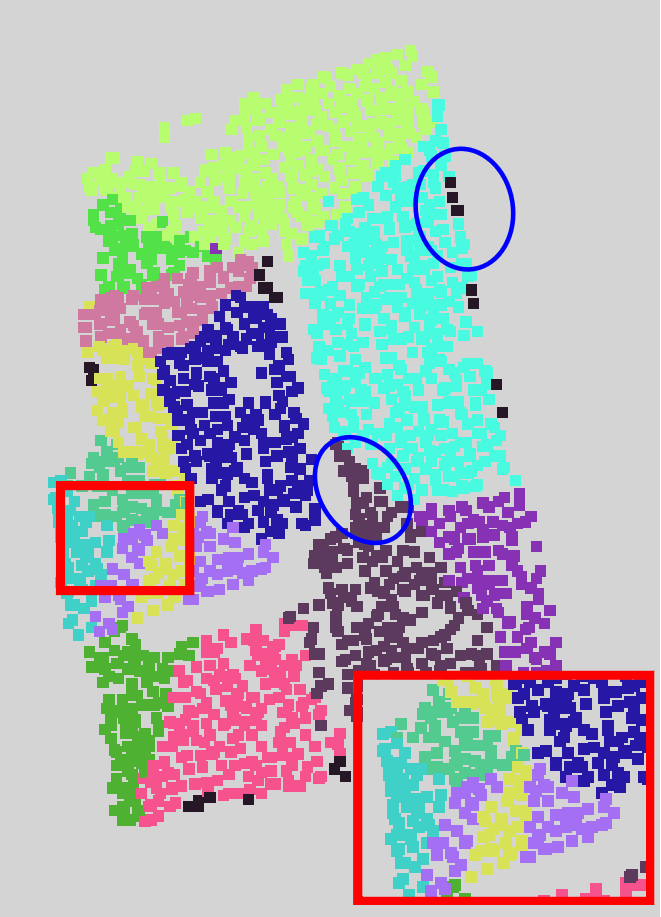}} &
		\multicolumn{1}{c}{\includegraphics[width=0.17\linewidth]{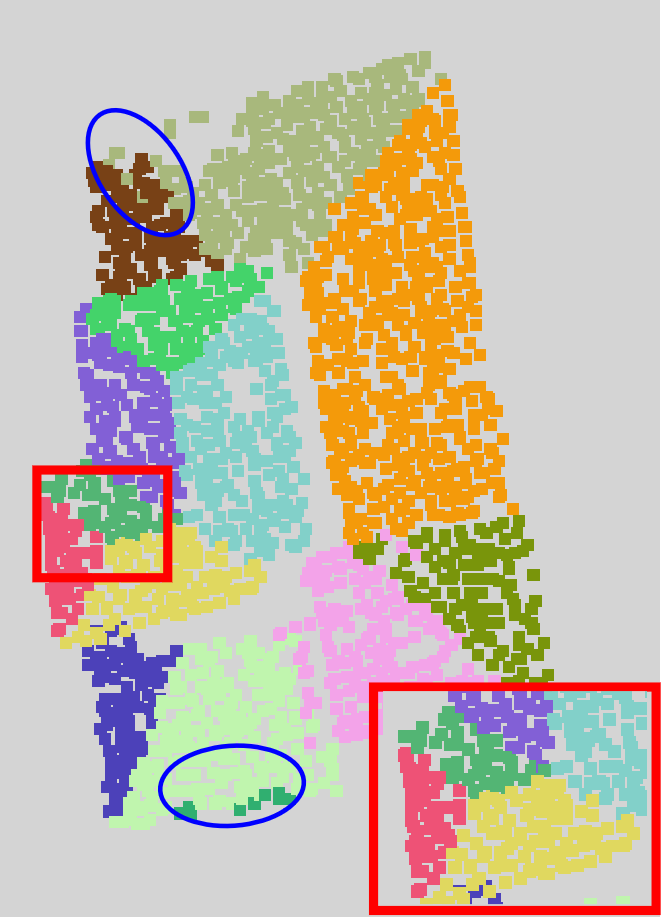}} &
		\multicolumn{1}{c}{\includegraphics[width=0.17\linewidth]{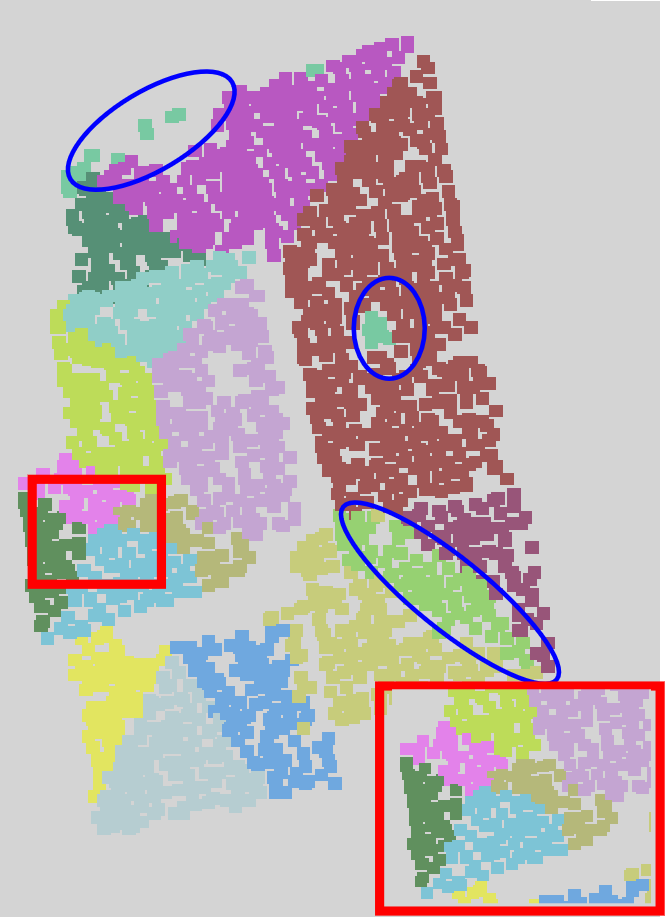}}
		\\
		($\text{A}_{2}$) Input & ($\text{B}_{2}$) Region growing & ($\text{C}_{2}$) RANSAC & ($\text{D}_{2}$) GoCoPP & ($\text{E}_{2}$) ParseNet
		\\
		
		\multicolumn{1}{c}{\includegraphics[width=0.17\linewidth]{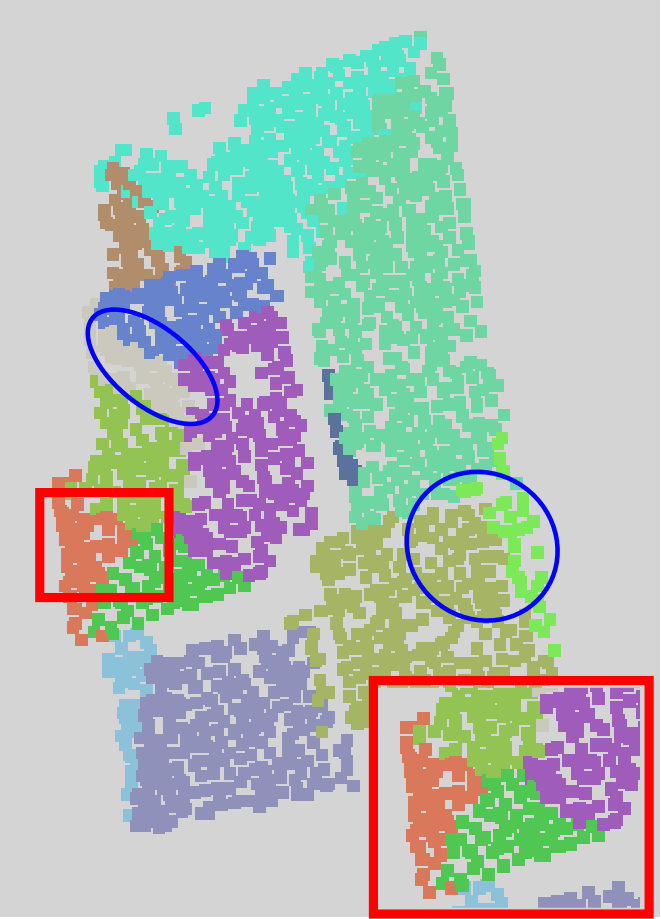}} &
		\multicolumn{1}{c}{\includegraphics[width=0.17\linewidth]{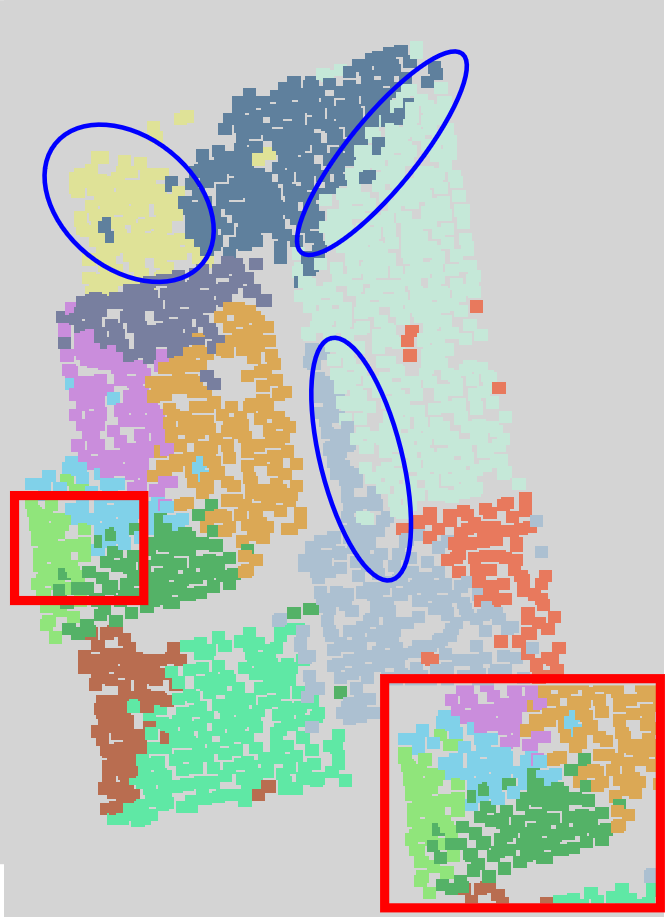}} &
		\multicolumn{1}{c}{\includegraphics[width=0.17\linewidth]{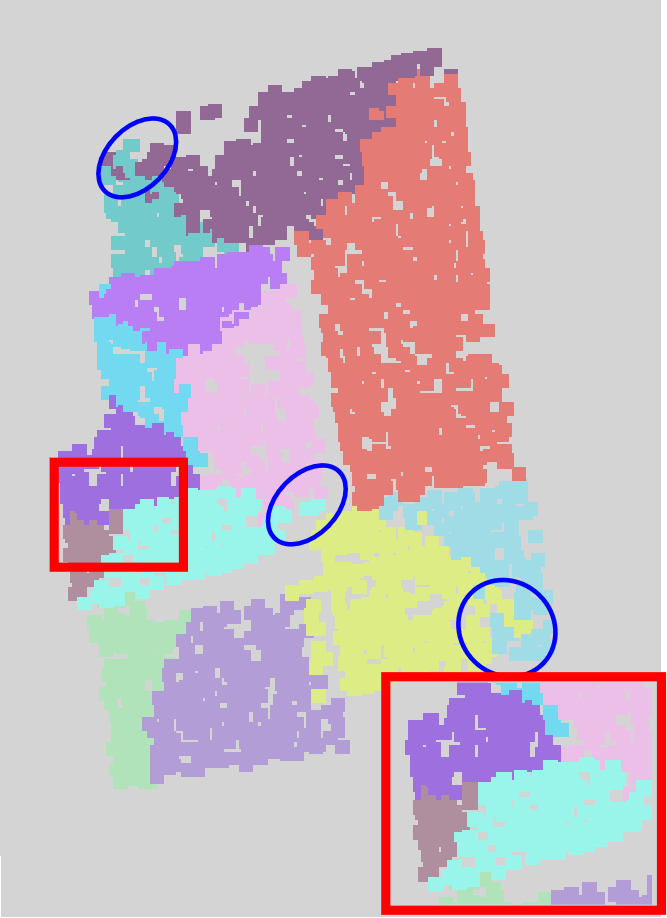}} &
		\multicolumn{1}{c}{\includegraphics[width=0.17\linewidth]{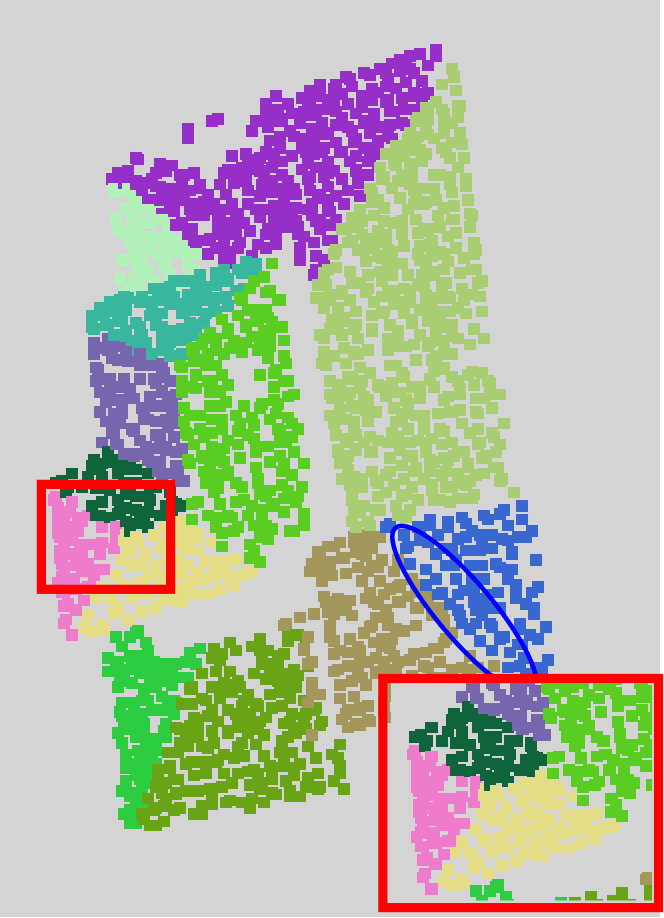}} &
		\multicolumn{1}{c}{\includegraphics[width=0.17\linewidth]{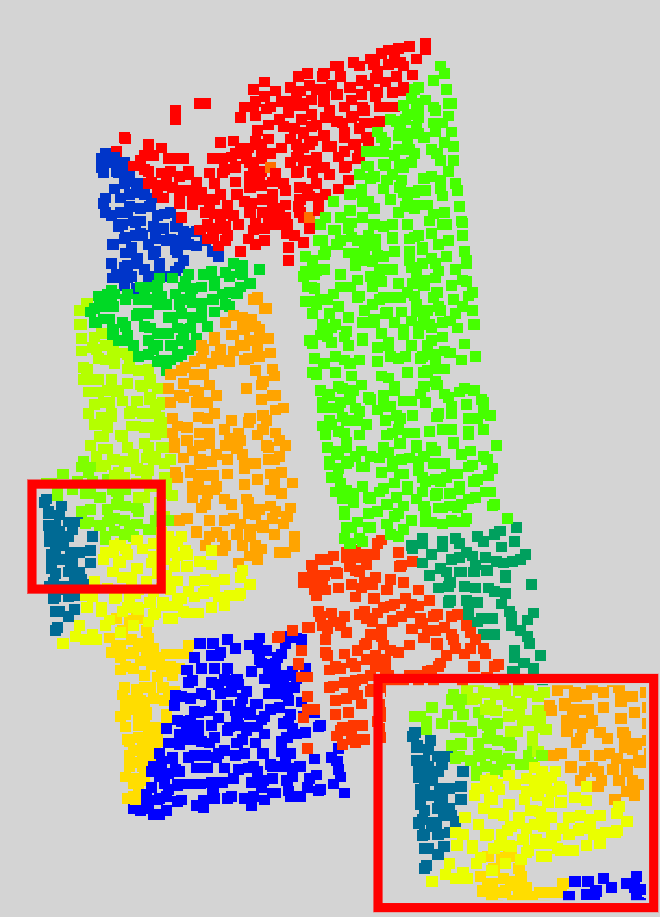}}
		\\
		($\text{F}_{2}$) HPNet & ($\text{G}_{2}$) Mask3D & ($\text{H}_{2}$) DeepRoofPlane & ($\text{I}_{2}$) RoofSeg (ours) & ($\text{J}_{2}$) GT
		
	\end{tabular} 
	\caption{Visual results of roof plane segmentation for all comparative approaches and RoofSeg on the Building3D benchmark. The blue ellipses point out some incorrectly segmented regions. The red wireframes point out some regions with under-segmentation.}
	\vspace{0.0em}
	\label{Fig:comparative_Building3D}
\end{figure}

\begin{figure}[h!]
	\scriptsize
	\centering
	\begin{tabular}{ccc} 	
		
		\multicolumn{3}{c}{\includegraphics[width= 0.98 \linewidth]{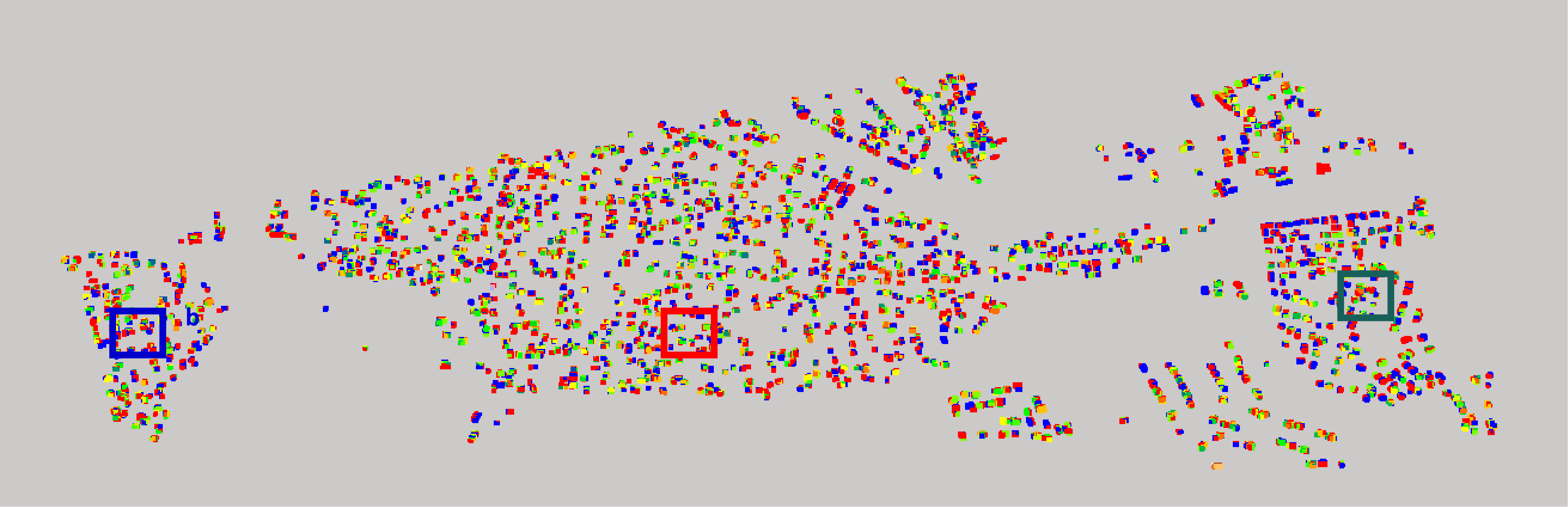}} 
		\\
		& (a) &
		\\
		\multicolumn{1}{c}{\includegraphics[width= 0.31 \linewidth]{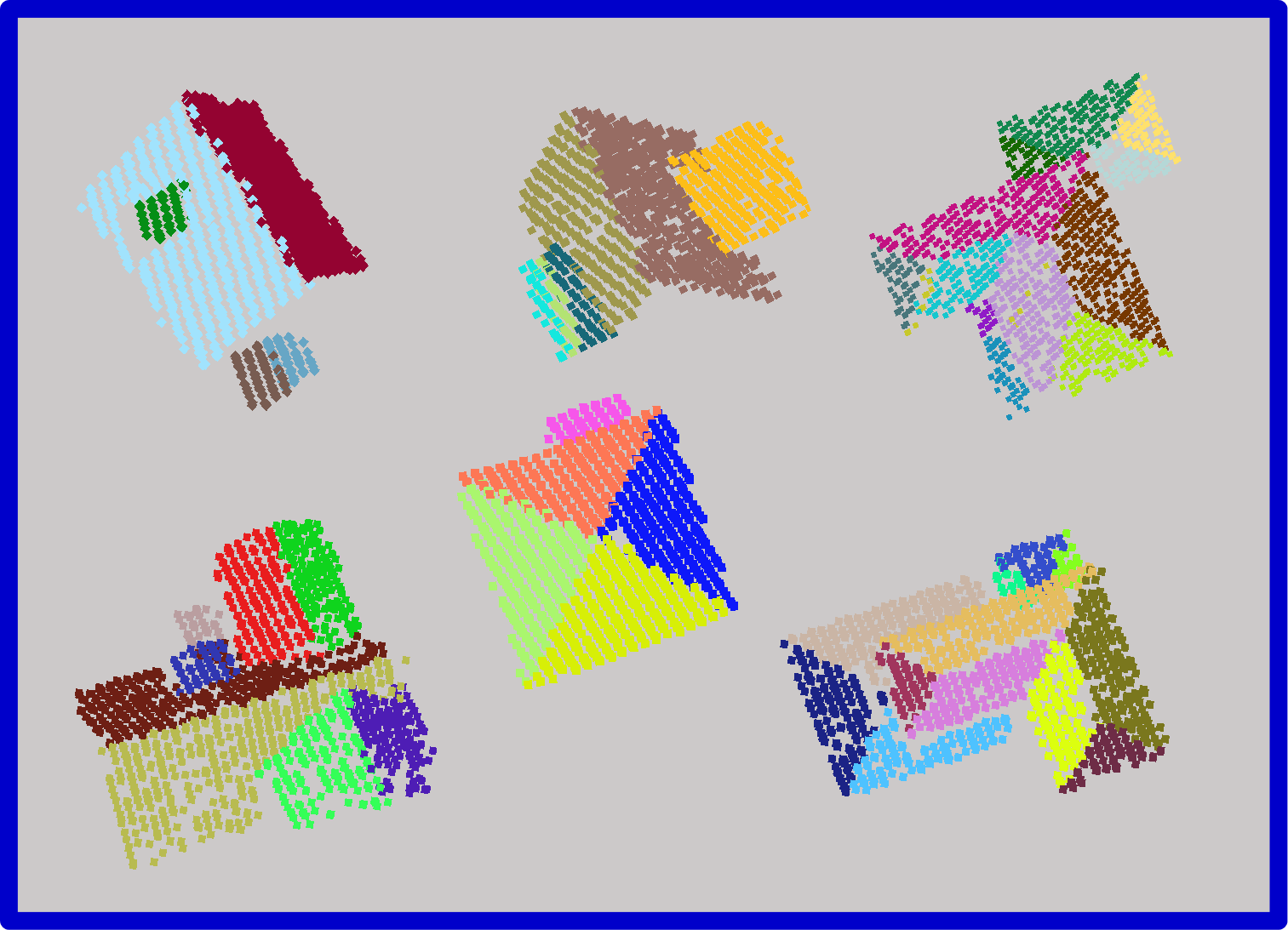}} &
		\multicolumn{1}{c}{\includegraphics[width= 0.31 \linewidth]{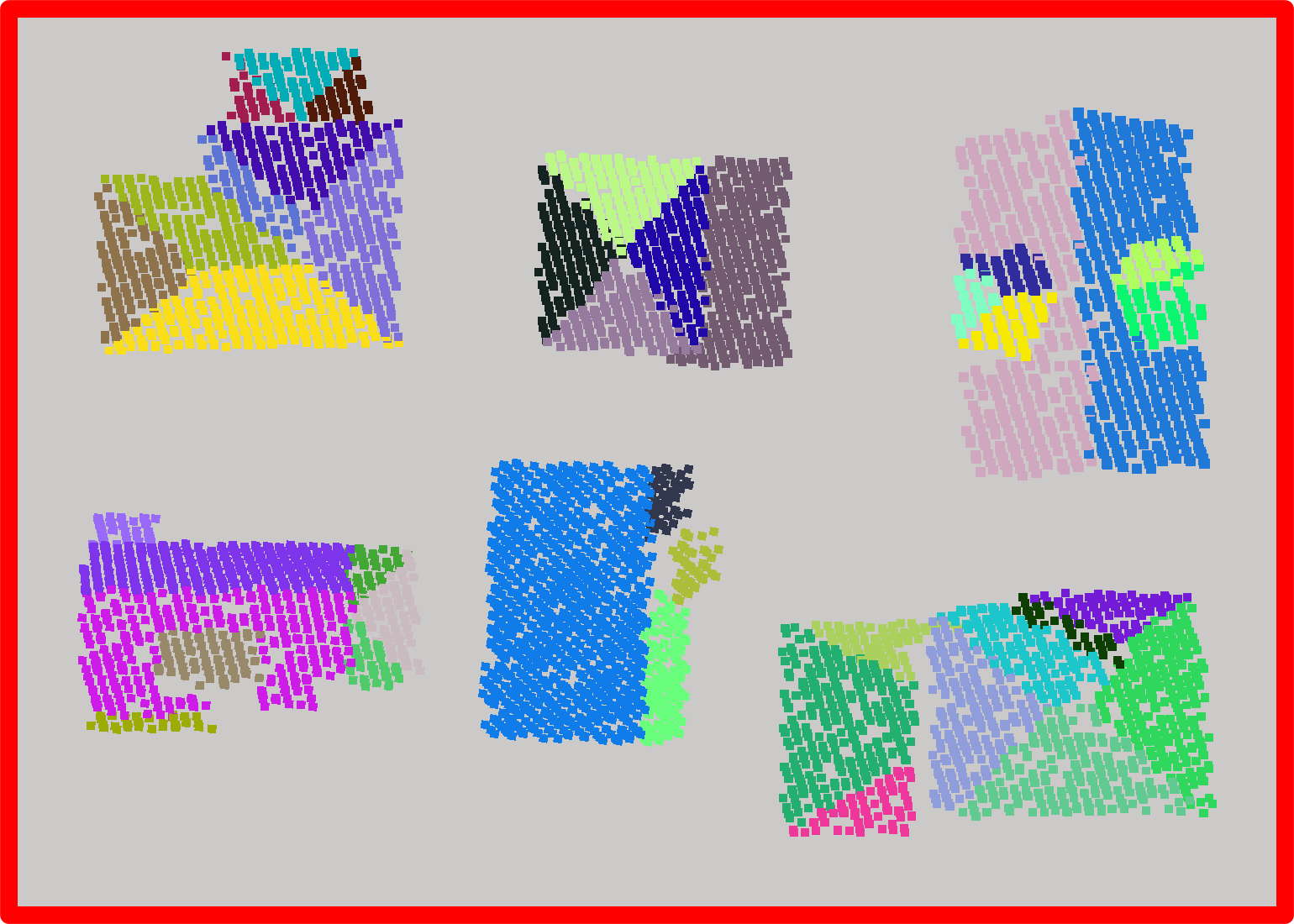}} &
		\multicolumn{1}{c}{\includegraphics[width= 0.31 \linewidth]{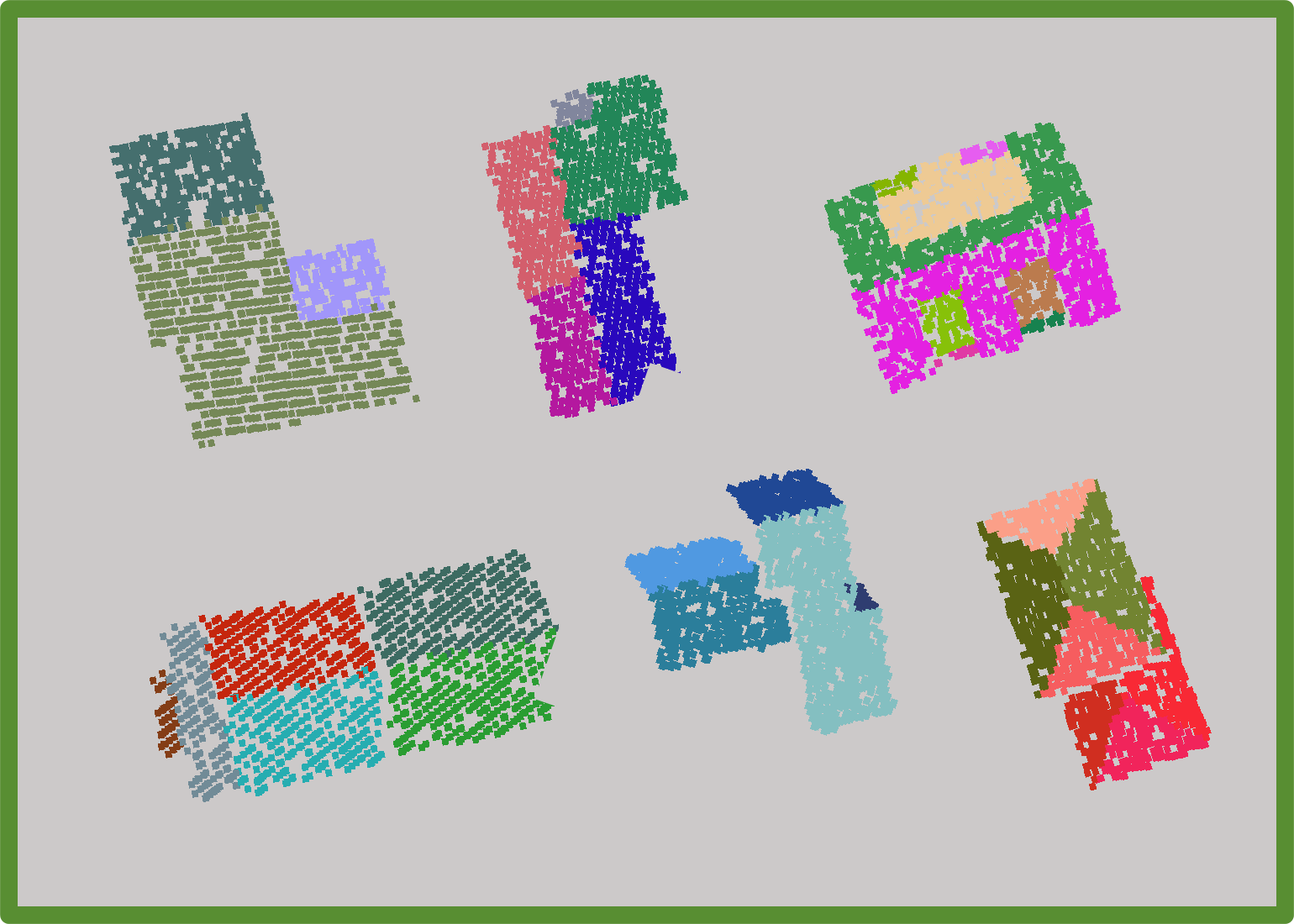}} 
		\\
		(b) & (c) & (d)

	\end{tabular} 
	\caption{Large scale segmentation results of RoofSeg on the real-world Building3D benchmark. In (a), we present all 1827 segmented roofs in the testing set. (b), (c) and (d) highlight the segmentation results of three local regions, respectively.}
	\vspace{0.0em}
	\label{Fig:Building3D_large_scale}
\end{figure}

To qualitatively evaluate different approaches, we present segmentation results of representative roof samples from three benchmarks in Figures 6, 7, and 8. For the Roofpc3D and RoofNTNU benchmarks, Region Growing~\citep{vo2015octree} tends to fail at edge regions between adjacent patches (Fig.~\ref{Fig:comparative_Roofpc3D} ($\text{B}_{1}$)) and produce under-segmented patches (Fig.~\ref{Fig:comparative_RoofNTNU} ($\text{B}_{2}$)). RANSAC~\citep{xu2015investigation} is prone to merging two approximately coplanar patches (Fig.~\ref{Fig:comparative_Roofpc3D} ($\text{C}_{1}$)) and generate spurious planes (Fig.~\ref{Fig:comparative_Roofpc3D} ($\text{C}_{2}$)). The results of GoCoPP~\citep{yu2022finding} tends to miss some small patches (Fig.~\ref{Fig:comparative_RoofNTNU} ($\text{D}_{1}$) and ($\text{D}_{2}$)). HPNet~\citep{yan2021hpnet} and ParseNet~\citep{sharma2020parsenet} struggle with over- and under-segmentation between adjacent patches (Fig.~\ref{Fig:comparative_Roofpc3D} ($\text{E}_{2}$) and ($\text{F}_{2}$)), and produce inaccurate edge points (Fig.~\ref{Fig:comparative_RoofNTNU} ($\text{E}_{1}$) and ($\text{F}_{1}$)). Mask3D~\citep{schult2023mask3d} generates some misclassified points across different patches (Fig.~\ref{Fig:comparative_Roofpc3D} ($\text{G}_{2}$)) and fail to correctly segment small planes (Fig.~\ref{Fig:comparative_RoofNTNU} ($\text{G}_{1}$)). DeepRoofPlane~\citep{li2024boundary} generates non-smooth edges at junctions and shows some over-segmented regions (Fig.~\ref{Fig:comparative_RoofNTNU} ($\text{H}_{1}$) and Fig.~\ref{Fig:comparative_RoofNTNU} ($\text{H}_{2}$)). RoofSeg achieves the highest-quality segmentation results, with accurate edges, minimal misclassified points and high geometric fidelity. For the more challenging Building3D benchmark (with complex roof structures), all comparative approaches show performance degradation with severe under- or over-segmentation issues and inaccurate edges. In contrast, RoofSeg achieves optimal results that maximally approximate the ground truth, demonstrating its effectiveness in handling challenging roof samples. Moreover, we also show the urban-scale segmented roofs of RoofSeg on the Building3D benchmark in Fig.~\ref{Fig:Building3D_large_scale}.\footnote{\noindent\parbox[t]{0.95\linewidth}{\raggedright
		The urban-scale segmentation results are publicly available at \url{https://github.com/Li-Li-Whu/RoofSeg/urban_results}.
}}

\begin{figure}[!htb]
	\scriptsize
	\centering
	\begin{tabular}{ccccc} 	
		
		\multicolumn{1}{c}{\includegraphics[width=0.18\linewidth]{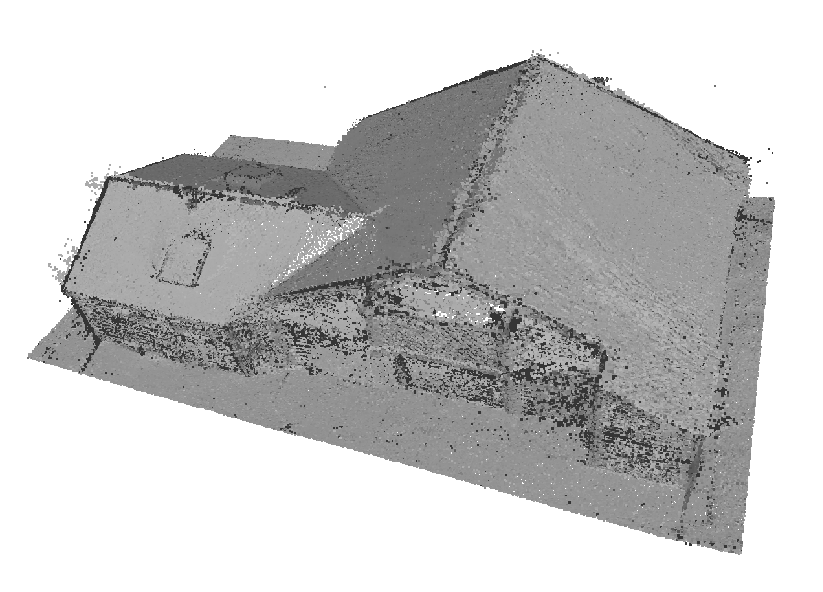}} &
		\multicolumn{1}{c}{\includegraphics[width=0.18\linewidth]{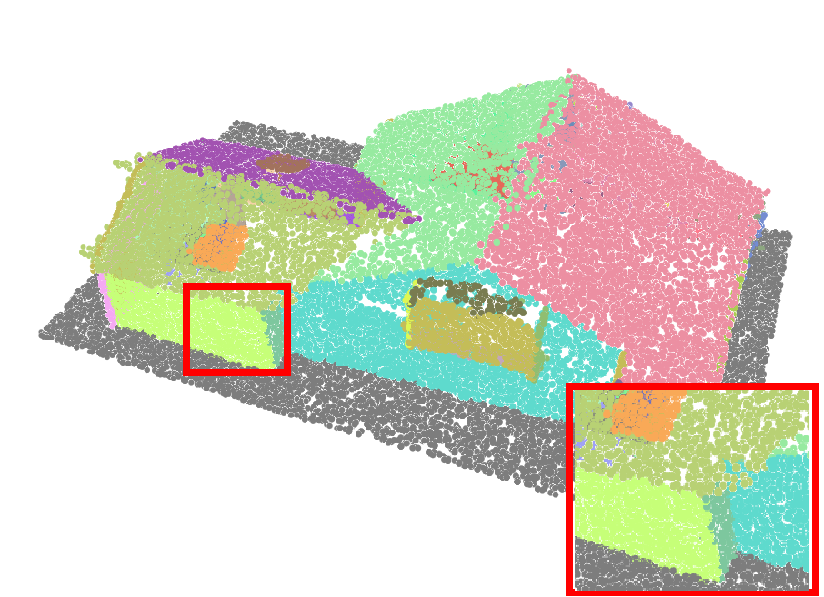}} &
		\multicolumn{1}{c}{\includegraphics[width=0.18\linewidth]{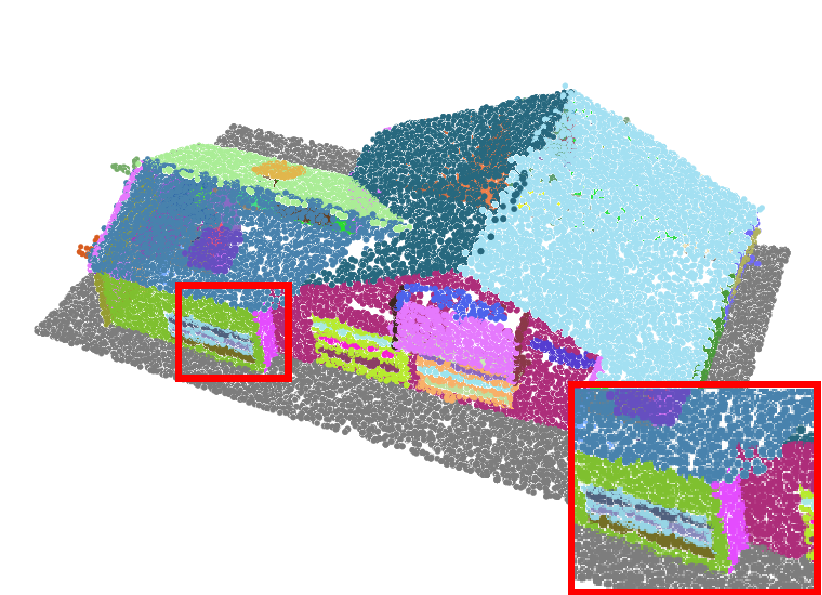}} &
		\multicolumn{1}{c}{\includegraphics[width=0.18\linewidth]{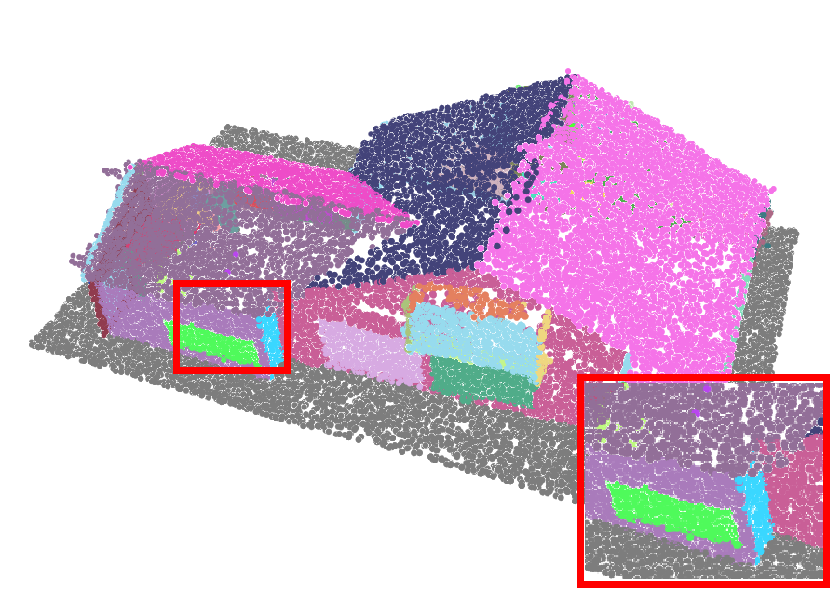}} &
		\multicolumn{1}{c}{\includegraphics[width=0.18\linewidth]{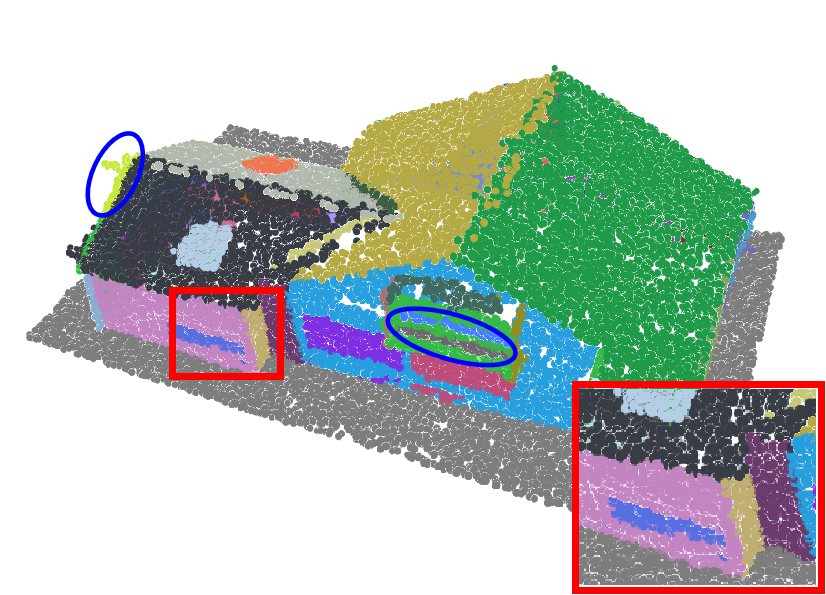}}
		\\
		($\text{A}_{1}$) Barn MVS & ($\text{B}_{1}$) Region growing & ($\text{C}_{1}$) RANSAC & ($\text{D}_{1}$) GoCoPP & ($\text{E}_{1}$) ParseNet
		\\
		
		\multicolumn{1}{c}{\includegraphics[width=0.18\linewidth]{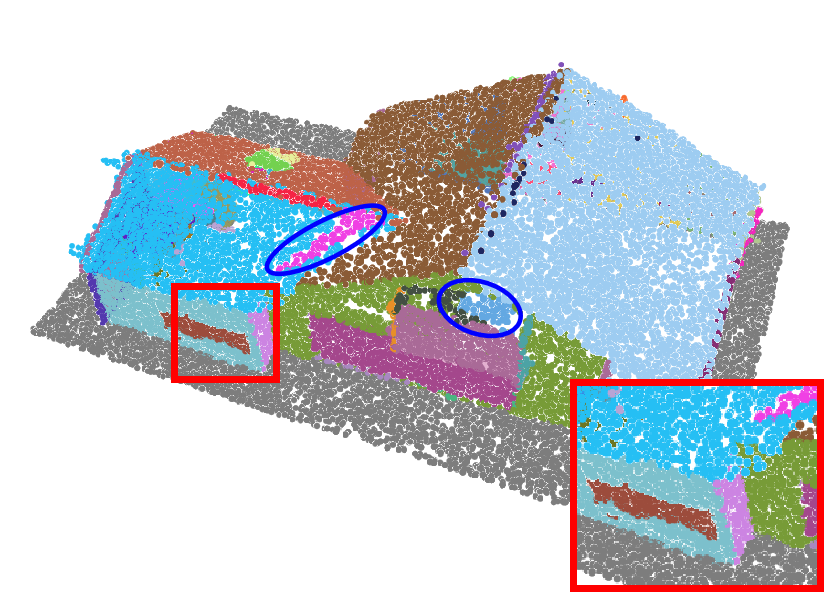}} &
		\multicolumn{1}{c}{\includegraphics[width=0.18\linewidth]{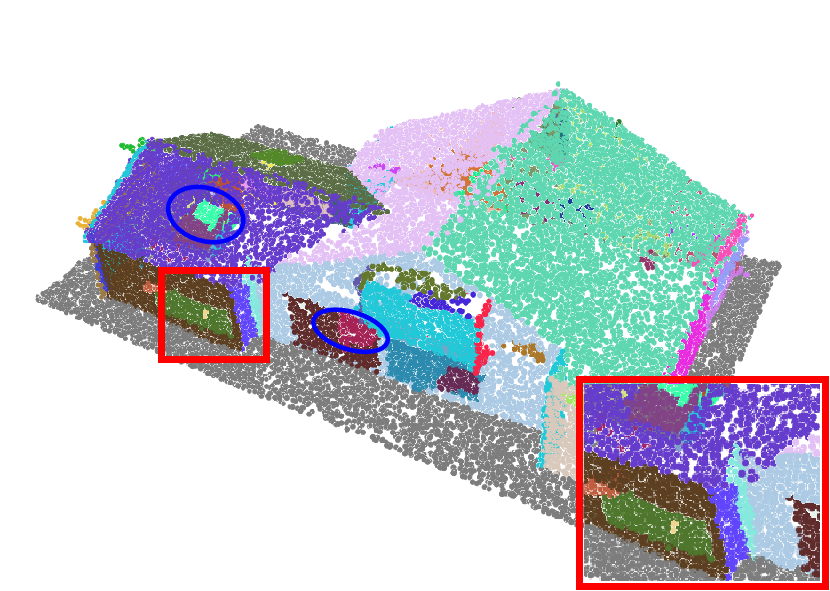}} &
		\multicolumn{1}{c}{\includegraphics[width=0.18\linewidth]{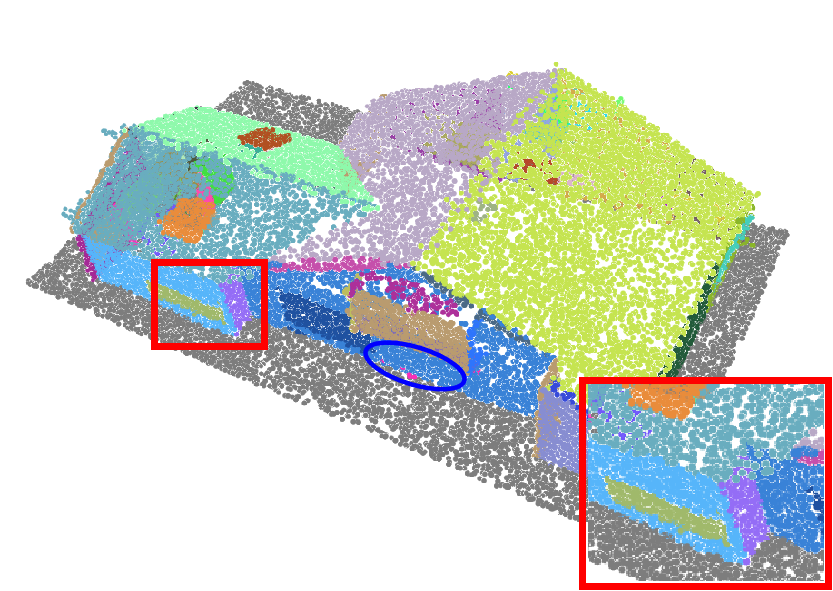}} &
		\multicolumn{1}{c}{\includegraphics[width=0.18\linewidth]{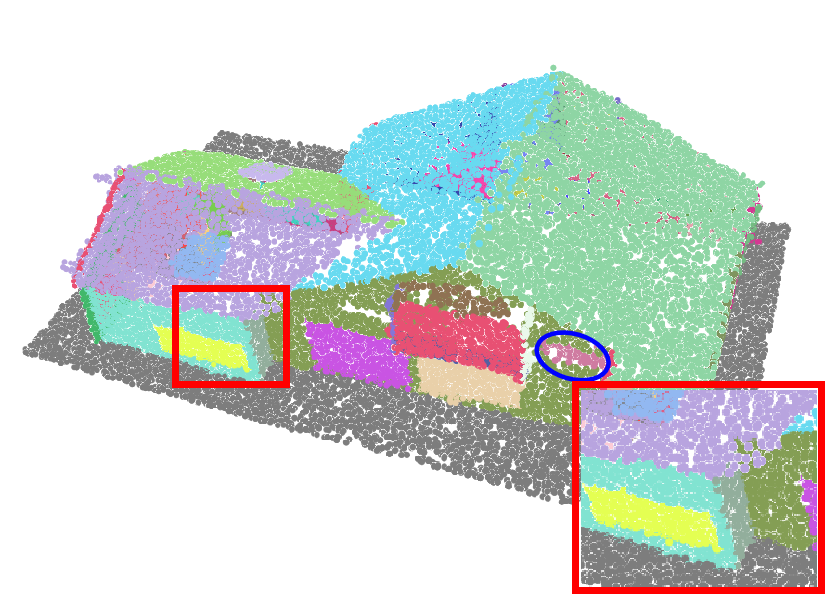}} &
		\multicolumn{1}{c}{\includegraphics[width=0.18\linewidth]{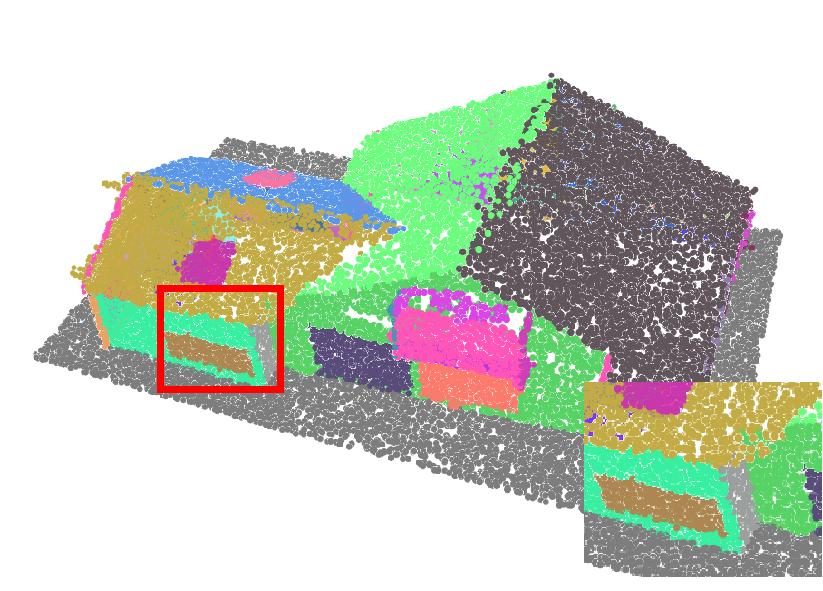}}
		\\
		($\text{F}_{1}$) HPNet & ($\text{G}_{1}$) Mask3D & ($\text{H}_{1}$) DeepRoofPlane & ($\text{I}_{1}$) RoofSeg (ours) & ($\text{J}_{1}$) GT
		\\
		
		& & & & \\
		\hdashline 
		& & & & \\
		
		\multicolumn{1}{c}{\includegraphics[width=0.18\linewidth]{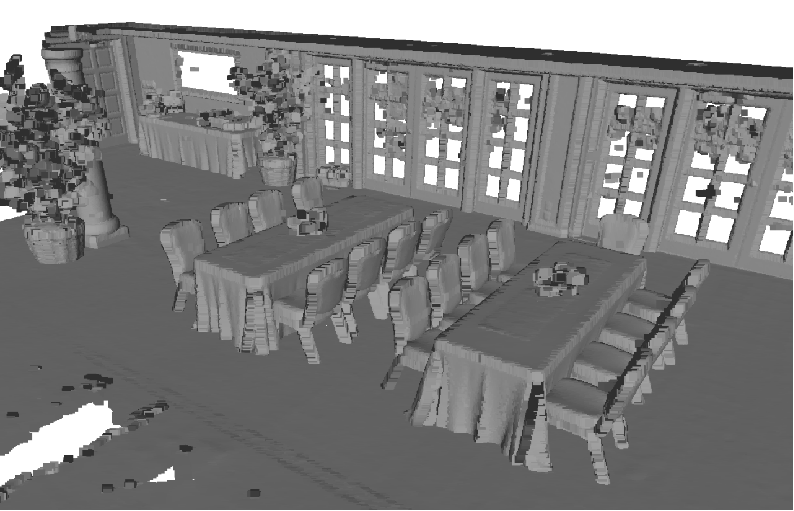}} &
		\multicolumn{1}{c}{\includegraphics[width=0.18\linewidth]{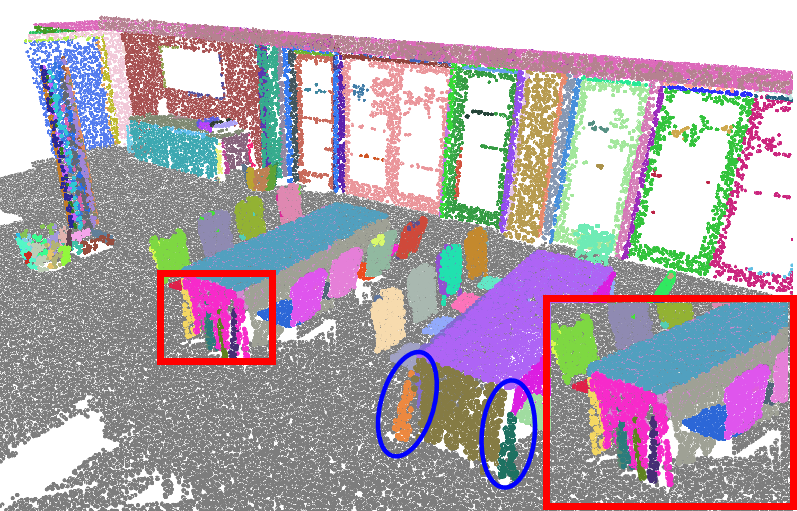}} &
		\multicolumn{1}{c}{\includegraphics[width=0.18\linewidth]{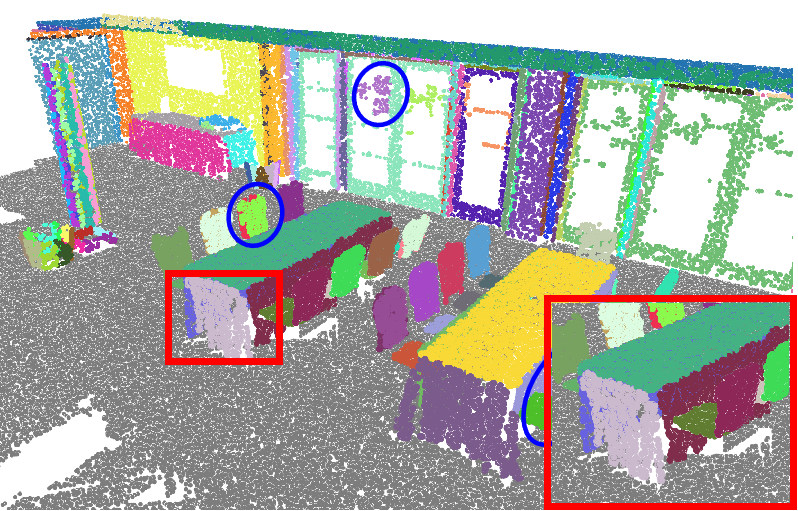}} &
		\multicolumn{1}{c}{\includegraphics[width=0.18\linewidth]{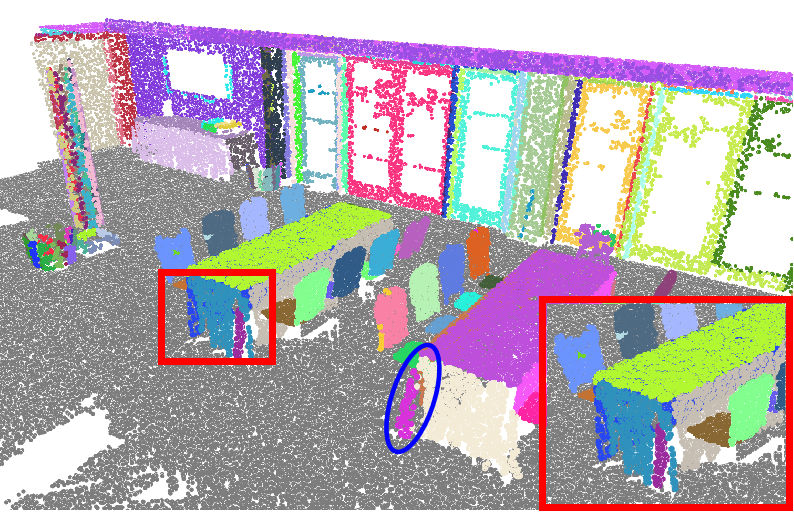}} &
		\multicolumn{1}{c}{\includegraphics[width=0.18\linewidth]{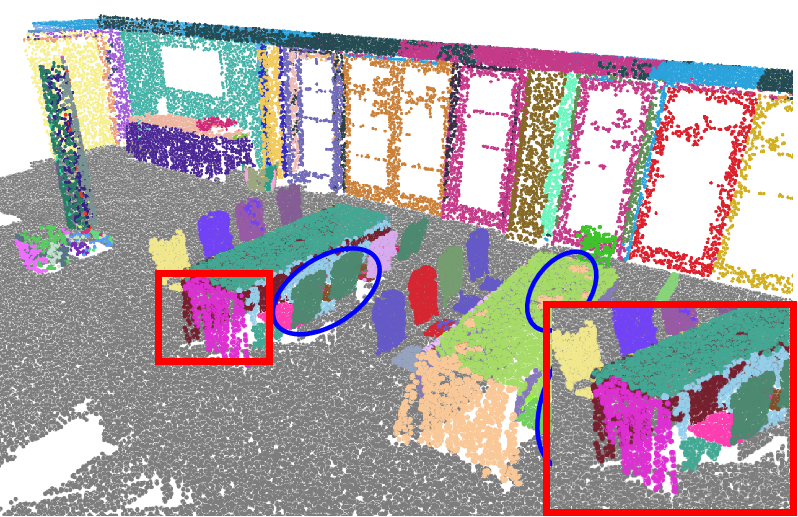}}
		\\
		($\text{A}_{2}$) Meeting room & ($\text{B}_{2}$) Region growing & ($\text{C}_{2}$) RANSAC & ($\text{D}_{2}$) GoCoPP & ($\text{E}_{2}$) ParseNet
		\\
		
		\multicolumn{1}{c}{\includegraphics[width=0.18\linewidth]{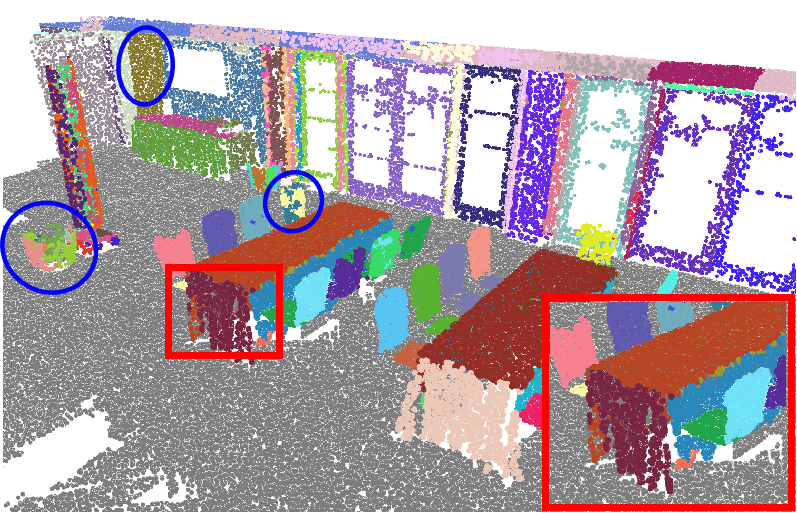}} &
		\multicolumn{1}{c}{\includegraphics[width=0.18\linewidth]{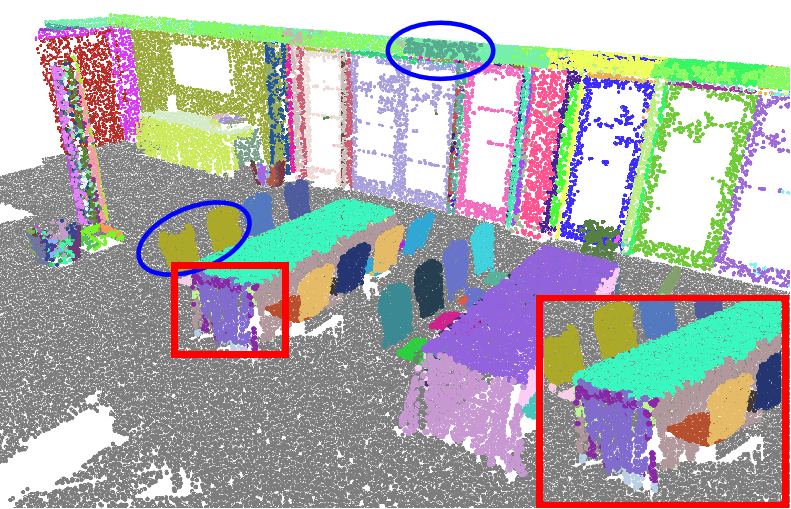}} &
		\multicolumn{1}{c}{\includegraphics[width=0.18\linewidth]{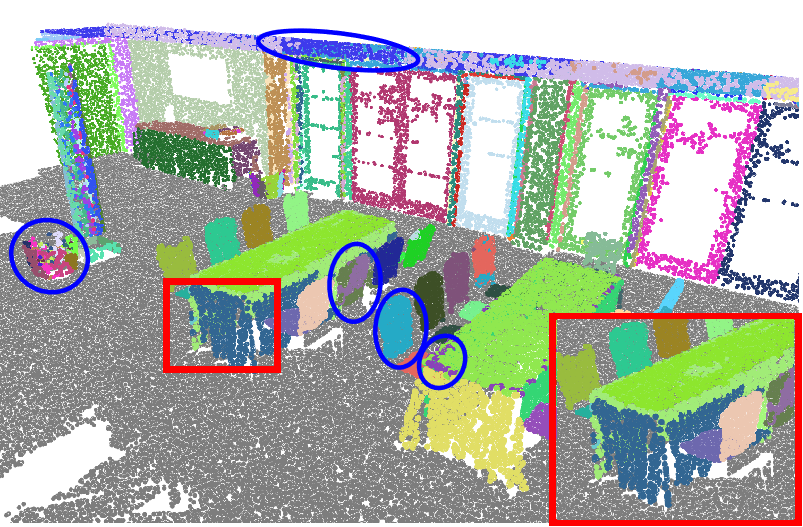}} &
		\multicolumn{1}{c}{\includegraphics[width=0.18\linewidth]{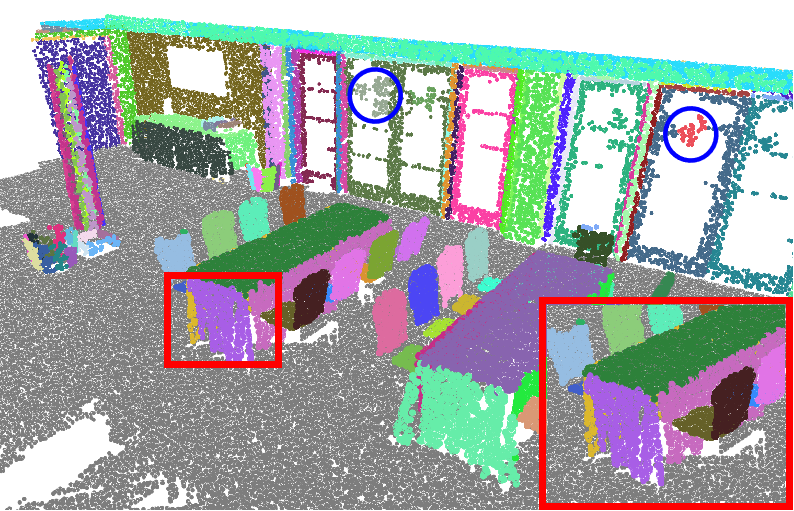}} &
		\multicolumn{1}{c}{\includegraphics[width=0.18\linewidth]{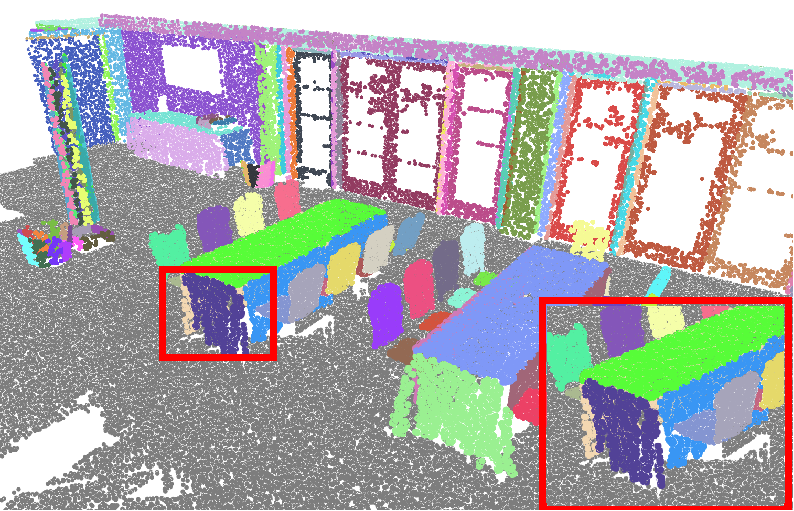}}
		\\
		($\text{F}_{2}$) HPNet & ($\text{G}_{2}$) Mask3D & ($\text{H}_{2}$) DeepRoofPlane & ($\text{I}_{2}$) RoofSeg (ours) & ($\text{J}_{2}$) GT
		
	\end{tabular} 
	\caption{Visual segmentation results of photogrammetric (barn MVS) and indoor-scene (meeting room) point clouds from the KSR42 benchmark. The blue ellipses point out some incorrectly segmented regions.}
	\vspace{0.0em}
	\label{Fig:comparative_MVS_indoor}
\end{figure}

\begin{figure}[!htb]
	\scriptsize
	\centering
	\begin{tabular}{ccccc} 	
		
		\multicolumn{1}{c}{\includegraphics[width= 0.18 \linewidth]{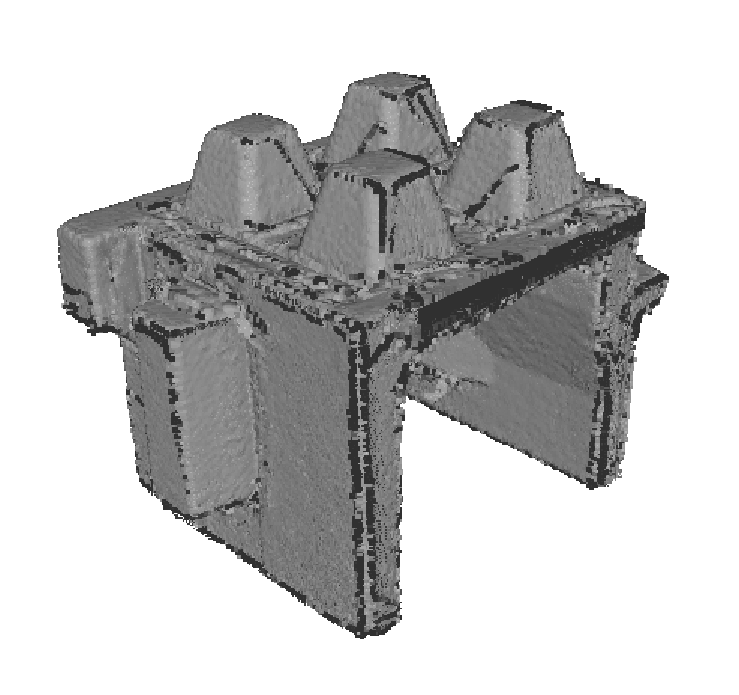}} &
		\multicolumn{1}{c}{\includegraphics[width= 0.18 \linewidth]{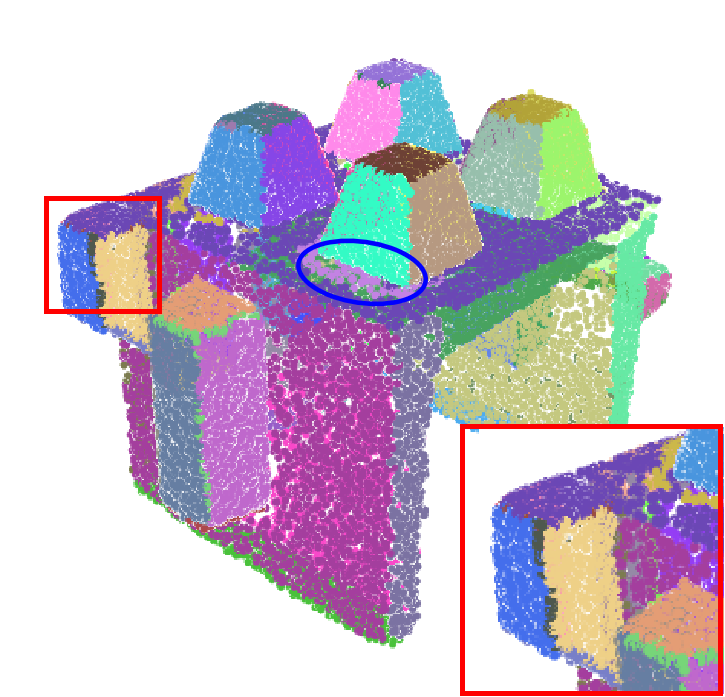}} &
		\multicolumn{1}{c}{\includegraphics[width= 0.18 \linewidth]{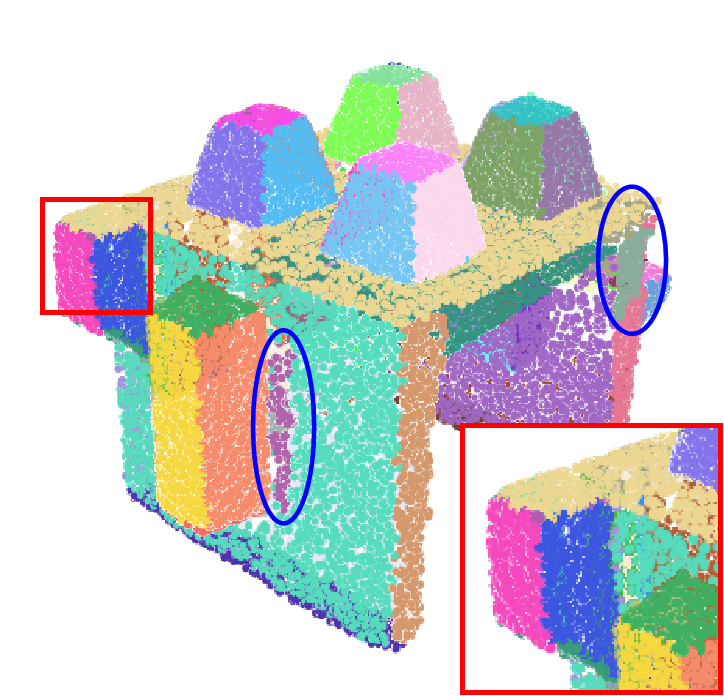}} &
		\multicolumn{1}{c}{\includegraphics[width= 0.18 \linewidth]{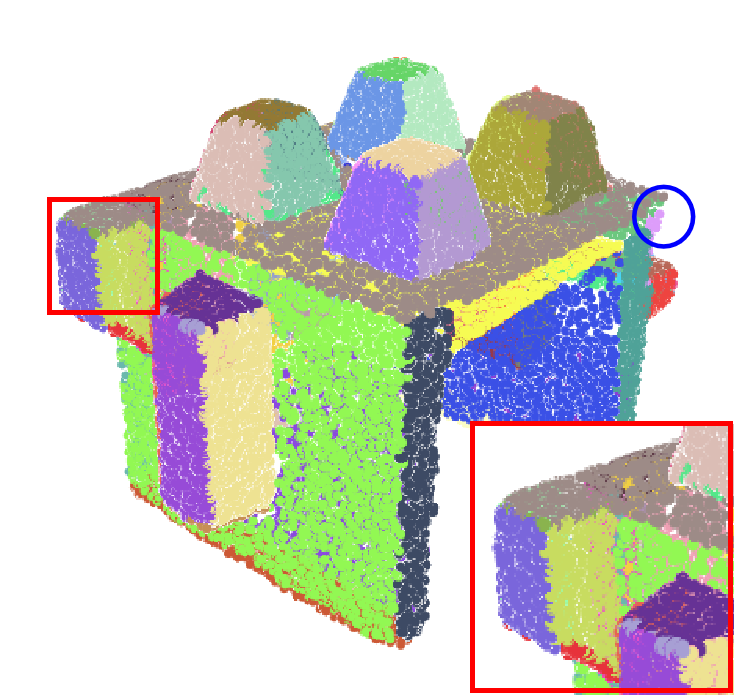}} &
		\multicolumn{1}{c}{\includegraphics[width= 0.18 \linewidth]{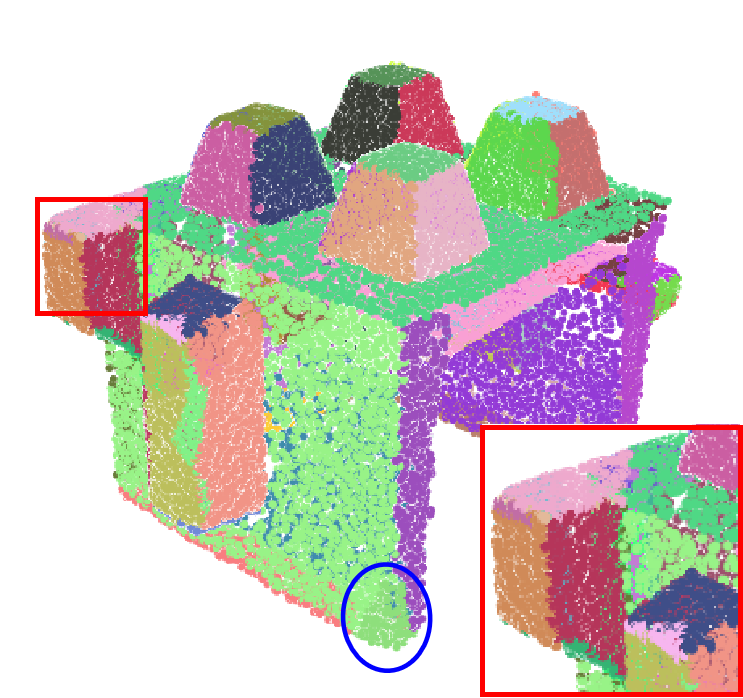}} 
		\\
		($\text{A}_{1}$) Initial mesh & ($\text{B}_{1}$) Region growing & ($\text{C}_{1}$) RANSAC & ($\text{D}_{1}$)  GoCoPP & ($\text{E}_{1}$) ParseNet 
		\\
		\multicolumn{1}{c}{\includegraphics[width= 0.18 \linewidth]{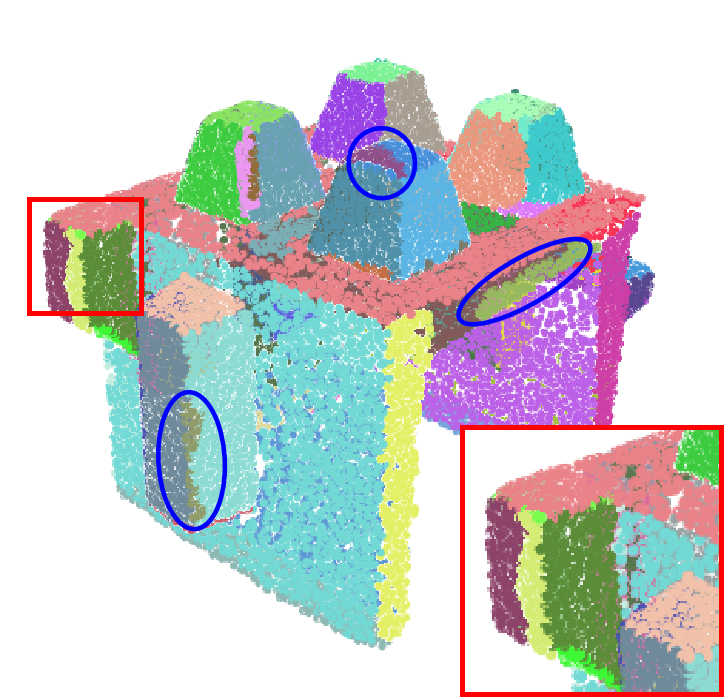}} &
		\multicolumn{1}{c}{\includegraphics[width= 0.18 \linewidth]{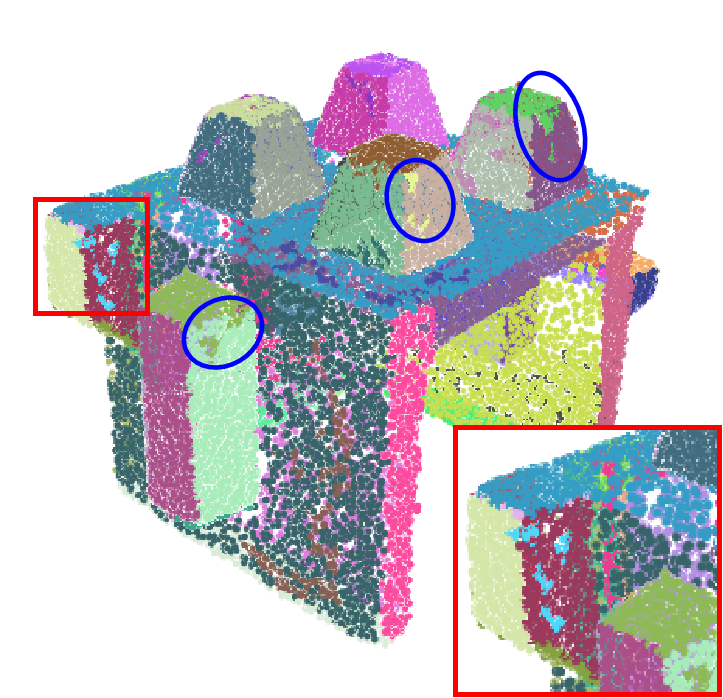}} &
		\multicolumn{1}{c}{\includegraphics[width= 0.18 \linewidth]{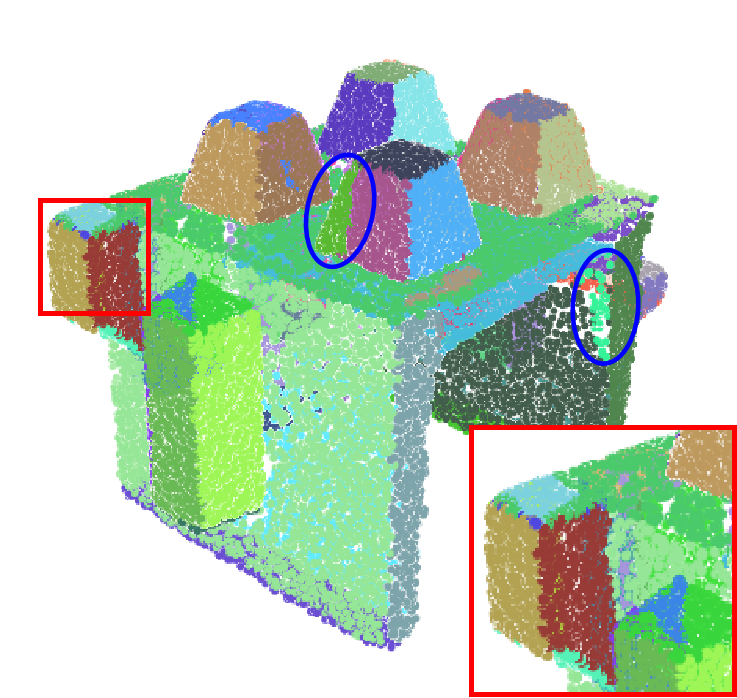}} &
		\multicolumn{1}{c}{\includegraphics[width= 0.18 \linewidth]{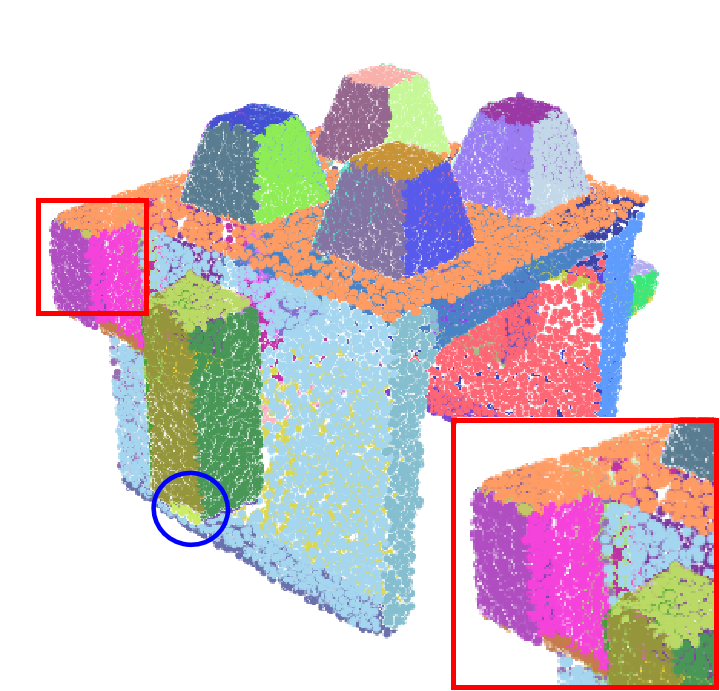}} &
		\multicolumn{1}{c}{\includegraphics[width= 0.18 \linewidth]{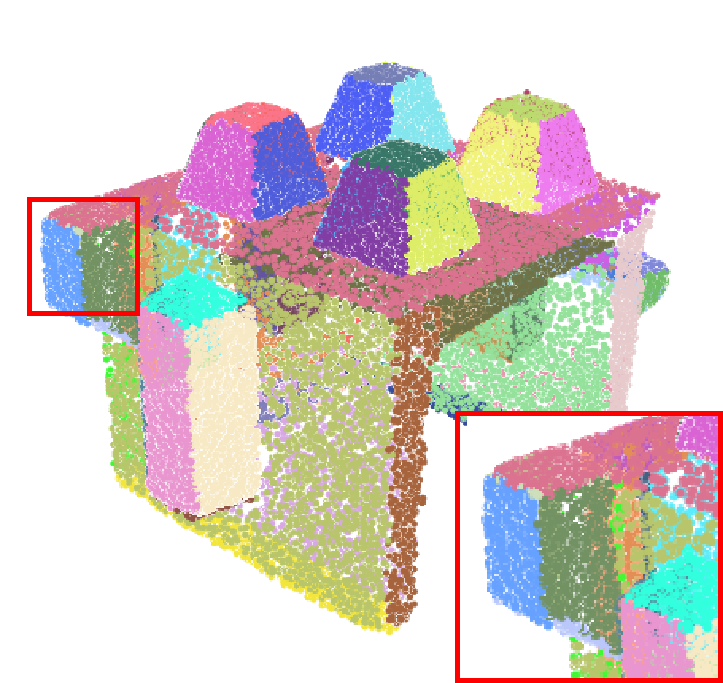}} 
		\\
		($\text{F}_{1}$) HPNet & ($\text{G}_{1}$) Mask3D & ($\text{H}_{1}$) DeepRoofPlane & ($\text{I}_{1}$) RoofSeg (ours) & ($\text{J}_{1}$) GT \\
		& & & & \\
		\hdashline 
		& & & & \\
		\multicolumn{1}{c}{\includegraphics[width= 0.18 \linewidth]{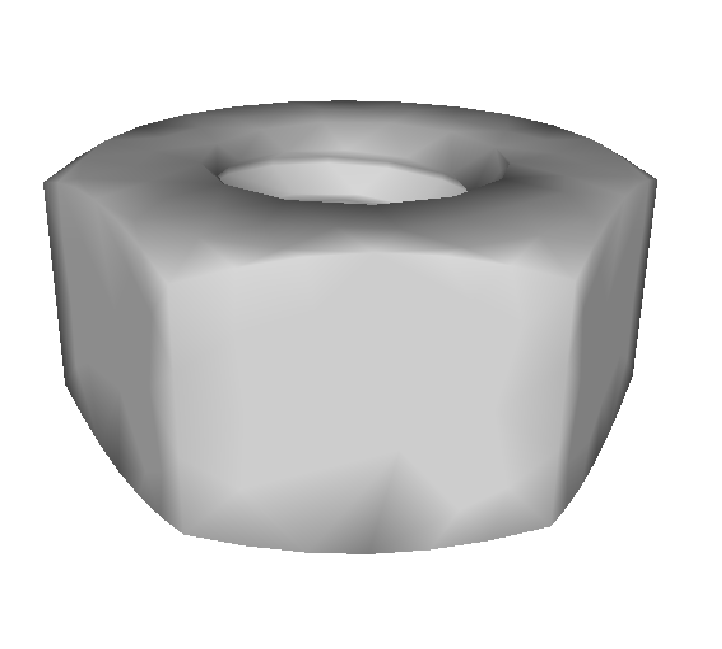}} &
		\multicolumn{1}{c}{\includegraphics[width= 0.18 \linewidth]{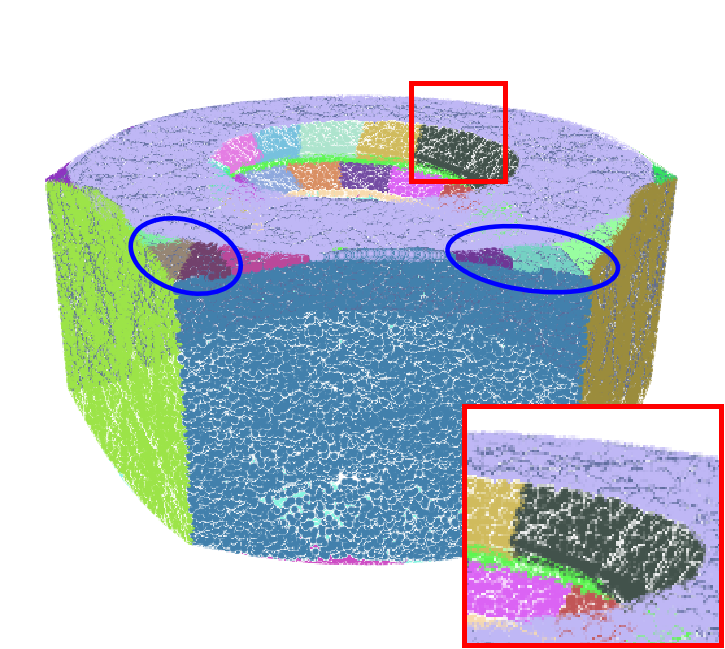}} &
		\multicolumn{1}{c}{\includegraphics[width= 0.18 \linewidth]{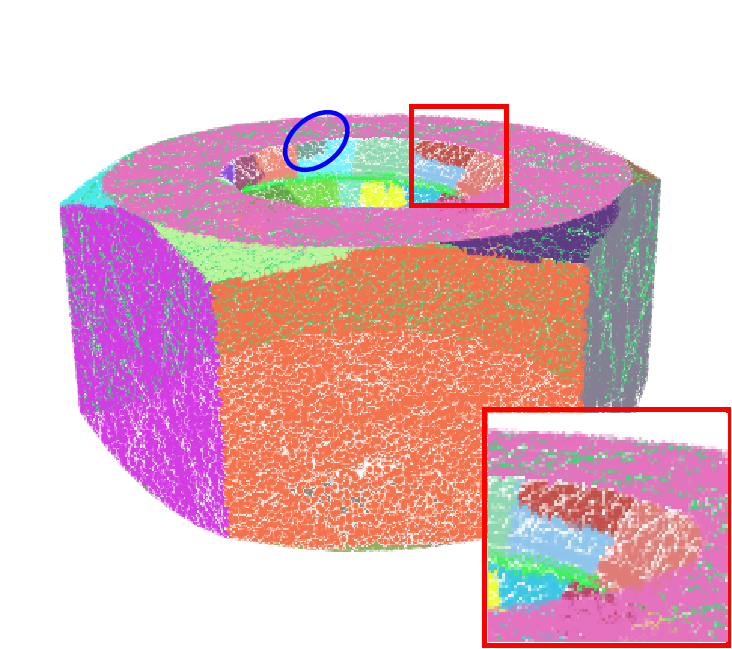}} &
		\multicolumn{1}{c}{\includegraphics[width= 0.18 \linewidth]{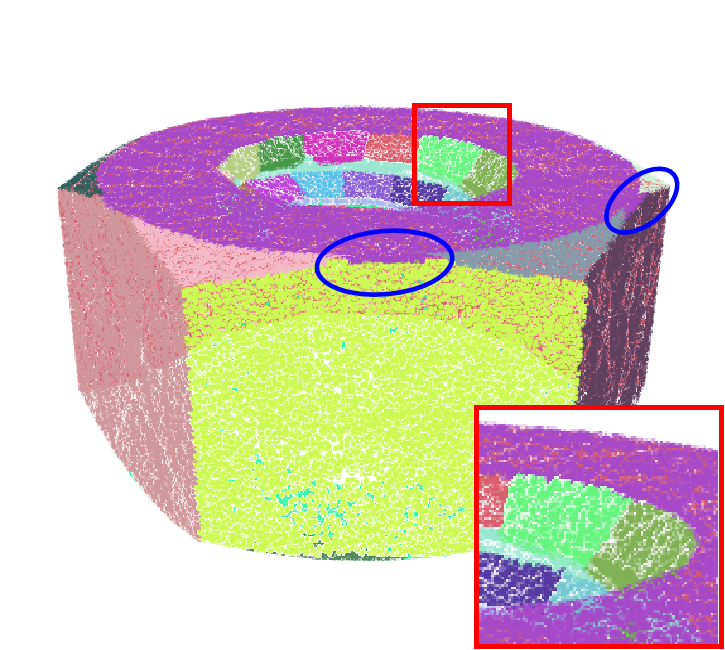}} &
		\multicolumn{1}{c}{\includegraphics[width= 0.18 \linewidth]{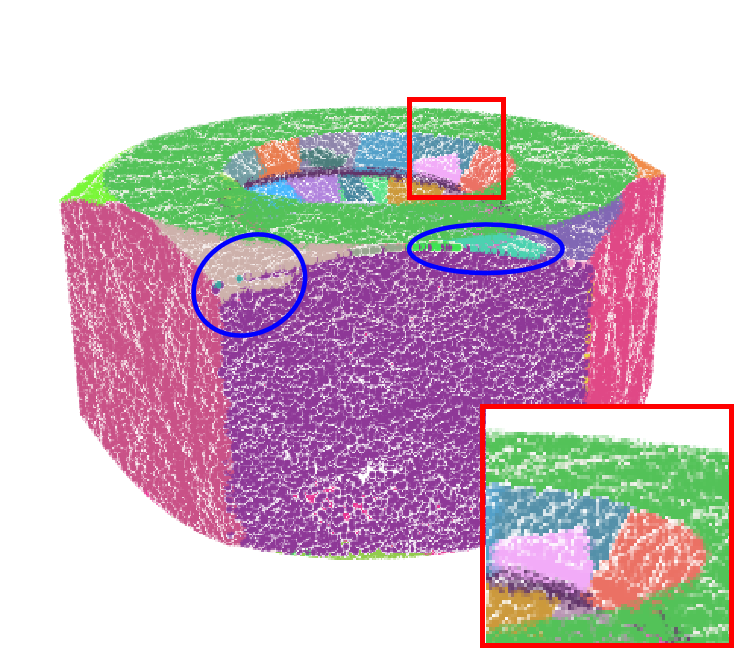}} 
		\\
		($\text{A}_{2}$) Initial mesh & ($\text{B}_{2}$) Region growing & ($\text{C}_{2}$) RANSAC & ($\text{D}_{2}$) GoCoPP & ($\text{E}_{2}$) ParseNet
		\\
		
		\multicolumn{1}{c}{\includegraphics[width= 0.18 \linewidth]{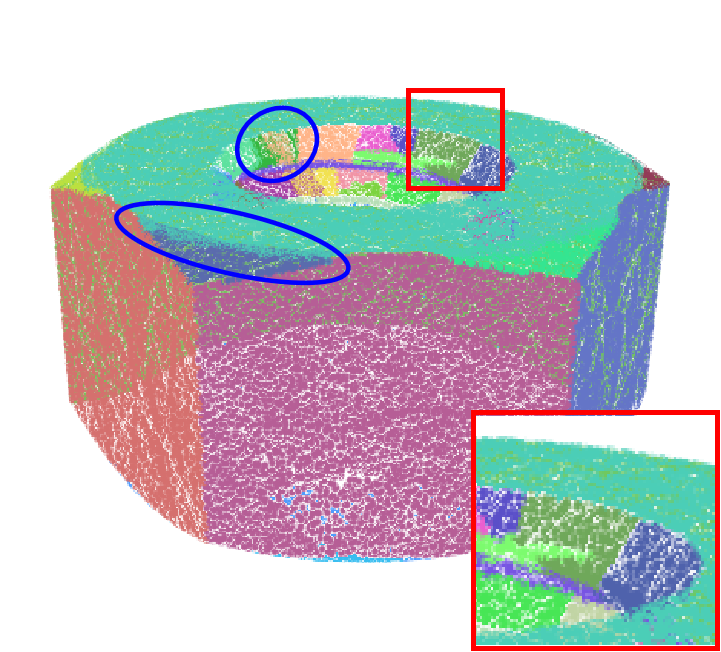}} &
		\multicolumn{1}{c}{\includegraphics[width= 0.18 \linewidth]{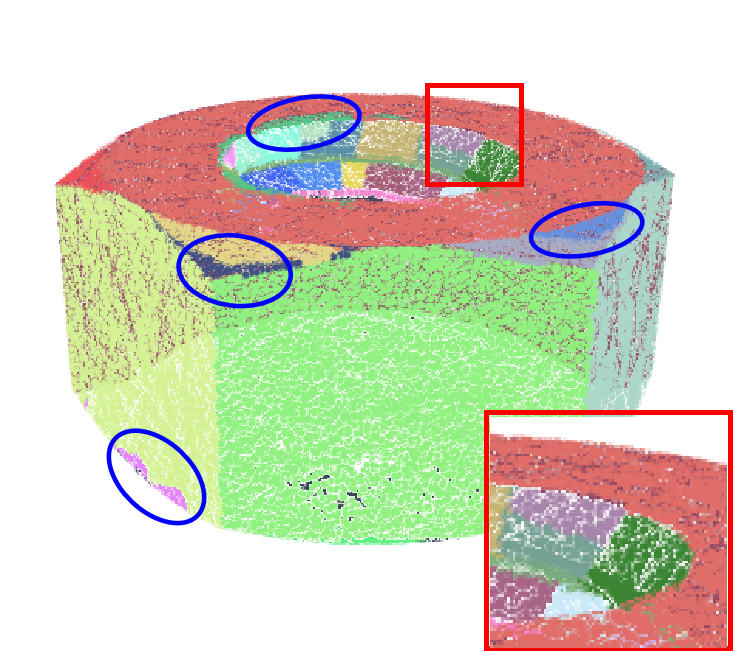}} &
		\multicolumn{1}{c}{\includegraphics[width= 0.18 \linewidth]{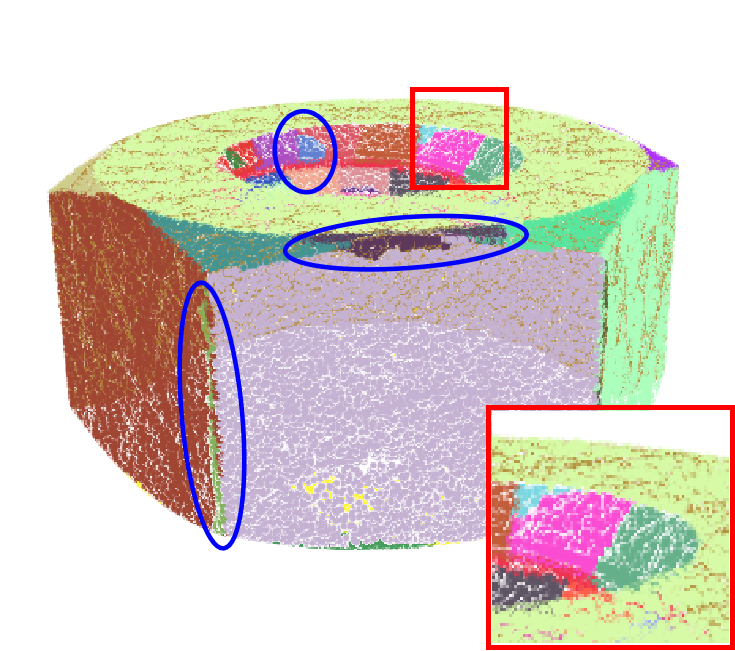}} &
		\multicolumn{1}{c}{\includegraphics[width= 0.18 \linewidth]{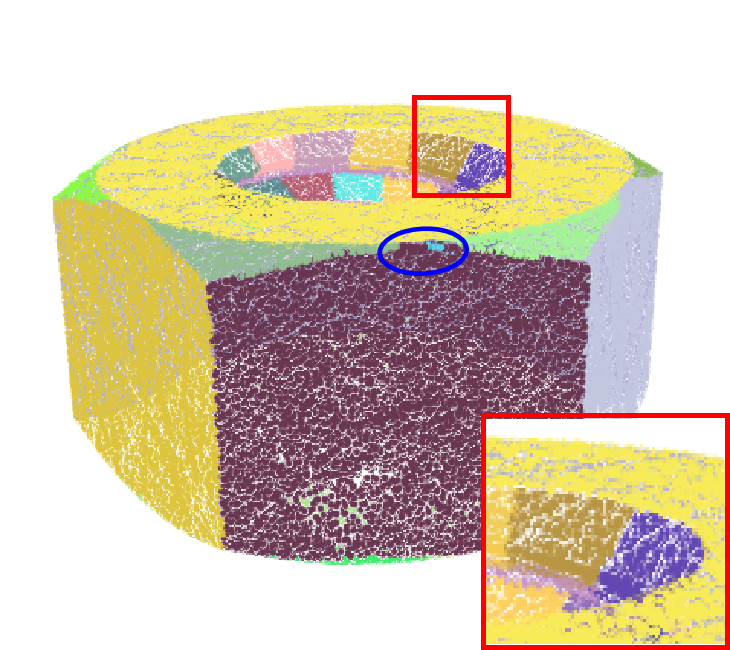}} &
		\multicolumn{1}{c}{\includegraphics[width= 0.18 \linewidth]{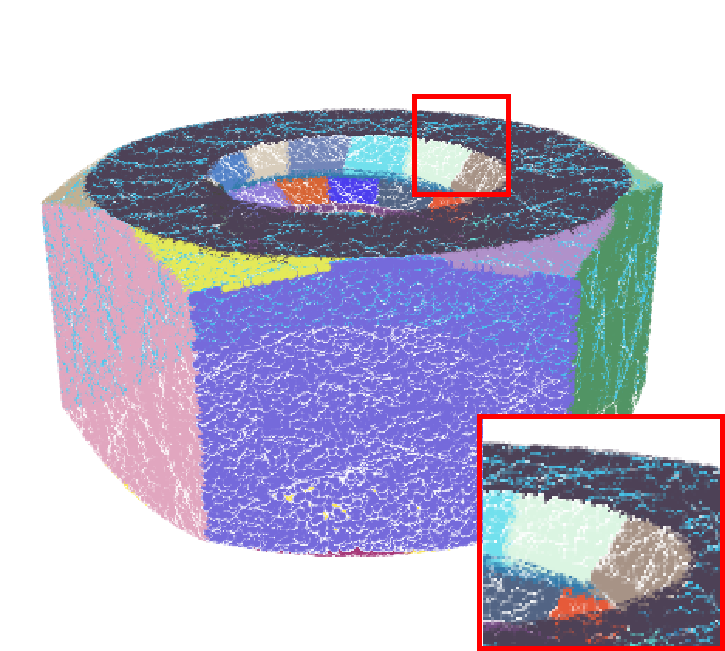}}
		\\
		($\text{F}_{2}$) HPNet & ($\text{G}_{2}$) Mask3D & ($\text{H}_{2}$) DeepRoofPlane & ($\text{I}_{2}$) RoofSeg (ours) & ($\text{J}_{2}$) GT \\ 
		
	\end{tabular} 
	\caption{Visual segmentation results of object-level point clouds from the KSR42 (rows 1-2) and ABC (rows 3-4) benchmark. The blue ellipses point out some incorrectly segmented regions.}
	\vspace{0.0em}
	\label{Fig:comparative_object_level}
\end{figure}

\textbf{Generalization capability.} To further validate the generalization ability of our network, we conducted additional evaluations on other two benchmarks: ABC~\citep{koch2019abc} and KSR42~\citep{yu2022finding}, which are widely used in shape detection and plane segmentation.The quantitative evaluation results are presented in Table~\ref{Tab:comparative_appraoches_KSR42} and Table~\ref{Tab:comparative_appraoches_ABC}.
From these two tables, it can be observed that our network outperforms the other comparative approaches for all metrics on both generic plane segmentation benchmarks. Compared with GoCoPP, which is the top‑performing approach besides ours, RoofSeg still maintains approximately 1.5 percentage points leading across all metrics. The small value of $O_{seg}$ and $U_{seg}$ demonstrate RoofSeg's robustness to achieve low over- and under segmentation, while the gained high score of $E_{acc}$ indicates its good performance in accurate edge segmentation. Moreover, as shown in Fig.~\ref{Fig:comparative_MVS_indoor} and Fig.~\ref{Fig:comparative_object_level}, despite some small errors occurred in local areas, our network also achieves the best visual segmentation results across various types of point cloud data (photogrammetric, indoor-scene and object-level), which demonstrates its good generalization capability for generic plane segmentation tasks.

\begin{table} [!htb] \renewcommand{\tabcolsep}{11.5 pt}
	\scriptsize
	\renewcommand{\arraystretch}{1.25}
	\newcommand{\tabincell}[2]{\begin{tabular}{@{}#1@{}}#2\end{tabular}}
	\begin{center}
		\caption{Quantitative assessment of all comparative approaches and RoofSeg on the KSR42 benchmark.}
		\label{Tab:comparative_appraoches_KSR42}
		\begin{tabular}{c|cccc|ccc} 
			\hline
			Different approaches 
			& mCov$\uparrow$        & mWCov$\uparrow$   &   mPrec$\uparrow$    &    mRec$\uparrow$       
			& $O_{seg}\downarrow$  & $U_{seg}\downarrow$  &  $E_{acc}\uparrow$      \\ 
			\hline
			Region growing  & 81.37  & 82.10 & 86.43 & 84.27 & 9.12 & 10.35 & 80.29  \\ 
			
			RANSAC  & 82.61  & 83.73 & 85.33 & 84.07 & 8.32 & 9.26 & 82.17  \\ 
			
			GoCoPP  & 88.45  & 89.35 & 93.74 & 92.16 & 5.27 & 6.19 & 87.04  \\ 
			
			ParseNet  & 82.53  & 83.74 & 87.24 & 85.23 & 7.32 & 8.97 & 81.56  \\ 
			
			HNet  & 83.62  & 84.20 & 86.39 & 83.13 & 7.40 & 7.01 & 82.93   \\ 
			
			Mask3D  & 85.23  & 86.45 & 90.14 & 89.34 & 5.61 & 6.97 & 84.76 \\ 
			
			DeepRoofPlane  & 87.92  & 88.45 & 92.18 & 90.32 & 5.94 & 5.87 & 86.15 \\ 
			
			RoofSeg (ours)  & \textbf{90.13}  & \textbf{91.35} & \textbf{95.74} & \textbf{92.38} & \textbf{4.18} & \textbf{5.76} & \textbf{88.76}   \\ 
			\hline
		\end{tabular}
	\end{center}
\end{table}

\begin{table} [!htb] \renewcommand{\tabcolsep}{11.5 pt}
	\scriptsize
	\renewcommand{\arraystretch}{1.25}
	\newcommand{\tabincell}[2]{\begin{tabular}{@{}#1@{}}#2\end{tabular}}
	\begin{center}
		\caption{Quantitative assessment of all comparative approaches and RoofSeg on the ABC benchmark.}
		\label{Tab:comparative_appraoches_ABC}
		\begin{tabular}{c|cccc|ccc} 
			\hline
			Different approaches 
			& mCov$\uparrow$       & mWCov$\uparrow$   &   mPrec$\uparrow$    &    mRec$\uparrow$       
			& $O_{seg}\downarrow$  & $U_{seg}\downarrow$  &  $E_{acc}\uparrow$      \\ 
			\hline
			Region growing  & 83.37  & 84.52 & 88.75 & 82.60 & 8.26 & 9.05 & 82.73  \\ 
			
			RANSAC  & 85.42  & 86.79 & 89.18 & 84.27 & 8.43 & 7.96 & 84.53  \\ 
			
			GoCoPP  & 89.86  & 90.72 & 93.15 & 88.42 & 5.33 & 4.97 & 89.24  \\ 
			
			ParseNet  & 86.75  & 87.42 & 89.52 & 86.47 & 7.28 & 6.41 & 85.64  \\ 
			
			HPNet  & 87.14  & 88.03 & 90.26 & 87.60 & 4.87 & 6.63 & 86.65   \\ 
			
			Mask3D  & 88.75  & 89.34 & 91.65 & 87.40 & 6.31 & 5.84 & 87.02 \\ 
			
			DeepRoofPlane  & 89.57  & 90.32 & 92.63 & 89.56 & 5.79 & 6.32 & 88.61 \\ 
			
			RoofSeg (ours)  & \textbf{91.34}  & \textbf{92.18} & \textbf{95.24} & \textbf{91.76} & \textbf{4.71} & \textbf{3.85} & \textbf{90.57}   \\ 
			\hline
		\end{tabular}
	\end{center}
\end{table}

In addition, some supplementary experimental results are provided in the Appendix, including cross-dataset evaluation (Appendix~\ref{appendix:cross-dataset_evaluation}), robustness analysis on challenging scenarios (Appendix~\ref{appendix:robustness_analysis}), efficiency and accuracy analysis (Appendix~\ref{appendix:efficiency_and_accuracy_analysis}), feature-space visualization and analysis to validate the effectiveness of edge-aware feature learning (Appendix~\ref{appendix:feature_space_analysis}), and evaluation of the effects on downstream building reconstruction tasks (Appendix~\ref{appendix:impact_on_building_reconstruction_tasks}).

\subsection{Ablations}
\textbf{The effects of different components.} To systematically evaluate the effectiveness of the Edge-Aware Mask Module (EAMM) and the proposed loss function with two key components ($\mathcal{L}_{mask}^{w}$ and $\mathcal{L}_{plane}$), we conducted the ablation experiments with 8 control groups (a-h). In these groups, (a-d) and (e-h) are employed to evaluate the network's performance without and with the EAMM, respectively. To study the impact of different loss components, we set up four loss controls for comparison, including two baselines (conventional mask loss $\mathcal{L}_{mask}$ and adaptive weighting mask loss $\mathcal{L}_{mask}^{w}$), along with their variants added with the plane geometric loss $\mathcal{L}_{plane}$. The quantitative evaluation results are presented in the Table~\ref{Tab:ablation_components}. From the table, it can be concluded that EAMM and the proposed loss function (with $\mathcal{L}_{mask}^{w}$ and $\mathcal{L}_{plane}$) can both effectively improve the segmentation accuracy. In terms of the Cov metric, using EAMM leads to an improvement of about 2 percentage points on the RoofNTNU and Roofpc3D benchmarks, while the improvement exceeds 3 percentage points on the Building3D benchmark. Besides, compared to the baseline loss that relies solely on $\mathcal{L}_{mask}$, using the proposed loss function can achieve an improvement of nearly 1.5 percentage points across the three benchmarks. To qualitatively demonstrate the impact of three components (EAMM, $\mathcal{L}_{mask}^{w}$ and $ \mathcal{L}_{plane}$), we presented the visual segmentation results of two roof samples (from RoofNTNU and Building3D benchmarks, respectively) with different components in Fig.~\ref{Fig:ablation}. Without EAMM, segmentation accuracy degrades significantly at edge regions. Without $\mathcal{L}_{mask}^{w}$, the segmentation results may produce a large number of misclassified points (outliers) across different patches. Without $\mathcal{L}_{plane}$, misclassified points still occur and under-segmentation could happen, where small patches are incorrectly merged with larger ones, resulting in plane segments with low geometric fidelity, as highlighted by the red wireframes in Fig.~\ref{Fig:ablation}.

\begin{table} [!htb] \renewcommand{\tabcolsep}{1.0 pt}
	\scriptsize
	\renewcommand{\arraystretch}{1.25}
	\newcommand{\tabincell}[2]{\begin{tabular}{@{}#1@{}}#2\end{tabular}}
	\begin{center}
		\caption{Quantitative assessment of RoofSeg using different components on three benchmarks. EAMM denotes Edge-Aware Mask Module, while $\mathcal{L}_{mask}$, $\mathcal{L}_{mask}^{w}$ and $\mathcal{L}_{plane}$ represent conventional mask loss, adaptive weighting mask loss and plane geometric loss. The values are presented as percentage differences from ground truth for clearer impact assessment.}
		\label{Tab:ablation_components}
		\begin{tabular}{cc|ccc|cccc|cccc|cccc} 
			\cline{2-5} \cline{6-9} \cline{10-13} \cline{14-17}
			&\multicolumn{4}{c|}{Different components} & \multicolumn{4}{c}{RoofNTNU benchmark}  &  \multicolumn{4}{c}{Roofpc3D benchmark} & \multicolumn{4}{c}{Building3D benchmark}\\ 
			\cline{2-5} \cline{6-9} \cline{10-13} \cline{14-17}
			&EAMM        & $\mathcal{L}_{mask}$    & $\mathcal{L}_{mask}^{w}$  &$\mathcal{L}_{plane}$ 
			& mCov      & mWCov  &   mPrec  &    mRec      
			& mCov      & mWCov  &   mPrec  &    mRec
			& mCov       & mWCov  &   mPrec  &    mRec    \\
			\cline{2-5} \cline{6-9} \cline{10-13} \cline{14-17}
			(a) & $\times$ & \checkmark & $\times$  & $\times$  & 6.29  & 5.15 & 4.27 & 5.48 & 5.86 & 4.62 & 3.87 & 5.26 & 9.93 & 6.69 & 5.80 & 6.86 \\ 
			(b) & $\times$ & $\times$ & \checkmark & $\times$ & 5.83  & 4.79 & 3.65 & 4.92 & 5.60 & 4.55 &2.98 & 4.84 & 9.12 & 6.17 & 5.16 & 5.32  \\ 
			(c) & $\times$ & \checkmark & $\times$ & \checkmark & 5.95  & 4.94 & 3.13 & 4.71 & 5.22 & 4.42 & 3.03 & 4.37 & 8.74 & 5.83 & 4.57 & 6.26    \\ 
			(d) & $\times$ & $\times$ & \checkmark & \checkmark & 5.52  & 4.57 & 2.49 & 3.72 & 4.87 & 3.79 & 2.71 & 3.75 & 8.25 & 5.54 & 4.83 & 5.89    \\ 
			\cline{2-5} \cline{6-9} \cline{10-13} \cline{14-17}
			(e) & \checkmark  & \checkmark & $\times$ & $\times$  & 5.28  & 4.54 & 1.54 & 2.81 & 4.92 & 3.23 & 1.50 & 1.97 & 7.92 & 4.91 & 4.11 & 5.15   \\ 
			(f) & \checkmark  & $\times$ & \checkmark & $\times$ & 4.88 & 3.65 & 1.35 & 2.97 & 4.32 & 2.99 & 0.85 & 2.11 & 7.33 & 4.08 & 3.52 & 4.61   \\ 
			(g) & \checkmark & \checkmark & $\times$ & \checkmark & 5.04  & 4.72 & 0.93 & 2.37 & 4.25 & 2.86 & 0.21 & 1.69 & 6.86 & 4.31 & 2.05 & 3.83     \\ 
			(h) & \checkmark & $\times$ & \checkmark & \checkmark & \textbf{4.11}  & \textbf{3.18} & \textbf{0.40} & \textbf{1.82} & \textbf{3.99} & \textbf{2.44} & \textbf{0.02} & \textbf{1.46} & \textbf{6.26} & \textbf{3.63} & \textbf{1.21} & \textbf{3.42}     \\ 
			\cline{2-5} \cline{6-9} \cline{10-13} \cline{14-17}

		\end{tabular}
	\end{center}
\end{table}

\begin{figure}[!htb] \renewcommand{\tabcolsep}{1.8 pt}
	\scriptsize
	\centering
	\begin{tabular}{ccccc} 	
		& Top View & Side View & Top View & Side View \\
		\raisebox{5.5\height}{w/o EAMM}  &
		\includegraphics[width= 0.21 \linewidth]{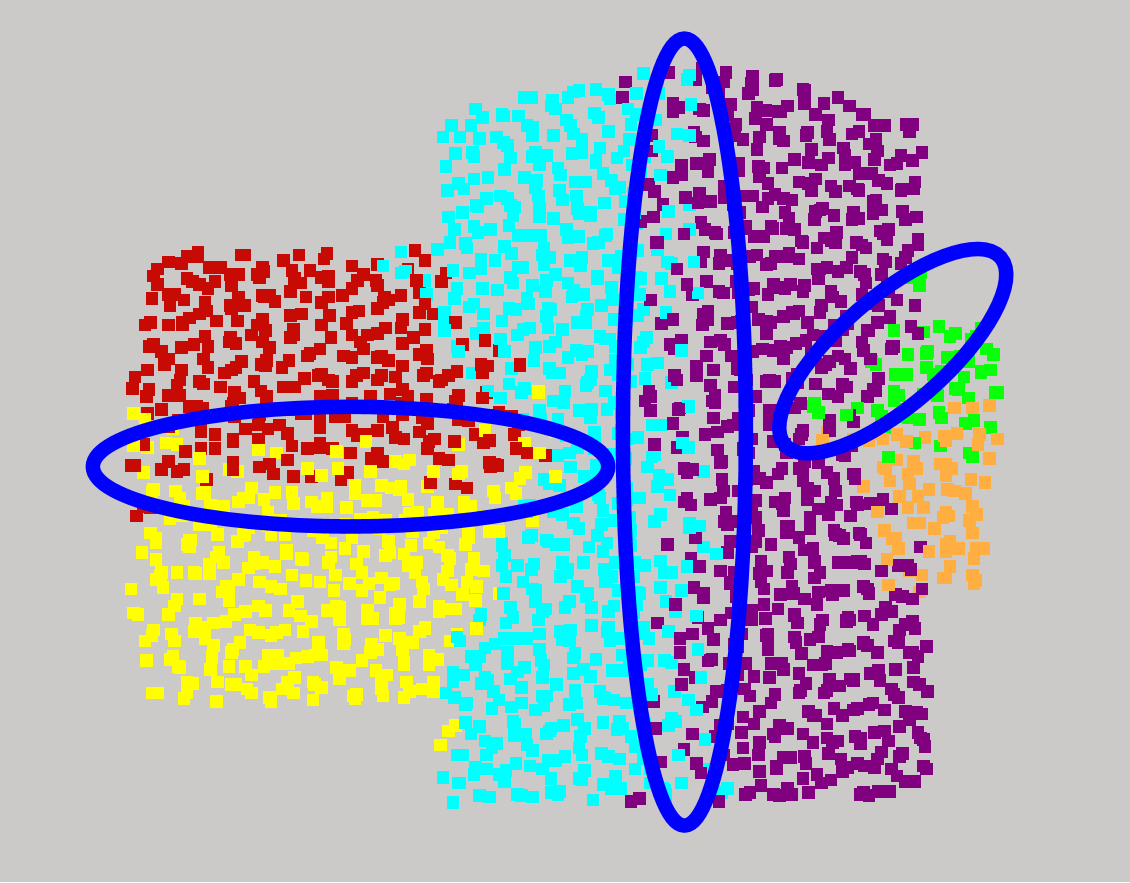} &
		\includegraphics[width= 0.21 \linewidth]{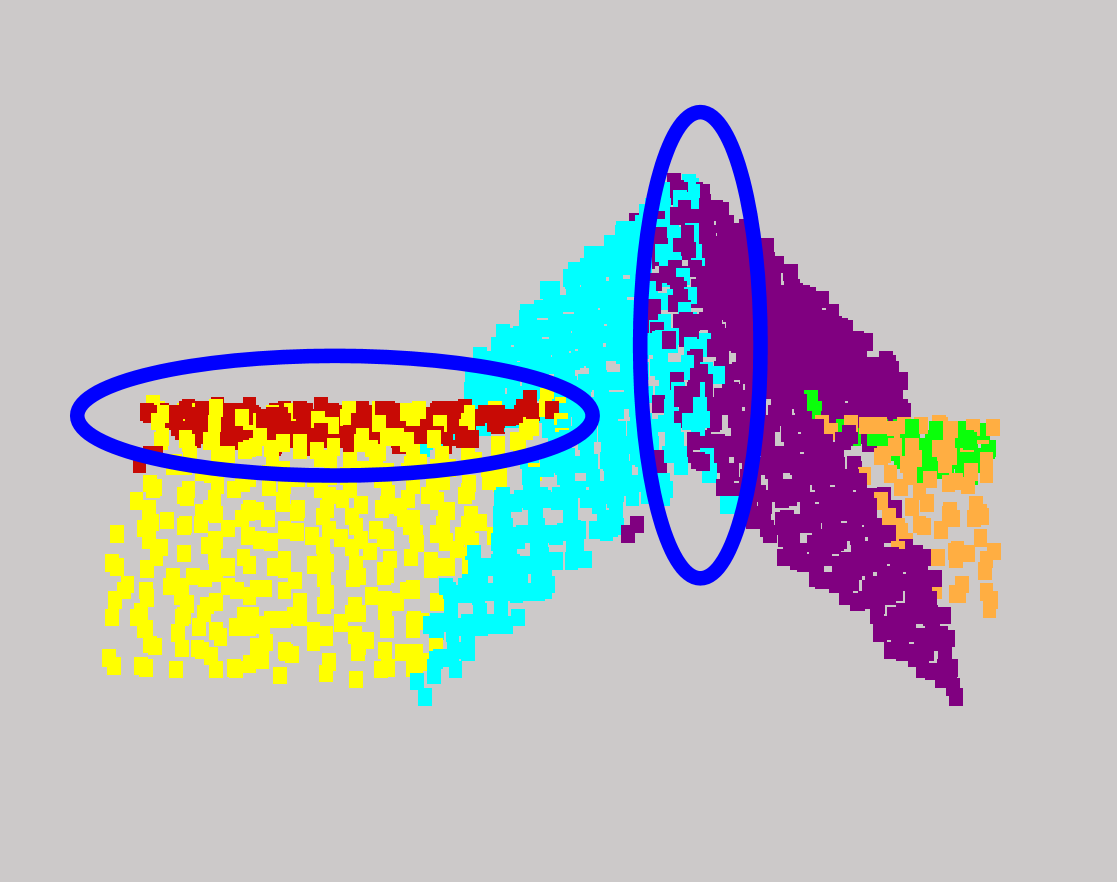} &
		\includegraphics[width= 0.21 \linewidth]{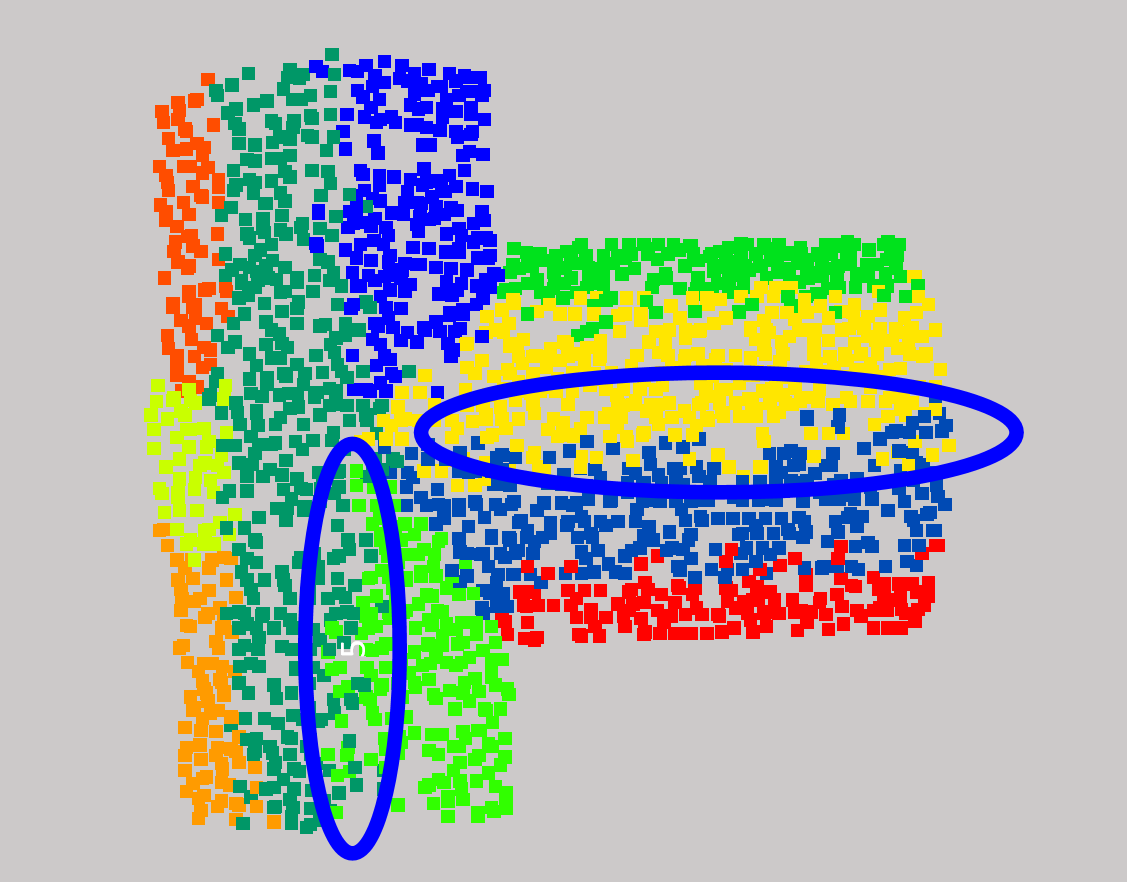} &
		\includegraphics[width= 0.21 \linewidth]{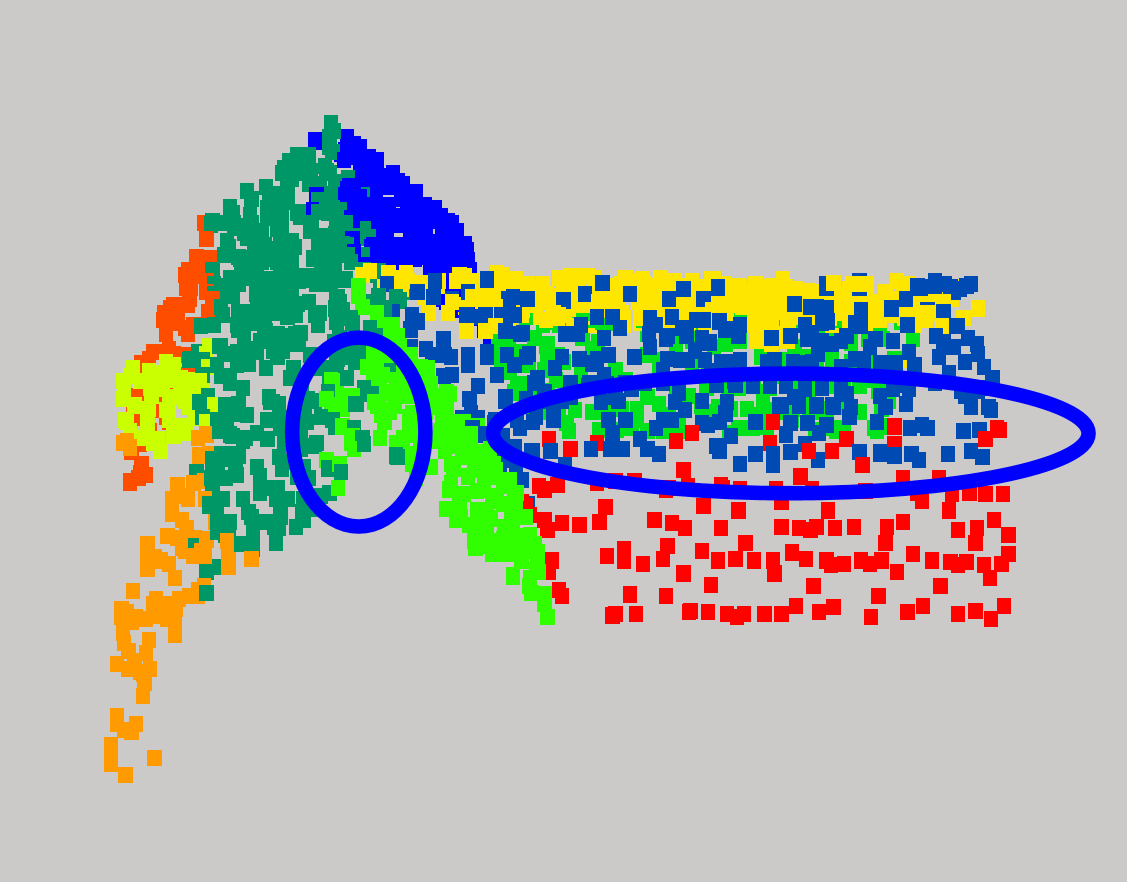}
		\\
		\raisebox{5.5\height}{w/o $\mathcal{L}_{mask}^{w}$} &
		\includegraphics[width= 0.21 \linewidth]{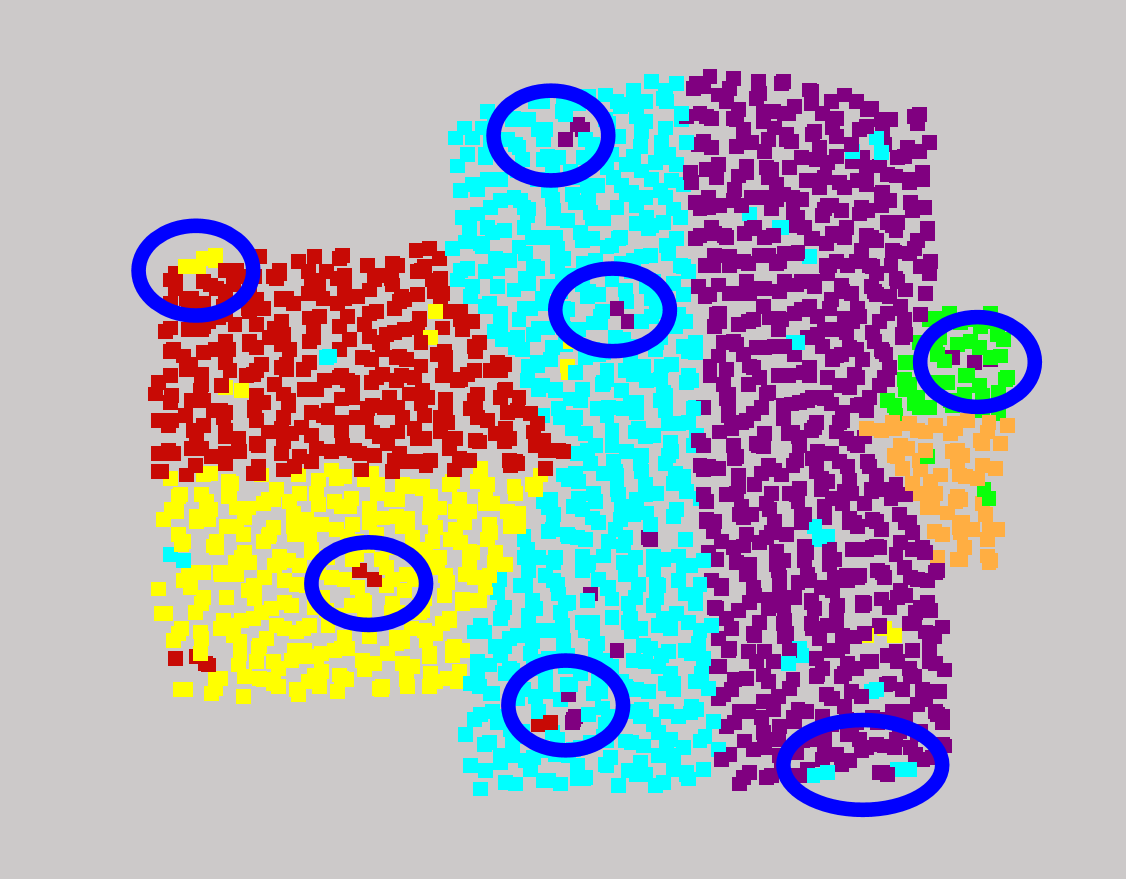} &
		\includegraphics[width= 0.21 \linewidth]{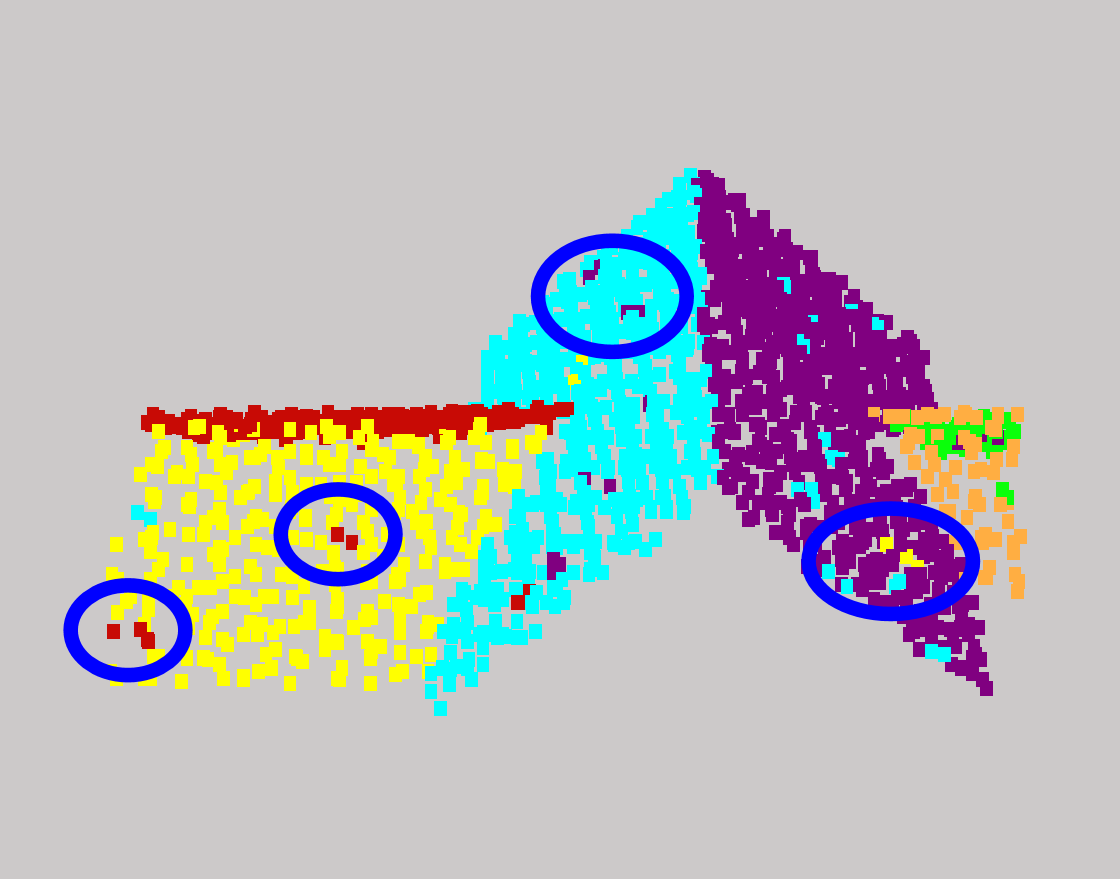} &
		\includegraphics[width= 0.21 \linewidth]{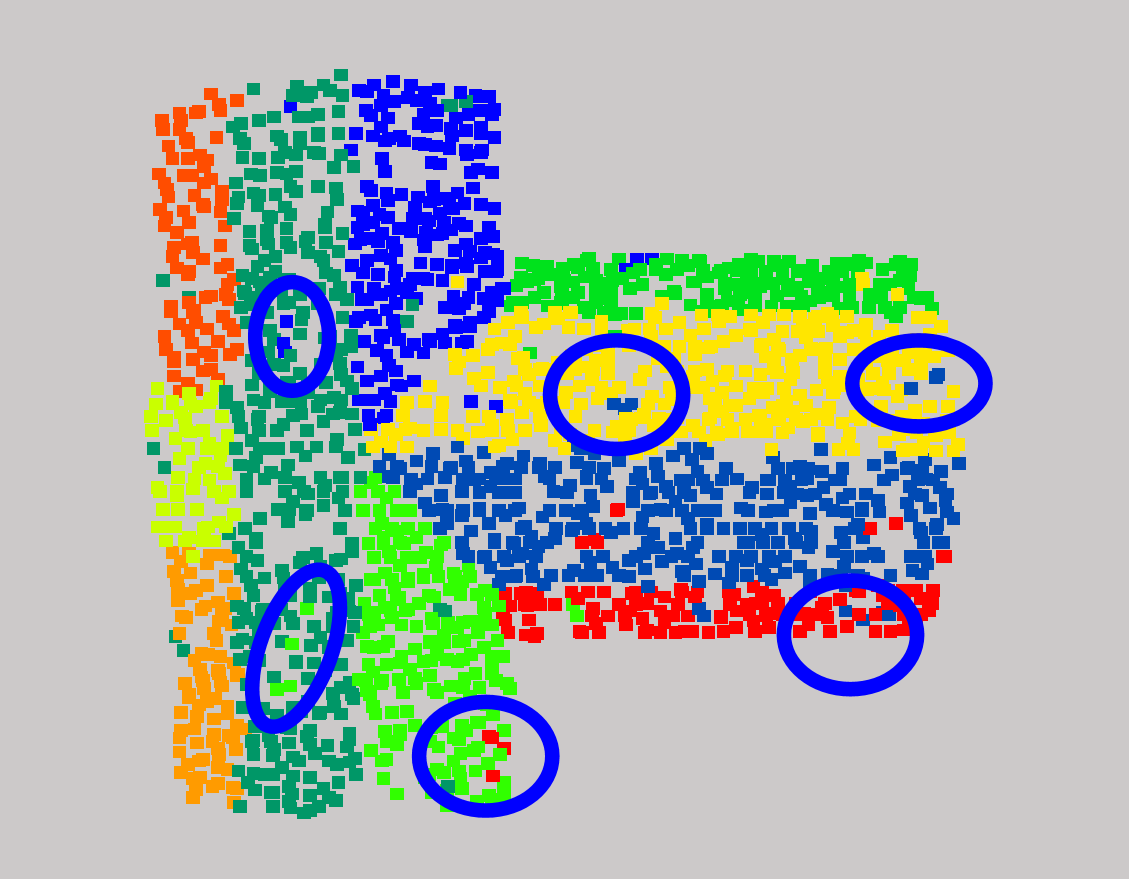} &
		\includegraphics[width= 0.21 \linewidth]{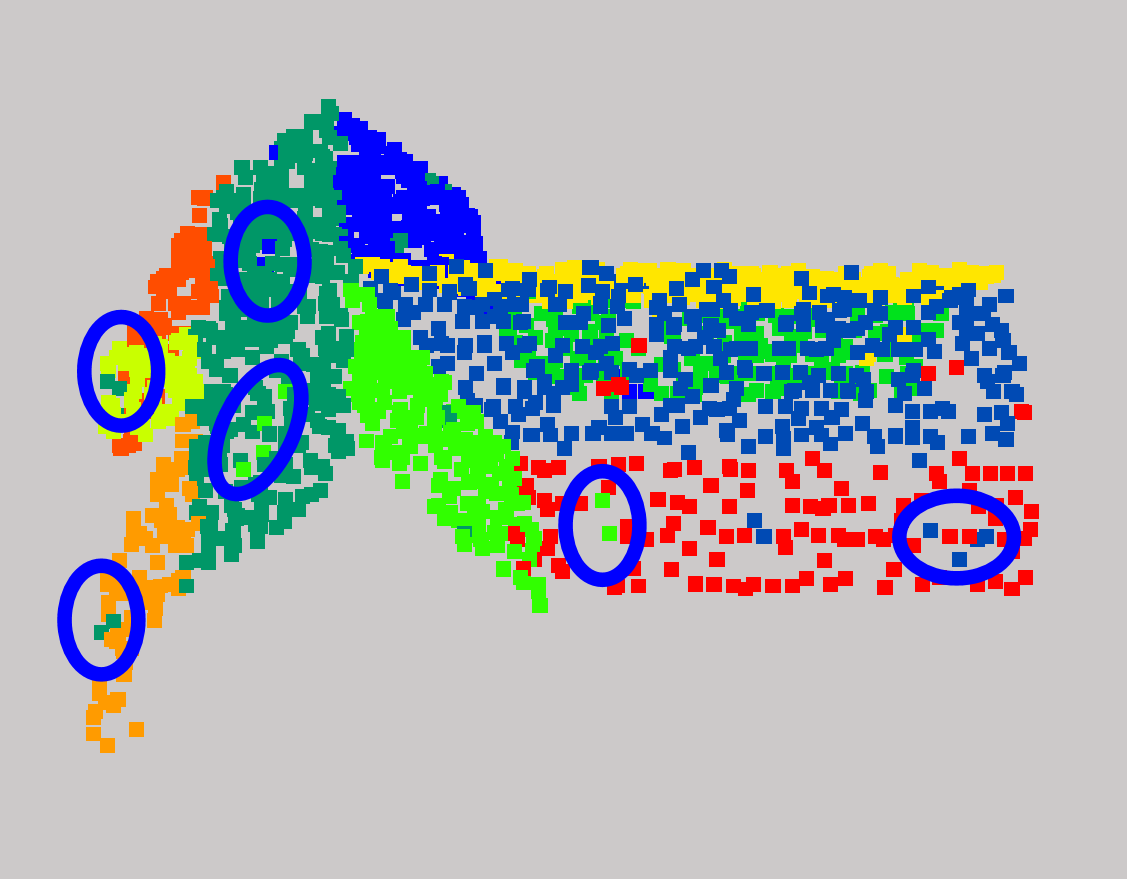} 
		\\
		\raisebox{5.5\height}{w/o $\mathcal{L}_{plane}$} &
		\includegraphics[width= 0.21 \linewidth]{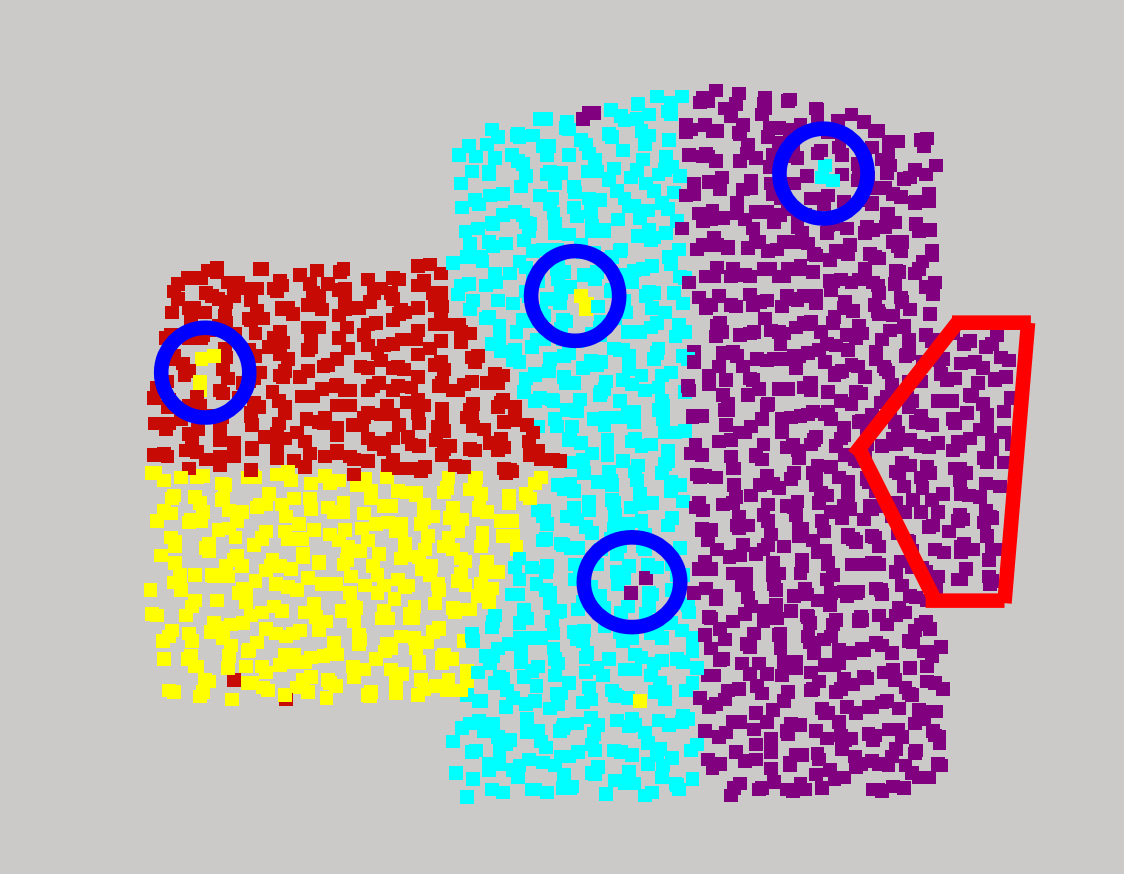} &
		\includegraphics[width= 0.21 \linewidth]{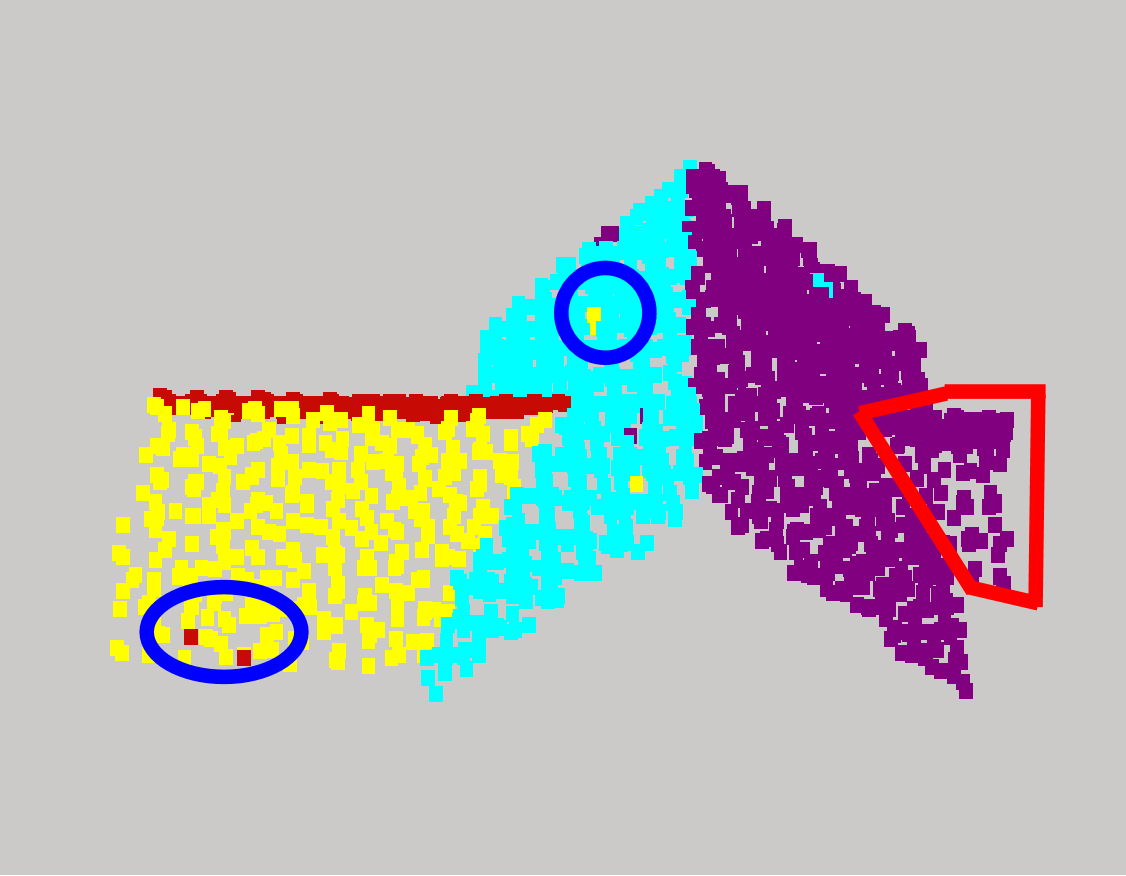} &
		\includegraphics[width= 0.21 \linewidth]{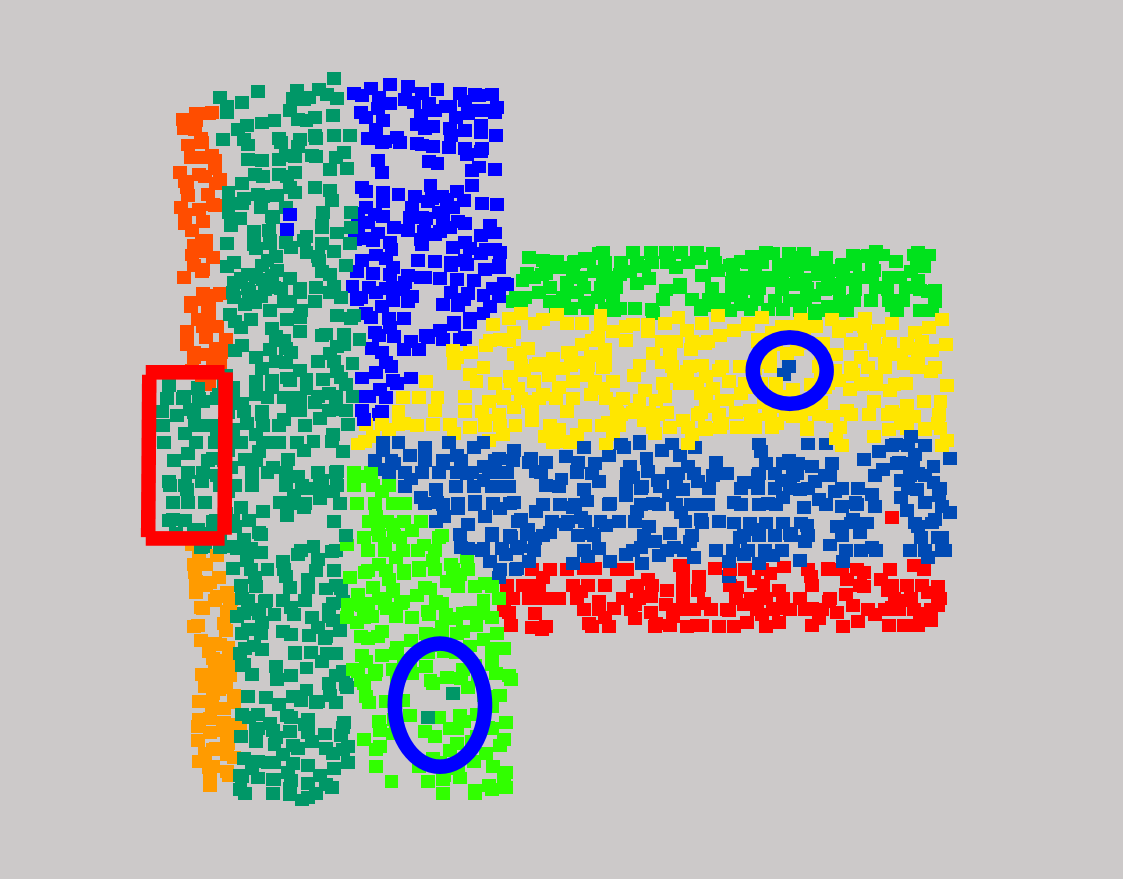} &
		\includegraphics[width= 0.21 \linewidth]{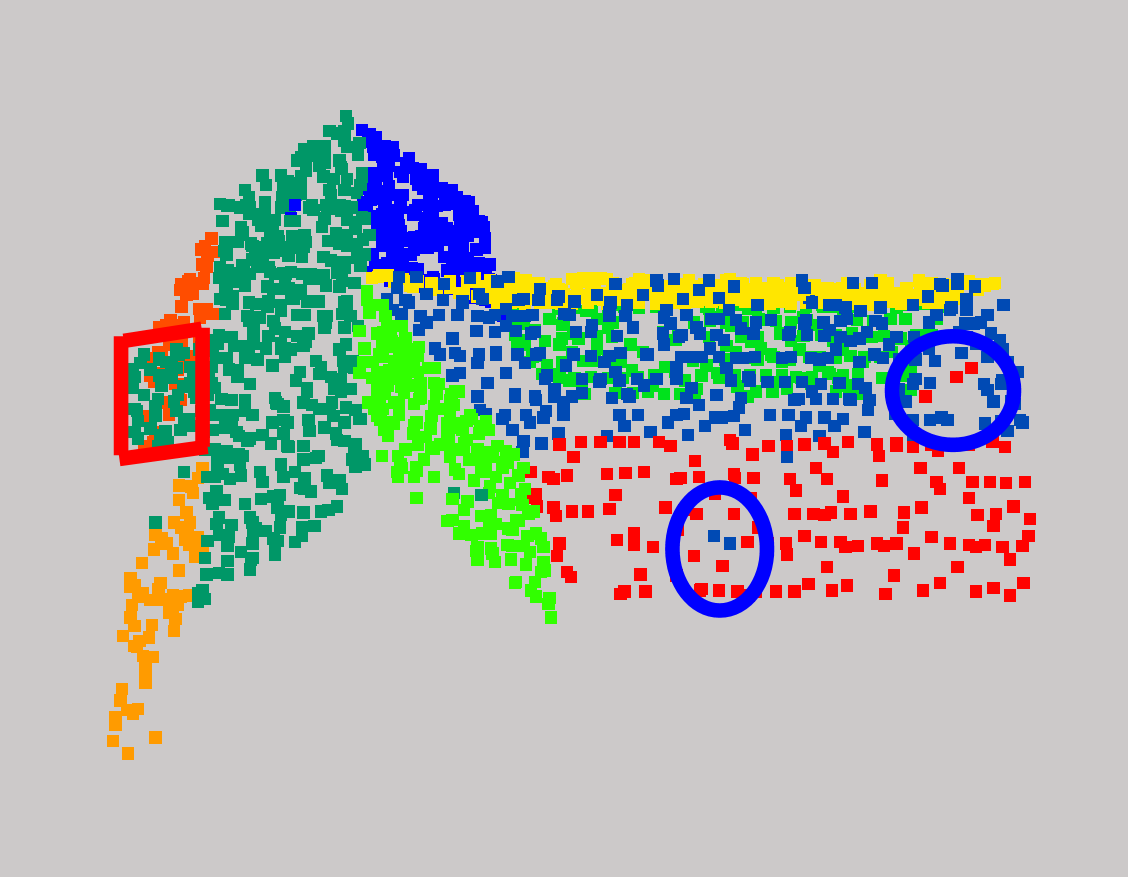}
		\\
		\raisebox{5.5\height}{RoofSeg} &
		\includegraphics[width= 0.21 \linewidth]{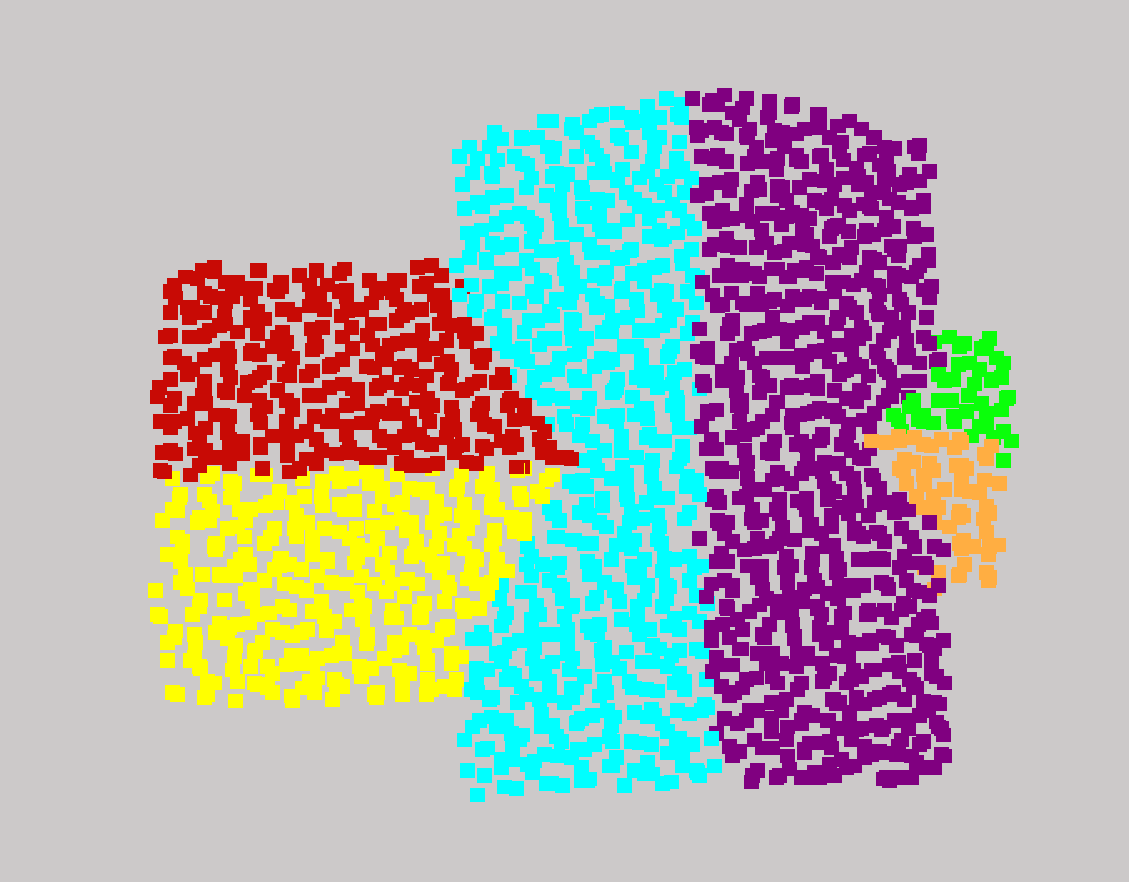} &
		\includegraphics[width= 0.21 \linewidth]{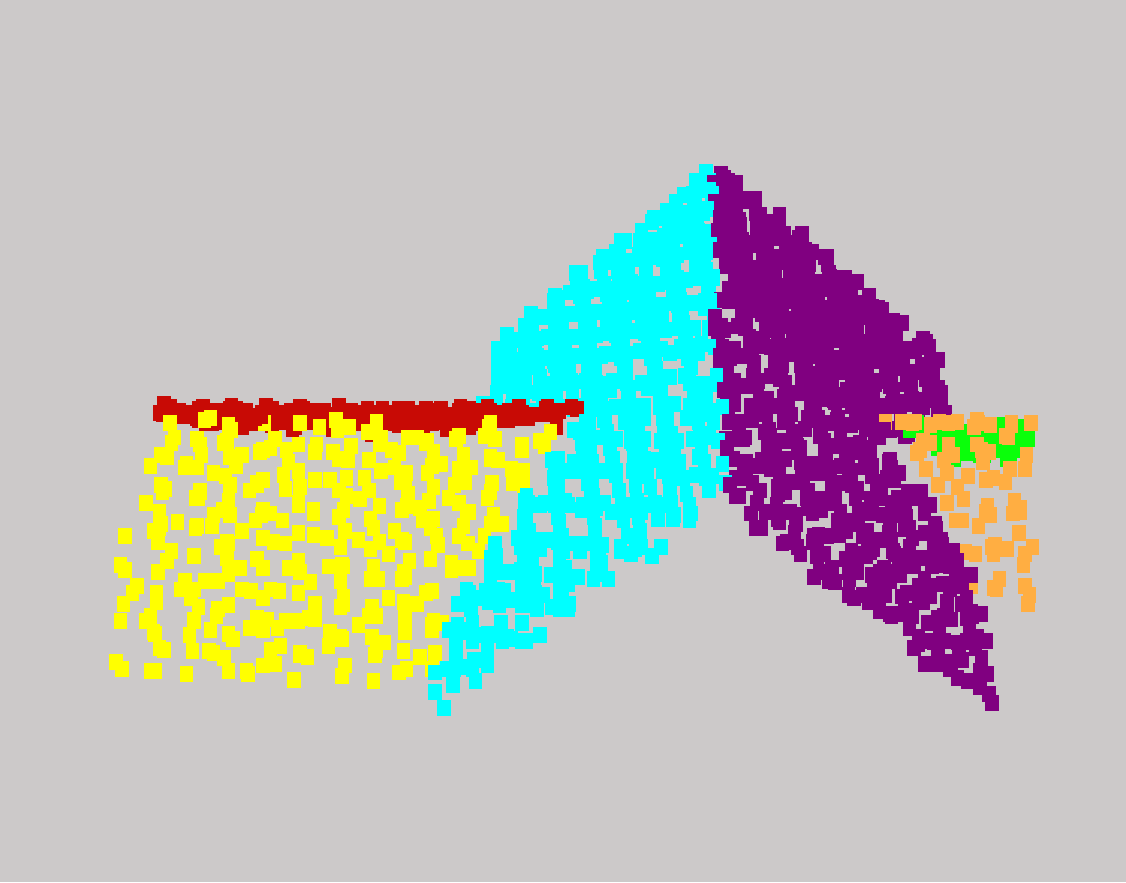} &
		\includegraphics[width= 0.21 \linewidth]{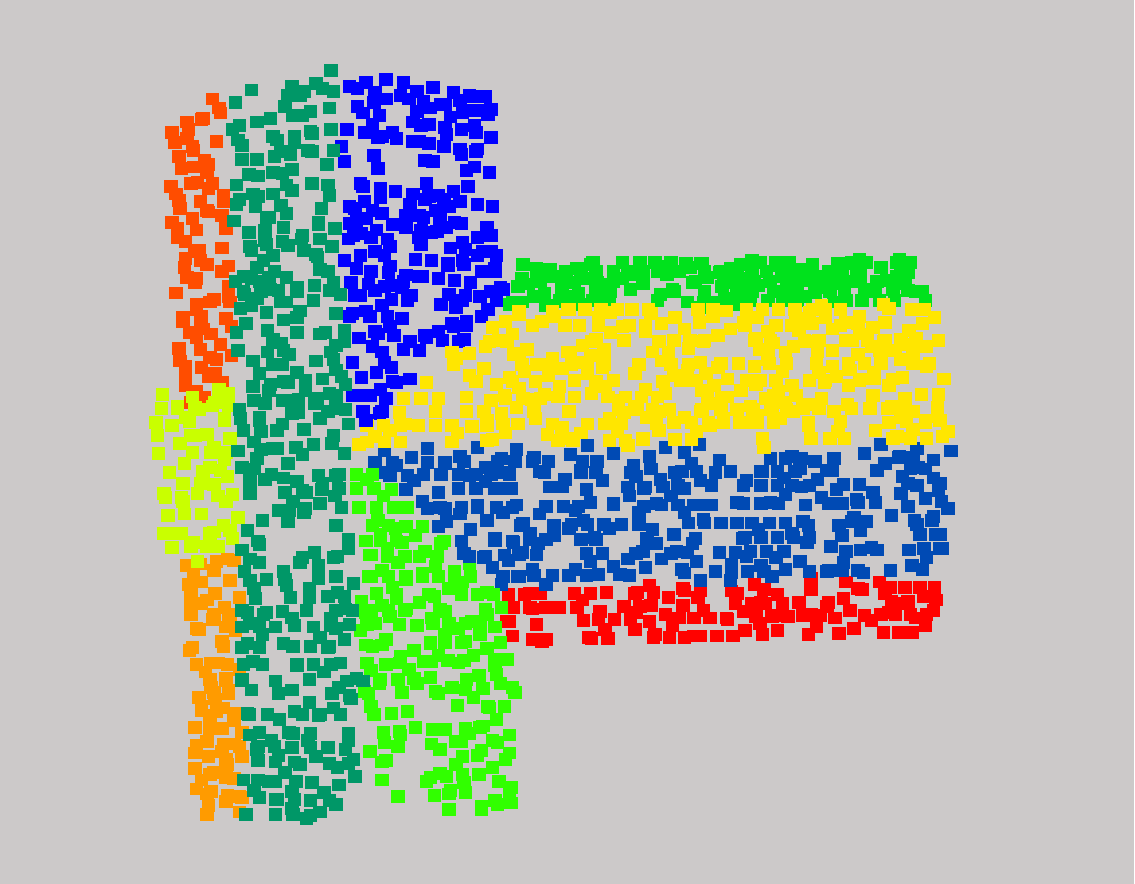} &
		\includegraphics[width= 0.21 \linewidth]{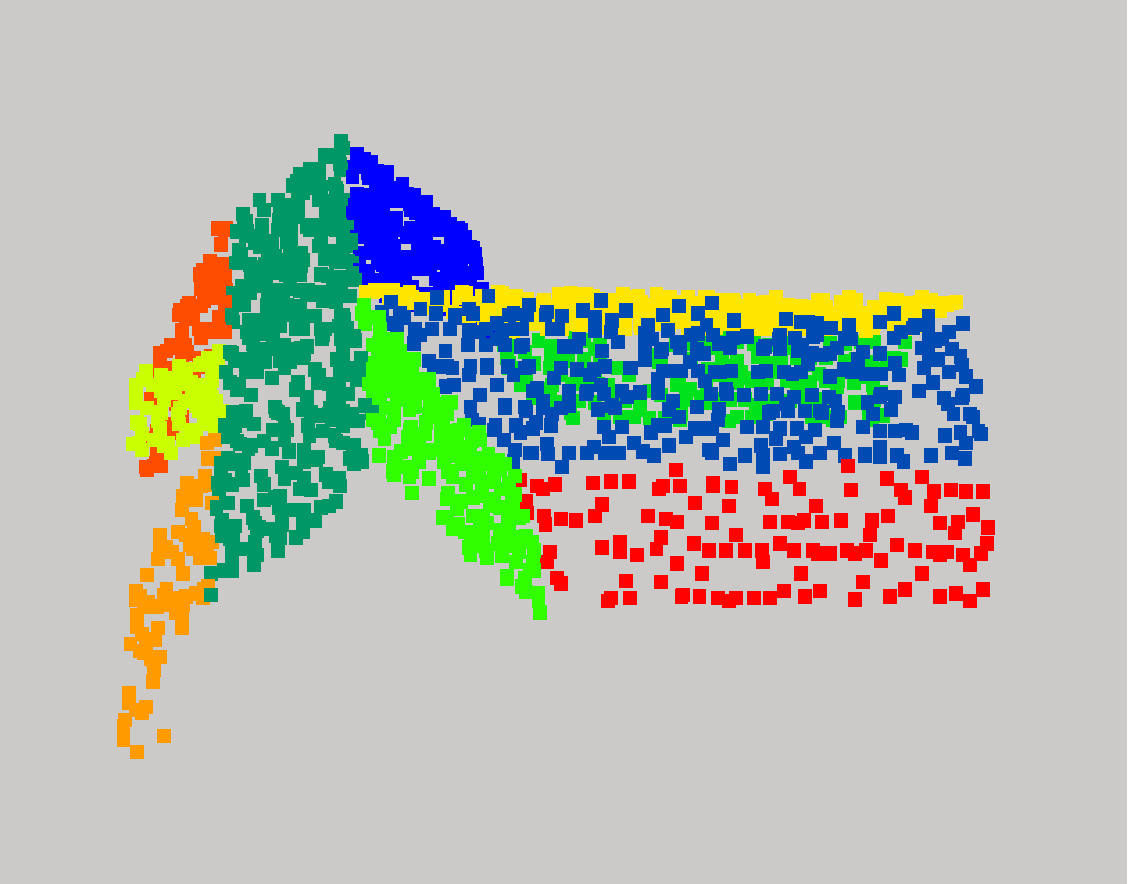}  
		\\
		\raisebox{5.5\height}{Ground Truth} &
		\includegraphics[width= 0.21 \linewidth]{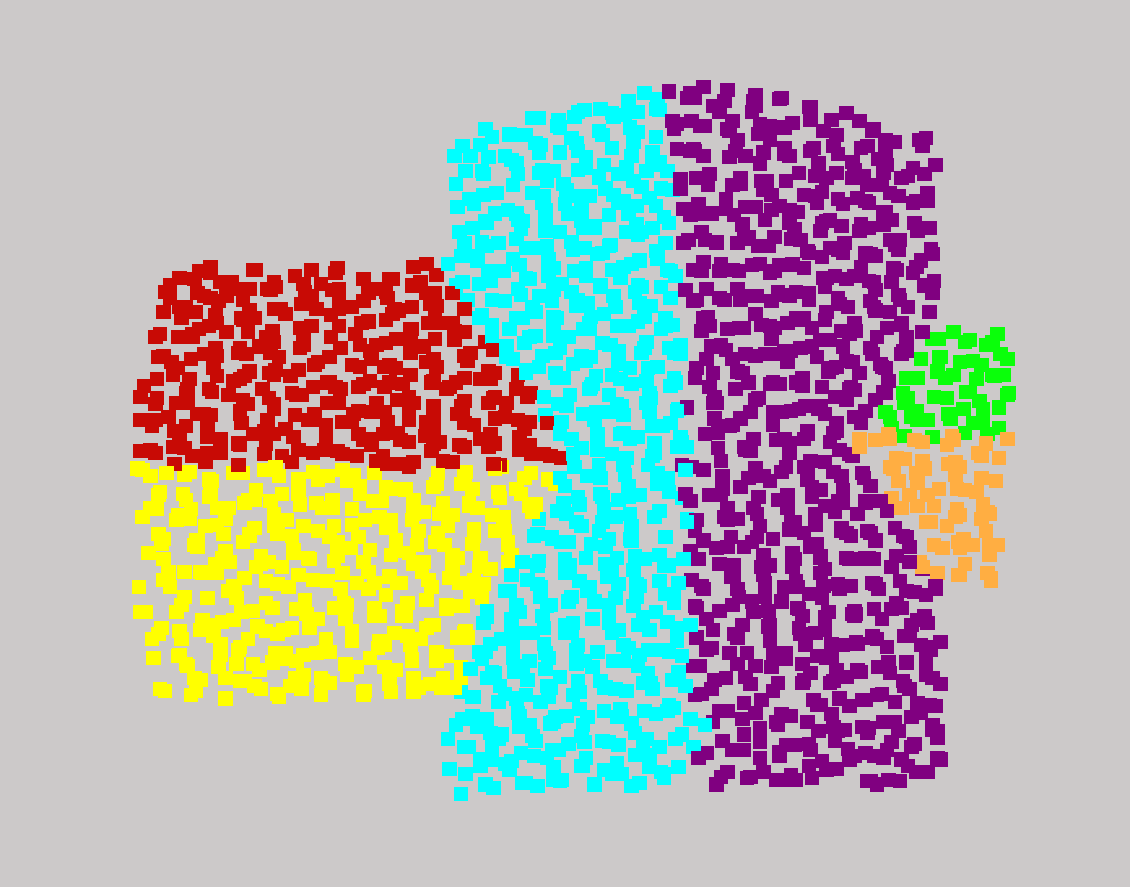} &
		\includegraphics[width= 0.21 \linewidth]{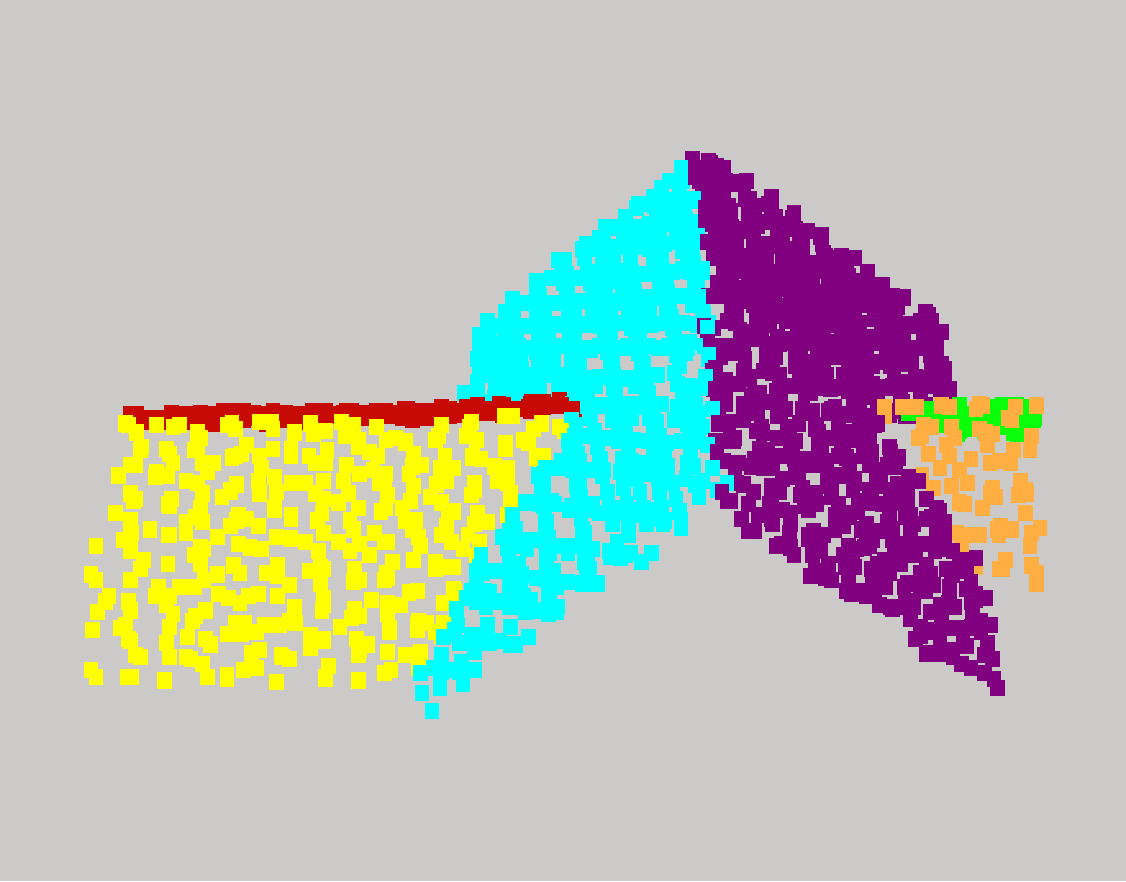} &
		\includegraphics[width= 0.21 \linewidth]{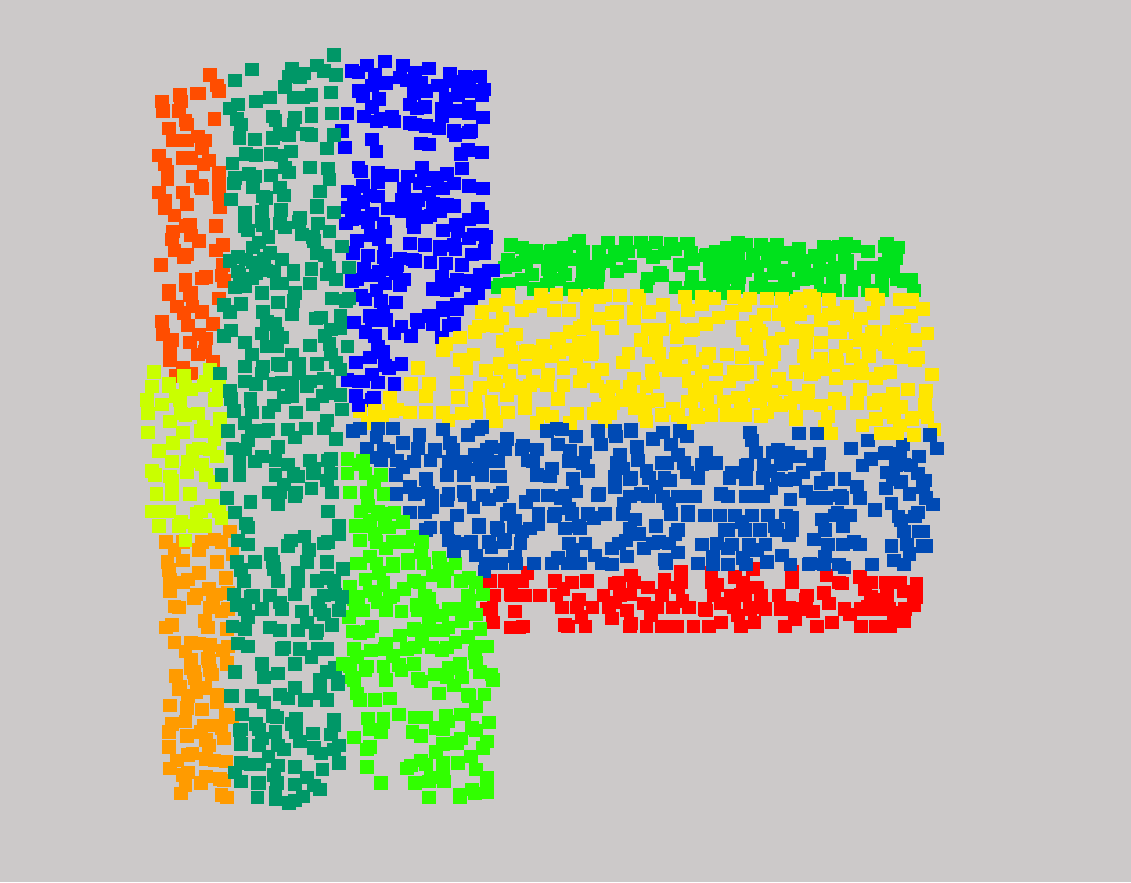} &
		\includegraphics[width= 0.21 \linewidth]{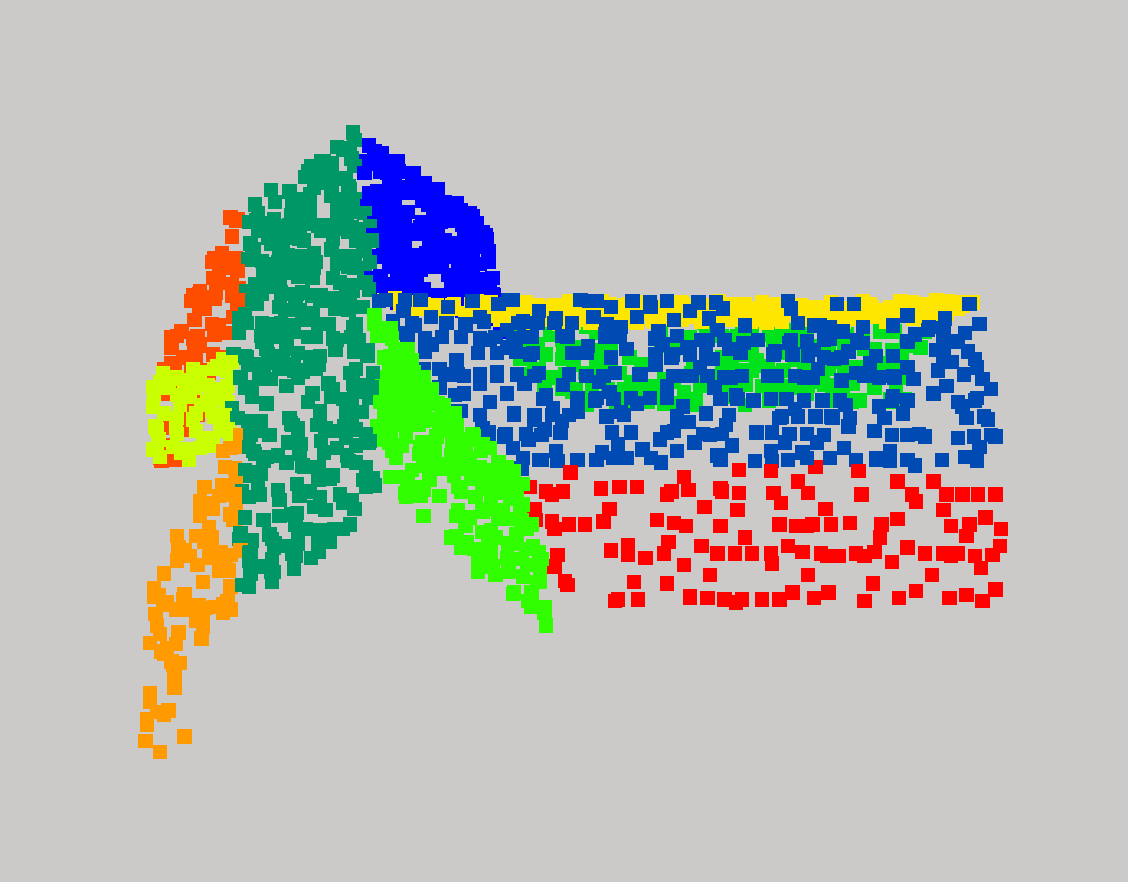}

	\end{tabular} 
	\caption{Segmentation results of two roof samples using the proposed network with different components. The first, second and third rows represent the segmentation results without using BAMM, adaptive mask loss $\mathcal{L}_{mask}^{w}$ and plane geometric loss $\mathcal{L}_{plane}$, respectively. The forth row represents the segmentation results using all three components, while the last row represents the Ground Truth. The blue ellipses and red wireframes point out some incorrectly segmented regions.}
	\vspace{0.0em}
	\label{Fig:ablation}
\end{figure}

\textbf{The effects of different backbones.} We evaluated several feature extraction backbones, including 3D U-Net~\citep{choy20194d}, PointNet++~\citep{qi2017pointnet++}, PointConvFormer~\citep{wu2023pointconvformer} and PointTransformer v3~\citep{wu2024point}, for a comprehensive comparison. For 3D U-Net, we used a grid size of $0.02$m for voxelization according to the point density of three benchmarks, while the other parameters are consistent with~\cite{choy20194d}. For PointNet++~\citep{qi2017pointnet++}, we implemented two variants: 1) the baseline with standard Feature Propagation (FP), and 2) the enhanced version with Attention-based Feature Propagation (AFP). PointConvFormer~\citep{wu2023pointconvformer} and PointTransformer v3~\citep{wu2024point} are the latest convolution-based and transformer-based backbones, respectively, which have demonstrated good performance across various point clouds segmentation tasks. The metric results of all backbones on three roof benchmarks are given in Table~\ref{Tab:ablation_backbone}. Compared to the 3D U‑Net and PointNet++ baselines, the PointNet++ variant with AFP achieves advantages of approximately 3 and 1 percentage points, respectively, across all metrics. While the two latest backbones also demonstrate strong performance, they remain overall comparable to the PointNet++ variant without significant superiority. Moreover, considering their relatively higher training costs, as reported in Table~\ref{Tab:ablation_backbone_efficency}, we select the PointNet++ variant as the backbone for RoofSeg after comprehensive evaluation.

\begin{table} [!htb] \renewcommand{\tabcolsep}{2.0 pt}
	\scriptsize
	\renewcommand{\arraystretch}{1.25}
	\newcommand{\tabincell}[2]{\begin{tabular}{@{}#1@{}}#2\end{tabular}}
	\begin{center}
		\caption{Quantitative assessment of RoofSeg using different backbones on three benchmarks. FP and AFP denote the standard and attention-based feature propagation applied in PointNet++, respectively.}
		\label{Tab:ablation_backbone}
		\begin{tabular}{c|c|cccc|cccc|cccc} 
			\hline
			\multicolumn{2}{c|}{\multirow{2}{*}{Different backbones}} & \multicolumn{4}{c}{RoofNTNU benchmark}  &  \multicolumn{4}{c}{Roofpc3D benchmark} & \multicolumn{4}{c}{Building3D benchmark}\\
			\cline{3-6} \cline{7-10} \cline{11-14} 
			\multicolumn{2}{c|}{} & mCov       & mWCov  &   mPrec   &    mRec      
			& mCov       & mWCov  &   mPrec   &    mRec   
			& mCov       & mWCov  &   mPrec   &    mRec    \\ 
			\hline
			\multicolumn{2}{c|}{3D U-Net}  & 92.48  & 94.23 & 96.70 & 94.73 & 92.09 & 94.55 & 97.86 & 95.35 & 90.41 & 92.74 & 95.97 & 93.32 \\ 
			\cline{1-2}
			\multirow{2}{*}{PointNet++} & w/ FP  & 95.03  & 96.21 & 98.99 & 97.57 & 95.28 & 96.64 & 99.25 & 98.01 & 92.83 & 95.40 & 97.90 & 96.02  \\ 
			& w/ AFP  & \textbf{95.89}  & 96.82 & 99.60 & \textbf{98.18} & 96.01 & 97.56 & \textbf{99.98} & 98.54 & \textbf{93.74} & 96.37 & \textbf{98.79} & 96.58     \\ 
			\cline{1-2}
			\multicolumn{2}{c|}{PointConvFormer}  & 95.37  & 96.44 & 99.28 & 97.61 & 95.85 & 96.97 & 99.42 & 98.36 & 93.15 & 95.94 & 98.02 & 96.31 \\ 
			\multicolumn{2}{c|}{PointTransformer v3}  & 95.84  & \textbf{96.85} & \textbf{99.71} & 98.12 & \textbf{96.05} & \textbf{97.61} & 99.82 & \textbf{98.60} & 93.69 & \textbf{96.40} & 98.77 & \textbf{96.62}\\ 
			\hline
		\end{tabular}
	\end{center}
\end{table}

\begin{table} [!htb] \renewcommand{\tabcolsep}{14.0 pt}
	\scriptsize
	\renewcommand{\arraystretch}{1.25}
	\newcommand{\tabincell}[2]{\begin{tabular}{@{}#1@{}}#2\end{tabular}}
	\begin{center}
		\caption{Efficiency evaluation of different backbones. \# Para. denotes the parameters of backbones. The last two columns represent the GPU memory consumption and training time cost (per epoch) with batch size = 24. Time costs are measured relative to the original PointNet++ (w/FP) as the baseline.}
		\label{Tab:ablation_backbone_efficency}
		\begin{tabular}{c|c|cccc} 
			\hline			
			\multicolumn{2}{c|}{Different backbones} & \multicolumn{1}{c}{Input}  &  \multicolumn{1}{c}{\# Paras.(MB)} & \multicolumn{1}{c}{GPU Mem.(MB)} & \multicolumn{1}{c}{Time/epoch}\\
			\hline
			\multicolumn{2}{c|}{3D U-Net}  & voxel  & 5.34 & 6201.70 & 1.84$\times$ \\ 
			\cline{1-2}
			\multirow{2}{*}{PointNet++} & w/ FP  & point  & 2.41 & 3651.17 & 1.00$\times$   \\ 
			& w/ AFP  & point  & 4.82 & 4984.60 & 1.35$\times$    \\ 
			\cline{1-2}
			\multicolumn{2}{c|}{PointConvFormer}  & voxel  & 6.85 & 7328.32 & 2.26$\times$ \\ 
			\multicolumn{2}{c|}{PointTransformer v3}  & point  & 8.76 & 8919.40 & 2.74$\times$  \\ 
			\hline
		\end{tabular}
	\end{center}
\end{table}

\textbf{The effects of adaptive weighting loss calculation with varied numbers of nearest neighbors.} As described in the Section~\ref{sec:loss functions}, to calculate the adaptive weighting mask loss, we first performed neighborhood analysis for each point based on the initial plane mask to identify potential misclassified points, whose labels do not align with the majority of their neighbors, and then adaptively computed their weights. To investigate the effect of selecting different numbers of nearest neighbors ($N_{knn}$), we conducted corresponding experiments on three benchmarks, with $N_{knn}$ set to 10, 20, 30, 40, and 50, respectively. As shown in Table~\ref{Tab:ablation_knn}, the results across all groups are very close on all four metrics, with the largest gap being less than 0.5 percentage points. From that observation, it can be concluded that our network is not sensitive to this hyperparameter. In experiments, we empirically set $N_{knn}$ to 30.

\begin{table} [!htb] \renewcommand{\tabcolsep}{3.5 pt}
	\scriptsize
	\renewcommand{\arraystretch}{1.4}
	\newcommand{\tabincell}[2]{\begin{tabular}{@{}#1@{}}#2\end{tabular}}
	\begin{center}
		\caption{Quantitative assessment of RoofSeg using varied numbers of nearest neighbors ($N_{knn}$) in adaptive weighting loss calculation on three benchmarks. }
		\label{Tab:ablation_knn}
		\begin{tabular}{c|cccc|cccc|cccc} 
			\hline
			\multirow{2}{*}{\makecell[c]{Variation in $N_{knn}$}} & \multicolumn{4}{c}{RoofNTNU benchmark}  &  \multicolumn{4}{c}{Roofpc3D benchmark} & \multicolumn{4}{c}{Building3D benchmark}\\
			\cline{2-5} \cline{6-9} \cline{10-13} 
			& mCov       & mWCov  &   mPrec   &    mRec      
			& mCov       & mWCov  &   mPrec   &    mRec   
			& mCov       & mWCov  &   mPrec   &    mRec    \\ 
			\hline
			10  & 95.76  & 96.39 & 99.06 & 97.69 & 95.60 & 97.23 & 99.47 & 98.06 & 93.25 & 95.88 & 98.40 & 95.96 \\ 
			
			20  & 95.73  & 96.52 & \textbf{99.63} & 98.02  & 96.00 & 97.34 & 99.50 & 98.43 & 93.57 & 95.97 & 98.51 & 96.09  \\ 
			30 & \textbf{95.89}  & \textbf{96.82} & 99.60 & \textbf{98.18} & \textbf{96.01} & \textbf{97.56} & \textbf{99.98} & \textbf{98.54} & \textbf{93.74} & 96.37 & \textbf{98.79} & \textbf{96.58}    \\ 
			
			40  & 95.69  & 96.75 & 99.54 & 97.93 & 95.87 & 97.50 & 99.87 & 98.32 & 93.29 & \textbf{96.39} & 98.57 & 96.25   \\ 
			
			50  & 95.47  & 96.45 & 99.37 & 97.74 & 95.71 & 97.11 & 99.41 & 98.15 & 93.38 & 96.05 & 98.39 & 96.13   \\ 
			\hline
		\end{tabular}
	\end{center}
\end{table}

\textbf{The effects of adaptive weighting loss calculation on edge regions.} To analyze whether the adaptive weighting mechanism affects edge regions, we conducted an additional ablation study. Edge points were defined based on the ground-truth KNN neighborhood: points whose neighborhoods contained different plane instances were treated as edge points. As shown in Table~\ref{Tab:ablation_edge_corner}, the full adaptive weighting strategy consistently outperforms the model without adaptive weighting on all three benchmarks. Moreover, compared with the setting where edge points are excluded from  adaptive weighting, the full adaptive weighting strategy still achieves better results. These results indicate that the proposed mechanism does not negatively affect edge points. Although some edge points may be identified as potential outliers due to their mixed-label neighborhoods, the mechanism only reweights the mask loss during training and does not perform label smoothing or post-processing. By increasing the loss contribution of these difficult points, the network is encouraged to better align these points with their ground truth labels.

\begin{table} [!htb] 
	\renewcommand{\tabcolsep}{2.0 pt}
	\scriptsize
	\renewcommand{\arraystretch}{1.4}
	\newcommand{\tabincell}[2]{\begin{tabular}{@{}#1@{}}#2\end{tabular}}
	\begin{center}
		\caption{Quantitative assessment of RoofSeg with different adaptive weighting strategies on three benchmarks.}
		\label{Tab:ablation_edge_corner}
		\begin{tabular}{c|cccc|cccc|cccc} 
			\hline
			\multirow{2}{*}{\makecell[c]{Weighting strategy}} 
			& \multicolumn{4}{c}{RoofNTNU benchmark}  
			& \multicolumn{4}{c}{Roofpc3D benchmark} 
			& \multicolumn{4}{c}{Building3D benchmark}\\
			\cline{2-5} \cline{6-9} \cline{10-13} 
			& mCov & mWCov & mPrec & mRec      
			& mCov & mWCov & mPrec & mRec   
			& mCov & mWCov & mPrec & mRec    \\ 
			\hline
			
			w/o adaptive weighting 
			& 94.96 & 95.28 & 99.07 & 97.63 
			& 95.75 & 97.14 & 99.79 & 98.31 
			& 93.14 & 95.69 & 97.95 & 96.17 \\ 
			
			Adaptive weighting w/o edge
			& 95.54 & 96.43 & 99.38 & 97.94  
			& 95.88 & 97.38 & 99.86 & 98.43 
			& 93.46 & 96.06 & 98.42 & 96.33 \\ 
			
			Full adaptive weighting 
			& \textbf{95.89} & \textbf{96.82} & \textbf{99.60} & \textbf{98.18} 
			& \textbf{96.01} & \textbf{97.56} & \textbf{99.98} & \textbf{98.54} 
			& \textbf{93.74} & \textbf{96.37} & \textbf{98.79} & \textbf{96.58} \\ 
			
			\hline
		\end{tabular}
	\end{center}
\end{table}

\textbf{The effects of adaptive weighting loss calculation with fixed outlier weights.} To further evaluate the weighting strategy, we replaced the adaptive point-wise weights with three constant outlier weights, i.e., C = 5, 10, and 20. The outlier detection process was kept unchanged. Detected outliers were assigned the fixed weight C, while the remaining points were assigned a weight of 1.
As shown in Table~\ref{Tab:ablation_fixed_weight}, the performance of constant weights varies across benchmarks. Although some fixed weights perform well on individual datasets, no single constant weight consistently achieves the best results. In contrast, the proposed adaptive weighting strategy achieves the best overall performance by dynamically adjusting the outlier weights according to both the global outlier ratio and local neighborhood inconsistency, demonstrating better robustness and generalization.

\begin{table} [!htb] 
	\renewcommand{\tabcolsep}{3.5 pt}
	\scriptsize
	\renewcommand{\arraystretch}{1.4}
	\newcommand{\tabincell}[2]{\begin{tabular}{@{}#1@{}}#2\end{tabular}}
	\begin{center}
		\caption{Quantitative assessment of RoofSeg using different fixed outlier weights in adaptive weighting loss calculation on three benchmarks.}
		\label{Tab:ablation_fixed_weight}
		\begin{tabular}{c|cccc|cccc|cccc} 
			\hline
			\multirow{2}{*}{\makecell[c]{Weight setting}} 
			& \multicolumn{4}{c}{RoofNTNU benchmark}  
			& \multicolumn{4}{c}{Roofpc3D benchmark} 
			& \multicolumn{4}{c}{Building3D benchmark}\\
			\cline{2-5} \cline{6-9} \cline{10-13} 
			& mCov & mWCov & mPrec & mRec      
			& mCov & mWCov & mPrec & mRec   
			& mCov & mWCov & mPrec & mRec    \\ 
			\hline
			
			$C=5$  
			& 95.55 & 96.49 & 99.42 & 97.83 
			& 95.32 & 96.76 & 99.34 & 97.66 
			& 92.81 & 95.24 & 98.02 & 95.41 \\ 
			
			$C=10$  
			& 95.29 & 96.25 & 99.18 & 97.52  
			& 95.76 & 97.21 & 99.70 & 98.17 
			& 93.41 & 96.02 & 98.54 & 96.23 \\ 
			
			$C=20$ 
			& 94.98 & 95.96 & 98.91 & 97.05 
			& 95.11 & 96.62 & 99.12 & 97.48 
			& 93.48 & 96.13 & 98.60 & 96.42 \\ 
			
			Adaptive weight 
			& \textbf{95.89} & \textbf{96.82} & \textbf{99.60} & \textbf{98.18} 
			& \textbf{96.01} & \textbf{97.56} & \textbf{99.98} & \textbf{98.54} 
			& \textbf{93.74} & \textbf{96.37} & \textbf{98.79} & \textbf{96.58} \\ 
			
			\hline
		\end{tabular}
	\end{center}
\end{table}

Additional ablation results, including the effects of using different numbers of plane queries for segmentation and the impact of varying sampling counts during training, are provided in Appendix~\ref{appendix:extended_ablation_results}.

\subsection{Limitation} 
Although the proposed approach demonstrates good performance in extensive experiments, it still has the following limitations. First, the training of the proposed network relies heavily on a sufficiently large and diverse dataset, while data annotation typically lead to high labor costs. Insufficient training data or poor annotation quality can significantly reduce the model's robustness. Therefore, the future work will focus on developing unsupervised or weakly supervised training strategies to reduce reliance on manually labeled data. Second, different from traditional approaches, it is difficult for the current network to handle large-scale point clouds due to limited computational resources, especially during multi-batch training. Although downsampling and block-wise processing can alleviate this issue, bottlenecks still remain. It is necessary to further streamline the network architecture, such as incorporating more efficient convolution or attention modules, to improve computational efficiency. Moreover, the fixed query‑number design remains a limitation for the current network. Future research should explore adaptive mechanisms to determine the query count per scene for avoiding the insufficient plane prediction.

\section{Conclusion} 

In this work, we presented RoofSeg, an end-to-end edge-aware transformer-based network for roof plane segmentation from airborne LiDAR point clouds. By directly predicting plane instance masks using learnable plane queries within a transformer encoder-decoder framework, RoofSeg demonstrates that fully end-to-end segmentation can overcome the limitations of multi-stage pipelines, including accumulated errors, parameter sensitivity, and high computational cost. The carefully designed Edge-Aware Mask Module (EAMM) demonstrates that incorporating geometric priors, namely point-to-plane distance, into network can effectively alleviate the ambiguity of edge features and enhance edge discriminability. In addition, the combination of an adaptive-weighting mask loss and a plane geometric loss introduces effective constraints at both point and instance levels, thereby mitigating misclassified points and enforcing high geometric fidelity. Experiments on roof-specific benchmarks demonstrate the superiority of RoofSeg over existing traditional and learning-based approaches, producing segmentation results with more accurate edges, fewer misclassified points, and higher geometric fidelity. Moreover, its strong performance on generic plane segmentation benchmarks further verifies its generalization capability beyond roof-specific scenarios.

Although the proposed RoofSeg achieves strong performance, it still has room for further improvement. In future work, we will explore unsupervised or weakly supervised training strategies to reduce the dependence on manually annotated datasets. We will also investigate adaptive query generation mechanisms to replace the current fixed-query-number strategy, and optimize the network architecture to improve scalability for large-scale point clouds. These directions are expected to enhance both the applicability and efficiency of end-to-end roof plane segmentation in practical 3D building reconstruction scenarios.

\section*{Acknowledgements}

This work was partially funded by the National Natural Science Foundation of China (No.U22A2009, No.42271445, No.42101440), the Natural Science Foundation of Hubei Province (No.2025AFB491), the Shenzhen Science and Technology Program (JCYJ2023\\0807090206013).

\appendix 

\renewcommand{\thetable}{\arabic{table}}
\renewcommand{\thefigure}{\arabic{figure}}

\renewcommand{\thesection}{Appendix \Alph{section}}
\renewcommand{\thesubsection}{\Alph{section}.\arabic{subsection}}

\section{Additional experimental details and results}

\renewcommand{\thesection}{\Alph{section}}

\subsection{Implementation details} 
\label{appendix:implementation_details}
\textbf{Network details.} The backbone, built on the PointNet++~\citep{qi2017pointnet++}, consists of four Set Abstraction (SA) layers and four Attention-based Feature Propagation (AFP) layers. The SA layers are applied to downsample the input points, with sampling radio of 1/4 in each layer, and aggregate the features with receptive radius of [0.05, 0.1], [0.1, 0.2], [0.2, 0.4] and [0.4, 0.8]. Then, AFP layers are applied to gradually upsample and generate the multi-scale point features. The attention-based modules (in AFP, QRD and EAMM) employ standard transformer blocks~\citep{vaswani2017attention}, with both input and output dimension set to 256, and the dropout threshold set to 0.3. The edge prediction branch and query semantic branch are essentially two MLPs, with the output demision set to 1. The dropout threshold is set to 0.1 in both branches.

\textbf{Training details.} The network training was performed under the PyTorch framework using a GeForce RTX 4090 GPU (24GB). The optimization process employs AdamW \\~\citep{loshchilov2017fixing} with the learning rate follows a cosine decay schedule to a minimum of $1 \times 10^{-6}$. RoofSeg are trained with the maximum 200 epochs, while the batch size is set to 24. To enable batched processing, all input point clouds are uniformly resampled to 2,048 points in training phase. 

\subsection{Dataset configuration} 
\label{appendix:dataset_configuration}
\begin{figure}[!htb]
	\scriptsize
	\centering
	\begin{tabular}{cccc} 	
		
		\multicolumn{1}{c}{\includegraphics[width= 0.22 \linewidth]{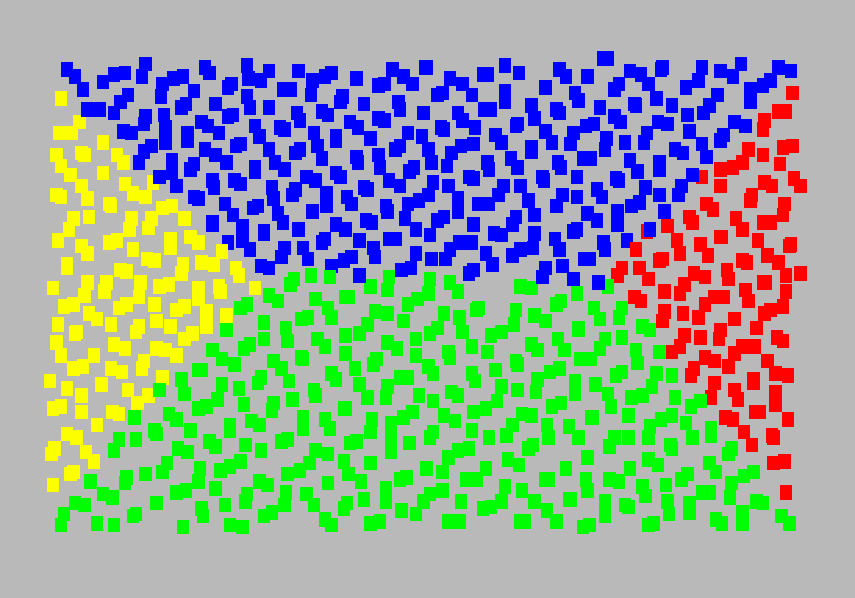}} &
		\multicolumn{1}{c}{\includegraphics[width= 0.22 \linewidth]{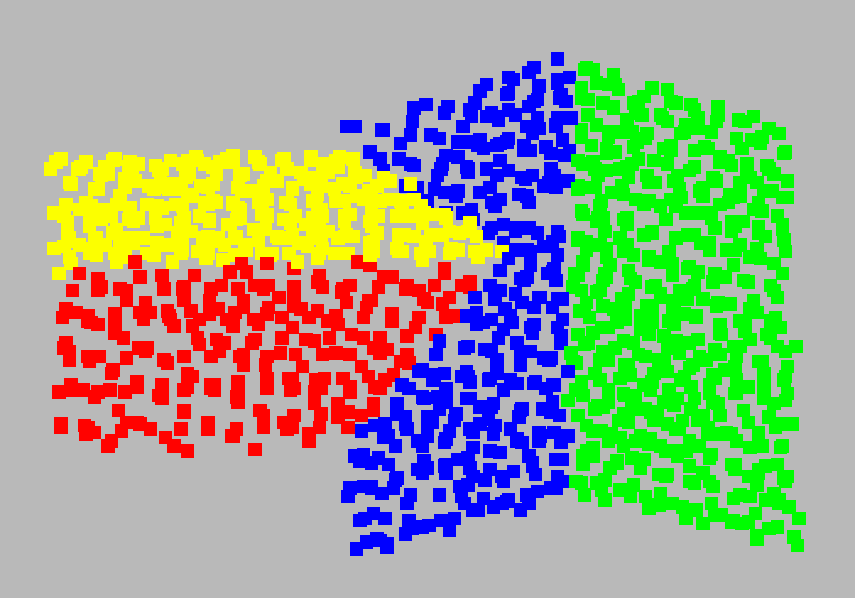}} &
		\multicolumn{1}{c}{\includegraphics[width= 0.22 \linewidth]{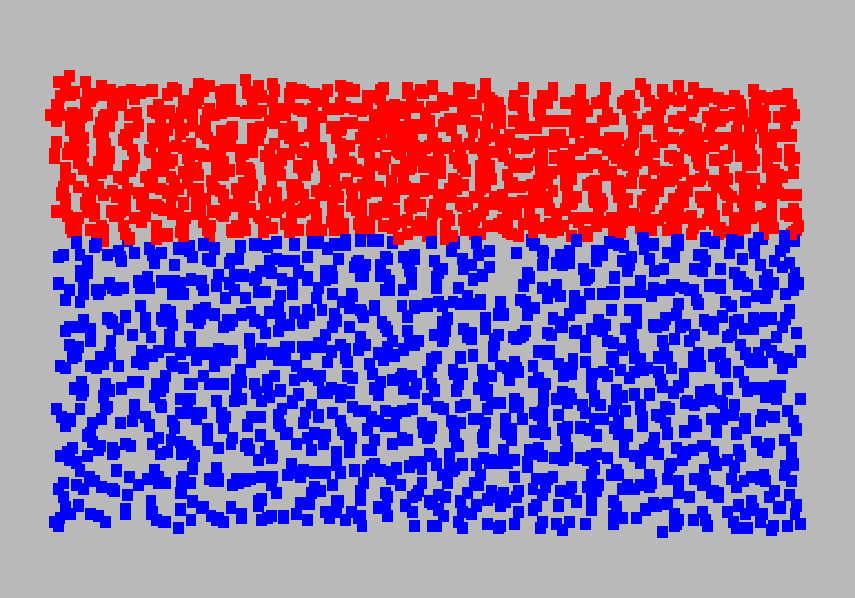}} &
		\multicolumn{1}{c}{\includegraphics[width= 0.22 \linewidth]{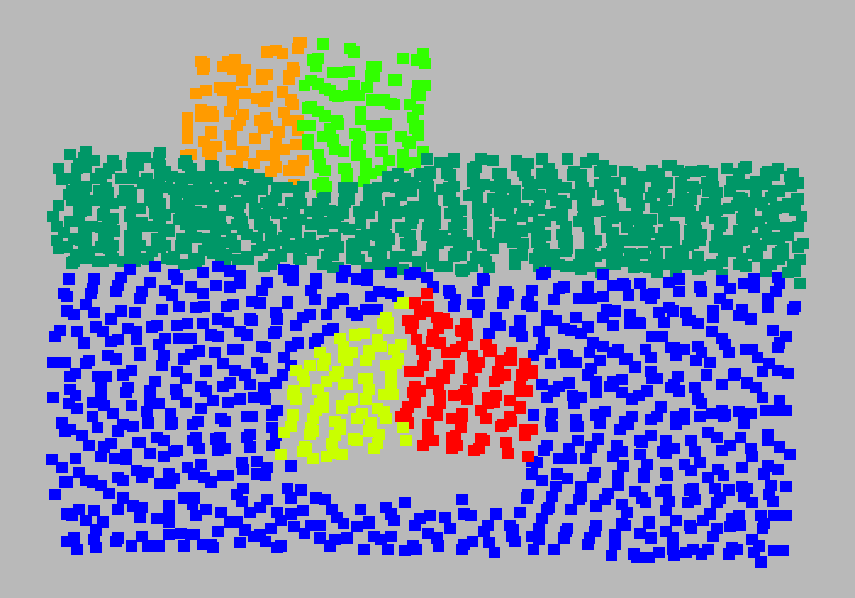}} 
		\\
		& & & 
		\\
		\multicolumn{1}{c}{\includegraphics[width= 0.22 \linewidth]{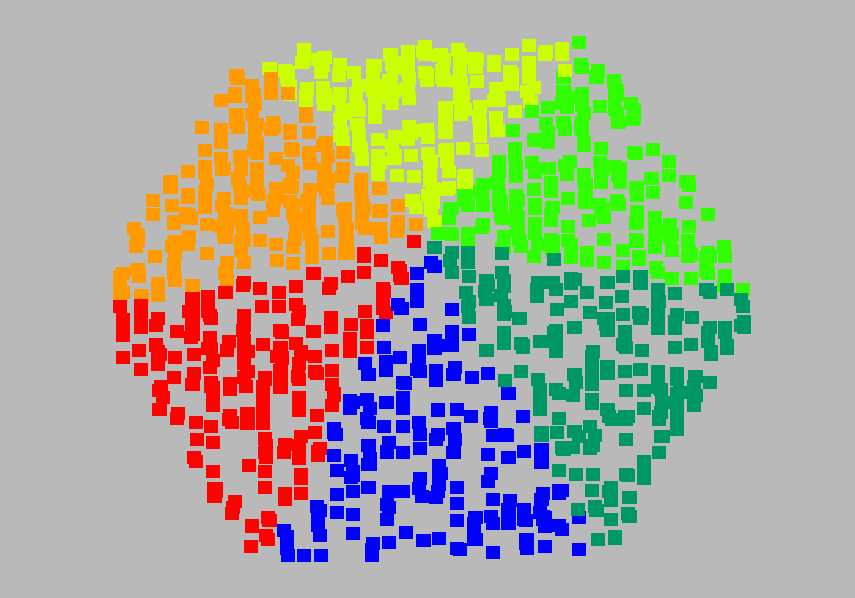}} &
		\multicolumn{1}{c}{\includegraphics[width= 0.22 \linewidth]{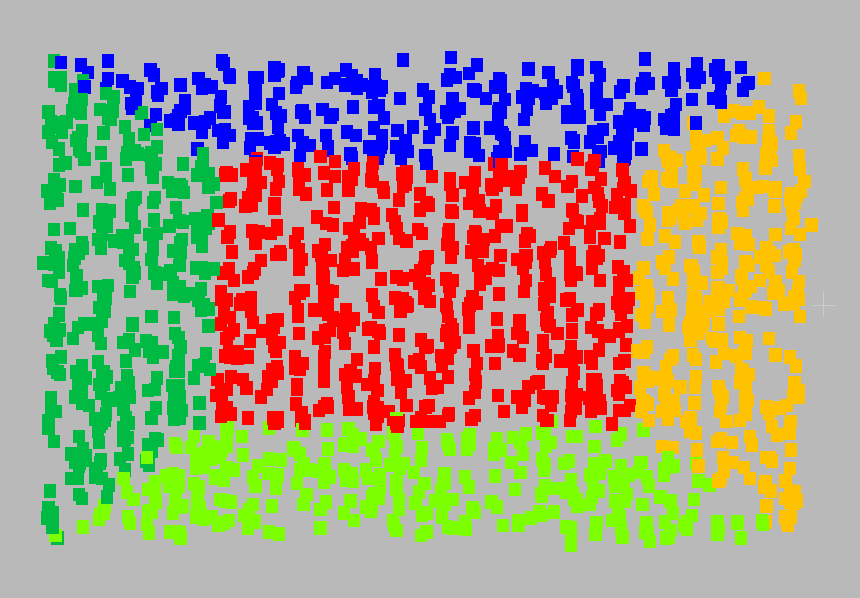}} &
		\multicolumn{1}{c}{\includegraphics[width= 0.22 \linewidth]{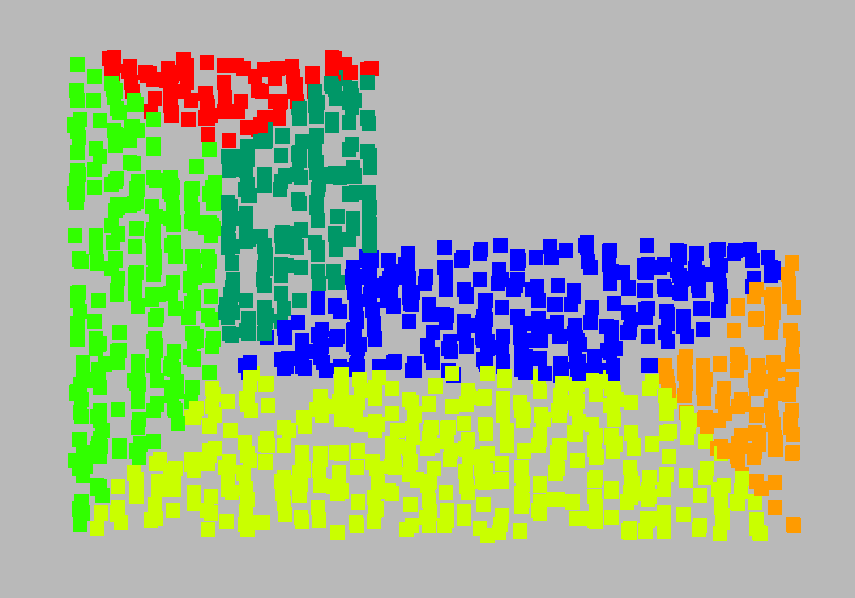}} &
		\multicolumn{1}{c}{\includegraphics[width= 0.22 \linewidth]{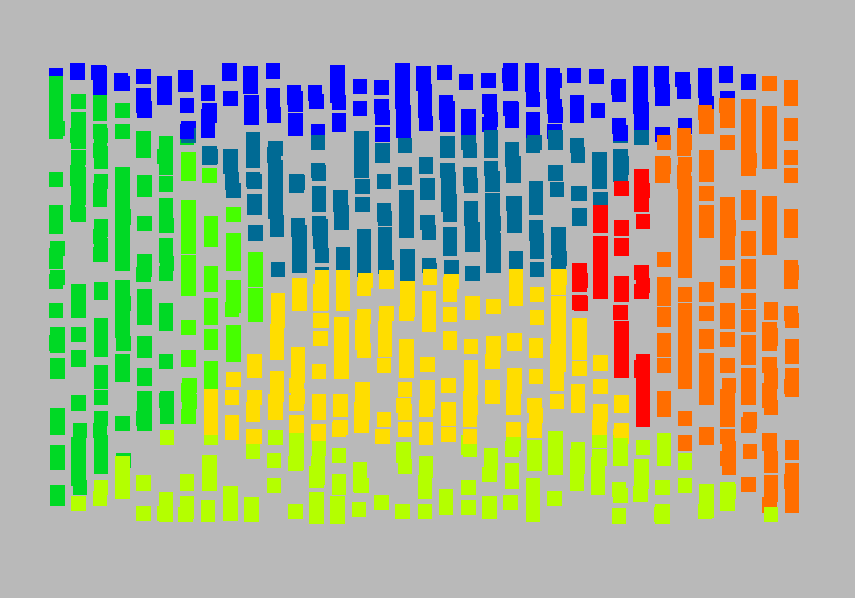}} 
		\\
		& & &
		\\
		\multicolumn{1}{c}{\includegraphics[width= 0.22 \linewidth]{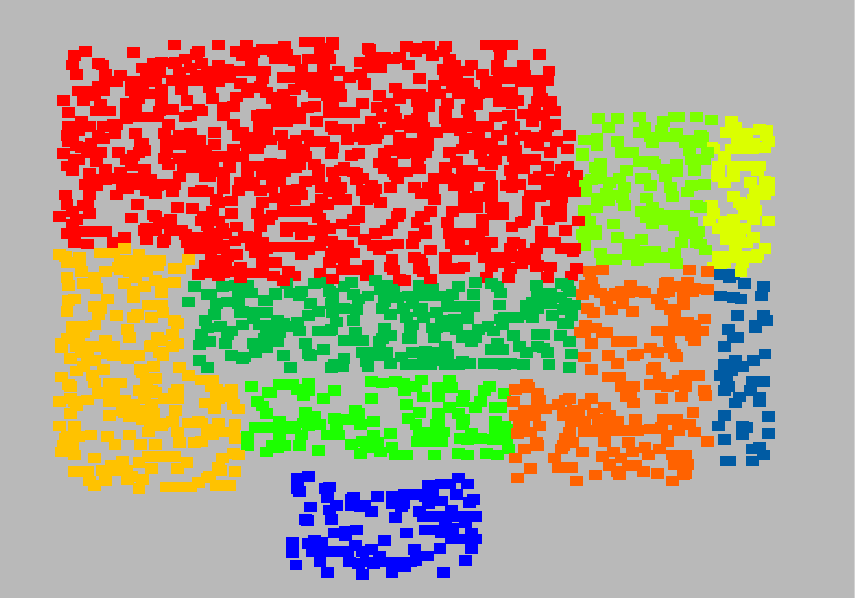}} &
		\multicolumn{1}{c}{\includegraphics[width= 0.22 \linewidth]{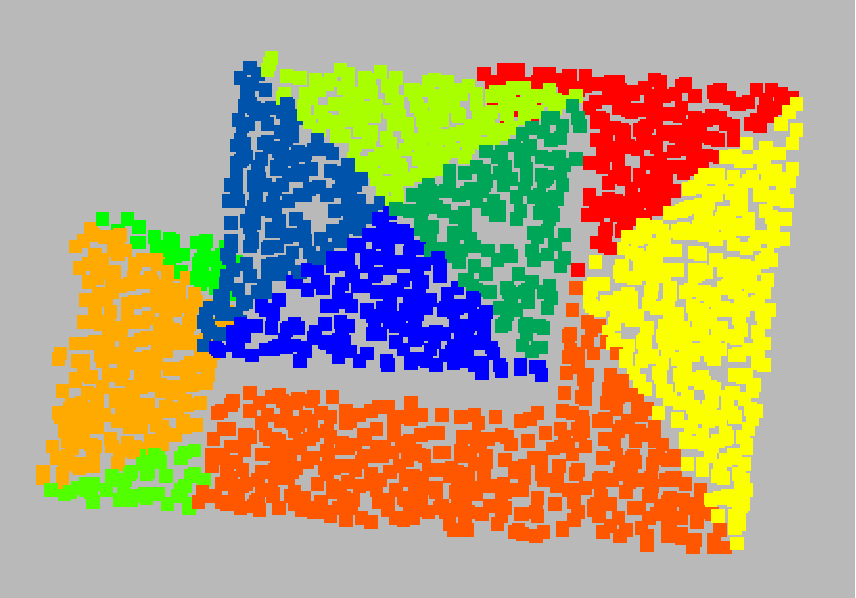}} &
		\multicolumn{1}{c}{\includegraphics[width= 0.22 \linewidth]{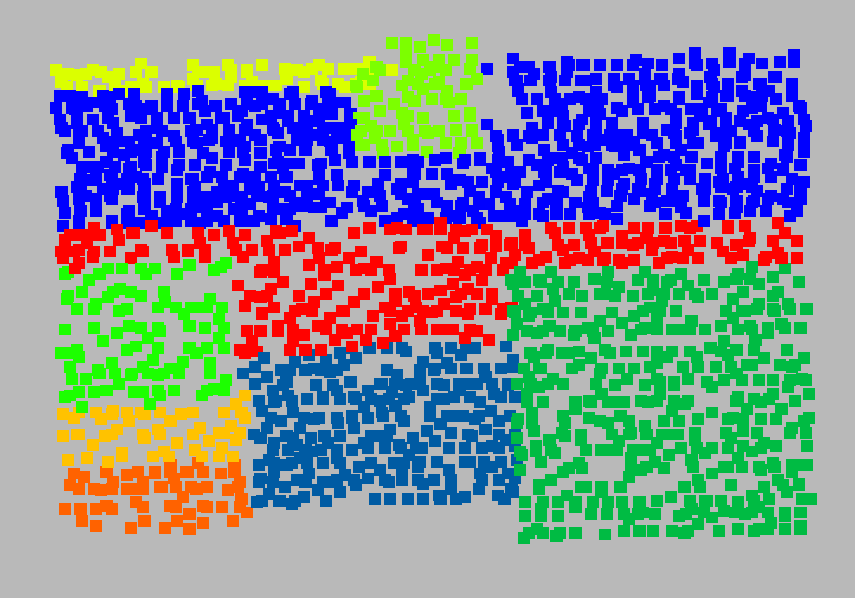}} &
		\multicolumn{1}{c}{\includegraphics[width= 0.22 \linewidth]{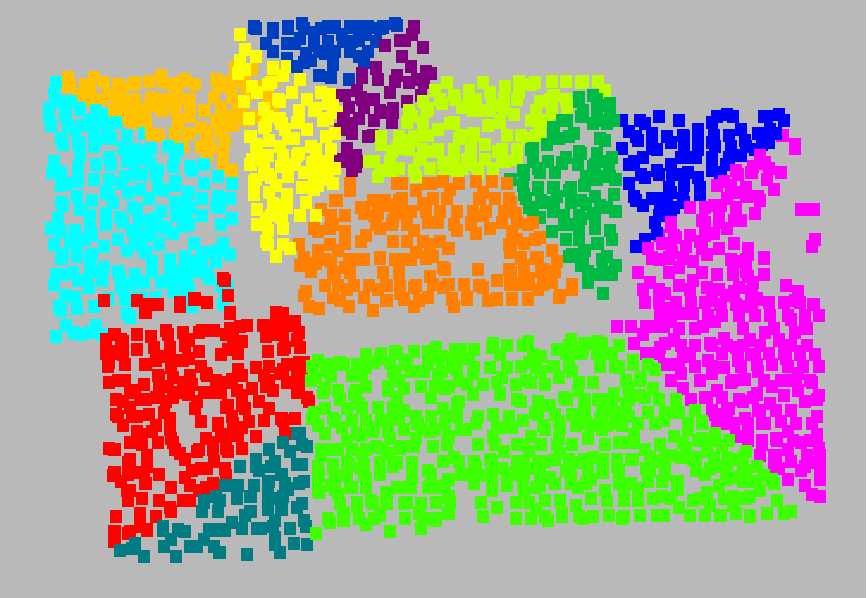}} 
		
	\end{tabular} 
	\caption{Visualization of some roof samples from RoofNTNU benchmark (the first line),  Roofpc3D benchmark (the second line) and Building3D benchmark (the third line). Different colored points on a roof sample denote different planar surfaces.}
	\vspace{0.0em}
	\label{Fig:Datasets}
\end{figure}

\textbf{RoofNTNU benchmark.} This is a small-scale real-world benchmark built by ~\cite{zhang2022improved}. It contains 1,032 building roofs from Trondheim city in Norway, which are manually segmented into 3,478 planes. The benchmark is divided into training, validation and test set in an 8:1:1 ratio. In this paper, we merge the training and validation sets, resulting in a combined training set of 933 roofs. Some roof samples of this benchmark are shown in Fig.~\ref{Fig:Datasets} (the first row).

\textbf{Roofpc3D benchmark.} It is a large-scale synthetic benchmark generated by~\cite{li2024boundary}, which contains 15,400 building roofs with annotated plane instance labels. It includes 14 types of roofs, such as pyramid, dormer and cross-hipped, with 1,100 samples of each type. In this benchmark, the maximum number of planes for a single roof sample is 8, while the minimum is 2. Following a ratio of 10:1, the benchmark is divided into a training set of 14,000 roofs and a test set of 1400 roofs. Some roof samples are presented in Fig.~\ref{Fig:Datasets} (the second row).

\textbf{Building3D benchmark.} This is a real-world benchmark constructed on large-scale 3D urban buildings for roof structure learning, created by~\cite{wang2023building3d}, which comprises the airborne point clouds, wireframe models and surface meshes of 32,618 building roofs from Tallinn city. Based on that,~\cite{li2024boundary} selected 18,708 roofs and manually annotated the ground truth plane labels, thus deriving a new benchmark for roof plane segmentation. The benchmark was split into 16,884 training and 1,824 test samples at an approximate 10:1 ratio. Notably, it is a challenging benchmark due to its more diverse and complex roof types. Some roof samples are presented in Fig.~\ref{Fig:Datasets} (the third row).

\textbf{ABC benchmark.} ABC~\citep{koch2019abc} is a large-scale public benchmark of 3D CAD models containing a variety of primitive shapes. Focusing on plane segmentation task, we selected 2,000 piece-wise planar models from a subset of the ABC (namely surface normal estimation benchmark constructed by~\cite{koch2019abc}). Given that these models are provided with accurate annotations for each planar face, we generated a point cloud for each model by sampling 500 points per face. All data were then split into training and test sets at a ratio of 10:1. 

\textbf{KSR42 benchmark.} The KSR42 benchmark, introduced by~\cite{yu2022finding} for plane detection, contains 42 challenging real-world models. Since point-level plane annotations are not available, we manually annotated 10 representative models, including meeting room, euler, barn MVS, barn laser, chair, couch, castle, foam box, full thing, and lans fine. For meeting room and euler, one area from each model was selected as the test set, and the remaining areas were randomly cropped for training. For the barn models, barn laser was used for training and barn MVS was used for testing. The remaining object-level models were used only for testing. Since KSR42 does not provide object-level training samples, the network was trained using the ABC benchmark along with the training portions of the manually annotated KSR42 models. Since KSR42 does not provide object-level training samples, the network was trained using the ABC benchmark along with the training portions of the manually annotated KSR42 models.

\subsection{Cross-dataset evaluation}
\label{appendix:cross-dataset_evaluation}

To evaluate the cross‑dataset performance of the proposed network, we tested the RoofSeg (trained on the Building3D benchmark) on the other two roof benchmarks, and compared it with the current best traditional approach (GoCoPP) and deep‑learning approach (DeepRoofPlane). As shown in Table~\ref{Tab:comparative_cross_dataset_1}, both RoofSeg and DeepRoofPlane exhibit significant performance degradation under the cross-dataset evaluation. Notably, this is an inevitable challenge for most deep learning approaches. Nevertheless, our network still shows a relatively smaller performance drop, and remains within 1 percentage point of the best traditional approach across all metrics. To mitigate this issue, fine-tuning is an effective technique. As shown in Table~\ref{Tab:comparative_cross_dataset_2}, with a small amount of training data from the two benchmarks (1/10 of the original training set) and a couple of iterations (within 10 epochs), our network can rapidly achieve significant performance gains, which demonstrates its strong transfer capability.

\begin{table} [!htb] \renewcommand{\tabcolsep}{5.5 pt}
	\scriptsize
	\renewcommand{\arraystretch}{1.25}
	\newcommand{\tabincell}[2]{\begin{tabular}{@{}#1@{}}#2\end{tabular}}
	\begin{center}
		\caption{Evaluation of the RoofSeg's cross-dataset performance against top-performing traditional (GoCoPP) and deep-learning (DeepRoofPlane) approaches. RoofSeg and DeepRoofPlane (both trained on the Building3D benchmark) are evaluated on the other two roof benchmarks. Values in parentheses denote deviations from their respective fully-trained performance on those benchmarks.}
		\label{Tab:comparative_cross_dataset_1}
		\begin{tabular}{c|c|cccc|ccc} 
			\hline
			Cross-dataset & Approaches &  
			mCov$\uparrow$       & mWCov$\uparrow$   &   mPrec$\uparrow$    &    mRec$\uparrow$       
			& $O_{seg}\downarrow$  & $U_{seg}\downarrow$  &  $E_{acc}\uparrow$      \\ 
			\hline
			\multirow{3}{*}{RoofNTNU} & GoCoPP  & 90.4  & 91.4 & 95.8 & 93.9 & 4.4 & 3.5 & 89.6  \\ 
			\cline{2-2}
			& DeepRoofPlane  & 81.5(-9.6)  & 86.3(-7.4) & 91.8(-5.9) & 88.3(-6.1) & 7.6(+3.8) & 8.0(+3.2) & 81.8(-8.9)  \\ 
			\cline{2-2}
			& RoofSeg   & 89.7(-6.2)  & 90.9(-5.9) & 94.9(-4.7) & 93.6(-4.6) & 4.7(+3.5) & 5.2(+2.8) & 89.1(-5.8)  \\ 
			\hline
			\multirow{3}{*}{Roofpc3D} &	GoCoPP  & 91.5  & 93.4 & 96.1 & 93.0 & 3.2 & 4.9 & 90.6  \\ 
			\cline{2-2}
			& DeepRoofPlane  & 82.1(-10.3)  & 85.9(-8.8) & 90.8(-6.5) & 87.9(-6.1) & 8.4(+2.6) & 9.3(+3.0) & 81.9(-9.8)  \\ 
			\cline{2-2}
			& RoofSeg   & 90.0(-6.0)  & 92.9(-4.7) & 95.8(-4.2) & 93.1(-5.4) & 4.4(+2.0) & 4.9(+2.2) & 88.5(-6.9)  \\ 
			\hline
		\end{tabular}
	\end{center}
\end{table}

\begin{table} [!htb] \renewcommand{\tabcolsep}{6.0 pt}
	\scriptsize
	\renewcommand{\arraystretch}{1.25}
	\newcommand{\tabincell}[2]{\begin{tabular}{@{}#1@{}}#2\end{tabular}}
	\begin{center}
		\caption{Effect of fine‑tuning on RoofSeg’s cross‑dataset performance. The table reports the performance of the network fine-tuned after 5 and 10 epochs on the subsets (1/10 of the training sets) of two benchmarks. Values in parentheses show the gaps relative to training the network from scratch on each benchmark.}
		\label{Tab:comparative_cross_dataset_2}
		\begin{tabular}{c|c|cccc|ccc} 
			\hline
			Cross-dataset & Fine-tuning &  
			mCov$\uparrow$       & mWCov$\uparrow$   &   mPrec$\uparrow$    &    mRec$\uparrow$       
			& $O_{seg}\downarrow$  & $U_{seg}\downarrow$  &  $E_{acc}\uparrow$      \\ 
			\hline
			\multirow{3}{*}{RoofNTNU} & w/o  & 89.7(-6.2)  & 90.9(-5.9) & 94.9(-4.7) & 93.6(-4.6) & 4.7(+3.5) & 5.2(+2.8) & 89.1(-5.8)  \\ 
			\cline{2-2}
			&epochs=5  & 93.8(-2.1)  & 94.7(-1.6) & 98.3(-1.3) & 96.7(-1.5) & 2.4(+1.2) & 3.3(+0.9) & 93.2(-1.7)  \\ 
			\cline{2-2}
			& epochs=10   & 95.6(-0.3)  & 96.4(-0.4) & 99.4(-0.2) & 98.1(-0.1) & 1.5(+0.3) & 2.5(+0.2) & 94.4(-0.5)  \\ 
			\hline
			\multirow{3}{*}{Roofpc3D} &	w/o  & 90.0(-6.0)  & 92.9(-4.7) & 95.8(-4.2) & 93.1(-5.4) & 4.4(+2.0) & 4.9(+2.2) & 88.5(-6.9)  \\
			\cline{2-2}
			& epochs=5  & 94.2(-1.8)  & 96.0(-1.6) & 97.8(-2.1) & 96.8(-1.7) & 3.6(+1.1) & 4.0(+1.3) & 93.1(-2.3)  \\ 
			\cline{2-2}
			& epochs=10   & 95.6(-0.4)  & 97.1(-0.5) & 99.4(-0.6) & 98.2(-0.3) & 2.7(+0.2) & 3.0(+0.3) & 95.0(-0.4)  \\ 
			\hline
		\end{tabular}
	\end{center}
\end{table}

\subsection{Robustness analysis}
\label{appendix:robustness_analysis}
To further analyze the capability of our network in handling common challenges of point cloud data, we conducted relevant experiments for in-depth validation of the network's robustness on the three roof benchmarks.

\begin{figure}[h!]
	\scriptsize
	\centering
	\begin{tabular}{ccccc} 	
		
		\multicolumn{1}{c}{\includegraphics[width= 0.175 \linewidth]{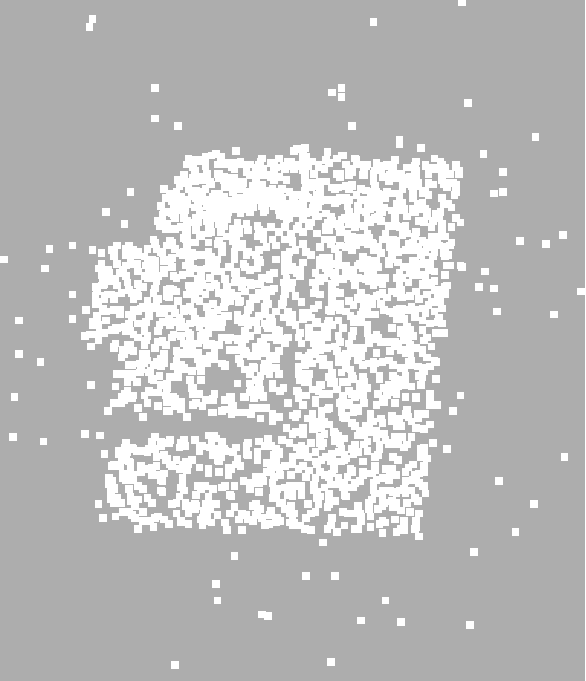}} &
		\multicolumn{1}{c}{\includegraphics[width= 0.175 \linewidth]{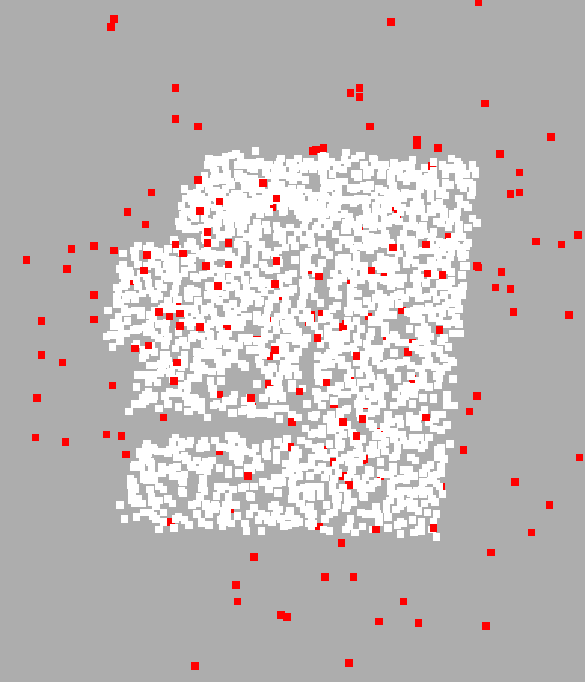}} &
		\multicolumn{1}{c}{\includegraphics[width= 0.175 \linewidth]{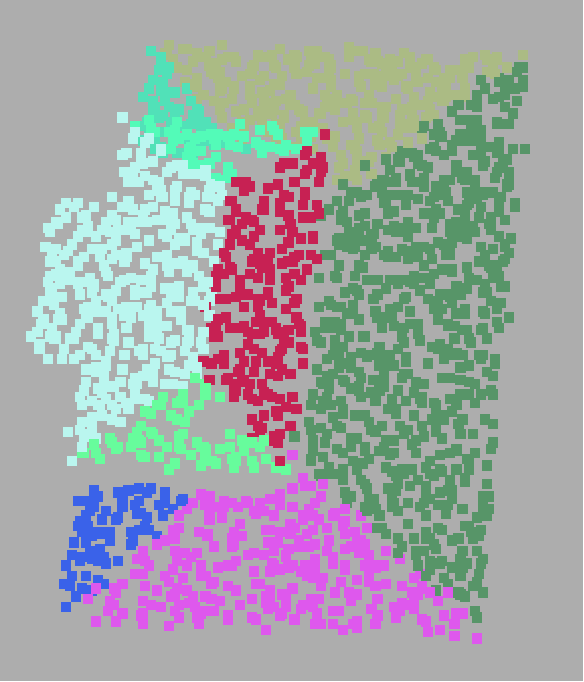}} &
		\multicolumn{1}{c}{\includegraphics[width= 0.175 \linewidth]{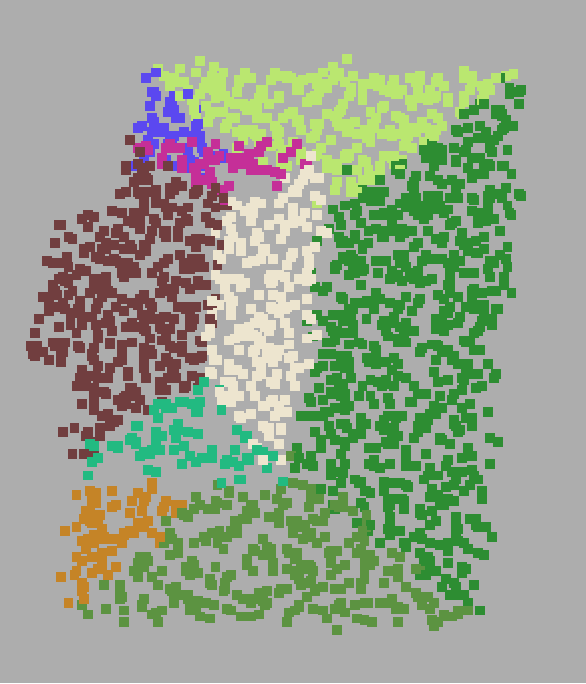}} &
		\multicolumn{1}{c}{\includegraphics[width= 0.175 \linewidth]{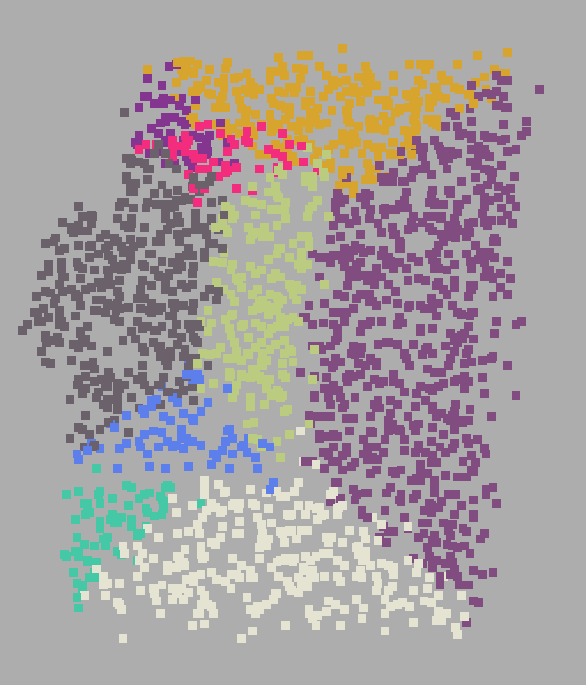}} 
		\\
		& & & &
		\\
		\multicolumn{1}{c}{\includegraphics[width= 0.175 \linewidth]{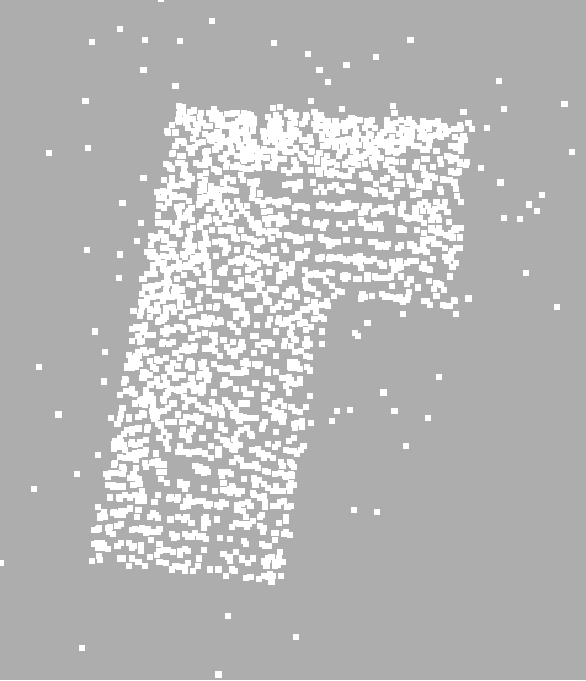}} &
		\multicolumn{1}{c}{\includegraphics[width= 0.175 \linewidth]{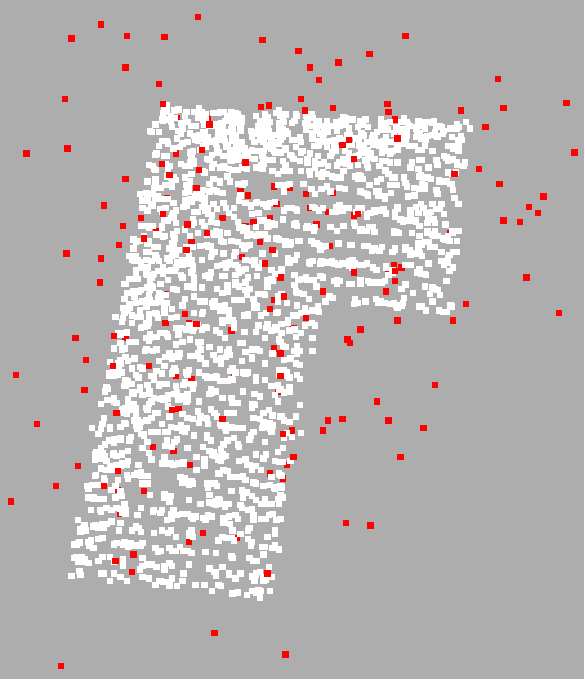}} &
		\multicolumn{1}{c}{\includegraphics[width= 0.175 \linewidth]{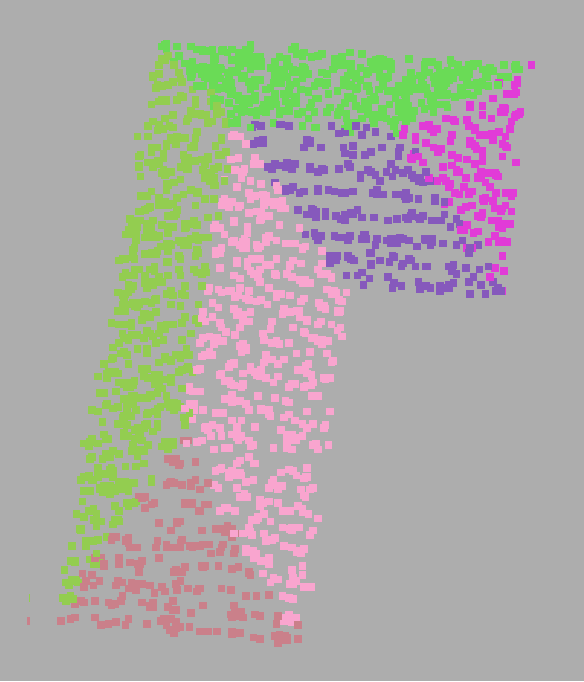}} &
		\multicolumn{1}{c}{\includegraphics[width= 0.175 \linewidth]{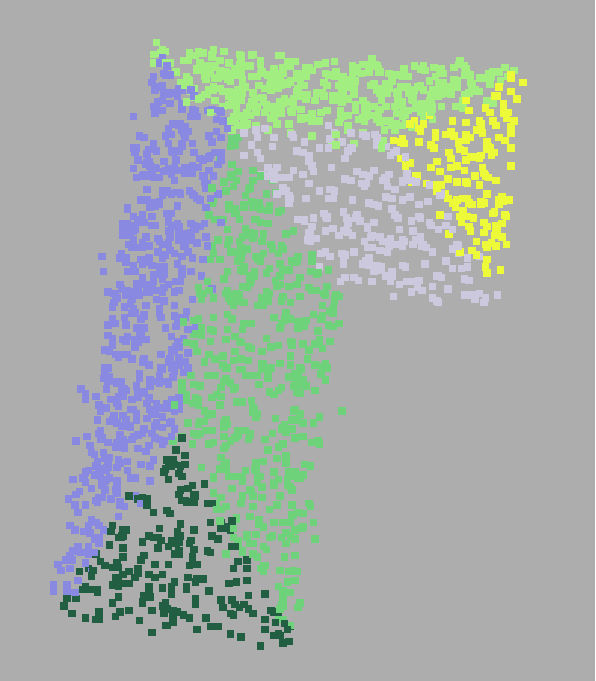}} &
		\multicolumn{1}{c}{\includegraphics[width= 0.175 \linewidth]{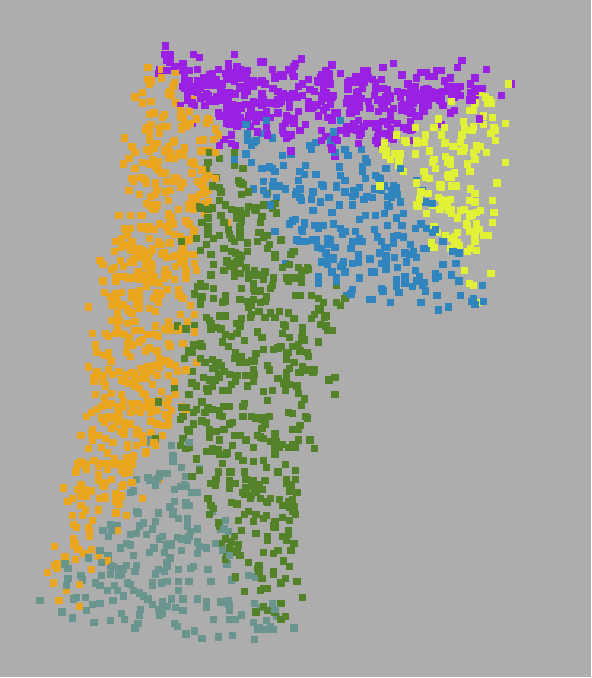}} 
		\\
		& & & &
		\\
		\multicolumn{1}{c}{\includegraphics[width= 0.175 \linewidth]{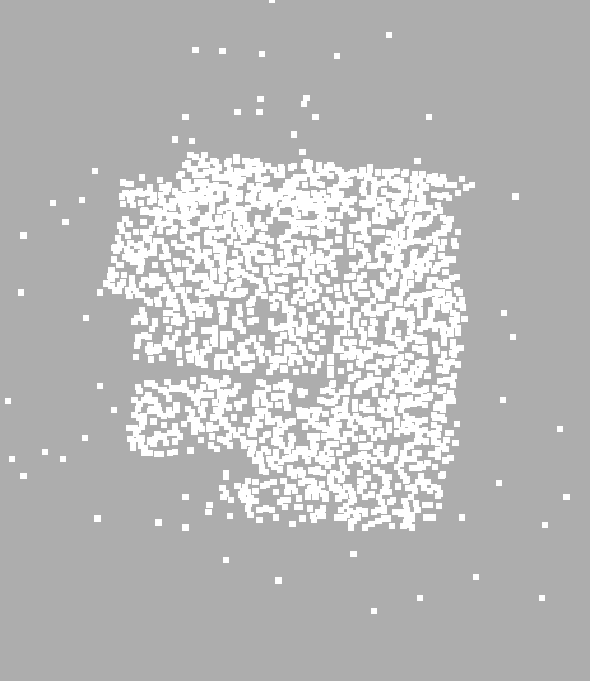}} &
		\multicolumn{1}{c}{\includegraphics[width= 0.175 \linewidth]{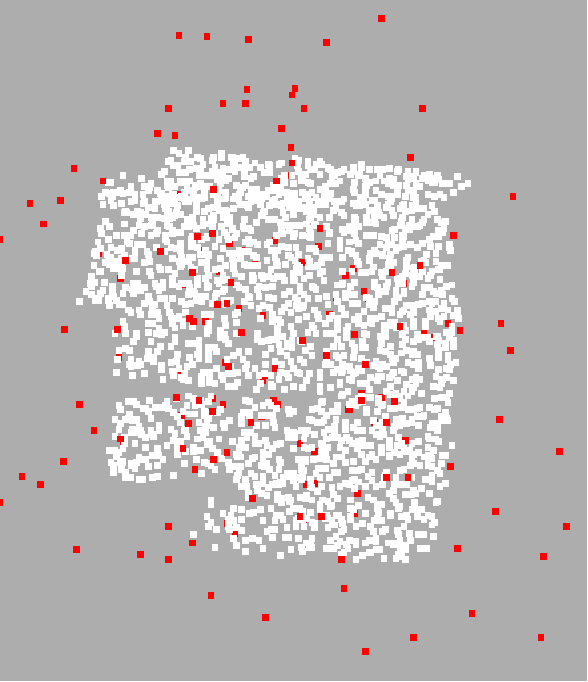}} &
		\multicolumn{1}{c}{\includegraphics[width= 0.175 \linewidth]{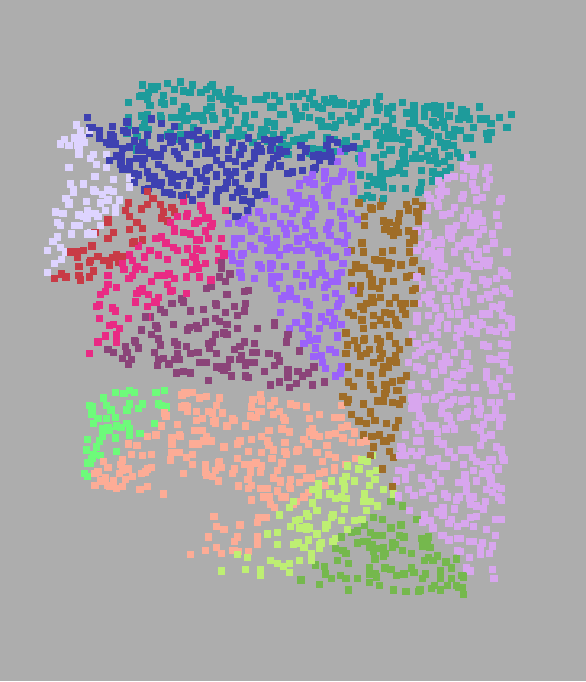}} &
		\multicolumn{1}{c}{\includegraphics[width= 0.175 \linewidth]{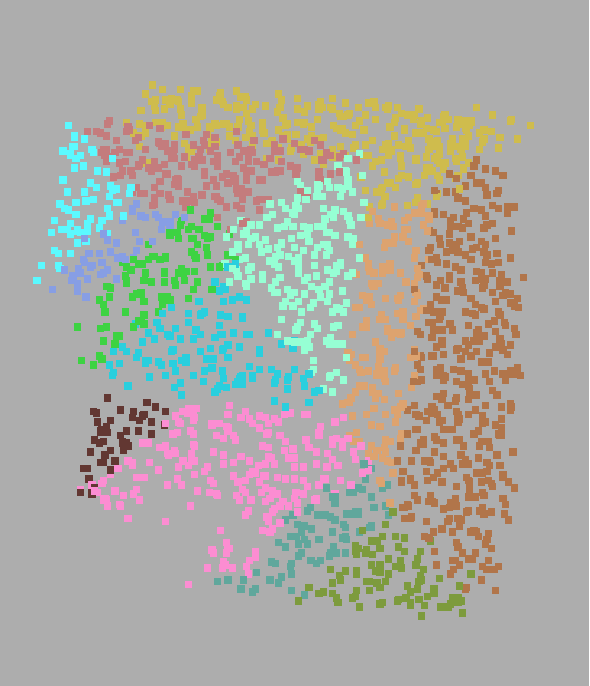}} &
		\multicolumn{1}{c}{\includegraphics[width= 0.175 \linewidth]{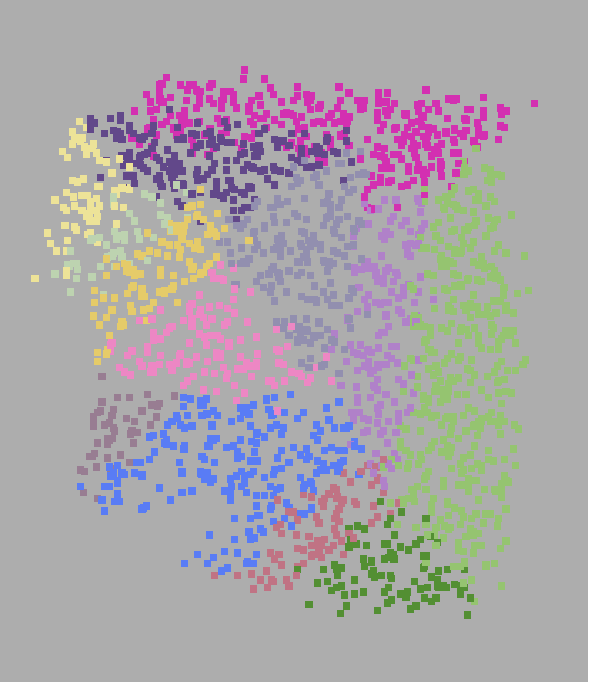}} 
		\\
		(a) Input & (b) Outlier detec. & (c) Noise (0.5\%) & (d) Noise (1.0\%) & (e) Noise (1.5\%) 
		
	\end{tabular} 
	\caption{Robustness to outliers and noises. (b) presents the potential outliers detected from the input point clouds (a). The (c), (d) and (e) shows the segmentation results with different levels of noise ($\alpha$ varied from 0.5\% to 1.5\%).}
	\vspace{0.0em}
	\label{Fig:Robust_noise}
\end{figure}

\textbf{Noise.} LiDAR point cloud noise can be categorized into two types: outliers that lie far from the underlying planes, and near-surface noise caused by ranging fluctuations or fine-scale geometric variations on surfaces. Notably, these two types of noise are handled with different strategies. Outliers should be correctly identified and removed rather than being assigned to any plane. This is handled by the network's outlier detection branch trained with $\mathcal{L}_{outlier}$. Near-surface noise, on the other hand, should be accurately assigned to its nearest plane despite slight perturbations. Due to the scarcity of real noise data, we introduce random Gaussian noise into the existing roof datasets to simulate the two types of noise described above. The threshold of noise $\sigma$ is defined as:
\begin{equation}
	\sigma = \alpha \cdot L_{\text{diag}},
\end{equation}
where $\alpha$ is the scaling factor, and $L_{\text{diag}}$ represents the diagonal length of the point cloud's bounding box. In practice, for generating outliers, we randomly selected a certain number (varied from 100 to 300) of points from the entire scene and added them with a larger noise (varied from $5\sigma$ to $15\sigma$), while the other points are added with a smaller noise (varied from $0$ to $\sigma$) to stimulate near-surface noises. In the experiments, we evaluated the performance of RoofSeg under different noise levels. As shown in the results of Table~\ref{Tab:Robust_noise} and ~\ref{Tab:Robust_noise_add_metrics}, although the network's performance degrades to some extent as the noise intensity increases, the overall metric scores remain excellent, demonstrating the robustness of our network against noise. This conclusion is further supported by the visual results presented in Fig.~\ref{Fig:Robust_noise}.

\begin{table} [!htb] \renewcommand{\tabcolsep}{3.0 pt}
	\scriptsize
	\renewcommand{\arraystretch}{1.25}
	\newcommand{\tabincell}[2]{\begin{tabular}{@{}#1@{}}#2\end{tabular}}
	\begin{center}
		\caption{Quantitative assessment (mCov, mWCov, mPrec and mRec) of RoofSeg's robustness to different noise levels on three roof benchmarks. The "baseline" represents the network's performance without added noises.}
		\label{Tab:Robust_noise}
		\begin{tabular}{c|cccc|cccc|cccc} 
			\hline
			\multirow{2}{*}{Varied noise levels} & \multicolumn{4}{c}{RoofNTNU benchmark}  &  \multicolumn{4}{c}{Roofpc3D benchmark} & \multicolumn{4}{c}{Building3D benchmark}\\
			\cline{2-5} \cline{6-9} \cline{10-13} 
			& mCov     & mWCov  &   mPrec   &    mRec      
			& mCov       & mWCov  &   mPrec   &    mRec   
			& mCov       & mWCov  &   mPrec   &    mRec    \\ 
			\hline
			baseline  & \textbf{95.89}  & \textbf{96.82} & \textbf{99.60} & \textbf{98.18} & \textbf{96.01} & \textbf{97.56} & \textbf{99.98} & \textbf{98.54} & \textbf{93.74} & \textbf{96.37} & \textbf{98.79} & \textbf{96.58}   \\ 
			\cline{1-1}
			$\alpha = 0.5\%$ & 95.34  & 96.10 & 99.37 & 98.09 & 95.89 & 97.05 & 99.36 & 98.32 & 93.29 & 95.98 & 98.32 & 96.27     \\
			\cline{1-1}
			$\alpha = 1.0\%$  & 94.83  & 95.78 & 98.92 & 97.63 & 94.98 & 96.45 & 98.01 & 97.55 & 92.64 & 95.07 & 97.88 & 95.62  \\ 
			\cline{1-1} 
			$\alpha = 1.5\%$  & 93.75  & 94.53 & 97.70 & 96.34 & 94.01 & 95.17 & 97.61 & 96.41 & 92.11 & 93.34 & 97.38 & 94.20 \\ 
			\hline
		\end{tabular}
	\end{center}
\end{table}

\begin{table} [!htb] \renewcommand{\tabcolsep}{8.0 pt}
	\scriptsize
	\renewcommand{\arraystretch}{1.25}
	\newcommand{\tabincell}[2]{\begin{tabular}{@{}#1@{}}#2\end{tabular}}
	\begin{center}
		\caption{Quantitative assessment ($O_{seg}$, $U_{seg}$ and $E_{acc}$) of RoofSeg's robustness to different noise levels on three roof benchmarks. The "baseline" represents the network's performance without added noises.}
		\label{Tab:Robust_noise_add_metrics}
		\begin{tabular}{c|ccc|ccc|ccc} 
			\hline
			\multirow{2}{*}{Varied noise levels} & \multicolumn{3}{c}{RoofNTNU benchmark}  &  \multicolumn{3}{c}{Roofpc3D benchmark} & \multicolumn{3}{c}{Building3D benchmark}\\
			\cline{2-4} \cline{5-7} \cline{8-10} 
			& $O_{seg}\downarrow$  & $U_{seg}\downarrow$   &   $E_{acc}\uparrow$      
			& $O_{seg}\downarrow$ & $U_{seg}\downarrow$   &   $E_{acc}\uparrow$  
			& $O_{seg}\downarrow$  & $U_{seg}\downarrow$   &   $E_{acc}\uparrow$      \\ 
			\hline			
			baseline   & \textbf{1.21}  & \textbf{2.39} & \textbf{94.90} & \textbf{2.45} & \textbf{2.69} & \textbf{95.36} & \textbf{3.57} & \textbf{4.38} & \textbf{92.53}  \\ 
			\cline{1-1}
			$\alpha = 0.5\%$  & 1.28  & 2.43 & 94.17 & 2.52 & 2.78 & 95.21 & 3.67 & 4.53 & 91.84  \\ 
			\cline{1-1}
			$\alpha = 1.0\%$  & 1.44  & 2.98 & 93.01 & 2.99 & 3.07 & 94.13 & 4.12 & 4.88 & 91.15  \\ 
			\cline{1-1} 
			$\alpha = 1.5\%$  & 1.98  & 3.33 & 92.74 & 3.49 & 4.61 & 93.28 & 4.75 & 5.43 & 90.06  \\ 
			\hline
		\end{tabular}
	\end{center}
\end{table}

\begin{figure}[h!]
	\scriptsize
	\centering
	\begin{tabular}{ccccc} 	
		\multicolumn{1}{c}{\includegraphics[width= 0.175 \linewidth]{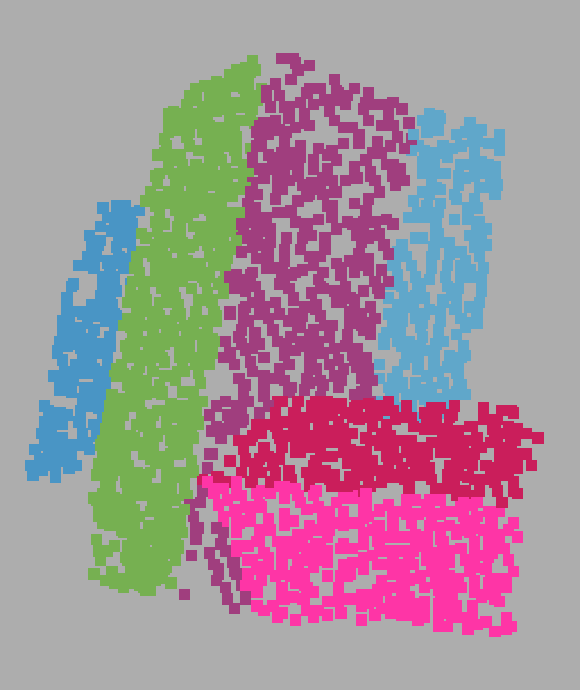}} &
		\multicolumn{1}{c}{\includegraphics[width= 0.175 \linewidth]{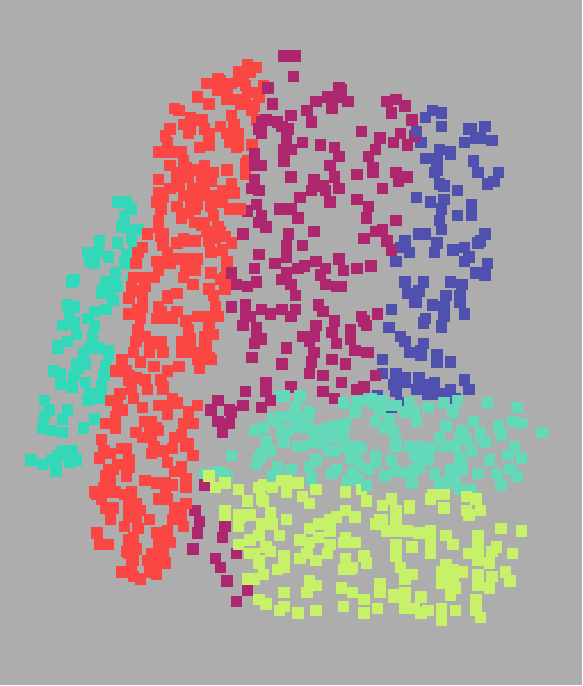}} &
		\multicolumn{1}{c}{\includegraphics[width= 0.175 \linewidth]{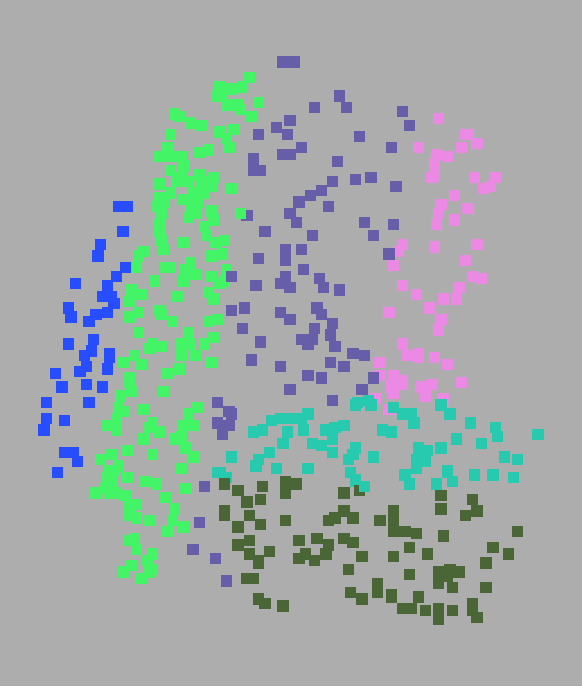}} &
		\multicolumn{1}{c}{\includegraphics[width= 0.175 \linewidth]{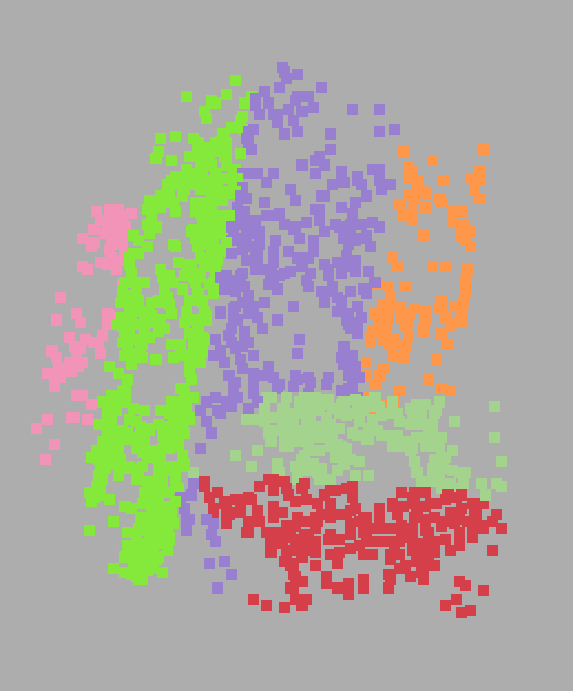}} &
		\multicolumn{1}{c}{\includegraphics[width= 0.175 \linewidth]{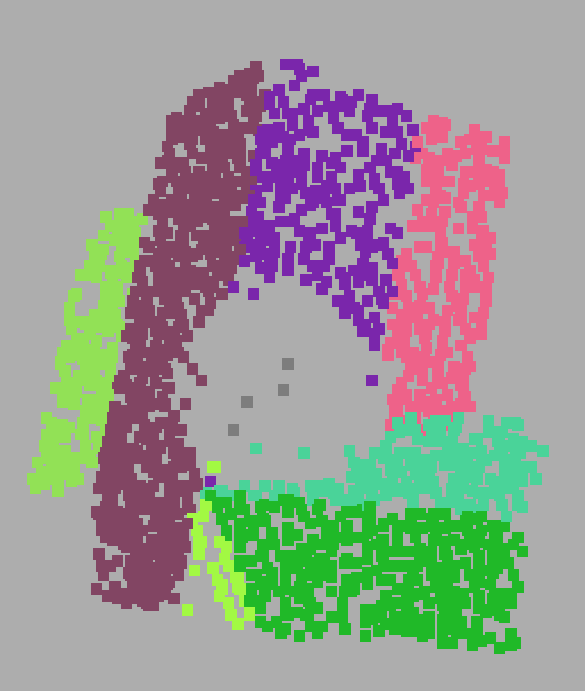}} 
		\\
		& & & &
		\\
		\multicolumn{1}{c}{\includegraphics[width= 0.175 \linewidth]{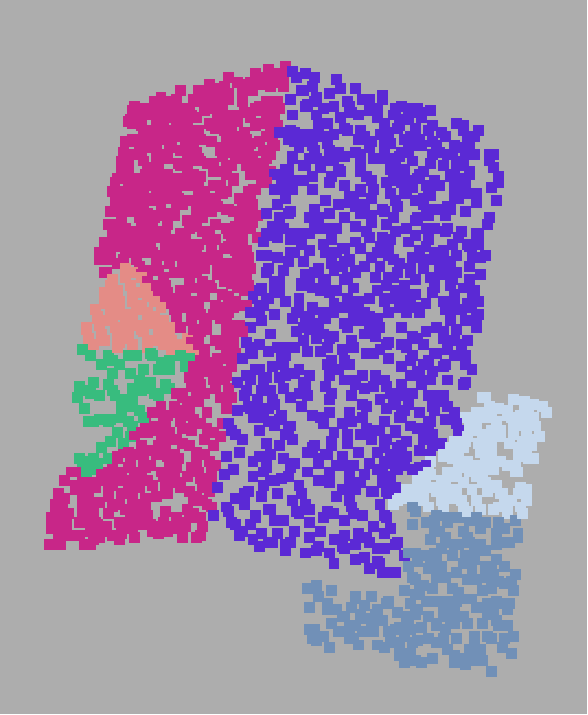}} &
		\multicolumn{1}{c}{\includegraphics[width= 0.175 \linewidth]{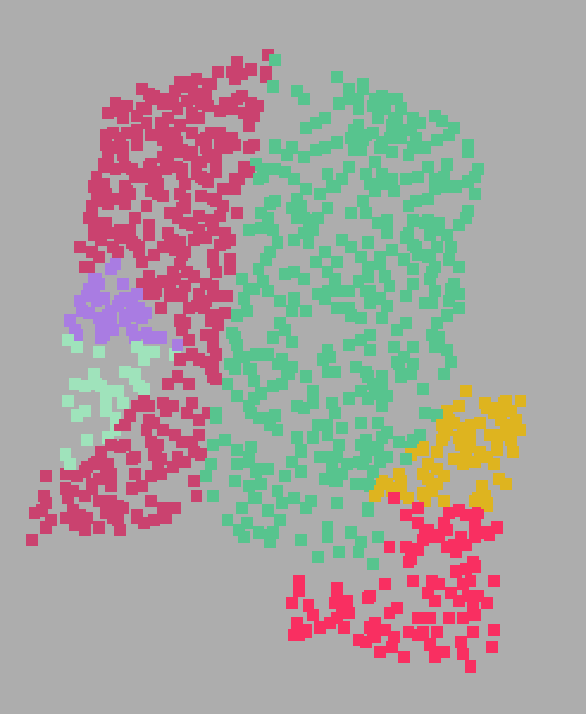}} &
		\multicolumn{1}{c}{\includegraphics[width= 0.175 \linewidth]{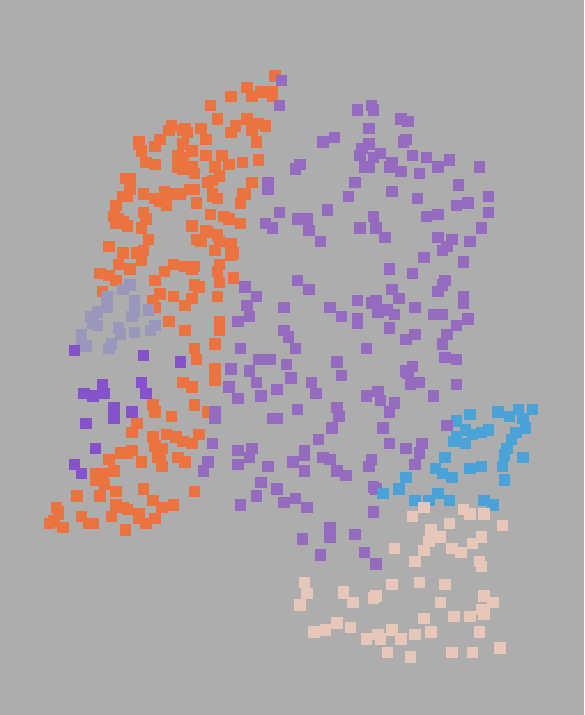}} &
		\multicolumn{1}{c}{\includegraphics[width= 0.175 \linewidth]{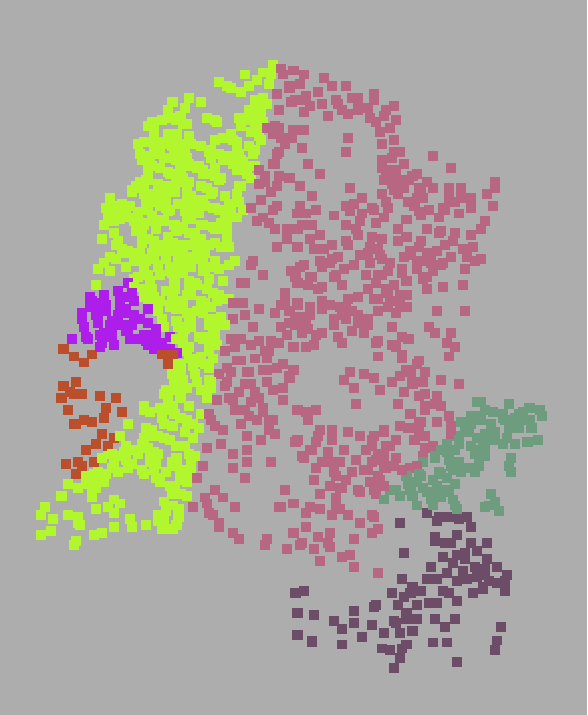}} &
		\multicolumn{1}{c}{\includegraphics[width= 0.175 \linewidth]{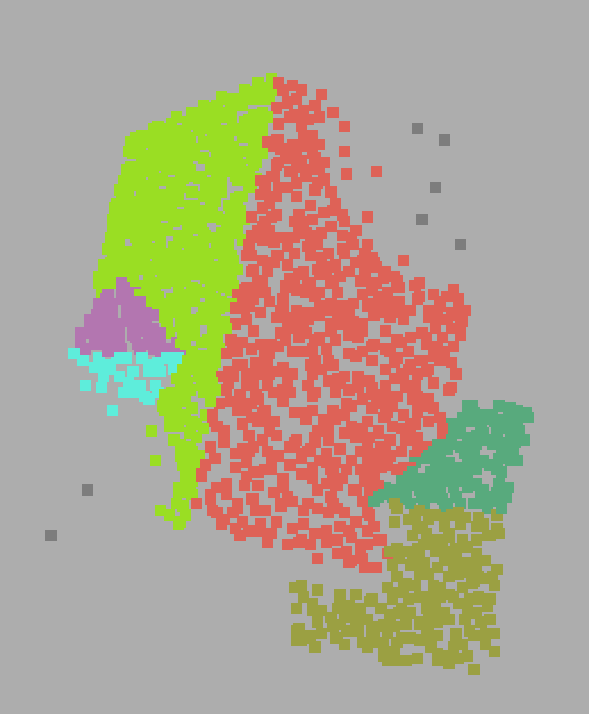}} 
		\\
		& & & &
		\\
		\multicolumn{1}{c}{\includegraphics[width= 0.175 \linewidth]{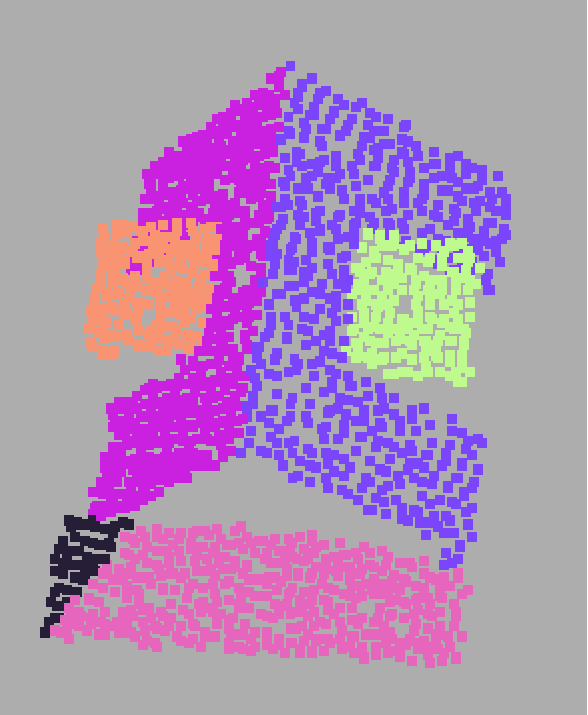}} &
		\multicolumn{1}{c}{\includegraphics[width= 0.175 \linewidth]{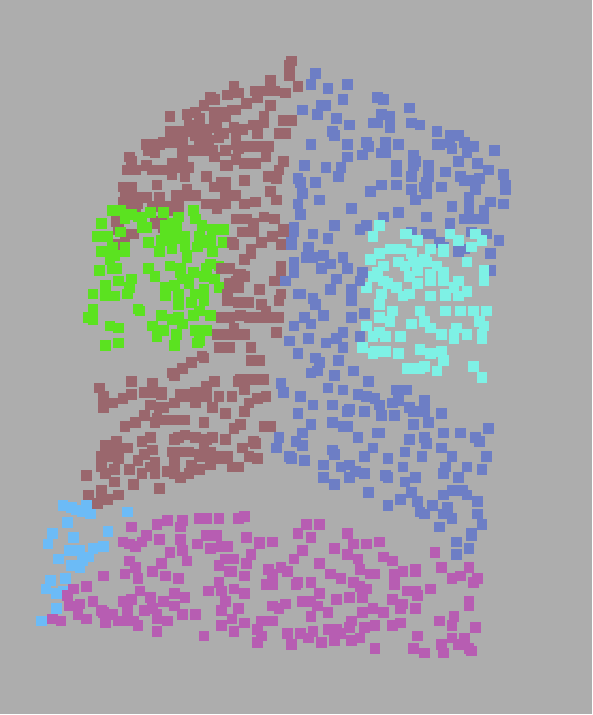}} &
		\multicolumn{1}{c}{\includegraphics[width= 0.175 \linewidth]{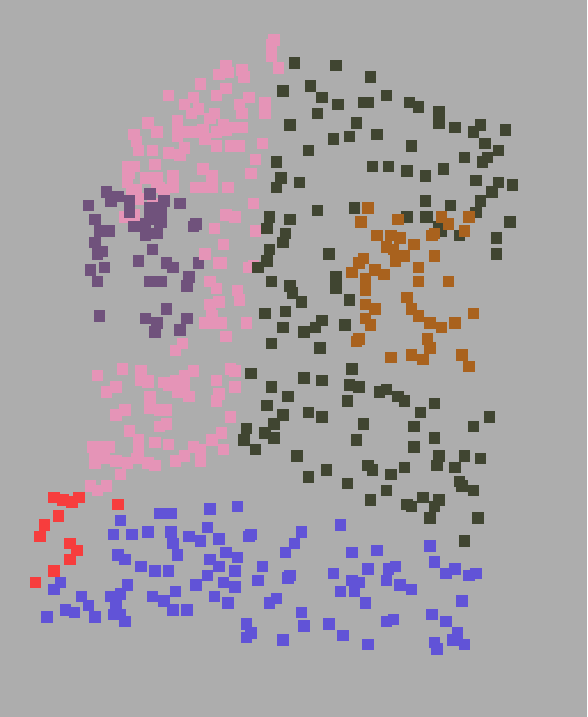}} &
		\multicolumn{1}{c}{\includegraphics[width= 0.175 \linewidth]{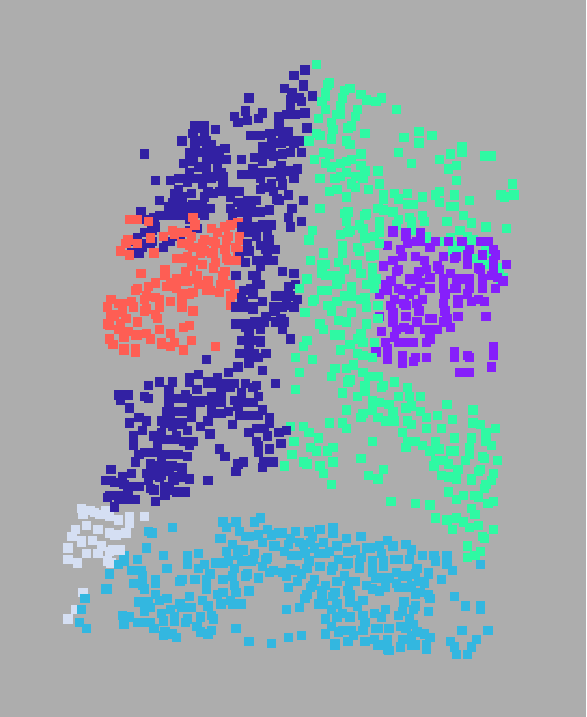}} &
		\multicolumn{1}{c}{\includegraphics[width= 0.175 \linewidth]{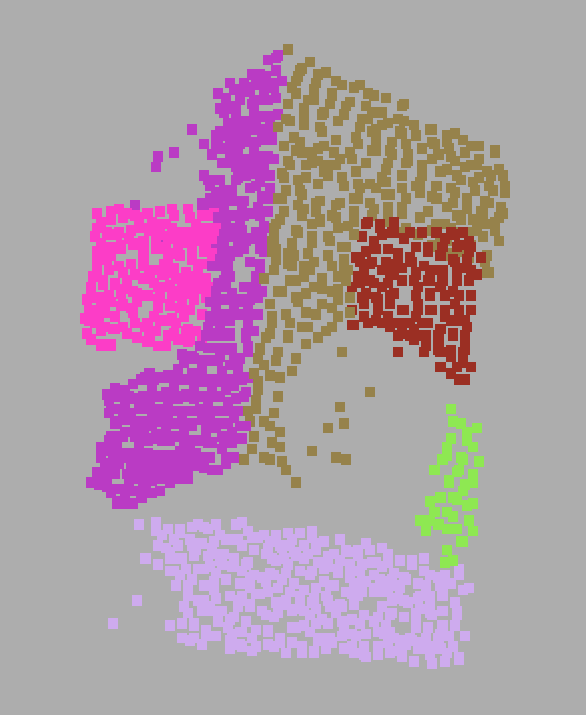}} 
		\\
		(a) Ground truth & (b) Density (50\%) & (c) Density (25\%) & (d) Non-uniform & (e) Occlusions \\
		& & & &\\
		\hdashline 
		& & & &\\
		\multicolumn{5}{c}{\includegraphics[width= 0.98 \linewidth]{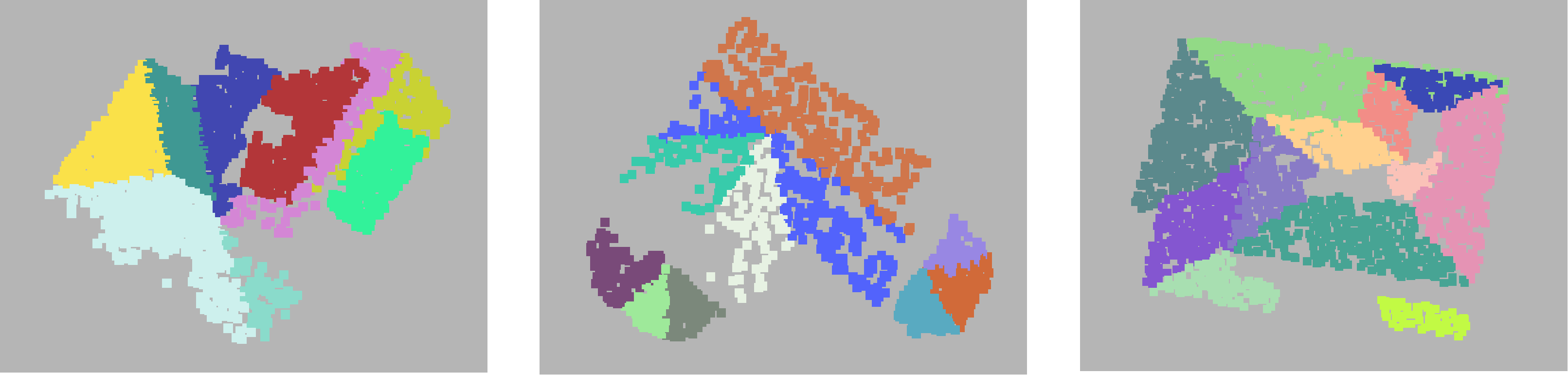}} 
		
	\end{tabular} 
	\caption{Robustness to varying point density, non-uniform sampling and occlusions. For the first three rows, we present three roof samples with ground truth (a) and the segmentation results under 50\% density (b), 25\% density (c), non-uniform sampling (d) and stimulated occlusions (e), respectively. At the last row, we provide segmentation results for occlusions occurred in the real-world benchmark.}
	\vspace{0.0em}
	\label{Fig:Robust_density}
\end{figure}

\textbf{Varying density, non-uniform sampling and occlusions.} These three issues can essentially be attributed to the sparsity of point clouds. To validate the robustness of our network against these three challenges, we designed four groups of comparative experiments. The first two groups apply uniform down‑sampling to the original inputs at ratios of 1/2 and 1/4, resulting in sparse point clouds with 50\% and 25\% of the original density, respectively. The third group performs non‑uniform sampling by allowing the local sampling ratio to vary randomly. The last group simulates occlusions caused by obstacles (e.g., trees or artificial structures) or limited field of view, which is achieved by randomly selecting seeds and then removing the neighboring points within a spherical neighborhood (radius = 1/4 $L_{\text{diag}}$). As the results reported in Table~\ref{Tab:Robust_density} and ~\ref{Tab:Robust_density_add_metrics}, 
RoofSeg demonstrates considerable robustness against all three types of sparsity issues. However, it should be noted that in cases of severe occlusion or excessively sparse points across large areas, the network may inevitably split a single plane into two or more patches, leading to over‑segmentation, as illustrated in the row 3, column 5 of Fig.~\ref{Fig:Robust_density}. This issue also remain challenging for current mainstream approaches.

\begin{table} [!htb] \renewcommand{\tabcolsep}{3.0 pt}
	\scriptsize
	\renewcommand{\arraystretch}{1.25}
	\newcommand{\tabincell}[2]{\begin{tabular}{@{}#1@{}}#2\end{tabular}}
	\begin{center}
		\caption{Quantitative assessment (mCov, mWCov, mPrec and mRec) of RoofSeg's robustness to varying point density, non-uniform sampling and occlusions on three roof benchmarks.}
		\label{Tab:Robust_density}
		\begin{tabular}{c|cccc|cccc|cccc} 
			\hline
			\multirow{2}{*}{Type of sparsity} & \multicolumn{4}{c}{RoofNTNU benchmark}  &  \multicolumn{4}{c}{Roofpc3D benchmark} & \multicolumn{4}{c}{Building3D benchmark}\\
			\cline{2-5} \cline{6-9} \cline{10-13} 
			& mCov     & mWCov  &   mPrec   &    mRec      
			& mCov       & mWCov  &   mPrec   &    mRec   
			& mCov       & mWCov  &   mPrec   &    mRec    \\ 
			\hline
			baseline & \textbf{95.89}  & \textbf{96.82} & \textbf{99.60} & \textbf{98.18} & \textbf{96.01} & \textbf{97.56} & \textbf{99.98} & \textbf{98.54} & \textbf{93.74} & \textbf{96.37} & \textbf{98.79} & \textbf{96.58}   \\ 
			\cline{1-1}
			Density 50\% & 95.19  & 96.42 & 99.20 & 98.33 & 95.67 & 97.55 & 99.18 & 97.95 & 93.01 & 96.17 & 98.25 & 96.32     \\
			\cline{1-1}
			Density 25\%  & 94.23  & 95.47 & 98.03 & 97.28 & 94.24 & 96.37 & 98.52 & 96.98 & 92.10 & 95.32 & 97.93 & 95.30  \\ 
			\cline{1-1} 
			Non-uniform  & 95.05  & 96.27 & 98.87 & 97.66 & 95.13 & 96.75 & 98.63 & 97.72 & 92.84 & 95.99 & 98.23 & 95.76 \\ 
			\cline{1-1}
			Occlusions  & 93.14  & 95.07 & 97.36 & 95.21 & 92.50 & 95.46 & 98.06 & 96.17 & 90.89 & 92.73 & 97.24 & 94.00 \\
			\hline
		\end{tabular}
	\end{center}
\end{table}

\begin{table} [!htb] \renewcommand{\tabcolsep}{8.0 pt}
	\scriptsize
	\renewcommand{\arraystretch}{1.25}
	\newcommand{\tabincell}[2]{\begin{tabular}{@{}#1@{}}#2\end{tabular}}
	\begin{center}
		\caption{Quantitative assessment ($O_{seg}$, $U_{seg}$ and $E_{acc}$) of RoofSeg's robustness to varying point density, non-uniform sampling and occlusions on three roof benchmarks.}
		\label{Tab:Robust_density_add_metrics}
		\begin{tabular}{c|ccc|ccc|ccc} 
			\hline
			\multirow{2}{*}{Type of sparsity} & \multicolumn{3}{c}{RoofNTNU benchmark}  &  \multicolumn{3}{c}{Roofpc3D benchmark} & \multicolumn{3}{c}{Building3D benchmark}\\
			\cline{2-4} \cline{5-7} \cline{8-10} 
			& $O_{seg}\downarrow$  & $U_{seg}\downarrow$   &   $E_{acc}\uparrow$      
			& $O_{seg}\downarrow$ & $U_{seg}\downarrow$   &   $E_{acc}\uparrow$  
			& $O_{seg}\downarrow$  & $U_{seg}\downarrow$   &   $E_{acc}\uparrow$      \\ 
			\hline			
			baseline   & \textbf{1.21}  & \textbf{2.39} & \textbf{94.90} & \textbf{2.45} & \textbf{2.69} & \textbf{95.36} & \textbf{3.57} & \textbf{4.38} & \textbf{92.53}  \\ 
			\cline{1-1}
			Density 50\%  & 1.38  & 2.97 & 93.21 & 2.83 & 3.28 & 94.87 & 3.98 & 4.71 & 91.05  \\ 
			\cline{1-1}
			Density 25\%  & 2.46  & 4.28 & 91.16 & 4.20 & 3.85 & 92.47 & 4.28 & 5.49 & 90.01  \\ 
			\cline{1-1} 
			Non-uniform  & 1.86  & 3.14 & 92.80 & 3.22 & 4.01 & 93.44 & 4.01 & 5.12 & 90.39  \\ 
			\cline{1-1}
			Occlusions  & 3.78  & 5.63 & 91.34 & 5.69 & 6.35 & 92.13 & 5.97 & 6.83 & 89.75  \\ 
			\hline
		\end{tabular}
	\end{center}
\end{table}

\begin{figure}[h!]
	\scriptsize
	\centering
	\begin{tabular}{cccc} 	
		\multicolumn{2}{c}{\includegraphics[width= 0.47 \linewidth]{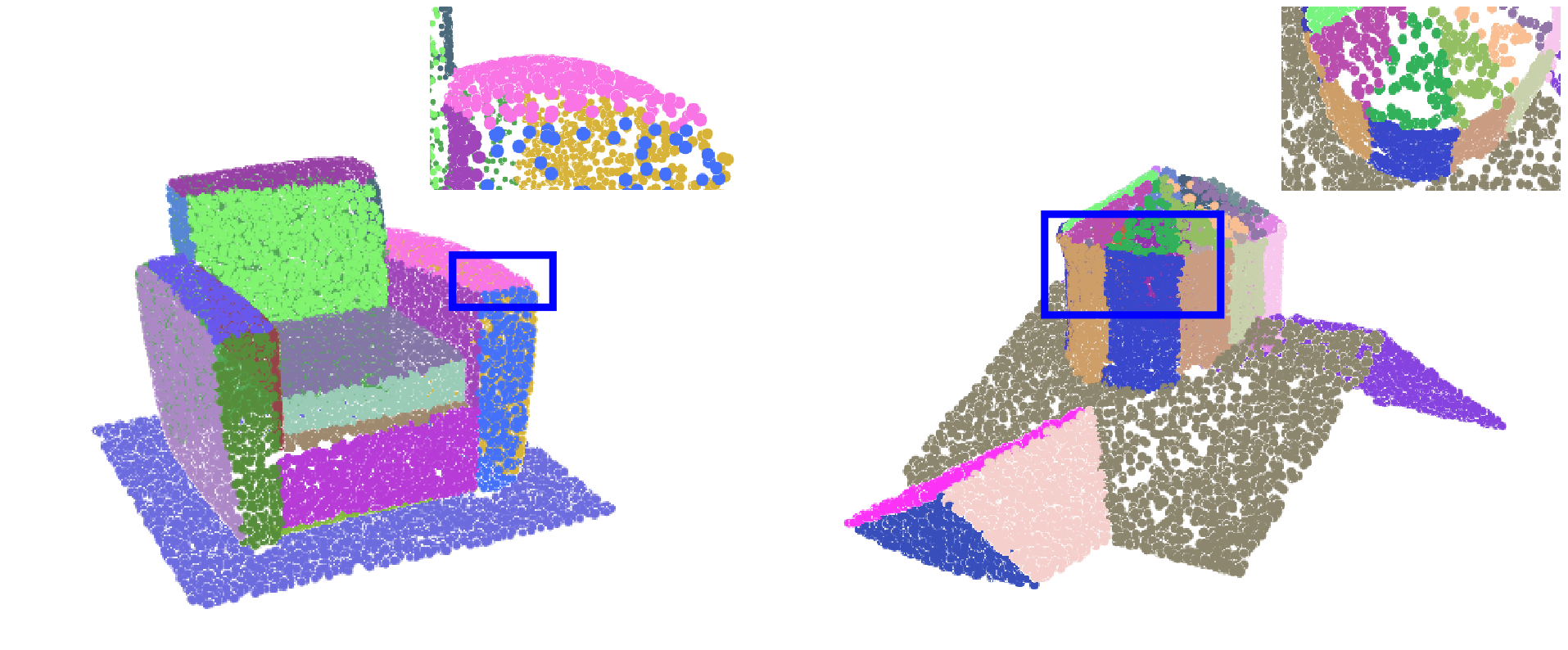}} & \multicolumn{2}{c}{\includegraphics[width= 0.47 \linewidth]{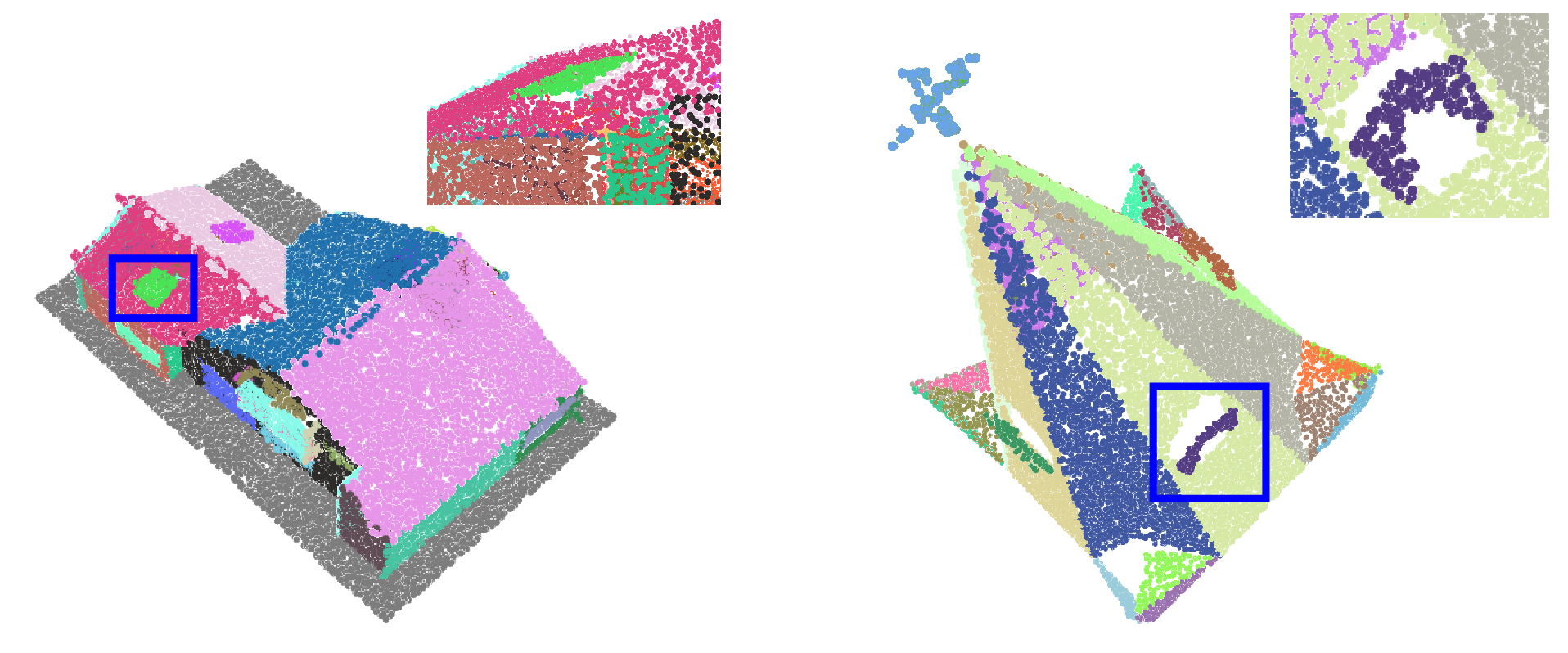}}
		\\
		\multicolumn{2}{c}{(a) Curvature} & \multicolumn{2}{c}{(b) Roof openings}
		\\
		& & & \\
		\hdashline 
		& & & \\
		\multicolumn{4}{c}{\includegraphics[width= 0.98 \linewidth]{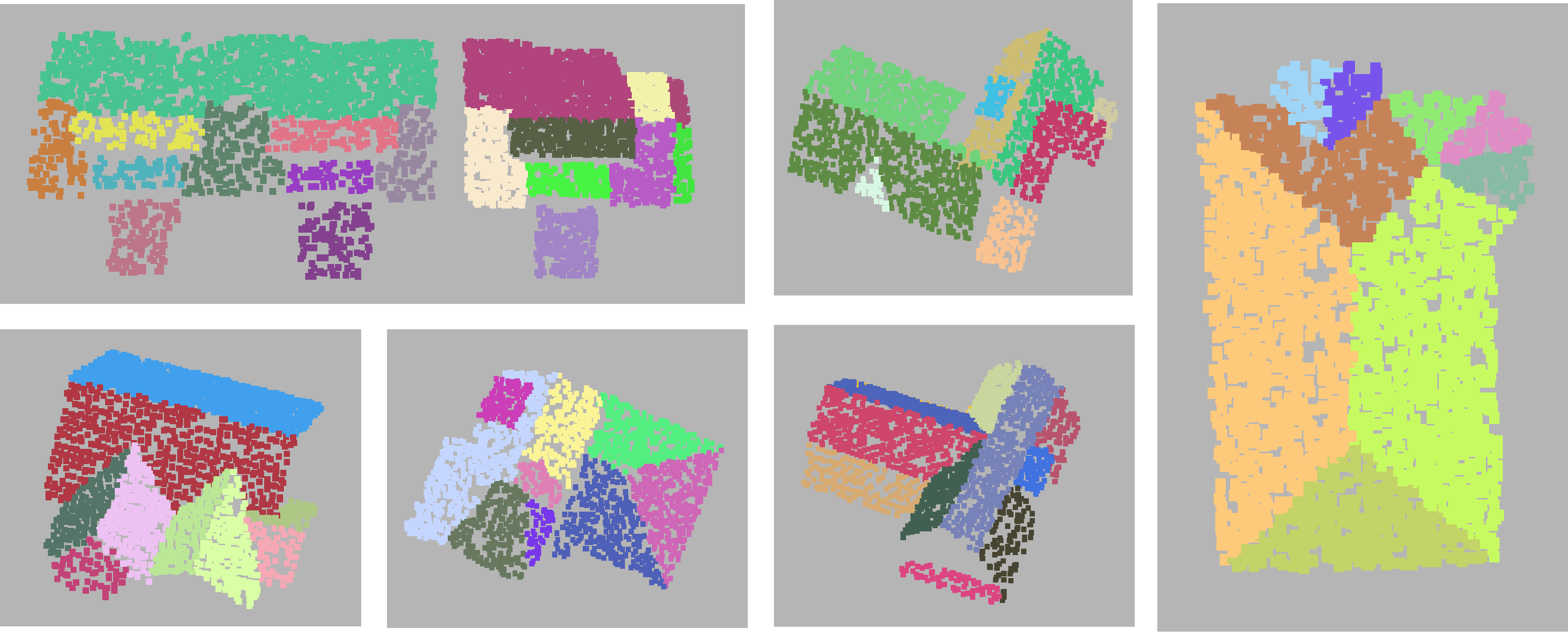}} 
		\\
		\multicolumn{4}{c}{(c) More complex structures} 
		
	\end{tabular} 
	\caption{Robustness to curvature and geometric ambiguities. (a) and (b) are segmentation results (on KSR42 benchmark) with curvature and roof openings, while (c) show several results for more complex roof structures on the Building3D benchmark.}
	\vspace{0.0em}
	\label{Fig:Robust_complex}
\end{figure}

\textbf{Curvature and geometric ambiguities.} Not limited to flat planar structures, RoofSeg is also effective in handling curved surface patches, as shown in Fig.~\ref{Fig:Robust_complex}(a). For geometrically ambiguous structures such as roof openings, our network still achieves accurate segmentation, as illustrated in Fig.~\ref{Fig:Robust_complex}(b). Furthermore, we present some segmentation results on challenging roof samples from the real-world benchmark in Fig.~\ref{Fig:Robust_complex}(c). It demonstrates the strong adaptability of our network to roof point clouds with complex geometries, especially in segmenting small patches.

\subsection{Efficiency and accuracy analysis}
\label{appendix:efficiency_and_accuracy_analysis}

\textbf{The effects of different numbers of Query Refinement Decoders (QRD).} In our network, we employ transformer decoders to hierarchically refine the plane queries with the multi-scale point features. To evaluate the effects of varied decoder numbers ($N_{dec}$) on the performance of RoofSeg, we conducted several experiments, with $N_{dec}$ set to 4, 8 and 12 respectively. As shown in Table~\ref{Tab:efficiency_decoder}, with the number of decoders increases, both the parameters and computational complexity of the network rise significantly, leading to increased time costs and resource consumption. From Table~\ref{Tab:accuarcy_decoder}, we observed that using 8 decoders can achieve better performance compared to using 4 decoders, while further increasing $N_{dec}$ does not provide a significant improvement. Thereby, we empirically set $N_{dec}$ to 8 for our network under the consideration of both computational efficiency and segmentation accuracy. 

\begin{table} [!htb] \renewcommand{\tabcolsep}{2.2 pt}
	\scriptsize
	\renewcommand{\arraystretch}{1.4}
	\newcommand{\tabincell}[2]{\begin{tabular}{@{}#1@{}}#2\end{tabular}}
	\newcolumntype{C}[1]{>{\centering\arraybackslash}p{#1}}
	\begin{center}
		\caption{Efficiency analysis of RoofSeg employing different numbers of Query Refinement Decoders (QRD). \# Para. represents the total parameter amount of RoofSeg. FLOPs represents the count of floating-point calculations performed during the forward pass in one second. Infer. time represents the average processing time per batch (batch size=24), while GPU mem. denotes the memory consumption on a single GPU during training. }
		\label{Tab:efficiency_decoder}
		\begin{tabular}{C{0.2\textwidth}|C{0.18\textwidth}C{0.18\textwidth}C{0.18\textwidth}C{0.18\textwidth}} 
			\hline
			Decoder numbers & \# Para.(MB)  &  FLOPs (GB) & Infer. time (s) & GPU mem. (MB)\\
			\hline 
			
			4  & 22.855  & 16.095 & 1.633 & 13303.05 \\ 
			
			8  & 32.926  & 18.118 & 2.944  & 14091.12 \\ 
			12  & 42.996  & 20.141 & 4.174 &  15262.72  \\ 
			\hline
		\end{tabular}
	\end{center}
\end{table}

\begin{table} [!htb] \renewcommand{\tabcolsep}{3.0pt}
	\scriptsize
	\renewcommand{\arraystretch}{1.4}
	\newcommand{\tabincell}[2]{\begin{tabular}{@{}#1@{}}#2\end{tabular}}
	
	\begin{center}
		\caption{Accuracy analysis of RoofSeg employing different numbers of Query Refinement Decoders (QRD) on three benchmarks.}
		\label{Tab:accuarcy_decoder}
		\begin{tabular}{c|cccc|cccc|cccc} 
			\hline
			\multirow{2}{*}{\makecell[c]{Decoder numbers}} & \multicolumn{4}{c}{RoofNTNU benchmark}  &  \multicolumn{4}{c}{Roofpc3D benchmark} & \multicolumn{4}{c}{Building3D benchmark}\\
			\cline{2-5} \cline{6-9} \cline{10-13} 
			& mCov       & mWCov  &   mPrec   &    mRec      
			& mCov       & mWCov  &   mPrec   &    mRec   
			& mCov       & mWCov  &   mPrec   &    mRec    \\ 
			\hline
			4  & 91.78  & 92.21 & 96.74 & 93.03 & 92.46 & 94.10 & 96.15 & 93.47 & 89.14 & 91.39 & 96.07 & 92.99 \\ 
			
			8  & 95.89  & \textbf{96.82} & 99.60 & \textbf{98.18} & \textbf{96.01} & \textbf{97.56} & 99.98 & 98.54 & \textbf{93.74} & 96.37 & \textbf{98.79} & \textbf{96.58}  \\ 
			12 & \textbf{95.93}  & 96.71 & \textbf{99.64} & 98.13  & 95.97 & 97.31 & \textbf{99.99} & \textbf{98.65} & 93.69 & \textbf{96.41} & 98.73 & 96.42   \\ 
			
			\hline
		\end{tabular}
	\end{center}
\end{table}

\textbf{The effects of different point features (multi-scale/full resolution) applied for cross-attention operation in QRD.} Notably, in each QRD, the standard cross-attention operation has quadratic computational complexity $O(KN)$ with respect to the query number $K$ and point number $N$,  where $K << N$. Therefore, the key factor affecting the computational efficiency of each QRD is the number of input points. Compared to use point features with full resolution in all decoders, using multi-scale features across different decoders can greatly reduce the computational costs, as demonstrated in the Table~\ref{Tab:efficiency_scale}. From the table, it can also be found that with the growth of input point number, the efficiency advantage of using multi-scale features becomes more evident.
Regarding the segmentation accuracy, these two strategies achieve very close performance, as shown in Table~\ref{Tab:accuarcy_scale}. Given these results, it can be concluded that using multi-scale features is a better choice, which can achieve competitive performance on segmentation accuracy while ensuring higher network efficiency.

\begin{table} [!htb] \renewcommand{\tabcolsep}{0.3pt}
	\scriptsize
	\renewcommand{\arraystretch}{1.25}
	\newcommand{\tabincell}[2]{\begin{tabular}{@{}#1@{}}#2\end{tabular}}
	\newcolumntype{C}[1]{>{\centering\arraybackslash}p{#1}}
	\begin{center}
		\caption{Efficiency analysis of RoofSeg using different point features (multi-scale/full resolution) applied for cross-attention operations in QRD. In this table, the batch size is set to 1, considering the maximum GPU memory may not enough for processing multi-batch inputs with large point numbers. }
		\label{Tab:efficiency_scale}
		\begin{tabular}{C{0.10\textwidth}|C{0.15\textwidth}C{0.15\textwidth}C{0.15\textwidth}|C{0.15\textwidth}C{0.15\textwidth}C{0.15\textwidth}} 
			\hline
			\multirow{2}{*}{\makecell[c]{Input point \\ numbers}} & \multicolumn{3}{c}{Multi-scale features}  &  \multicolumn{3}{c}{Full resolution features} \\
			\cline{2-7}
			& FLOPs (GB)  & Infer. time (s)  &  GPU mem. (MB)   &  FLOPs (GB)  &   Infer. time (s)  &  GPU mem. (MB)     \\ 
			\hline
			2048  & 17.935 & 0.457 & 164.91 & 23.820 & 0.598 & 274.08  \\ 
			
			4096  & 36.247  & 0.571 & 555.01 & 47.356  & 0.656 & 798.12   \\ 
			6144  & 58.754  & 0.689 & 1201.11 & 78.198  & 0.789 & 1652.25   \\ 
			
			8192  & 97.106  & 0.705 & 2103.21 & 121.627  & 0.823 & 3287.30 \\
			
			\hline
		\end{tabular}
	\end{center}
\end{table}

\begin{table} [!htb] \renewcommand{\tabcolsep}{3.5pt}
	\scriptsize
	\renewcommand{\arraystretch}{1.4}
	\newcommand{\tabincell}[2]{\begin{tabular}{@{}#1@{}}#2\end{tabular}}
	
	\begin{center}
		\caption{Accuracy analysis of RoofSeg using different point features (multi-scale/full resolution) applied for cross-attention operations in QRD on three benchmarks. }
		\label{Tab:accuarcy_scale}
		\begin{tabular}{c|cccc|cccc|cccc} 
			\hline
			\multirow{2}{*}{\makecell[c]{Different point \\ features}} & \multicolumn{4}{c}{RoofNTNU benchmark}  &  \multicolumn{4}{c}{Roofpc3D benchmark} & \multicolumn{4}{c}{Building3D benchmark}\\
			\cline{2-5} \cline{6-9} \cline{10-13} 
			& mCov       & mWCov  &   mPrec   &    mRec      
			& mCov       & mWCov  &   mPrec   &    mRec   
			& mCov       & mWCov  &   mPrec   &    mRec    \\ 
			\hline
			Multi-scale feat.  & \textbf{95.89}  & 96.82 & \textbf{99.60} & \textbf{98.18} & 96.01 & 97.56 & \textbf{99.98} & \textbf{98.54} & 93.74 & \textbf{96.37} & \textbf{98.79} & 96.58  \\ 
			Full resolution feat. & 95.87  & \textbf{96.83} & 99.48 & 98.07 & \textbf{96.05} & \textbf{97.61} & 99.70 & 98.39 & \textbf{93.80} & 96.18 & 98.64 & \textbf{96.62}    \\ 
			
			\hline
		\end{tabular}
	\end{center}
\end{table}

\textbf{Performance and scalability.} To further evaluate the network's performance and scalability, we measured the time cost of the network with different-sized point sets input (varied from 5k to 25k points). As the results reported in Table.~\ref{Tab:performance_tradition}, compared to traditional algorithms running on the CPU, our network achieves a 3–5× speedup. Since our approach does not rely on normal estimation, this advantage becomes more pronounced as the number of points increases. When deployed on a GPU, the time cost can be further reduced significantly.
Compared to deep learning-based approaches relying on post-processing steps, such as clustering or voting mechanism, our end-to-end architecture reduces runtime by several tens of times, as shown in Table.~\ref{Tab:performance_deeplearning}. Even compared with Mask3D, which is also end-to-end, RoofSeg retains a slight advantage in both inference speed and resource use, demonstrating the network's good performance and scalability.

\begin{table} [!htb] \renewcommand{\tabcolsep}{2.2 pt}
	\scriptsize
	\renewcommand{\arraystretch}{1.25}
	\newcommand{\tabincell}[2]{\begin{tabular}{@{}#1@{}}#2\end{tabular}}
	\newcolumntype{C}[1]{>{\centering\arraybackslash}p{#1}}
	\begin{center}
		\caption{Performance comparison with the traditional approaches. The input data only contains 3D coordinates of the points, without normal vectors. For traditional approaches, Norm. est. time, Proc. time and Optim. time denotes the time cost of normal vector estimation, algorithm procession and energy optimization (GoCoPP). For our network, we provide the time cost of inference on CPU and GPU, respectively. Except for this table, Infer. time mentioned elsewhere is computed on GPU.}
		\label{Tab:performance_tradition}
		\begin{tabular}{C{0.15\textwidth}|C{0.20\textwidth}|C{0.11\textwidth}C{0.11\textwidth}C{0.11\textwidth}C{0.11\textwidth}C{0.11\textwidth}} 
			\hline
			\multicolumn{2}{c}{Input point numbers} & \multicolumn{1}{c}{5k}  &  \multicolumn{1}{c}{10k} & \multicolumn{1}{c}{15k} & \multicolumn{1}{c}{20k} & \multicolumn{1}{c}{25k} \\
			\hline 
			\multirow{3}{*}{\makecell[c]{Region growing}} & Norm. est. time (s) & 1.90 & 4.05 & 7.59 & 10.33 & 14.48 \\
			\cline{2-2} 
			& Proc. time (s) & 1.65 & 2.59 & 3.92 & 4.44 & 5.12 \\
			\cline{2-2} 
			& Total time (s) & 3.55 & 6.64 &11.51 & 14.77 & 19.60 \\
			\hline
			\multirow{3}{*}{\makecell[c]{RANSAC}} & Norm. est. time (s) & 2.16 & 3.97 & 7.68 & 10.38 & 14.23 \\
			\cline{2-2} 
			& Proc. time (s) & 1.15 & 1.76 & 2.59 & 3.09 & 4.26 \\
			\cline{2-2} 
			& Total time(s) & 3.31 & 5.93 &10.27 & 13.47 & 18.49 \\
			\hline
			\multirow{4}{*}{\makecell[c]{GoCoPP}} & Norm. est. time (s) & 2.02 & 3.98 & 7.73 & 10.28 & 14.33 \\
			\cline{2-2} 
			& Proc. time (s) & 1.74 & 2.48 & 3.87 & 4.50 & 5.35 \\
			\cline{2-2} 
			& Optim. time (s) & 1.91 & 4.20 & 8.52 & 12.26 & 16.18 \\
			\cline{2-2} 
			& Total time(s) & 5.67 & 10.66 & 20.12 & 27.04 & 35.86 \\
			\hline
			\multirow{2}{*}{\makecell[c]{RoofSeg}} & CPU Infer. time (s) & 1.04 & 2.13 & 3.22 & 4.73 & 6.25 \\
			\cline{2-2} 
			& GPU Infer. time (s) & \textbf{0.67} & \textbf{0.73} & \textbf{0.78} & \textbf{0.83} & \textbf{0.87} \\
			\hline
			
		\end{tabular}
	\end{center}
\end{table}

\begin{table} [!htb] \renewcommand{\tabcolsep}{2.2 pt}
	\scriptsize
	\renewcommand{\arraystretch}{1.25}
	\newcommand{\tabincell}[2]{\begin{tabular}{@{}#1@{}}#2\end{tabular}}
	\newcolumntype{C}[1]{>{\centering\arraybackslash}p{#1}}
	\begin{center}
		\caption{Performance comparison with the deep-learning-based approaches. Post. time denotes the time cost of post-processing step (clustering/voting mechanism). Notably, the end-to-end pipelines (Mask3D and our network) are only reported with the inference time.}
		\label{Tab:performance_deeplearning}
		\begin{tabular}{C{0.15\textwidth}|C{0.17\textwidth}|C{0.12\textwidth}C{0.12\textwidth}C{0.12\textwidth}C{0.12\textwidth}C{0.12\textwidth}} 
			\hline
			\multicolumn{2}{c}{Input point numbers} & \multicolumn{1}{c}{5k}  &  \multicolumn{1}{c}{10k} & \multicolumn{1}{c}{15k} & \multicolumn{1}{c}{20k} & \multicolumn{1}{c}{25k} \\
			\hline 
			\multirow{3}{*}{\makecell[c]{ParseNet}} & Infer. Time (s) & 0.54 & 0.67 & 0.82 & 1.00 & 1.27 \\
			\cline{2-2} 
			& Post. Time (s) & 31.58 & 74.72 & 153.57 & 231.29 & 330.75 \\
			\cline{2-2} 
			& GPU Mem. (MB) & 792.55 & 2298.96 & 5157.61 & 9164.59 & 14314.52 \\
			\hline 
			\multirow{3}{*}{\makecell[c]{HPNet}} & Infer. Time (s) & 0.62 & 0.79 & 0.93 & 1.15 & 1.32 \\
			\cline{2-2} 
			& Post. Time (s) & 42.34 & 111.72 & 198.70 & 283.25 & 405.28 \\
			\cline{2-2} 
			& GPU Mem. (MB) & 849.33 & 2337.18 & 5260.52 & 9351.94 & 14420.19 \\
			\hline
			\multirow{3}{*}{\makecell[c]{RoofDeepPlane}} & Infer. Time (s) & 0.49 & 0.55 & 0.67 & 0.72 & 0.79 \\
			\cline{2-2} 
			& Post. Time (s) & 15.34 & 42.10 & 89.57 & 136.35 & 208.22 \\
			\cline{2-2} 
			& GPU Mem. (MB) & 512.48 & 1933.61 & 3679.43 & 7868.50 & 11474.98 \\
			\hline
			\multirow{2}{*}{\makecell[c]{Mask3D}} & Infer. Time (s) & 0.71 & 0.79 & 0.86 & 0.97 & 1.10 \\
			\cline{2-2} 
			& GPU Mem. (MB) & 933.15 & 3899.42 & 8289.16 & 13911.28 & 19875.66 \\
			\hline
			\multirow{2}{*}{\makecell[c]{RoofSeg}} & Infer. Time (s) & \textbf{0.67} & \textbf{0.73} & \textbf{0.78} & \textbf{0.83} & \textbf{0.87} \\
			\cline{2-2} 
			& GPU Mem. (MB) & 809.70 & 3112.60 & 6943.49 & 12298.39 & 18179.12 \\
			\hline
			
		\end{tabular}
	\end{center}
\end{table}

\subsection{Extended ablation results}
\label{appendix:extended_ablation_results}
\textbf{The effects of using different numbers of plane queries for segmentation.} We trained the network with 16, 32, 64 and 128 plane queries for three benchmarks, respectively. The metric results are provided in Table~\ref{Tab:ablation_query}. On the RoofNTNU and Roofpc3D benchmarks, the network trained with 32 plane queries achieves the highest metric scores, while more plane queries (64) are required for achieving the best results on the Building3D benchmark, which contains complex roof structures with more planar patches. In addition, from the Table~\ref{Tab:ablation_query}, we observed that a small number (16) or a large number (128) of plane queries could weaken the performance of the network. This is because with few queries, under-segmentation may occur, potentially missing small planar patches. On the other hand, with excessive queries, redundant predictions are likely to arise, thus leading to over-segmentation. According to this observation, the number of queries K should be determined based on the number of planes present in the dataset. When scenes contain a large number of planes, a larger K is preferable; otherwise, a smaller K should be used.

\begin{table} [!htb] \renewcommand{\tabcolsep}{3.0 pt}
	\scriptsize
	\renewcommand{\arraystretch}{1.4}
	\newcommand{\tabincell}[2]{\begin{tabular}{@{}#1@{}}#2\end{tabular}}
	\begin{center}
		\caption{Quantitative assessment of RoofSeg using different numbers of plane queries on three benchmarks. }
		\label{Tab:ablation_query}
		\begin{tabular}{c|cccc|cccc|cccc} 
			\hline
			\multirow{2}{*}{\makecell[c]{Different numbers of \\ plane queries}} & \multicolumn{4}{c}{RoofNTNU benchmark}  &  \multicolumn{4}{c}{Roofpc3D benchmark} & \multicolumn{4}{c}{Building3D benchmark}\\
			\cline{2-5} \cline{6-9} \cline{10-13} 
			& mCov       & mWCov  &   mPrec   &    mRec      
			& mCov       & mWCov  &   mPrec   &    mRec   
			& mCov       & mWCov  &   mPrec   &    mRec    \\ 
			\hline
			16  & 92.78  & 94.12 & 97.43 & 96.19 & 93.88 & 95.52 & 98.40 & 95.38 & 91.36 & 93.83 & 97.01 & 94.57 \\ 
			
			32  & \textbf{95.89}  & \textbf{96.82} & \textbf{99.60} & \textbf{98.18} & \textbf{96.01} & \textbf{97.56} & \textbf{99.98} & \textbf{98.54} & 93.01 & 95.90 & 98.20 & 96.13 \\ 
			64  & 94.38  & 95.45 & 97.92 & 96.72 & 95.06 & 96.14 & 98.22 & 96.80 & \textbf{93.74} & \textbf{96.37} & \textbf{98.79} & \textbf{96.58}   \\ 
			
			128  & 90.24  & 93.05 & 95.72 & 92.38 & 91.75 & 93.86 & 96.90 & 92.28 & 89.12 & 91.79 & 95.87 & 92.04   \\ 
			\hline
		\end{tabular}
	\end{center}
\end{table}

\textbf{The effects of different sampling counts during the training process.}As mentioned in Section 4.1, during training, we uniformly downsample point clouds to 2048 points to ensure batch‑processable inputs. However, during testing, the model is evaluated directly on the original full‑resolution point clouds without downsampling process. Considering the variations in scale and geometric complexity of building roofs, we further investigate the impact of different training‑time sampling count on RoofSeg's performance. Specifically, we conduct comparative experiments on 836 simple samples ($< 10$ planes) and 415 complex samples ($> 15$ planes) from the Building3D benchmark. As shown in Table~\ref{Tab:ablation_sampling_count}, as the sampling count increases, the performance on simple samples remains stable, while a marginal improvement (within 0.5 percentage points) is gained on complex samples. These results indicate that our network exhibits good robustness to different sampling counts.

\begin{table} [!htb] \renewcommand{\tabcolsep}{6.5 pt}
	\scriptsize
	\renewcommand{\arraystretch}{1.25}
	\newcommand{\tabincell}[2]{\begin{tabular}{@{}#1@{}}#2\end{tabular}}
	\begin{center}
		\caption{Effect of different sampling counts (during the training step) on roof samples with different complexities.}
		\label{Tab:ablation_sampling_count}
		\begin{tabular}{c|c|cccc|ccc} 
			\hline
			\multirow{2}{*}{Different complexities} & \multirow{2}{*}{\makecell[c]{Sampling count \\ in training step}} & \multicolumn{7}{c}{Building3D benchmark} \\
			\cline{3-9}
			& & mCov$\uparrow$       & mWCov$\uparrow$   &   mPrec$\uparrow$    &    mRec$\uparrow$       
			& $O_{seg}\downarrow$  & $U_{seg}\downarrow$  &  $E_{acc}\uparrow$      \\ 
			\hline
			\multirow{3}{*}{\makecell[c]{Simple samples \\ ($<10$ planes)}} & 2048  & 95.32 & 97.15 & 99.21 & 98.14 & 1.25 & 1.83 & 94.70    \\ 
			\cline{2-2}
			& 3072  & 95.29 & 97.03 & 99.39 & 98.05 & 1.18 & 1.95 & 94.68    \\ 
			\cline{2-2}
			& 4096  & 95.39 & 95.40 & 99.13 & 98.24 & 1.28 & 1.71 & 94.73    \\ 
			\hline
			\multirow{3}{*}{\makecell[c]{Complex samples \\ ($>15$ planes)}} &	2048  & 92.36 & 93.27 & 96.34 & 94.41 & 3.12 & 4.48 & 91.76    \\ 
			\cline{2-2}
			& 3072   & 92.77 & 93.65 & 96.75 & 94.52 & 2.93 & 4.14 & 91.89    \\ 
			\cline{2-2}
			& 4096   & 92.81 & 93.78 & 96.83 & 94.48 & 2.79 & 4.03 & 91.96    \\ 
			\hline
		\end{tabular}
	\end{center}
\end{table}

\subsection{Feature space analysis}
\label{appendix:feature_space_analysis}

\begin{figure*}[!htb]
	\centering
	\def\imgwidth{0.90\textwidth} 
	
	\begin{tabular}{c@{\hspace{2pt}}c@{\hspace{4pt}}c}
		& & 
		{\small
			\begin{tabularx}{\imgwidth}{*{4}{>{\centering\arraybackslash}X}}
				Plane label & Edge label & Traditional feats. & Learned feats. \\
				(GT) & (GT) & (Normal vector) & (RoofSeg)
		\end{tabularx}} \\
		\noalign{\vspace{2pt}} 

		$\vcenter{\hbox{\rotatebox{90}{w./o. noise}}}$ &
		$\left[ \vphantom{\vcenter{\hbox{\includegraphics[width=\imgwidth]{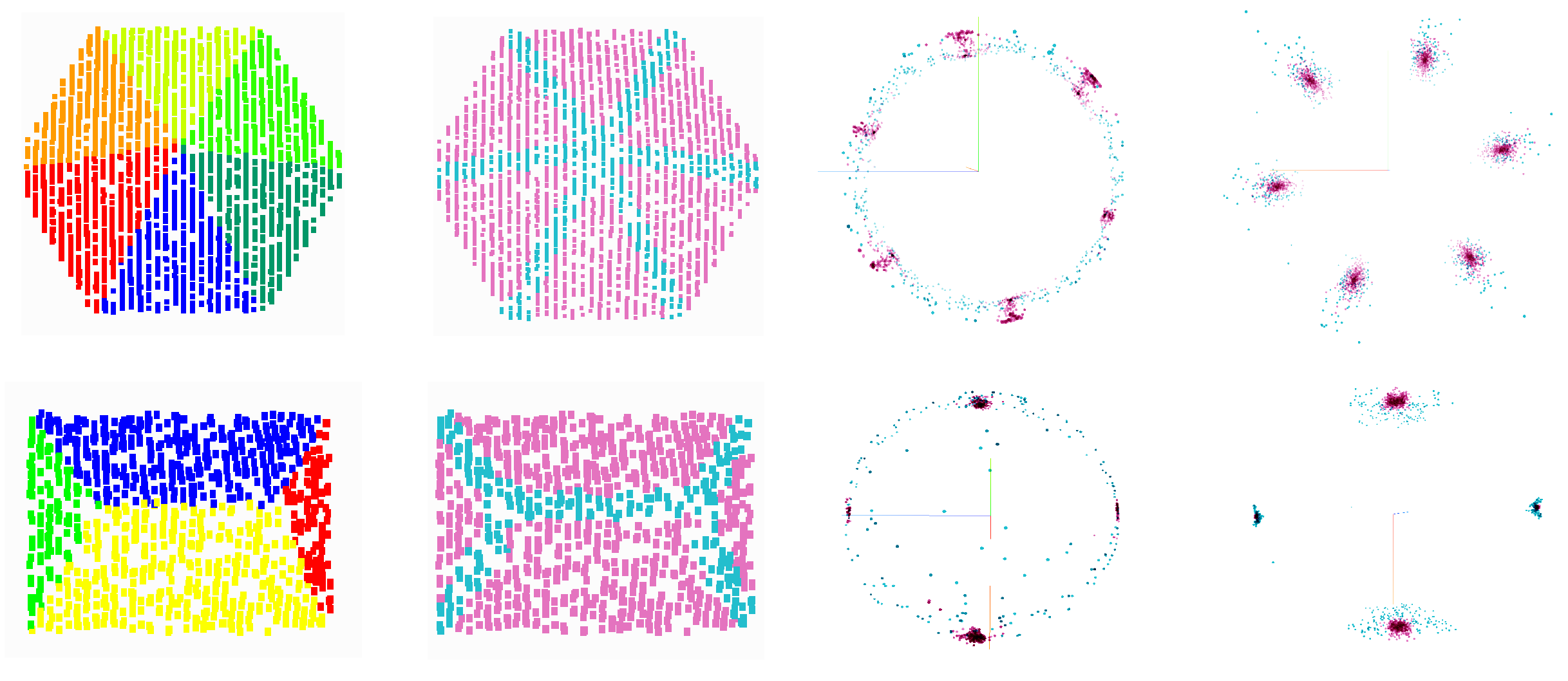}}}} \right.$ &
		$\vcenter{\hbox{\includegraphics[width=\imgwidth]{feats_wo_noise.png}}}$ \\
		
		\noalign{\vspace{3mm}}
		\cdashline{1-3}[4pt/3pt] 
		\noalign{\vspace{3mm}}
		
		$\vcenter{\hbox{\rotatebox{90}{w. noise}}}$ &
		$\left[ \vphantom{\vcenter{\hbox{\includegraphics[width=\imgwidth]{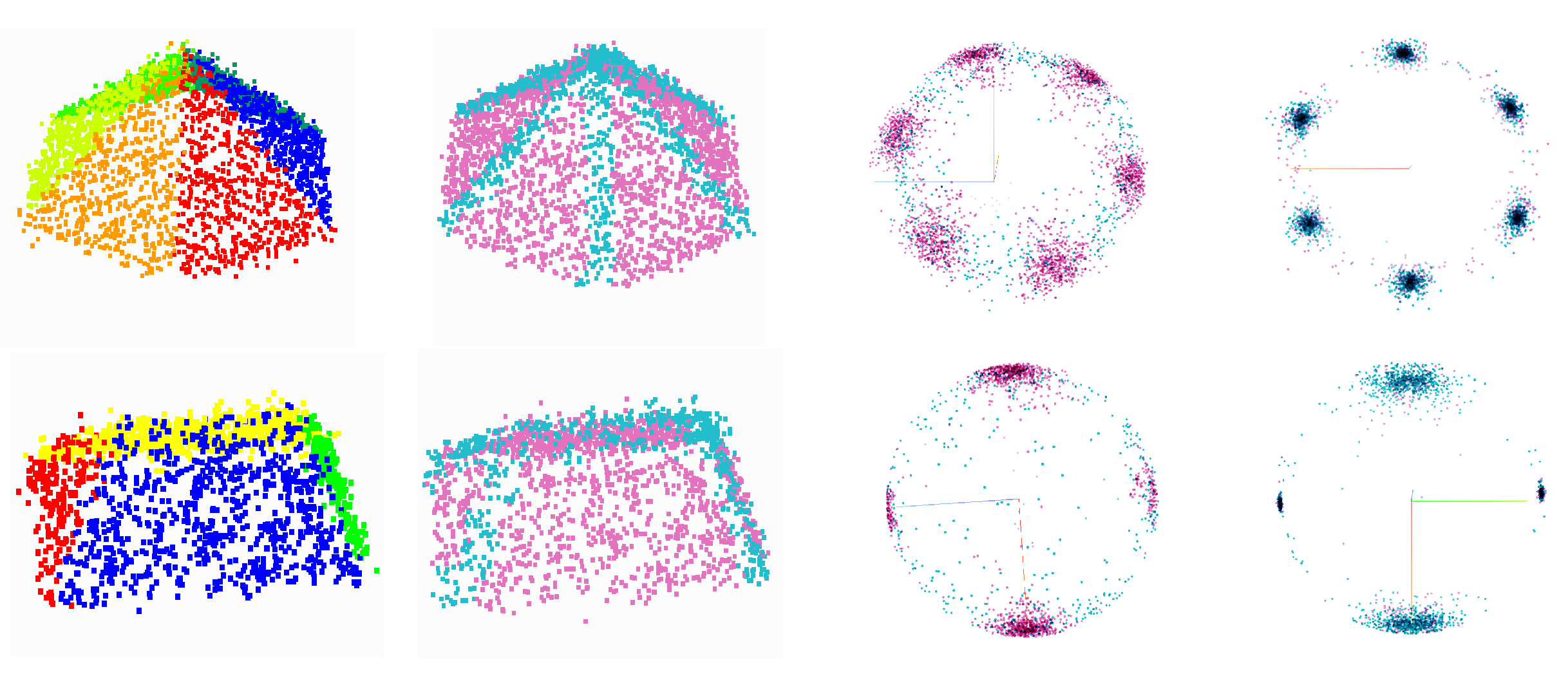}}}} \right.$ &
		$\vcenter{\hbox{\includegraphics[width=\imgwidth]{feats_w_noise.png}}}$ \\
	\end{tabular}
	
	\caption{Visualization comparison between traditional geometric features, and learned features of RoofSeg under noise-free and noisy conditions. The cyan points denote ground-truth edges, while pink points represent non-edge (interior) points.}
	\label{Fig:comparison_feats}
\end{figure*}

To rigorously justify the necessity of the proposed deep learning architecture over traditional handcrafted-feature-based approaches, we provide a comparative visualization of the feature spaces. As illustrated in Fig.~\ref{Fig:comparison_feats}, the high-dimensional feature spaces are visualized using Google's open-source Embedding Projector\footnote{\url{https://projector.tensorflow.org/}} under both clean and noisy conditions. In these visualizations, cyan points denote ground-truth edges, while pink points represent non-edge (interior) points.

\textbf{Limitations of traditional geometric features.} Traditional Principal Component Analysis (PCA) is effective at extracting locally consistent segments. As shown in Column 3 of Fig.~\ref{Fig:comparison_feats}, traditional normal features form tight, distinct clusters for interior points. However, in roof edge regions, these features become highly ambiguous. The edge points (cyan) are widely scattered and heavily mixed with interior points. This ambiguity arises because traditional PCA relies on a fixed local neighborhood (e.g., K-Nearest Neighbors, KNN). Near edges, a point's neighborhood inevitably contains points from multiple intersecting planes, leading to a mixed or averaged normal estimation that belongs to neither plane. Furthermore, this feature confusion is severely amplified under noisy conditions (Column 3, bottom rows of Fig.~\ref{Fig:comparison_feats}).

\textbf{Plane-prior-guided edge-aware features in RoofSeg.} The fundamental limitation of traditional methods at edges is their lack of instance-level plane priors. RoofSeg overcomes this by learning plane-prior-guided edge-aware features. Instead of relying solely on local neighborhoods, RoofSeg first predicts initial masks for individual planes, which provide a global instance-level prior. For potential edge points (predicted by the edge branch), the network explicitly calculates the Euclidean distance from the point to the fitted plane derived from the predicted mask. Then, through the proposed Edge Aware Mask Model (EAMM), these critical geometric priors are fused with the initially learned edge features, thereby enforcing edge discriminability.

As demonstrated in Column 4 of Fig.~\ref{Fig:comparison_feats}, the learned features of RoofSeg successfully distinguish edge points from interior points, maintaining high discriminability even under severe noise. The network is forced to learn how geometrically consistent an edge point is with a specific candidate plane instance derived from the predicted plane mask, rather than merely averaging local point features. This specific, instance-aware geometric feature is inherently impossible to capture using traditional bottom-up PCA methods.

\subsection{Impact on building reconstruction task}
\label{appendix:impact_on_building_reconstruction_tasks}
To further evaluate the impact of roof segmentation quality on downstream 3D building reconstruction, we conducted an additional reconstruction experiment on the Building3D benchmark. The segmentation results generated by Region Growing, RANSAC, GoCoPP, and the proposed RoofSeg were used as input primitives for PolyFit~\citep{nan2017polyfit}-based building reconstruction. In addition, the ground-truth plane segments were also used for PolyFit reconstruction to provide an upper-bound reference under ideal segmentation conditions.

Given Building3D provides roof-only point clouds, we completed them before PolyFit reconstruction for generating closed building model. The height of bottom plane was obtained from the ground-truth mesh. In practical applications, it can instead be determined from the local ground elevation when ground points are available. The footprint was estimated from vertically projected roof points using 2D alpha-shape detection algorithm implemented in CGAL.\footnote{CGAL, "2D Alpha Shapes", \url{https://doc.cgal.org/latest/Alpha_shapes_2/index.html}} We then uniformly sampled points on the bottom and façade planes and combined them with the original roof points. The same completion procedure was applied to all approaches to ensure a fair comparison.

\textbf{Evaluation metrics.} Following the evaluation metrics used in Building3D~\citep{wang2023building3d} and Point2Building~\citep{liu2024point2building}, we adopted five metrics to quantitatively compare the reconstructed meshes, including Mean Distance Error (MDE), Chamfer Distance (CD), Hausdorff Distance (HD), mesh mIoU, and RMS point-to-mesh error. For surface-based distance evaluation, 10,000 points were uniformly sampled from the ground-truth mesh surface and compared with the reconstructed mesh surface. MDE measures the average geometric deviation between the reconstructed and ground-truth surfaces, CD evaluates the bidirectional average closest-point distance, HD measures the maximum surface deviation, mesh mIoU evaluates the overlap between the predicted and ground-truth meshes, and RMS measures the average point-to-mesh fitting error. For MDE, CD, HD, and RMS, lower values indicate better reconstruction quality, whereas for mesh mIoU, higher values indicate better performance.

\textbf{Quantitative results.} As shown in Table~\ref{Tab:reconstruction_comparison}, the reconstruction results based on RoofSeg achieve better geometric accuracy than those based on traditional segmentation methods. Compared with Region Growing, RANSAC, and GoCoPP, RoofSeg obtains lower MDE, CD, HD, and RMS values, while achieving higher mesh mIoU. Specifically, RoofSeg achieves 0.189 MDE, 0.384 CD, 1.344 HD, 0.958 mIoU, and 0.0316 RMS. These results are closer to those obtained using the ground-truth plane segments, indicating that RoofSeg provides more accurate and reconstruction-friendly roof plane segments.

\begin{figure*}[htbp]
	\centering
	\def\imgwidth{0.90\textwidth} 
	
	\begin{tabular}{c@{\hspace{2pt}}c@{\hspace{4pt}}c}
		& & 
		{\small 
		\begin{tabularx}{\imgwidth}{*{5}{>{\centering\arraybackslash}X}}
			\mbox{Region growing} & GoCoPP & DeepRoofPlane & RoofSeg & \mbox{Ground truth}
			\end{tabularx}} \\
		\noalign{\vspace{2pt}} 
		
		$\vcenter{\hbox{\rotatebox{90}{view1}}}$ &
		$\left[ \vphantom{\vcenter{\hbox{\includegraphics[width=\imgwidth]{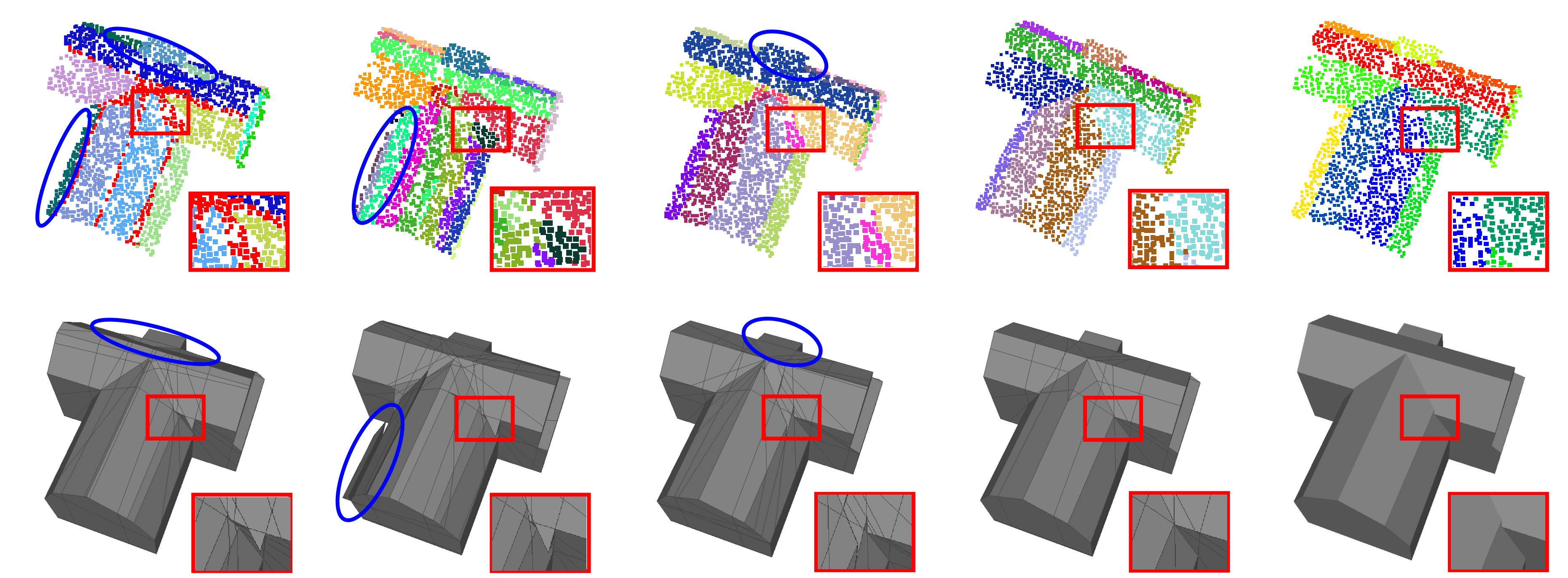}}}} \right.$ &
		$\vcenter{\hbox{\includegraphics[width=\imgwidth]{model_top_view_1.png}}}$ \\
		\noalign{\vspace{2pt}}
		
		$\vcenter{\hbox{\rotatebox{90}{view2}}}$ &
		$\left[ \vphantom{\vcenter{\hbox{\includegraphics[width=\imgwidth]{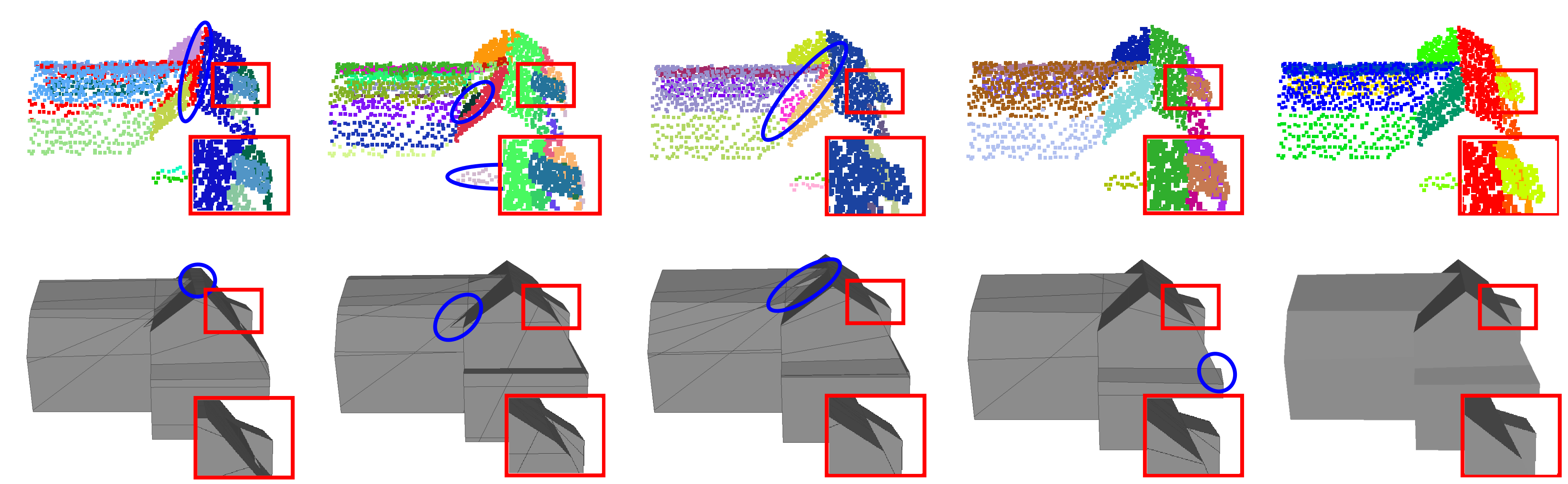}}}} \right.$ &
		$\vcenter{\hbox{\includegraphics[width=\imgwidth]{model_top_view_2.png}}}$ \\
		
		\noalign{\vspace{3mm}}
		\cdashline{1-3}[4pt/3pt] 
		\noalign{\vspace{3mm}}
		
		$\vcenter{\hbox{\rotatebox{90}{view1}}}$ &
		$\left[ \vphantom{\vcenter{\hbox{\includegraphics[width=\imgwidth]{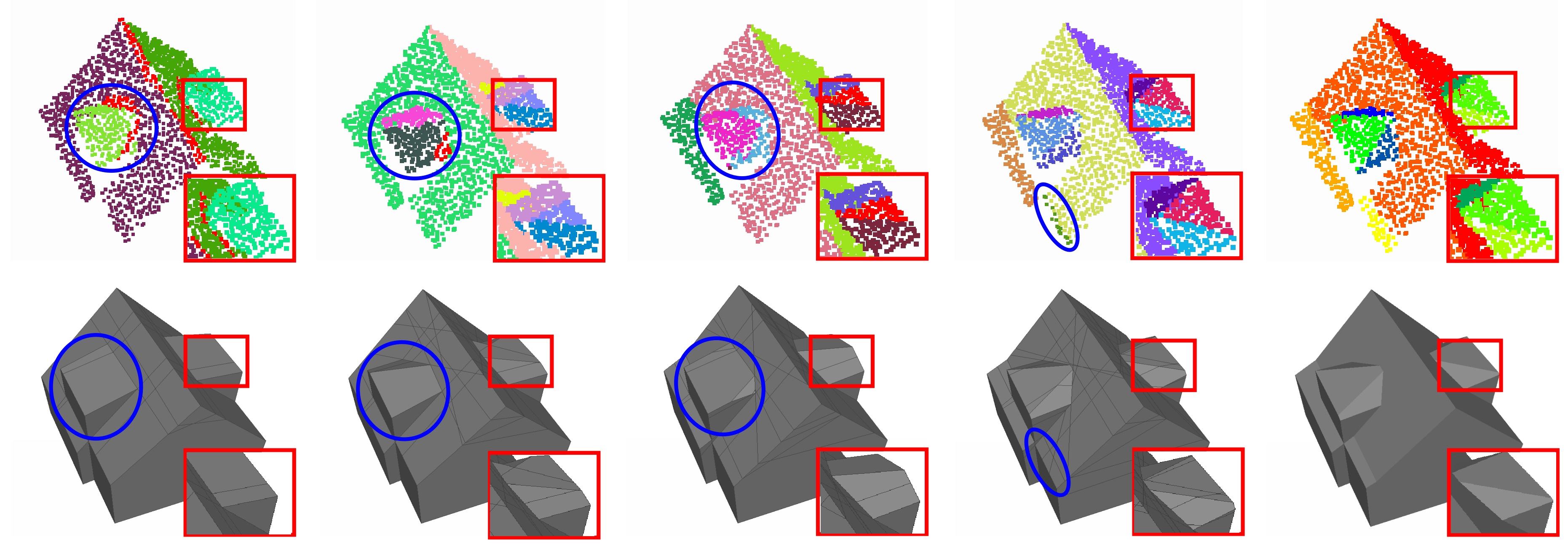}}}} \right.$ &
		$\vcenter{\hbox{\includegraphics[width=\imgwidth]{model_bottom_view_1.png}}}$ \\
		\noalign{\vspace{2pt}}
		
		$\vcenter{\hbox{\rotatebox{90}{view2}}}$ &
		$\left[ \vphantom{\vcenter{\hbox{\includegraphics[width=\imgwidth]{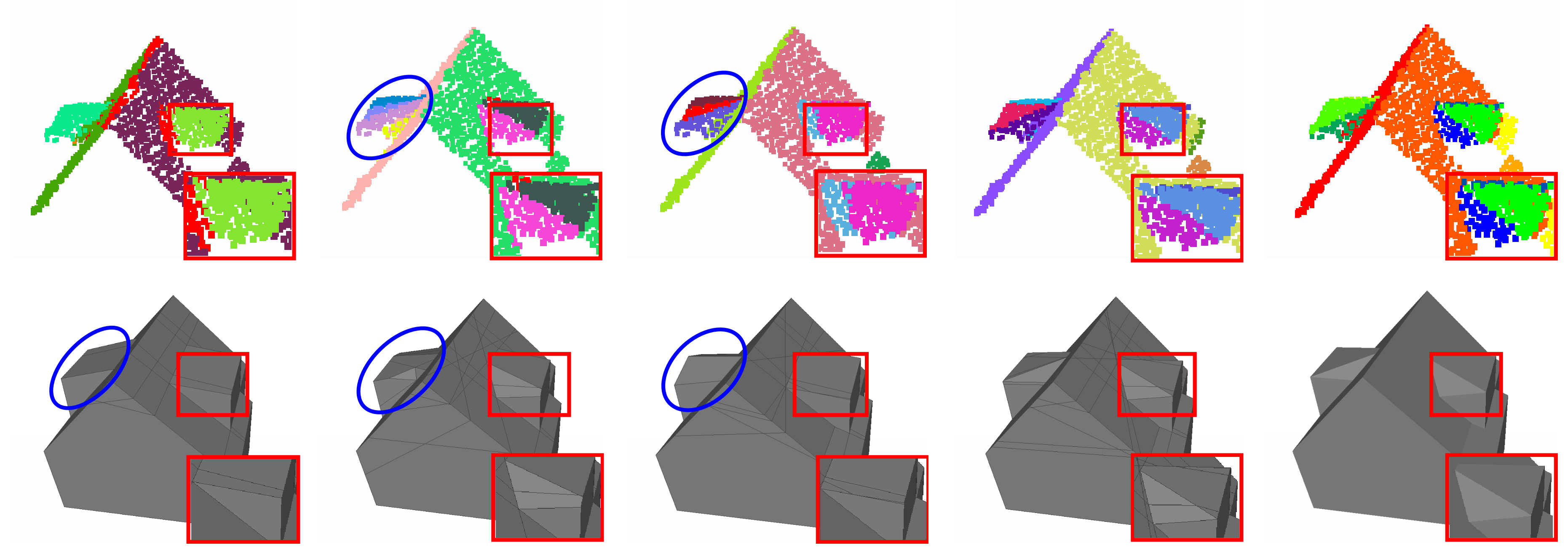}}}} \right.$ &
		$\vcenter{\hbox{\includegraphics[width=\imgwidth]{model_bottom_view_2.png}}}$ \\
	\end{tabular}
	
	\caption{Comparison of building reconstruction results based on different segmentation approaches with PolyFit algorithm on Building3D benchmark.}
	\label{Fig:reconstruction_comparison}
\end{figure*}

\textbf{Qualitative results.} Fig.~\ref{Fig:reconstruction_comparison} further provides qualitative comparisons of reconstructed building models using different segmentation results as input. Reconstruction results based on traditional approaches exhibit missing roof planes, inaccurate plane merging, and distorted geometric structures, especially in complex roof regions. In contrast, the models reconstructed from RoofSeg segmentation results preserve roof topology more faithfully and show cleaner surface geometry. These results demonstrate that the proposed segmentation approach not only improves plane segmentation accuracy, but also benefits downstream building reconstruction task.

\begin{table}[!htb]
	\renewcommand{\tabcolsep}{9 pt}
	\scriptsize
	\renewcommand{\arraystretch}{1.3}
	\begin{center}
		\caption{Quantitative comparison of reconstructed model quality with the PolyFit algorithm on the Building3D benchmark. Values in parentheses indicate the deviation from the results reconstructed using GT segments.}
		\label{Tab:reconstruction_comparison}
		\begin{tabular}{c|ccccc}
			\hline
			Approaches & MDE$\downarrow$ & CD$\downarrow$ & HD$\downarrow$ & mIoU$\uparrow$ & RMS$\downarrow$ \\
			\hline
			Region growing & 0.415 (+0.280) & 0.648 (+0.305) & 1.711 (+0.395) & 0.920 (-0.042) & 0.078 (+0.051) \\
			RANSAC         & 0.395 (+0.261) & 0.582 (+0.239) & 1.680 (+0.365) & 0.921 (-0.041) & 0.074 (+0.047) \\
			GoCoPP         & 0.264 (+0.129) & 0.463 (+0.120) & 1.501 (+0.185) & 0.947 (-0.015) & 0.054 (+0.026) \\
			ParseNet       & 0.335 (+0.200) & 0.529 (+0.186) & 1.612 (+0.296) & 0.933 (-0.029) & 0.067 (+0.039) \\
			HPNet          & 0.327 (+0.192) & 0.519 (+0.177) & 1.598 (+0.282) & 0.935 (-0.027) & 0.065 (+0.037) \\
			Mask3D         & 0.293 (+0.158) & 0.490 (+0.147) & 1.548 (+0.232) & 0.941 (-0.021) & 0.058 (+0.030) \\
			DeepRoofPlane  & 0.252 (+0.117) & 0.469 (+0.127) & 1.487 (+0.171) & 0.945 (-0.017) & 0.053 (+0.026) \\
			RoofSeg (ours) & \textbf{0.189} (+0.054) & \textbf{0.384} (+0.042) & \textbf{1.344} (+0.029) & \textbf{0.958} (-0.004) & \textbf{0.032} (+0.004) \\
			\hline
			GT segments    & 0.135 & 0.343 & 1.316 & 0.962 & 0.027 \\
			\hline
		\end{tabular}
	\end{center}
\end{table}

\bibliographystyle{elsarticle-harv} 
\bibliography{refs}

\end{document}